\documentclass[a4paper,12pt]{article}

\usepackage[T1]{fontenc}
\usepackage{times}
\usepackage[english]{babel}
\usepackage[utf8]{inputenc}
\usepackage{dtklogos}
\usepackage{wallpaper}
\usepackage[absolute]{textpos}
\usepackage[top=2cm, bottom=2.5cm, left=3cm, right=3cm]{geometry}
\usepackage{appendix}
\usepackage[nottoc]{tocbibind}
\usepackage[colorlinks=true,
            linkcolor=black,
            urlcolor=blue,
            citecolor=black]{hyperref}

\usepackage{algorithm}
\usepackage{algpseudocode}
\usepackage{amsmath}
\usepackage{sectsty}
\usepackage{svg}
\usepackage{booktabs} 
\usepackage{longtable} 
\usepackage{array}     
\usepackage{ragged2e}  
\usepackage{tabularx} 
\usepackage{enumitem}

\sectionfont{\fontsize{14}{15}\selectfont}
\subsectionfont{\fontsize{12}{15}\selectfont}
\subsubsectionfont{\fontsize{12}{15}\selectfont}

\usepackage{csquotes} 

\newsavebox{\mybox}
\newlength{\mydepth}
\newlength{\myheight}

\newenvironment{sidebar}%
{\begin{lrbox}{\mybox}\begin{minipage}{\textwidth}}%
{\end{minipage}\end{lrbox}%
 \settodepth{\mydepth}{\usebox{\mybox}}%
 \settoheight{\myheight}{\usebox{\mybox}}%
 \addtolength{\myheight}{\mydepth}%
 \noindent\makebox[0pt]{\hspace{-20pt}\rule[-\mydepth]{1pt}{\myheight}}%
 \usebox{\mybox}}

\newcommand\BackgroundPic{
    \put(-2,-3){
    \includegraphics[keepaspectratio,scale=0.3]{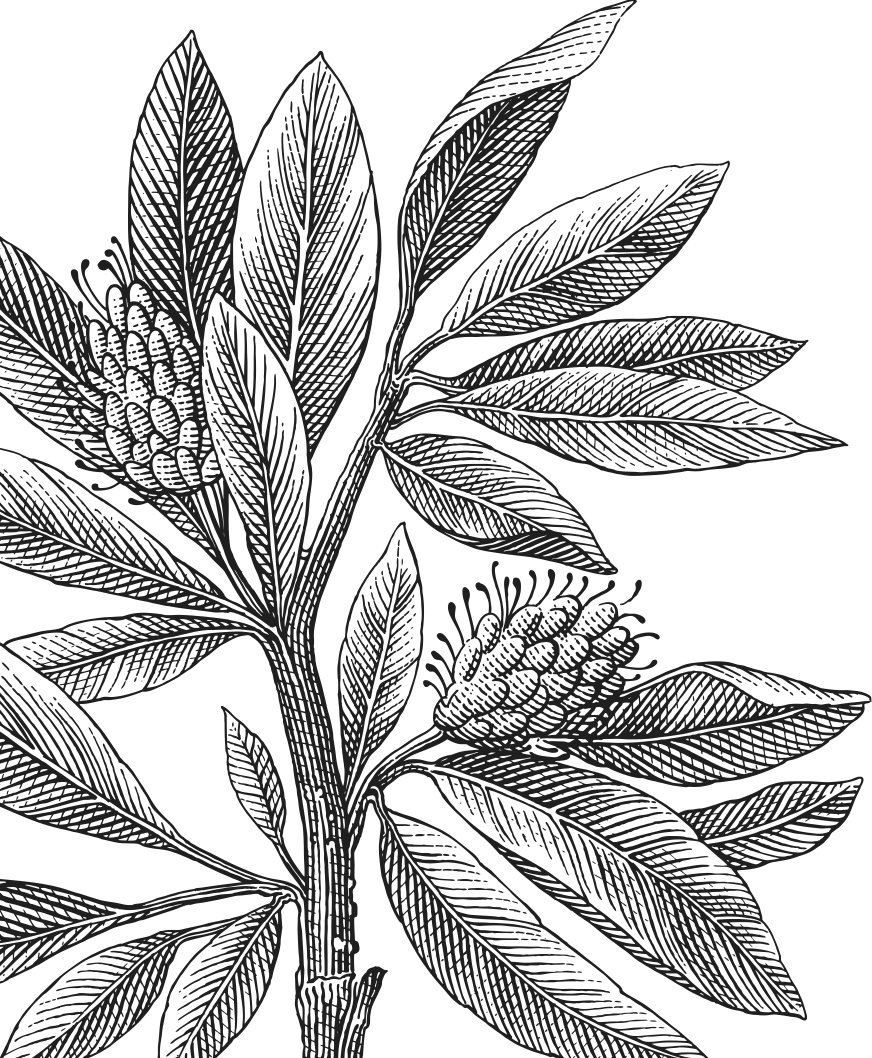} 
    }
}
\newcommand\BackgroundPicLogo{
    \put(30,740){
    \includegraphics[keepaspectratio,scale=0.10]{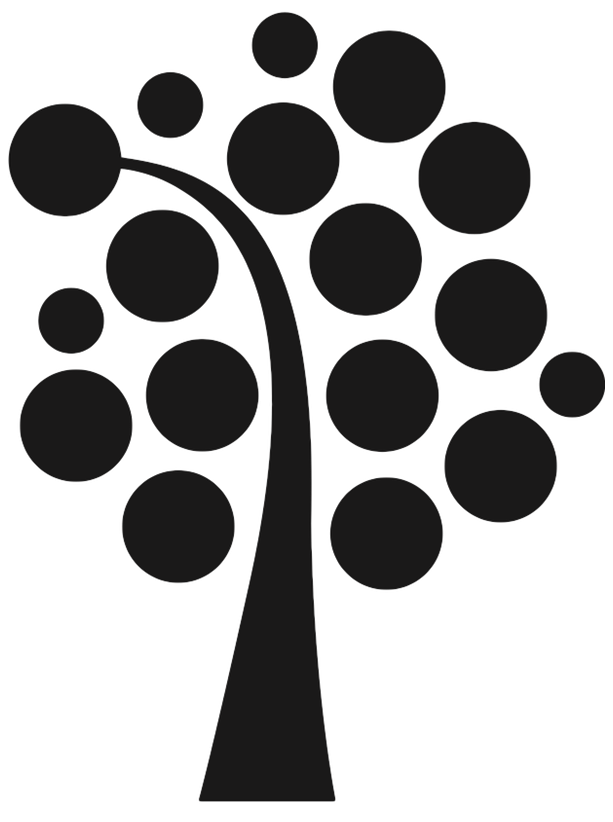} 
    }
}

\title{	
\vspace{-8cm}
\begin{sidebar}
    \vspace{10cm}
    \normalfont \normalsize
    \Huge Masters Degree Project \\
    \vspace{-1.3cm}
\end{sidebar}
\vspace{3cm}
\begin{flushleft}
    \huge VILOD: A Visual Interactive Labeling Tool for Object Detection\\ 
    \it \LARGE
\end{flushleft}
\null
\vfill
\begin{textblock}{6}(10,12)
\begin{flushright}
\begin{minipage}{\textwidth}
\begin{flushleft} \large
\emph{Author(s):} Isac Holm\\ 
\emph{Supervisor(s):} Amilcar Soares\\ 
\emph{Semester:} HT 2025\\ %
\emph{Course:} 5DT90E\\ %
\emph{Subject:} Computer Science\ 
\end{flushleft}
\end{minipage}
\end{flushright}
\end{textblock}
}

\date{} 

\begin{document}
\pagenumbering{gobble}
\newgeometry{left=5cm}
\AddToShipoutPicture*{\BackgroundPic}
\AddToShipoutPicture*{\BackgroundPicLogo}
\maketitle
\restoregeometry
\clearpage
\selectlanguage{english}
\begin{abstract}
\noindent The advancement of Object Detection (OD) using Deep Learning (DL) is often hindered by the significant challenge of acquiring large, accurately labeled datasets, a process that is time-consuming and expensive. While techniques like  Active Learning (AL) can reduce annotation effort by intelligently querying informative samples, they often lack transparency, limit the strategic insight of human experts, and may overlook informative samples not aligned with an employed query strategy. To mitigate these issues, Human-in-the-Loop (HITL) approaches integrating human intelligence and intuition throughout the machine learning life-cycle have gained traction. Leveraging Visual Analytics (VA), effective interfaces can be created to facilitate this human-AI collaboration. This thesis explores the intersection of these fields by developing and investigating \textbf{VILOD: A Visual Interactive Labeling tool for Object Detection}, a system designed to support expert users in the OD annotation process by integrating interactive visualizations, with AL suggestions and guidance. VILOD utilizes components such as a t-SNE projection of image features, together with uncertainty heatmaps and model state views. Enabling users to explore data, interpret model states, AL suggestions, and implement diverse sample selection strategies within an iterative HITL workflow for OD. An empirical investigation using comparative use cases demonstrated how VILOD, through its interactive visualizations, facilitates the implementation of distinct labeling strategies by making the model's state and dataset characteristics more interpretable (RQ1). Furthermore, the study showed that different visually-guided labeling strategies employed within VILOD result in competitive OD performance trajectories compared to an automated uncertainty sampling AL baseline (RQ2). Specifically, the \textit{Balanced Guidance Integration} strategy, which synthesized information from all available cues, achieved the highest final model performance. Suggesting that human guidance effectively supported by interactive VA can offer an edge in performance and provide important benefits such as quality control and strategic depth in the OD annotation process. This work contributes a novel tool and empirical insight into making the HITL-AL workflow for OD annotation more transparent, manageable, and potentially more effective. 

\end{abstract}



\newpage

\pagenumbering{gobble}
\tableofcontents 
\newpage
\pagenumbering{arabic}

%
%

\section{Introduction}
\label{introduction}


\noindent The rapid advancement of Artificial Intelligence (AI), particularly in the domain of Computer Vision, has led to increasing capabilities in analyzing and understanding visual data. 
One application is Object Detection (OD) \cite{zou2023object}, a technology enabling machines to not only identify objects within images or videos but also to precisely locate them. This capability is used in many critical systems, ranging from autonomous vehicles navigating complex environments to medical imaging systems aiding in disease diagnosis \cite{wang2024comprehensive}. The success of modern OD systems is largely attributed to the power of Deep Learning (DL), a subset of Machine Learning (ML) that utilizes deep neural networks to learn intricate patterns from vast amounts of data. However, this reliance on data presents a significant challenge: the creation of large, accurately labeled datasets required to train these models is time-consuming, expensive, and often requires specialized human expertise \cite{zou2023object}.

Techniques like Active Learning (AL) \cite{settles2009active} offer a way to intelligently select the most informative samples for labeling, thereby reducing the overall annotation effort. 
It provides little insight into their selection process and limits the potential for human experts to contribute their domain knowledge strategically. Furthermore, purely algorithmic selection might overlook samples that do not align with simple predefined metrics like model uncertainty. 

To address these limitations, the concept of Human-in-the-Loop (HITL) \cite{wu2022survey} machine learning has gained popularity, advocating for a collaborative approach where human intelligence is integrated throughout the ML lifecycle. 
Within this, Visual Analytics (VA) \cite{ThomasCook2005VA,abreu2021trajectory} and Interactive Machine Learning (IML) \cite{amershi2014power} offer powerful tools and methodologies to create effective interfaces that facilitate human-AI collaboration. 
By leveraging interactive visualizations, VA can make complex processes like AL more transparent and understandable for the user. 

This thesis explores the intersection of these fields, focusing on the development and investigation of an interactive visual tool designed to support expert users in the annotation process for object detection models. Specifically, it introduces "VILOD: A Visual Interactive Labelling tool for Object Detection", a system that integrates AL suggestions with interactive visualizations to facilitate more efficient and strategically informed data labeling. 
The tool aims not merely to optimize final model accuracy according to automated metrics but to investigate how such a visually-guided system could enhance the user's ability to understand the model's state, interpret AL suggestions, and implement a mixed sample selection strategy within an iterative HITL workflow. 
This work demonstrates the potential of visual interactive systems to transform the task of OD annotation into a more manageable and ultimately more effective process.

\subsection{Background}

\noindent To fully understand the context of this thesis, it is essential to understand the fundamental concepts and technologies involved. 
This chapter provides the necessary background on key concepts such as: Object Detection, Deep Learning, Data annotation, Active Learning, Human-in-the-Loop systems, Visual Analytics, Interactive Machine Learning, and Transfer Learning.

\subsubsection{Object Detection}

Object detection stands as a cornerstone technology within the broader field of computer vision. The goal of an object detection model is not only to predict the type of object present within an image or video (classification), but also to determine where they are located (localization)  \cite{zou2023object}. 
This typically involves predicting a bounding box, which is a rectangular region enclosing each detected object along with a class label for that object. More advanced types of object detection exist, such as segmentation. 
This involves determining precise pixel-level boundaries of each object, offering a more granular understanding of a scene \cite{s25010214}. 
However, this thesis mainly focuses on bounding box predictions.

The ability to accurately detect and localize objects is critical for a vast array of applications that impact various sectors of society and industry. In autonomous vehicles, OD systems are used to understand the environment, such as reading traffic signs or detecting pedestrians. Furthermore, OD systems are used in the medical industry to identify anomalies such as tumors or fractures in scans like X-rays, CTs, or MRIs to aid in early diagnosis and treatment planning \cite{wang2024comprehensive}. Surveillance systems utilize OD for monitoring public spaces, detecting suspicious activities, and enhancing security \cite{s25010214}. 

\subsubsection{Deep Learning in Object Detection}

Deep Learning \cite{lecun2015deep} is a subfield of machine learning that uses neural networks with multiple layers to learn hierarchical representations of data. 
Unlike traditional ML models that often use networks with only one or two layers, deep learning architectures, particularly Convolutional Neural Networks (CNNs), have proven effective for visual tasks. CNNs are specifically designed to process grid-like data, such as images, by automatically learning relevant spatial hierarchies of features directly from the input pixels \cite{s25010214}. 
CNNs form the backbone of most state-of-the-art object detection frameworks. These frameworks can be broadly categorized into two-stage and single-stage detectors. Two-stage detectors, like the R-CNN family (R-CNN, Fast R-CNN, Faster R-CNN), first propose regions of interest in the image that likely contain objects and then classify the objects within those regions \cite{8825470, s25010214}. 

Single-stage detectors, exemplified by the YOLO (You Only Look Once) \cite{redmon2016you}  family and SSD (Single Shot MultiBox Detector) \cite{liu2016ssd}, try to streamline the process by performing localization and classification simultaneously in a single pass through the network \cite{8825470}. 
YOLO, introduced by Redmon et al. \cite{redmon2016you}, divides the input image into a grid system. 
Each grid cell is responsible for predicting bounding boxes, confidence scores, and class probabilities for objects whose centers fall within that cell. 
This approach makes YOLO models fast and suitable for real-time applications where speed is critical. The YOLO architecture has undergone numerous iterations, with each version introducing improvements in accuracy, speed, and the ability to detect objects at different scales. 
This thesis utilizes the latest YOLO11 \cite{yolo11_ultralytics} architecture and uses the pre-trained YOLO11n, a specific variant within the YOLO family, as the base detector within the interactive system.  

The performance of OD models is typically evaluated using metrics that consider both classification accuracy and localization precision. Key metrics include \textit{Intersection over Union} (IoU), which measures the overlap between a predicted bounding box and the ground-truth box. \textit{Precision}, the fraction of correct detections among all detections made. \textit{Recall}, the fraction of actual objects that were correctly detected. Finally, the \textit{Mean Average Precision} (mAP), a summary metric that averages precision across different recall levels and object classes, serves as a standard benchmark for comparing OD models \cite{8825470,UltralyticsPerfMetrics}.

\subsubsection{Data Annotation}

Data annotation, or labeling, is the process of adding informative metadata to raw data to make it understandable for machine learning algorithms \cite{soares2019vista}. 
For object detection, this typically involves drawing bounding boxes around each object instance in an image and assigning the correct class label. 
Creating these labeled datasets presents a bottleneck in the development of Object Detection systems. Several challenges contribute to this problem \cite{li2024survey,zou2023object}:

\begin{itemize}
    \item \textbf{Scale and Cost:} State-of-the-art models often require very large datasets of labeled images to achieve robust performance. 
    Manually annotating data at this scale is incredibly time-consuming and expensive, often requiring significant human resources \cite{junior2017analytic}. 
    The cost is further amplified when domain expertise is required, such as in medical image analysis, where annotations must be performed by trained radiologists or pathologists \cite{wang2024comprehensive}.
    \item \textbf{Complexity:} Annotating objects in real-world scenes can be complex. Objects may be partially hidden, appear at vastly different scales, or be present in challenging lighting or weather conditions. Accurately drawing tight bounding boxes in such scenarios requires attention to detail.
    \item \textbf{Subjectivity and Quality:} Annotation can sometimes involve subjective judgment, especially for defining object boundaries or classifying instances. Different annotators might interpret guidelines differently, leading to inconsistencies in the labels. Ensuring high quality and consistency across large datasets requires clear instructions, quality control processes, and possibly multiple rounds of review, adding further to the cost and complexity.
\end{itemize}

\noindent This annotation bottleneck significantly slows down the progress and practical application of OD and other supervised learning models, motivating the search for more efficient data labeling strategies.

\subsubsection{Active Learning}

Active Learning is a strategy to combat the problem of extensive data annotations. It's a machine learning technique where the learning model itself plays an active role in selecting the data from which it learns \cite{settles2009active}. 
Instead of receiving a large, pre-labeled dataset, an AL algorithm interactively queries a human annotator to label specific data points that are deemed most beneficial for improving the model's performance. 

The goal of AL is to achieve high model accuracy with significantly fewer labeled samples compared to traditional supervised learning or random sampling approaches. It operates on the premise that not all data points are equally informative. By strategically selecting the most informative unlabeled instances for annotation, AL aims to maximize the model's learning efficiency and reduce the overall labeling cost. 

\noindent The typical pool-based AL process follows an iterative cycle:

\begin{enumerate}
    \item Train an initial model on a small set of labeled data ($D_{l,0}$)
    \item Use the model to predict the labels of a pool of unlabeled data ($D_u$)
    \item A \textit{Query Strategy} is employed to analyze the model's predictions and selects a subset of unlabeled instances from $D_u$ deemed most informative. 
    \item The selected samples are presented to the human for manual labeling/annotation.
    \item Newly labeled samples are added to the labeled set $D_l$
    \item The model is retrained using the expanded $D_l$
    \item Steps 2-6 are repeated until a predefined stopping criterion is met. 
\end{enumerate}

\begin{figure}[ht!]
\begin{center}
\includegraphics*[width=0.9\columnwidth]{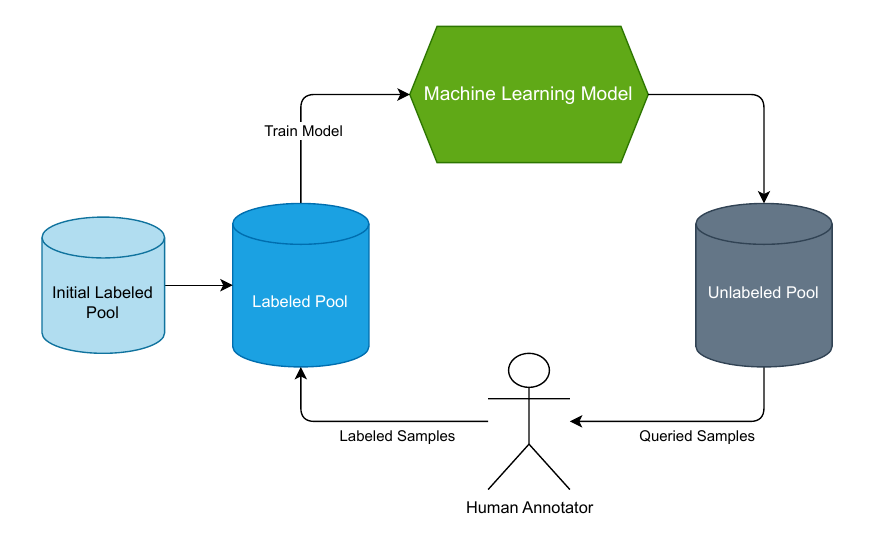}
\end{center}
\caption{Pool Based Active Learning Cycle.}
\label{AciveLearningCycle}
\end{figure}

\noindent The effectiveness of the AL relies on the query strategy used to select the samples. Li et al. \cite{li2024survey} defines five groups of common querying strategies: \\

\paragraph{Uncertainty-Based Strategies.} Uncertainty-based methods focus on selecting samples where the current deep learning model exhibits the highest ambiguity or lowest confidence in its predictions. Common techniques measure uncertainty through predictive entropy, least confidence scores, the smallest margin between top predictions, or disagreement among a committee of models. The idea is that labeling these uncertain instances provides the model with the most information to refine its decision boundaries. However, potential drawbacks include selecting redundant samples, overlooking dataset coverage, sensitivity to outliers, and task-specific designs with limited generalizability.

\paragraph{Representative-Based Strategies} These strategies try to select samples that best represent the underlying structure or distribution of the entire unlabeled dataset. They often fall into two sub-categories: density-based methods, which select prototypical points like cluster centers or those maximizing probability coverage, and diversity-based methods, which prioritize samples that are different from already labeled instances, often using distance metrics, networks, or clustering. These methods are effective at capturing the data distribution, but can have a high computational complexity, especially with large datasets. 

\paragraph{Influence-Based Strategies.} Influence-based methods select unlabeled samples predicted to have the greatest impact on the model's performance or parameters if they were to be labeled. This impact can be estimated by measuring expected changes in model gradients, loss functions, or prediction errors. Alternatively, some approaches use reinforcement learning policies to learn which samples yield the highest rewards (e.g., performance improvement) or train separate models to estimate the potential influence of labeling a given sample. A challenge for these methods is often the significant computational cost associated with estimating influence or training auxiliary policies. 

\paragraph{Bayesian Strategies.} These methods leverage Bayesian principles, employing models like Bayesian Neural Networks (BNNs) or Gaussian Processes, to quantify uncertainty and guide sample selection. Bayesian approaches aim to minimize classification errors and improve model beliefs by selecting samples based on metrics derived from Bayesian inference, such as maximizing the mutual information between model predictions and parameters. An advantage is their ability to provide probabilistic uncertainty estimates that can offer explanations for sample selection.

\paragraph{Hybrid Strategies.} Hybrid strategies attempt to combine the strengths of two or more different query criteria, such as uncertainty and representativeness (diversity), to achieve a better trade-off. These can offer flexibility and potentially overcome the limitations of individual strategies. However, designing effective hybrid methods and determining the optimal balance between different criteria can be complex and often requires careful tuning.

\subsubsection{Human-in-the-Loop}

While AL focuses on optimizing which data points a human should label, HITL represents a broader approach that emphasizes the collaborative synergy between humans and AI systems throughout the machine learning lifecycle. 
The core concept is to combine the computational power and pattern-recognition capabilities of machines with the domain expertise, contextual understanding, and judgment of humans. \cite{wu2022survey} 

HITL systems are designed to leverage human input at various stages, including data preprocessing, annotation, model training, evaluation, and deployment. In the context of data annotation, HITL often involves workflows where AI models first generate preliminary labels or predictions, which are then reviewed, corrected, or validated by human annotators. 
This iterative feedback loop allows the AI model to learn from human corrections and improve its performance over time. Active learning is frequently employed within HITL systems as the mechanism for intelligently selecting which AI outputs or data points require human review, thereby optimizing the use of human expertise. 

The integration of humans in the loop offers several advantages. 
Humans excel at handling ambiguity, understanding context, identifying subtle errors, and dealing with rare or unforeseen edge cases where purely automated systems might falter \cite{wu2022survey}. 
Human oversight can also be good for mitigating biases that might be present in the data or learned by the model, as well as building trust in AI systems, particularly in crucial applications like healthcare or finance.

The approach taken in this thesis aligns with the HITL paradigm, where the user is actively involved in the iterative loop of selecting samples, annotating them, and retraining the model.

\subsubsection{Visual Analytics \& Interactive Machine Learning}

Effectively managing the interaction between humans and machine learning models within an AL or HITL framework requires sophisticated interfaces. Visual Analytics provides the scientific foundation for designing such interfaces. VA is defined as "the science of analytical reasoning facilitated by interactive visual interfaces" \cite{ThomasCook2005VA,abreu2021local}. It combines automated analysis techniques with interactive visualizations to empower users to explore complex data, discover patterns, gain insights, and make informed decisions. Key benefits include the ability to identify trends and outliers visually, explore data from multiple perspectives interactively, and ultimately arrive at actionable insights more quickly than through purely computational or manual methods. 

Interactive Machine Learning \cite{amershi2014power} is a closely related field that focuses specifically on developing techniques and tools that allow users, often without deep ML expertise, to interact with learning algorithms. 
IML systems enable users to provide data, feedback, or corrections through intuitive interfaces, thereby guiding the learning process and personalizing the resulting model \cite{dudley2018review}. The process is typically iterative. The user provides input, the ML system updates and presents results, the user evaluates and provides further feedback, and the cycle repeats until a satisfactory outcome is achieved. 

\subsubsection{Dimensionality Reduction}

High-dimensional data, such as the feature vectors extracted from images by deep learning models, poses challenges for direct analysis and visualization. 
Dimensionality Reduction is a methodology designed to transform data from a high-dimensional space into a lower-dimensional space while retaining meaningful properties and structure of the original data \cite{rajaraman2011mining}. 
The goals are often to simplify the data for easier visualization (typically in 2D or 3D), reduce computational complexity for subsequent processing, or mitigate issues related to the "curse of dimensionality".

One widely used method for non-linear dimensionality reduction is t-Distributed Stochastic Neighbor Embedding (t-SNE) \cite{maaten2008visualizing}. 
T-SNE models the similarity between high-dimensional data points as conditional probabilities and computes similar probabilities for the corresponding low-dimensional points. 
It then optimizes the low-dimensional embedding to minimize the divergence (specifically, Kullback-Leibler divergence) between these two sets of probabilities, with a focus on preserving the local structure and revealing clusters within the data.

Another technique is Uniform Manifold Approximation and Projection (UMAP) \cite{mcinnes2018umap}. 
UMAP is based on manifold learning theory and constructs a high-dimensional graph representation of the data, then optimizes a low-dimensional graph layout to be as structurally similar as possible. Compared to t-SNE, UMAP is often considerably faster computationally and can be more effective at preserving both local neighborhood structure and the broader global structure of the data \cite{mcinnes2018umap}.

Visualizing extracted image features using techniques like t-SNE or UMAP is central to many VA systems for machine learning. 
It allows users to see relationships between images based on their learned features, revealing clusters, outliers, and other patterns relevant for understanding the dataset and guiding the sample selection process in interactive labeling scenarios. In this work, t-SNE will be further explored in such scenario. 

\subsubsection{Transfer Learning}

Training large deep learning models, such as YOLO \cite{redmon2016you} from scratch requires large amounts of labeled data and computational resources. 
Transfer Learning (TL) is a technique to mitigate this requirement \cite{zhuang2020comprehensive,bappee2021examining}. 
The core idea of TL is to leverage knowledge gained from training a model on a large source task or dataset (e.g., image classification on ImageNet) and apply it to a different but related target task (e.g., object detection on a specific custom dataset). 

Instead of initializing the model weights randomly, TL starts with the weights of a pre-trained model. 
This pre-trained model has already learned general features from the source data (e.g., edges, textures, basic shapes) that can be relevant and transferable to the target task. 
The model is then adapted, via fine-tuning on the smaller target dataset. 
Fine-tuning involves resuming the training process, allowing the models' weights to adjust to the target task and data distribution.  

The primary advantages of TL are reduced training time and computational cost, and the ability to achieve good performance even with relatively small labeled datasets for the target task. 
By reusing the learned features from the pre-trained model, the network requires less data and fewer training iterations to converge on the target task. 
Furthermore, starting with a pre-trained model helps mitigate the 'cold start' problem often encountered in active learning, where initial models trained on very few labels perform poorly, leading to ineffective sample selection in early cycles \cite{li2024survey}. 
This makes TL particularly valuable in the context of this thesis. 

In this work, using a pre-trained YOLOv11n model as the starting point makes the iterative retraining process computationally feasible and allows the model to benefit from the knowledge embedded in its initial weights. 
Additionally, the feature representations learned by the pre-trained model can be leveraged for other tasks, such as extracting high-dimensional feature vectors from images in the target dataset, which can then be reduced using previously discussed techniques like t-SNE or UMAP to create visualizations supporting the interactive labeling workflow. 

\subsubsection{Summary}

The progression from the success of Deep Learning to the challenge of data annotation, and subsequently to the development of techniques like Active Learning, Human-in-the-Loop, Visual Analytics, and Transfer Learning, forms a coherent narrative to motivate this project. 
Deep Learning's demand for data created the annotation bottleneck. 
Active Learning offers efficiency gains but can lack transparency and strategic depth. 
Human-in-the-Loop brings human expertise into the process, but requires effective management. 
Visual Analytics provides the necessary tools and paradigms for creating transparent and steerable human-AI collaboration interfaces. 
Finally, Transfer Learning acts as a crucial enabler, making the iterative retraining inherent in these interactive approaches computationally practical. 
This interconnectedness underscores the motivation for developing systems that integrate these concepts to create more effective and efficient workflows for tasks like object detection, annotation, and model training.

\subsection{Problem Formulation}
\label{problemFormulation}

Despite the advancements in object detection models and techniques like Active Learning, the process of acquiring sufficient high-quality labeled data for training robust OD models remains a critical bottleneck. 
This thesis addresses the specific challenges and limitations inherent in current approaches to OD annotation and model training, particularly concerning the integration of human expertise within an active learning framework that also facilitates object detection annotation through an integrated drawing interface. 

The specific issue being investigated is the inefficiency of the standard active learning process when applied to the complex task of object detection annotation, especially when the goal is to leverage the strategic insights of expert users. 
While manual annotation is expensive and unscalable for large datasets, and fully automated methods often lack the necessary accuracy, conventional AL approaches present their own set of limitations:

\begin{enumerate}
    \item \textbf{Lack of Transparency and Control:} Standard AL algorithms typically select samples based on internal metrics (e.g., uncertainty) and present them to the user for labeling without much context or explanation. This "black box" nature prevents users from understanding why certain samples are chosen, hindering their ability to trust the process or strategically intervene based on their own domain knowledge or labeling goals.
    
    \item \textbf{Oversimplification of Informativeness:} Many AL strategies rely on simple heuristics like uncertainty sampling. While useful, these metrics may not capture the full spectrum of what makes a sample informative. An expert might identify valuable samples based on criteria like representativeness of rare classes, visual diversity, or coverage of specific features, which might be overlooked by purely uncertainty-driven selection.
    
    \item \textbf{Focus on Automated Metrics:} The primary goal of traditional AL is often narrowly defined as optimizing a specific model performance metric (e.g., mAP) with the minimum number of labels. This overlooks the importance of the user's experience, their understanding of the model's learning process, and their ability to pursue different strategic objectives during annotation, for example, prioritizing initial rapid improvement vs. focusing on robustness against specific errors.
    
    \item \textbf{Inadequate Support for Expert Strategies in OD:} Existing HITL annotation systems often provide basic interfaces for verification or correction but lack the sophisticated visual guidance needed to support users in implementing diverse and informed sample selection strategies within an AL loop, specifically for object detection. The unique challenges of OD (combining classification and localization, multiple objects per image) further complicate the design of effective guidance mechanisms.
\end{enumerate}

\noindent These limitations point to a gap in existing work. There is a lack of interactive visual systems designed to effectively mediate the HITL-AL process for OD annotation. 
Making it more transparent, interpretable, and strategically manageable for expert users, but also providing an interface for annotation by drawing bounding boxes and assigning labels. 
The interplay between different forms of visual guidance, expert-driven selection strategies, and the resulting OD model performance trajectories within an iterative retraining loop remains underexplored. 

Addressing this problem is significant because overcoming the annotation bottleneck is crucial for advancing OD technology. 
Enabling experts to work more efficiently and strategically with AL systems can accelerate the development of more accurate and reliable models for critical applications. 
Furthermore, understanding how to effectively design human-AI collaborative systems for complex tasks like OD annotation contributes valuable knowledge to the fields of HCI, VA, and ML. 
This thesis aims to fill the identified gap by developing and evaluating a visual interactive tool that encourages users to navigate the AL process for OD annotation more effectively. 
This is formalized into two main research questions:

\begin{description}
    \item[RQ1] How can a Visual Analytics system, incorporating interactive t-SNE projections, uncertainty heatmaps, and model state visualizations, facilitate the implementation of distinct, expert-driven labeling strategies (Exploration-focused, Uncertainty-focused, and Balanced Guidance) within an iterative HITL workflow for Object Detection?
    \item[RQ2] How do the Object Detection performance, resulting from distinct, visually-guided labeling strategies facilitated by VILOD compare against each other and against a conventional automated uncertainty sampling baseline?
\end{description}



\subsection{Motivation}

The investigation into visual-guided active learning for object detection annotation is driven by compelling motivations, such as scientific advancement, industrial application, and societal impact. The core drive lies in transforming the data labeling process from a purely cost-minimization exercise, often driven by automatic algorithms, into a more transparent, strategic, and collaborative effort between human experts and machine learning models. 

From a scientific research perspective, this work combines several fields: Active Learning, Human-in-the-Loop systems, Visual Analytics, and Object Detection. 
It aims to explore how these fields interact. While AL aims for efficiency and HITL incorporates human judgment, the mechanism of their interaction, especially for complex tasks like OD, is often underdeveloped. This thesis investigates how interactive visualizations, grounded in VA principles, can serve as a mediating layer, making the underlying AL suggestions and model state interpretable to the user. 
It moves beyond only applying AL to OD, but also investigates how visual guidance facilitates the process of an expert implementing diverse labeling strategies, such as focusing on uncertain samples, exploring underrepresented data regions, or balancing both. 
This contributes to a deeper understanding of human-AI collaboration in the context of iterative model training and provides empirical insights into the effectiveness of visually-enabled deep active learning workflows. 
Rather than only optimizing model metrics as in typical AL, the motivation here centers on enhancing the collaborative process itself, empowering the human expert through a visual interface that encourages understanding and strategic control. 

For industry experts involved in developing and deploying object detection models, the motivations are mainly practical. The current data annotation bottleneck represents a restriction to project timelines and budgets. Tools and workflows that can demonstrably increase the efficiency and effectiveness of annotation are highly valuable. 
This thesis proposes a system designed to make the annotation process easier for everyone involved, but also facilitate those who have knowledge about the data and the application context to leverage their insights more directly and strategically within the labeling process. 
By providing visual tools to understand model uncertainty, data distribution, and class balance, the system aims to enable experts to make more informed decisions about which data to label next, potentially leading to better model performance with fewer labels or achieving desired performance levels faster. 

On a broader societal level, advancing the efficiency and reliability of object detection development has significant implications. 
OD is a technology used for many applications aimed at improving safety, health, and productivity. Including autonomous driving, medical diagnosis, environmental monitoring, and security systems. Making the development process for these critical AI systems more efficient, transparent, and trustworthy can offer great value. 
By facilitating better human oversight and strategic input during the crucial data labeling stage, the approaches explored in this thesis contribute to the responsible development and deployment of AI technologies that can positively impact society.

\subsection{Objectives}
\label{objectives}
To address the identified problem and contribute to the motivated goals, this thesis outlines the following specific and measurable objectives:

\begin{enumerate}
    \item \textbf{Investigate existing approaches to object detection data annotation, active learning, human-in-the-loop systems, and visual analytics for guiding machine learning processes.} This objective involves a review of the relevant literature to understand the state-of-the-art, identify established methodologies, and pinpoint existing gaps. This objective corresponds primarily to the work presented in Chapter 2.
    
    \item \textbf{Develop a prototype: "Visual Interactive Labeling tool for Object Detection (VILOD)."} This goal focuses on the design and implementation of the technical contribution of the thesis. This objective corresponds primarily to the work presented in Chapter 3. The tool aims to integrate several key components:
    \begin{itemize}
        \item A base object detection model (YOLOv11n) capable of iterative retraining using transfer learning.
        \item An Active Learning component suggesting informative samples based on model uncertainty.
        \item A Data View featuring a t-SNE projection of image features, visually encoding model uncertainty, and highlighting AL-suggested samples.
        \item A Model View displaying relevant model state information, such as prediction confidence distributions and the class balance within the currently labeled dataset.
        \item Interactive capabilities allowing users to select single or multiple images (e.g., via lasso selection in the Data View) and perform bounding box annotations.
    \end{itemize}
        \item \textbf{Evaluate the utility of the developed tool through illustrative use cases and performance comparison.} This objective aims to empirically assess the effectiveness of the tool in supporting different expert-driven annotation strategies. 
        It involves designing and executing the different use cases, where the author utilizes the tool to label data according to predefined strategies (e.g., exploration-focused, uncertainty-focused, balanced approach). The performance trajectories resulting from these strategies will be compared against each other and a baseline AL strategy. This objective corresponds primarily to the work presented in Chapter 4.
\end{enumerate}

\noindent Based on these objectives, the expected outcomes includes a functional prototype of the Visual Interactive Labeling tool for Object Detection, quantitative evidence comparing the performance trajectories achieved by different expert-driven strategies facilitated by the tool versus a pure AL baseline, and finally a demonstration of the potential for such interactive systems to make the HITL-AL workflow more transparent, manageable, and strategically adaptable for OD tasks.

\subsection{Contributions of the Work}

This thesis makes several contributions to the fields of machine learning, computer vision, visual analytics, and human-computer interaction, primarily centered around improving the process of data annotation for object detection through visually guided active learning. The main contributions can be categorized as follows: 

\paragraph{Methodological and Technical Contribution.} The primary contribution is the design, implementation, and investigation of a novel interactive visual analytics system specifically tailored to support and guide the Human-in-the-Loop Active Learning (HITL-AL) process for object detection annotation.

\paragraph{Empirical Contribution.} The thesis provides empirical insights through a comparative case study evaluating the utility of the proposed visual tool. This study analyzes the model performance trajectories achieved by the author, acting as an expert user, employing different AL strategies (exploration-focused, uncertainty-focused, balanced) facilitated by the tool's visualizations and interactions. These strategies are benchmarked against each other and a Pure AL (uncertainty sampling) baseline. This empirical evaluation offers concrete data on the effectiveness of different human-guided approaches within a visually interactive OD annotation context, contributing evidence to the discussion about the relative merits of user-centered versus model-centered labeling strategies. The focus extends beyond prior work by specifically examining how an expert user can leverage the visual interface to enact different strategies, specifically for sampling and annotating in an OD environment. 

\paragraph{Conceptual Contribution.} This work contributes to the conceptual understanding of how integrated visual analytics can enhance the transparency, interpretability, and strategic manageability of the HITL-AL workflow, particularly for complex tasks like object detection involving expert users. It can demonstrate how VA can move beyond model analysis or data exploration to become an active component in guiding the interactive learning process itself. By providing visual representations of data relationships, model uncertainty, and overall model state, the tool aims to make the user more informed and allow for strategic decisions during sample selection, encouraging a more effective human-AI collaboration within the annotation loop. 

\subsection{Scope and Limitations}

To ensure focus and feasibility of this work, the boundaries and scope of this project are defined in this section. 
The primary task focus is on OD using bounding box annotations. Related tasks such as instance segmentation or keypoint detection are excluded, though some key principles might be adaptable. The investigation centers around a specific underlying deep learning model, YOLOv11n \cite{yolo11_ultralytics}, leveraging transfer learning for efficient iterative retraining. 
Consequently, findings related to model performance and user interaction might not directly generalize to different OD architectures. 

The built-in Active Learning component provides suggestions based solely on uncertainty sampling (the lowest average prediction confidence for object detection in an image). 
While users can interactively employ other strategies, the system's inherent suggestion mechanism is limited to this approach, and the interplay with other AL strategies, like diversity-based, is not explicitly explored. 

Regarding visualizations, the study implements and evaluates a specific set of visualizations. A scatterplot of t-SNE projected 2D datapoints of image features, and model state visualizations that utilize a box plot for confidence distributions and a stacked bar chart for class balance. A key limitation is that the t-SNE visualization depends on hyperparameters (e.g., perplexity) and can sometimes introduce distortions or misleading interpretations of high-dimensional structure. This study uses a fixed configuration and does not deeply investigate the impact of t-SNE parameter choices or alternative dimensionality reduction techniques. 

The evaluation method employs a comparative case study with the author acting as an expert user, implementing predefined strategies (exploration, uncertainty, balanced) on a specific dataset. 
While offering qualitative insights, this approach inherently limits the quantitative generalizability due to the single user sample, as well as a single dataset being used. 
The effectiveness of visualizations and AL strategies can be dataset-dependent, and user behavior might vary with different experts or task complexities.

Finally, the research goal emphasis is on investigating how the tool facilitates the AL process and enables the comparison of strategic approaches, rather than solely optimizing for state-of-the-art accuracy.




\subsection{Thesis Structure}

This thesis is organized into six chapters, systematically presenting the research from background and motivation through methodology, results, discussion, and conclusions.

\begin{itemize}
    \item \textbf{Chapter 1 Introduction:} This chapter provides a comprehensive overview of the research topic. It establishes the background by introducing key concepts such as Object Detection, Deep Learning, Active Learning, Human-in-the-Loop, Visual Analytics, and Transfer Learning. It formulates the specific problem addressed, the need for better tools to support expert-driven active learning for OD annotation. The chapter outlines the motivation, specific objectives, main contributions, scope and limitations, and target audience, concluding with this overview of the thesis structure.
    \item \textbf{Chapter 2 Related Work:} This chapter explores the existing literature relevant to the thesis. It reviews prior research in the core areas, including object detection techniques, challenges in data annotation, active learning strategies, Human-in-the-Loop systems for annotation, the use of Visual Analytics and Interactive Machine Learning for guiding ML processes, and features of existing interactive annotation tools. It compares different approaches, identifies key research gaps that this thesis aims to fill, and positions the current work within the broader academic landscape.
    \item \textbf{Chapter 3 Methodology:} This chapter details the methods and techniques employed in this research. It describes the design and implementation of the proposed interactive visual annotation tool, including its architecture, the specific visualization components (Data View with t-SNE and uncertainty heatmap, Model View with confidence/balance plots), and the interactive features (lasso selection, annotation interface). It specifies the underlying OD model (YOLOv11n) and the AL suggestion mechanism. Finally, it outlines the experimental design for the use case study, including the dataset used, the expert-driven strategies evaluated, the baseline condition, and the quantitative and qualitative metrics collected.
    \item \textbf{Chapter 4 Result and Analysis:} This chapter presents the findings obtained from the empirical investigation described in Chapter 3. It begins with detailed narrative walkthroughs for each of the three distinct VILOD guided labeling strategies: \textit{Use Case 1: Exploration \& Structure Focus}, \textit{Use Case 2: Uncertainty-Driven Focus}, and \textit{Use Case 3: Balanced Guidance Integration}. For each use case, qualitative observations on the interaction process and VILOD's utility are presented. Furthermore, a quantitative analysis of model performance trajectories across the three guided strategies and the automated AL baseline is conducted.
    \item \textbf{Chapter 5 Discussion:} This chapter interprets the results presented in Chapter 4. It discusses the significance of the findings concerning the utility of VILOD and the performance of different labeling strategies in relation to the research questions and objectives. The findings are contextualized with the related work reviewed in Chapter 2, and the implications of the study for researchers and practitioners are explored. The chapter also includes a critical assessment of the study's limitations and their potential impact on the generalizability of the results.
    \item \textbf{Chapter 6 Conclusion and Future Works:} This final chapter summarizes the entire thesis. It reiterates the main contributions, specifically how VILOD addresses challenges in OD annotation, and the key findings regarding the effectiveness of visually-guided HITL strategies. Concluding remarks are provided on the value of interactive visual systems for such tasks, and promising directions for future research and development are suggested based on the insights gained and limitations identified.
\end{itemize}

\newpage

\section{Related Work}
\label{relatedWork}

This chapter provides a comprehensive review of the existing research relevant to the thesis topic.
The goal is to situate this work within the context of prior advancements in OD, data annotation methodologies, AL for OD, HITL systems, VA for ML, and interactive annotation tools. 
By summarizing key contributions, comparing different approaches, and identifying existing gaps in the literature, this chapter establishes the foundation and justification for the research presented in subsequent chapters.

\subsection{Review of Existing Research}

Unlike Chapter 1.1, which introduced the core concepts, this chapter examines specific prior works, systems, and methodologies to position the current thesis within the scientific landscape, identify existing gaps, and highlight how this work builds upon or diverges from previous efforts.

\subsubsection{Active Learning and Visual Interactive Labeling Systems}

VA leverages interactive visual interfaces to facilitate analytical reasoning. In the context of machine learning, VA aims to make models more transparent, interpretable, and steerable. IML specifically focuses on enabling users, often non-experts, to guide learning algorithms through intuitive interfaces. 
Visual Interactive Labeling (VIL) is a subset of IML where users employ visualizations to explore data and select instances for labeling. 

Yang et al. \cite{yang2018visually} present a comprehensive review investigating the intersection of AL, VA, and DL, with a particular focus on applications in geo-text and image classification. 
The paper surveys the state-of-the-art in these fields, highlighting the challenge of acquiring large labeled datasets required for DL and positioning AL as a key technique to mitigate this bottleneck. 
Furthermore, the authors strongly advocate for human-in-the-loop approaches, arguing that integrating VA with AL (AL+VA) and Active Deep Learning (ADL) can significantly enhance model performance, reduce labeling effort, and improve the interpretability and trustworthiness of complex models. 
The review synthesizes influential works across these domains, discusses challenges, and identifies research opportunities, particularly within GIScience and Remote Sensing. 
The review provides significant context and motivation for the research undertaken in this thesis. 
It establishes the academic landscape and the recognized potential of combining interactive visualizations with active learning, the core paradigm explored here. 
While \cite{yang2018visually} offers a broad survey, this thesis contributes a specific implementation and in-depth case study evaluation of a novel interactive system designed for object detection, utilizing specific visual mechanisms (t-SNE, uncertainty heatmaps, model feedback charts) to facilitate expert-driven AL strategies. 
Thus, this thesis explores and evaluates a concrete instantiation within the AL+VA domain systematically reviewed by Yang et al. \cite{yang2018visually}. 

A foundational study directly comparing user-centered VIL with model-centered AL was conducted by Bernard et al. \cite{bernard2017comparing}. 
They investigate whether VIL, enhanced with different visualization techniques overlaid on dimensionality-reduced data (primarily t-SNE), could compete with various automated AL strategies in classification tasks. 
Their experimental study involving expert participants demonstrated that VIL strategies, particularly when using visualizations that expose aspects of the model state (like class colors or convex hulls), can match and sometimes outperform standard AL, especially in the early labeling stages, suggesting VIL's potential to mitigate AL's cold-start problem. 
Furthermore, they identified distinct data-centered and model-centered strategies adopted by users during the VIL process. The work by Bernard et al. \cite{bernard2017comparing} significantly informed the motivation and approach of this thesis. 
It provides empirical evidence validating the potential of human-in-the-loop visual approaches as an alternative to purely algorithmic AL. 
While their work focused on comparing different static visual encodings for classification, this thesis builds upon this by developing an interactive tool specifically for object detection that integrates dynamic uncertainty visualization (heatmaps) and explicit AL suggestions within the VIL framework. 
Inspired by their findings, that users employ different strategies, this thesis uses a comparative case study methodology to explore how an expert user leverages the developed tool's specific guidance features when adopting distinct predefined strategies, comparing these processes and outcomes against a Pure AL baseline, further investigating the synergy between human intuition and visual guidance hinted at by Bernard et al. \cite{bernard2017comparing}. 

Another work that builds upon the concepts of VIL is VisGIL, presented by Grimmeisen et al \cite{grimmeisen2023visgil}. 
VisGIL is a visual analytics approach designed to enhance interactive labeling of images by explicitly guiding users toward informative samples. 
Addressing the weaknesses of purely model-driven AL and purely user-driven VIL systems, VisGIL calculates a utility score for unlabeled instances based on both model uncertainty and data representativeness. 
This utility is conveyed through visual cues overlaid on a t-SNE scatter plot. 
Instances deemed more useful by the model are rendered with larger diameters, and diverse, high-utility recommendations are highlighted with distinct star icons. 
While users retain final control over selection, this visual guidance aims to steer them efficiently towards samples beneficial for model training. 
A user study comparing different levels of visual guidance found that more explicit cues (including size and recommendations) led to significantly higher final classifier accuracy and increased user confidence compared to less explicit guidance. 
The VisGIL system shares significant conceptual ground with the work presented in this thesis, as both investigate the synergy between AL, VIL, and visual guidance within an interactive framework. 
Both approaches utilize a dimensionality-reduced scatter plot enhanced with model-derived information to support user decision-making in sample selection. 
However, whereas VisGIL encodes sample utility primarily through icon size and shape, the tool developed in this thesis explores alternative encodings, such as a continuous uncertainty heatmap overlay combined with distinct markers for AL suggestions. The positive results reported for VisGIL regarding the impact of visual guidance provide strong motivation for the current investigation into these different visualization strategies and for comparing the effectiveness of expert-driven, visually-guided interaction against automated baselines, in an OD environment. 

\begin{figure}[ht!]
\begin{center}
\includegraphics*[width=0.8\columnwidth]{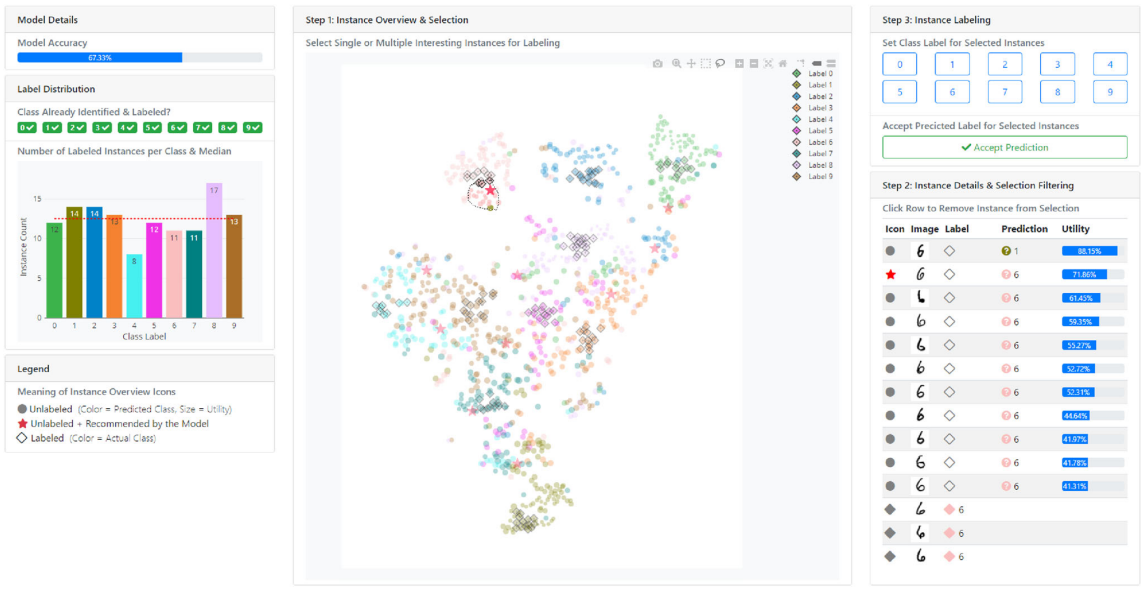}
\end{center}
\caption{VisGIL workspace by Grimmeisen et al. \cite{grimmeisen2023visgil}}
\label{visgilworkspace}
\end{figure}

Klaassen \cite{Klaassen2024Interactive} propose an interactive visualization dashboard combining principles from VIL and Multi-target Active Learning (MTAL). 
The system aimed to reduce labeling effort while aiding user understanding of model predictions, particularly when data features are complex. 
Key features included a dynamic dimensionality reduction visualization (scatterplot) updated based on model predictions and data features, glyph-based AL guidance indicating model uncertainty across multiple targets, an AL-informed lasso tool, linked views for feature attributes and label distributions, and the use of secondary targets (e.g., music danceability, energy) to explain a primary target (e.g., user preference). 
A user study (N=5) comparing dynamic and non-dynamic versions against baselines found the approach outperformed a pure MTAL baseline on average and received positive feedback on usability and the novel interactions.  

\begin{figure}[ht!]
\begin{center}
\includegraphics*[width=0.8\columnwidth]{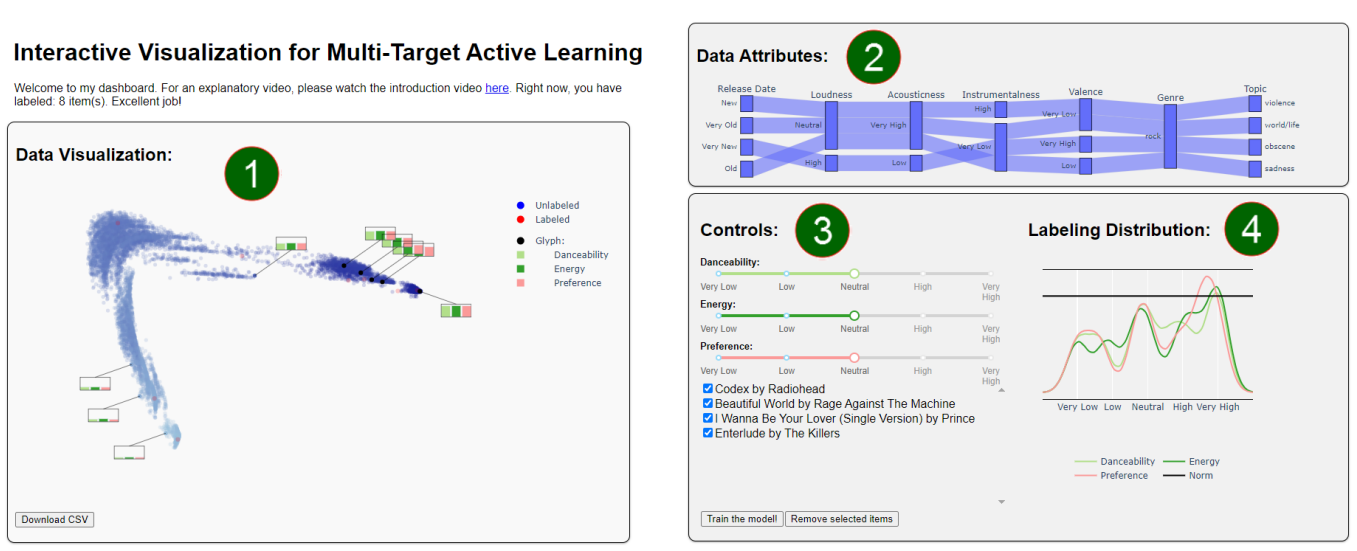}
\end{center}
\caption{Overview of the dashboard by Klaassen \cite{Klaassen2024Interactive}}
\label{klaassenDashboard}
\end{figure}

Qian et al. \cite{qian2021slamvis} introduced SLAMVis, an interactive visual analytics system designed for smart labeling of multidimensional data by combining AL and VIL. 
The system facilitates an iterative labeling process where algorithms are tightly integrated with an interactive visual interface composed of multiple coordinated views, such as a pattern view based on UMAP and glyphs, and detailed instance views. 
A key contribution of SLAMVis is its novel pattern-based query strategy, which employs SOINN combined with K-means to identify data patterns and recommend informative candidate instances for labeling, aiming to improve model effectiveness. 
SLAMVis is relevant to this thesis as it presents another concrete example of a system synergizing AL and VIL principles to support complex data labeling tasks. 
While SLAMVis is designed for general multidimensional data and introduces a unique pattern recognition approach for its query strategy, its focus on interactive guidance, iterative learning, and coordinated visualizations resonates with the objectives of the tool developed in this thesis. 
The approach in SLAMVis provides a valuable comparative point, particularly concerning its methods, for instance, suggestion and data representation in a non-object-detection context, further highlighting the diverse strategies being explored in the broader field of interactive machine learning. 

\begin{figure}[ht!]
\begin{center}
\includegraphics*[width=0.8\columnwidth]{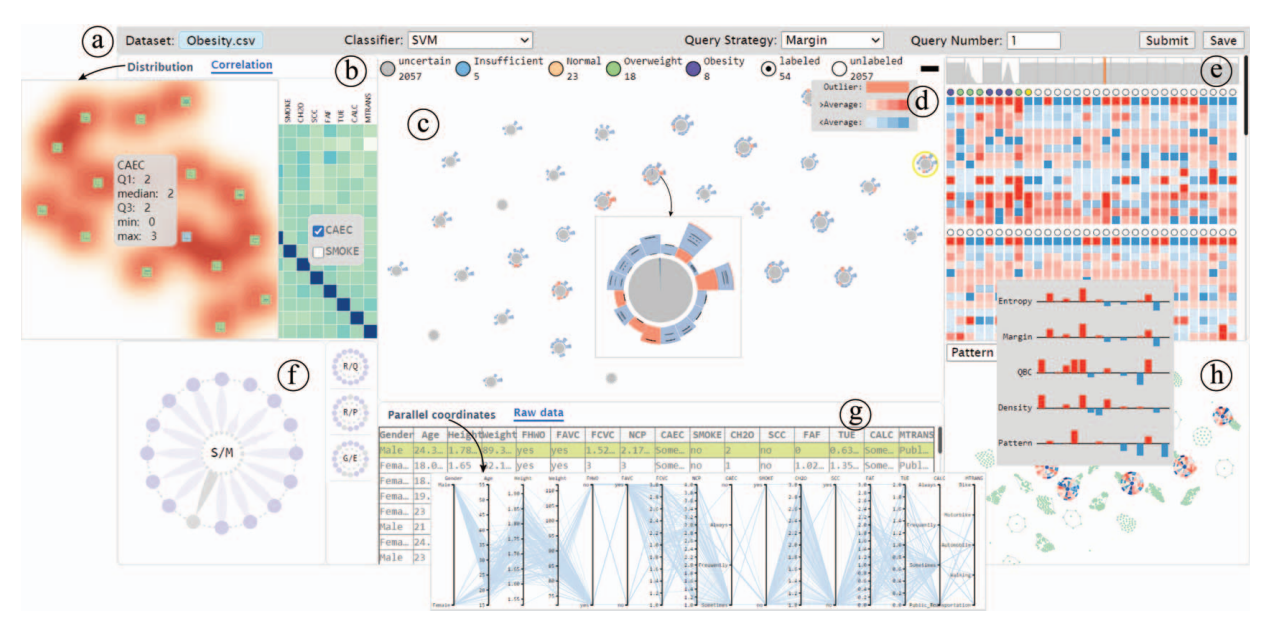}
\end{center}
\caption{The Interactive Visual Interface of SLAMVis by Qian et al. \cite{qian2021slamvis}}
\label{slamVIS}
\end{figure}

Other related works have explored interactive visualization specifically for understanding the AL process itself. 
Iwata et al. \cite{iwata2013active} investigated the use of active learning to improve the process of interactively refining data visualizations. They addressed the problem where automatically generated layouts (using methods like Laplacian Eigenmaps) might not align with a user's preferred spatial arrangement of data points. Their proposed "Active Visualization" framework selects which specific data point the user should manually reposition within the visualization space at each step. The selection criterion was based on information theory, aiming to choose the point whose relocation would reduce the system's uncertainty about the desired locations of all other unlabeled points. Through experiments, they demonstrated that their active selection method allowed the visualization layout to converge towards a target configuration with fewer user interactions compared to random selection or simple uncertainty sampling. While the application goal in \cite{iwata2013active} differs from the goal of this thesis, their work provides valuable context by demonstrating the successful application of active learning principles to guide user interaction directly within a visualization interface to optimize user effort. It shares the core concept of an iterative loop where the system suggests an action to the user based on an algorithmic criterion to more efficiently reach a desired state, highlighting the broader application use for AL to enhance interactive visual systems beyond labeling tasks. 

\subsubsection{Active Learning and Object Detection}

Brust et al. \cite{brust2018activelearningdeepobject} specifically address active learning for deep object detectors like YOLO, proposing uncertainty-based metrics suitable for scenarios where new data is continually explored. 
They focus on adapting classification-based uncertainty measures, such as margin sampling (the difference between the top two class scores), to the object detection task by aggregating the uncertainty scores from individual detections within an image to produce a single image-level score for selection. 
They compare three aggregation strategies: Sum, Average, and Maximum uncertainty across detections. Their experiments on PASCAL VOC 2012 demonstrated that all three aggregation methods outperformed random sampling, with the Sum aggregation generally performing best in their setup. Furthermore, they introduced and showed the benefit of a weighting scheme applied prior to aggregation to counteract class imbalance by prioritizing uncertain detections belonging to underrepresented classes. 
The work done by Brust et al. \cite{brust2018activelearningdeepobject} is relevant to this thesis as it provides foundational methods for applying uncertainty-based AL to deep object detectors. 
While this thesis uses average detection confidence as its base metric and employs average aggregation, the work in \cite{brust2018activelearningdeepobject} validates the concept of aggregating detection scores for image-level AL selection and offers alternative aggregation methods (Sum, Max) for context.

Further advancing AL techniques for object detection, Chen et al. \cite{chen2023semi} propose a framework that combines active learning with semi-supervised learning (SSL). 
Their approach aims to minimize labeling costs by strategically selecting samples for manual annotation while also leveraging the large amounts of unlabeled data through pseudo-labeling. 
A key aspect of their AL strategy is the assessment of image uncertainty from two perspectives: classification stability and localization stability, derived by comparing model outputs on original and augmented image views. 
Images deemed to have low overall stability are prioritized for manual annotation. For the SSL component, images exhibiting high stability are assigned pseudo-labels. 
The authors introduce a novel pseudo-label mining strategy featuring a Stability Aggregation Score (SAS), which considers both classification and localization quality for pseudo-boxes, and a Dynamic Adaptive Threshold (DAT) to adjust filtering thresholds per category, thereby mitigating class imbalance issues. 
This work by Chen et al. \cite{chen2023semi} is significant as it presents a purely algorithmic framework for efficient learning in object detection, explicitly addressing the dual nature (classification and localization) of the task in both its AL sample selection and SSL pseudo-labeling mechanisms. 
It provides a valuable counterpoint to visually-driven labeling systems by showcasing how advanced AL and SSL methodologies can be integrated, offering important insights into state-of-the-art sample selection and pseudo-labeling quality control for object detection that could inform the design of interactive tools. \\

\noindent Therefore, the works by Brust et al. \cite{brust2018activelearningdeepobject} and Chen et al. \cite{chen2023semi} provide important algorithmic groundwork and highlight current trends in AL for object detection. 
Brust et al. \cite{brust2018activelearningdeepobject} validate uncertainty aggregation for image-level selection, while Chen et al. \cite{chen2023semi} introduce a comprehensive AL+SSL framework addressing both classification and localization stability and sophisticated pseudo-labeling. 
Together, these studies demonstrate the algorithmic efforts to define informativeness and manage labeling efficiency in OD. 
While these focus on algorithmic performance and automated selection/pseudo-labeling, this thesis explores how such AL principles can be integrated and complemented within a visual interactive environment to synergize algorithmic suggestions with expert user intuition. 

\subsection{Comparison of Approaches}

This section analyzes AL, VA, and HITL, focusing on their methodologies, strengths, limitations, and applicability to the problem of enhancing expert-driven OD annotation. \\

\noindent \textbf{Methodological Similarities and Differences.} 
A common feature across many Visual Interactive Labeling (VIL) systems, such as those inspired by Bernard et al. \cite{bernard2017comparing} and implemented in \cite{grimmeisen2023visgil,Klaassen2024Interactive,qian2021slamvis}, is the use of dimensionality reduction techniques to provide users with a visual overview of the dataset. This allows users to leverage their pattern recognition abilities to select samples. 
These systems often integrate model-derived information (e.g., uncertainty, class predictions) as visual overlays to guide user attention. However, they differ in the specifics of these visual encodings (e.g., icon size/shape in \cite{grimmeisen2023visgil}, glyphs in \cite{Klaassen2024Interactive}, continuous heatmaps as proposed in this thesis) and the degree of explicit AL integration versus user-driven exploration. 
For instance, Bernard et al. directly compared user-driven VIL against model-driven AL, while systems in \cite{grimmeisen2023visgil,qian2021slamvis} aim for a hybrid approach, providing explicit AL-based recommendations within the visual interface. \cite{iwata2013active} presents a different approach, using AL to refine the visualization itself, highlighting AL's versatility in interactive contexts. 

In contrast, AL methodologies specifically for OD, such as those by Brust et al. \cite{brust2018activelearningdeepobject} and Chen et al. \cite{chen2023semi}, are primarily algorithmic. Brust et al. \cite{brust2018activelearningdeepobject} focus on adapting classification-based uncertainty for OD by aggregating detection-level scores, while Chen et al. \cite{chen2023semi} propose a more comprehensive framework combining AL (based on classification and localization stability) with Semi-Supervised Learning (SSL), including a more sophisticated pseudo-label mining. 
These approaches typically operate without a direct, visual, interactive sample selection interface for the user, focusing instead on optimizing the selection strategy itself. 
The survey by Yang et al. \cite{yang2018visually} covers the broader intersection of AL, VA, and Deep Learning, providing a high-level view that encompasses both visually-driven and algorithmically-driven advancements. \\

\noindent \textbf{Strengths and Limitations.}
VIL systems in \cite{bernard2017comparing,grimmeisen2023visgil,Klaassen2024Interactive,qian2021slamvis} excel in transparency and leveraging human intuition. 
They allow experts to apply domain knowledge and identify strategically important samples that purely algorithmic methods might miss. Bernard et al. \cite{bernard2017comparing} demonstrated that VIL can even outperform standard AL in early labeling stages. 
The primary limitation often lies in scalability if purely reliant on user exploration, and the potential for user-introduced bias if not carefully guided. 
Furthermore, many existing VIL tools are primarily evaluated on classification tasks, and their direct applicability or optimal design for the complexities of OD (multiple objects, localization precision) is less explored. 

Algorithmic AL approaches for OD \cite{brust2018activelearningdeepobject,chen2023semi} are strong in their potential for automation and more complex definitions of informativeness (e.g., uncertainty, stability). 
Chen et al. \cite{chen2023semi} advances this by explicitly considering both classification and localization aspects. However, as noted in Chapter 1.2, they often lack transparency (the "black box" issue) and direct user control over the selection strategy beyond initial parameter settings. This can limit the engagement of expert users who may have insights not captured by the specific heuristics employed. \\

\noindent \textbf{Applicability to the investigated problem.}
The core problem this thesis addresses is the need for an interactive visual system that makes the HITL-AL process for OD annotation more transparent and manageable for expert users, including an integrated annotation interface. 
VIL systems for classification  \cite{bernard2017comparing,grimmeisen2023visgil} offer foundational interaction and visualization principles that are highly applicable but require adaptation for OD specifics (e.g., visualizing bounding boxes,  uncertainty, handling multiple instances per image effectively in the visual summary). 
More general VA-driven labeling systems like \cite{Klaassen2024Interactive,qian2021slamvis} demonstrate the power of coordinated views and integrated AL suggestions, which is highly relevant. However, SLAMVis's pattern-based query strategy and Klaassen's multi-target focus, while informative, differ from the uncertainty-visualization focus for OD in this thesis. 

The OD related approaches in \cite{brust2018activelearningdeepobject,chen2023semi} provide insights into what constitutes informativeness in OD data (e.g., aggregated uncertainty, classification/localization stability). 
While not interactive themselves, their metrics can inform the type of information that should be visualized in an interactive system like VILOD to guide the user. 
For instance, the uncertainty heatmaps in VILOD are inspired by the need to convey model uncertainty, a core theme in these AL papers. 
None of the existing approaches, when considered individually, fully address the specific combination of transparent, expert-steerable AL for OD within a visually rich, integrated annotation environment that supports diverse selection strategies. 

The strategies, limitations and key aspects of the different approaches in previous work are summarized in Table \ref{tab:comparison_approaches_revised}

\begin{table}[htbp] 
  \centering 
  \caption{Comparison of Selected Existing Approaches}
  \label{tab:comparison_approaches_revised}
  \scriptsize 
  \begin{tabularx}{\textwidth}{ 
    >{\RaggedRight}p{2.2cm} 
    >{\RaggedRight\arraybackslash}X 
    >{\RaggedRight}p{1.8cm} 
    >{\RaggedRight\arraybackslash}X 
    >{\RaggedRight\arraybackslash}X 
    >{\RaggedRight\arraybackslash}X 
    >{\RaggedRight\arraybackslash}X 
  }
    \toprule
    Author(s) (Year) & Primary Task / Focus & Interaction Style & Key Visualizations / Guidance & AL Strategy Used / Type & HITL Aspect & Limitations for Current Thesis's Problem \\
    \midrule

    Bernard et al. \cite{bernard2017comparing} (2017)& 
    Image Classification & 
    User-driven VIL vs. AL & 
    t-SNE, model state overlays (class colors, convex hulls) & 
    Compared VIL with standard AL (e.g., uncertainty) & 
    Expert labeling, strategy analysis & 
    Classification-focused; static encodings; does not explore dynamic uncertainty viz for OD. \\
    \midrule

    Grimmeisen et al. \cite{grimmeisen2023visgil} (2023) & 
    Image Classification & 
    AL-guided VIL & 
    t-SNE with utility scores (size, star icons for recommendations) & 
    Utility score (uncertainty + representativeness) & 
    User selection guided by model suggestions & 
    Classification-focused; specific icon-based encoding may not suit OD uncertainty representation. \\
    \midrule

    Klaassen \cite{Klaassen2024Interactive} (2024) & 
    Multi-target Classification & 
    AL-guided VIL & 
    Dynamic scatterplot, glyphs, linked views & 
    MTAL (Multi-target AL) & 
    User understanding \& labeling with AL aid & 
    Multi-target, not OD specific; glyphs differ from proposed heatmap. \\
    \midrule

    Qian et al. \cite{qian2021slamvis} (2021) & 
    Multidimensional Data Labeling (General) & 
    AL-guided VIL & 
    UMAP with glyphs, coordinated views & 
    Pattern-based (SOINN+K-means), also supports standard AL & 
    User interactive labeling with AL suggestions, customizable strategies & 
    Pattern strategy different from thesis focus on uncertainty visualization for OD. \\
    \midrule

    Brust et al. \cite{brust2018activelearningdeepobject} (2018) & 
    Object Detection & 
    Algorithmic AL & 
    N/A (Algorithmic) & 
    Uncertainty aggregation (Sum, Avg, Max) for OD & 
    Human annotates selected samples (standard AL loop) & 
    Non-interactive selection \\
    \midrule

    Chen et al. \cite{chen2023semi} (2023) & 
    Object Detection & 
    Algorithmic AL + SSL & 
    N/A (Algorithmic) & 
    Classification \& Localization Stability; SAS \& DAT for SSL & 
    Human annotates AL-selected samples (standard AL loop) & 
    Non-interactive AL selection; highly algorithmic, lacks visual user interface for strategic exploration. \\
    \bottomrule
  \end{tabularx}
\end{table}

\newpage
\subsection{Identified Gaps and Positioning of this Work}

The review and comparison of existing research reveal several gaps that this thesis, through the development and investigation of VILOD, aims to address:

\begin{itemize}
    \item \textbf{Lack of OD-Specific Interactive AL Tools Supporting Expert Strategies.} While VIL systems like \cite{grimmeisen2023visgil} and SLAMVis \cite{qian2021slamvis} demonstrate the benefits of VA in AL for classification or general multidimensional data, there is a less explored niche for tools specifically designed for the nuances of object detection that also explicitly aim to support an expert user in implementing diverse, strategic sample selection approaches. Many AL for OD methods \cite{brust2018activelearningdeepobject,chen2023semi} are purely algorithmic, lacking the interactive visual interface that can guide users.
    \item \textbf{Limited Support for Integrated Annotation and Iterative Exploration in OD-VIL.} While some interactive tools support labeling, the seamless integration of (a) visual exploration of the dataset informed by model state, (b) flexible selection of samples (individually or in groups based on visual patterns), (c) an embedded OD annotation interface (bounding box drawing and labeling), and (d) immediate iterative model retraining and feedback within a single environment specifically for OD is not widely addressed. This thesis proposes VILOD to create such an integrated workflow.
    \item \textbf{Underexplored Role of Visual Guidance in Modulating Expert Labeling Strategies for OD:} While Bernard et al. \cite{bernard2017comparing} showed users adopt different strategies in VIL for classification, there is a need for more focused investigation into how different visual guidance mechanisms within an interactive tool can help an expert specifically formulate and execute distinct labeling strategies (e.g., exploration-focused, uncertainty-focused, balanced approach) for OD tasks, and how these visually-guided strategies compare in terms of process and outcomes.
\end{itemize}

\noindent This thesis positions VILOD to address these gaps between purely algorithmic AL approaches and user-centric visual analytics, offering a novel solution tailored to the specific challenges and expert needs in object detection.

\subsection{Summary}

This chapter has reviewed existing research pertinent to visually guided active learning for object detection annotation. Section 2.1 surveyed key works in Active Learning and Visual Interactive Labeling systems \cite{yang2018visually,bernard2017comparing,grimmeisen2023visgil,Klaassen2024Interactive,qian2021slamvis,iwata2013active}, highlighting foundational concepts and representative tools, and then focused on specific Active Learning advancements for Object Detection \cite{brust2018activelearningdeepobject,chen2023semi}. These studies underscore a trend towards integrating machine intelligence with human expertise to combat the data annotation bottleneck. 

Despite these advancements, the comparison in Section 2.2 and the gap analysis in Section 2.3 revealed that work remains for developing interactive visual systems specifically tailored to the complexities of object detection annotation. 
Key limitations in prior work include a focus on classification tasks in many VIL systems, a lack of transparency and control in purely algorithmic AL approaches for OD, and insufficient support for experts wishing to implement diverse, strategically informed sample selection strategies within an integrated annotation and retraining loop for OD. 

By addressing the identified gaps, this research aims to contribute a novel tool (VILOD) and empirical insights into how human-AI collaboration can be effectively mediated by visual analytics to make the object detection annotation process more efficient, strategic, and comprehensible. This review sets the stage for Chapter 3, which will detail the design and implementation of the VILOD system, and lay the groundwork of the methodology used to evaluate it.

\newpage 
\section{Methodology}
\label{Method}

This chapter details the methodology employed to address the research objectives outlined in Chapter 1. 
It describes the overall research approach, the development and key features of the VILOD tool prototype (addressing Objective 2), the procedures for data collection including the baseline simulation and the execution of illustrative use cases utilizing the tool (addressing Objective 3), the subsequent data analysis techniques, and considerations regarding the reliability, validity, and ethical conduct of the research. 
The method provides a structured approach to investigate how a visual interactive tool can support different informed strategies within a human-in-the-loop active learning workflow for object detection.

\subsection{Research Approach}
The research approach adopted for this thesis is a comparative analysis of use cases, executed by the author acting as an informed user simulating expert interaction. The use cases are designed to answer RQ1 and RQ2. 
The methodology combines quantitative performance evaluation, comparing model learning curves under different conditions, addressing RQ2, with qualitative process analysis examining how the tool was used and how strategies were implemented, addressing RQ1. 
While only the author interacted with the tool may limit generalizability, it allows for consistent application of predefined strategies and detailed insight into the tool's potential functionality and workflow implications.

\subsection{The suggested VILOD Pipeline}
\label{activeLearningPipelineInVILODSection}
\begin{figure}[ht!]
\begin{center}
\includegraphics*[width=0.6\columnwidth]{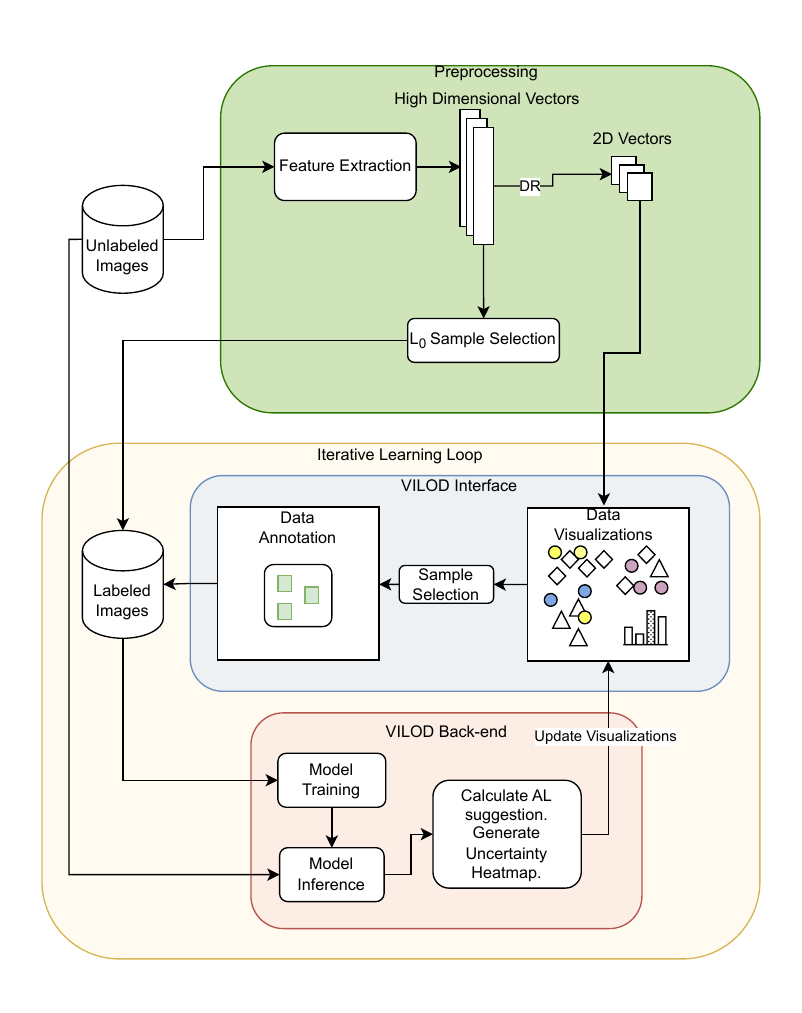}
\end{center}
\caption{Overview of the suggested approach in the VILOD system.}
\label{activeLearningPipelineInVILOD}
\end{figure}

To implement the research approach described above, the VILOD system proposes a design built around an iterative, human-in-the-loop (HITL) active learning pipeline. This process, illustrated in Figure \ref{activeLearningPipelineInVILOD}, integrates human expertise directly into the model training cycle, transforming the annotation process into a strategic, collaborative effort.

The pipeline is initiated with an initial model ($M_0$ ), which is trained on a small, pre-selected data subset ($L_0$). This model's predictions are then used to populate the VILOD user interface, rendering the key visualizations like the uncertainty heatmap and AL suggestions for the entire pool of unlabeled images. From there, the expert user interacts with these visualizations to explore the dataset and strategically perform sample selection. Following this selection, the user proceeds with data annotation, labeling a new batch of 30 images within the VILOD interface. A labeling budget of 30 was chosen as it represents a feasible annotation effort for each iteration while still being substantial enough to impact model performance. Once the user confirms the batch of newly annotated images, they are added to the growing labeled dataset, and a retraining process is triggered. A new model version is trained by fine-tuning the weights from the previous iteration. After training is complete, the back-end uses the updated model for inference on the entire unlabeled pool, generating new predictions and recalculating the AL suggestions and uncertainty heatmap. The system then updates all visualizations in the front-end, presenting the new model state to the user. This entire HITL cycle, from user selection and annotation to model retraining and visual feedback, is repeated, allowing the model to improve iteratively with expert guidance until a stopping criterion is met.

The following sections detail the specific components and procedures used to implement this pipeline, from dataset preparation to the technical architecture of the VILOD tool, together with detailed descriptions of the UI components.

\newpage
\subsection{Data Collection}
Data collection involved preparing the necessary dataset and machine learning models, developing the VILOD tool itself as the primary instrument, simulating an automated baseline, and executing the descriptive use case scenarios using the tool.

\subsubsection{Dataset Preparation}

This study utilized the "African Wildlife" dataset, publicly available and originally contributed to Ultralytics documentation\footnote{https://docs.ultralytics.com/datasets/detect/african-wildlife/\#dataset-yaml}. 
This dataset was mainly chosen for its recognizable class definitions that do not require extensive prior domain knowledge for annotation. The four classes represented in the dataset are Rhino, Elephant, Zebra, and Buffalo. The dataset also includes predefined data splits. The integrity of these splits was maintained throughout the research to ensure methodological soundness. 
Specifically, the original $test$ split of 227 images with corresponding ground truth labels in YOLO format was reserved for the evaluation of all trained models in our pipeline and was entirely excluded from any training, validation, or active learning selection processes. 
The original $val$ split, containing 225 images and labels, was designated as the fixed validation set. 
This set was consistently used during the training phase of all models (including the initial model and subsequent iterations in both the baseline simulation and human-led studies) solely for monitoring performance and selecting the best-performing weights $(best.pt)$ for each training iteration. 
The largest partition, the original train split, containing 1052 images and labels, served as the comprehensive pool for active learning. 
This pool, hereafter referred to as $U_{pool}$, was the exclusive source for selecting the initial training set ($L_0$) and all subsequent training sampling and annotation by the user. 

\subsubsection{Initial Model Preparation}

To establish a consistent starting point for both the automated baseline and the use case scenarios, an initial model ($M_0$) was prepared. 
The objective was to create a model, based on the standard pre-trained YOLOv11n architecture (yolov11n.pt), that possessed a foundational understanding of the target dataset's classes beyond the generic COCO pre-training, using a small but representative initial training set $L_0$. 
This $L_0$ set, containing 40 images, was selected exclusively from the $U_0$ using a diversity sampling strategy designed to maximize coverage of the feature space. 
Specifically, high-dimensional feature vectors were extracted from the YOLOv11n backbone for all images in $U_0$. To implement the diversity sampling, K-Means clustering was employed on these extracted features. The number of clusters, K, was set to 20. This value was chosen to strike a balance between capturing sufficient diversity of the groupings present in the feature space. Subsequently, two image samples closest to each of these cluster centroids were chosen. This was determined to be an adequate starting point for the initial model, but no extensive search was made beyond making sure every class was represented. Subsequently, the $M_0$ model was generated by fine-tuning the base YOLOv11n model. 
This training utilized the 40 images and their true labels $L_0$ as the training dataset and the full, fixed 225-image validation set for performance monitoring. 

Based on preliminary testing and observations regarding model stabilization, training was conducted for a fixed 50 epochs, using an image size of $imgsz = 640$ and a fixed random seed ($seed=42$) to ensure reproducibility. 
The model weights yielding the best performance on the validation set during this process were saved as $M_0$. 
Initial predictions from $M_0$ were then generated for initializing the interactive tool's state.

\subsubsection{Feature Extraction and Dimensionality Reduction}

To provide a visual overview of the entire training dataset within the VILOD tool's Data View, a two-dimensional representation of the images was generated. 
This process involved two main stages: feature extraction followed by dimensionality reduction.

Initially, high-dimensional feature vectors were extracted for every image in the training dataset. 
This was accomplished using the base pre-trained YOLOv11n model. 
The model's \textit{embedded} method was employed to process each image and generate a corresponding high-dimensional embedding. 
By default, the method returns the embeddings of the second-to-last layer of the model, and the output is a vector of size (1, 256) for each image. 
These embeddings capture the rich visual characteristics of the images as learned by the model on the COCO dataset. 

Once the high-dimensional embeddings were obtained for all training images, dimensionality reduction was used to transform these vectors into a two-dimensional representation suitable for interactive visualization. 
For this task, t-Distributed Stochastic Neighbor Embedding (t-SNE) was utilized due to its common application and effectiveness in revealing underlying structures in high-dimensional datasets and preserving local similarities.
The t-SNE algorithm was configured with parameters for basic interpretability and reproducibility. The number of components was set to 2, as the goal was to generate a 2D visual representation. The perplexity parameter was set to 12 after visualizing the reduced data points for a sample of different perplexities. However, no exhaustive hyperparameter search was conducted. The visual assessments of the scatter plot with perplexity of 12 were deemed good enough for this work. The transformation mapped each high-dimensional image embedding to a pair of (x, y) coordinates. These resulting 2D points, representing the entire training dataset, were saved and subsequently used to render the interactive scatterplot in the Data View of the VILOD system, forming the basis for visual exploration and sample selection.

\subsubsection{Visual Interactive Labeling tool for Object Detection (VILOD)}

This section presents and details the user interface of the Interactive Visual Labeling tool for Object Detection
developed for in this work, designed to address Objective 2 (Section \ref{objectives}). 
The tool serves as the platform for the expert-led Use Cases, comparing different human-in-the-loop strategies against an automated baseline. 
The design and implementation choices were guided by the goal of creating a system that not only enables active learning for object detection but also enhances the user’s ability to understand model uncertainty, interpret visual guidance, and confidently execute selection strategies. Subsequent Sections \ref{SystemArchitecture} and \ref{activeLearningPipelineInVILODSection} will cover the system architecture, including important algorithmic procedures implemented and utilized in VILOD.

\begin{figure}[ht!]
\begin{center}
\includegraphics*[width=1\columnwidth]{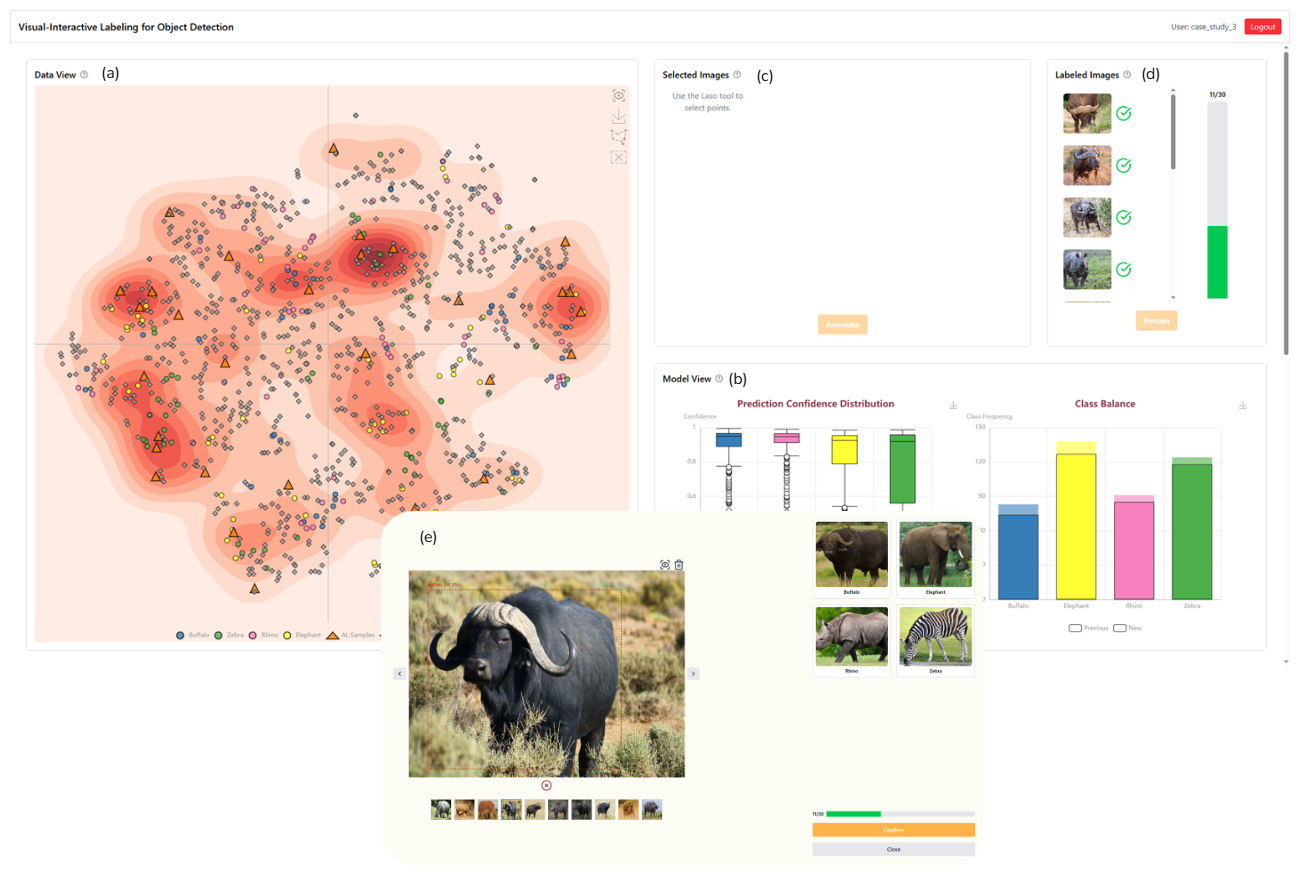}
\end{center}
\caption{Dashboard of Visual Interactive Labeling tool for Object Detection. (a) Displays the main Data View, visualizing the dataset. (b) Shows the Model View, giving insight into the trained model. (c) Shows the Selected Images View, illustrating the selected images through the use of the lasso tool in the Data View. (c) Labeled View, which keeps track of the annotated images in this iteration and labeling budget progress. (e) The annotation modal.}
\label{maindashboard}
\end{figure}

\begin{figure}[ht!]
\begin{center}
\includegraphics*[width=0.8\columnwidth]{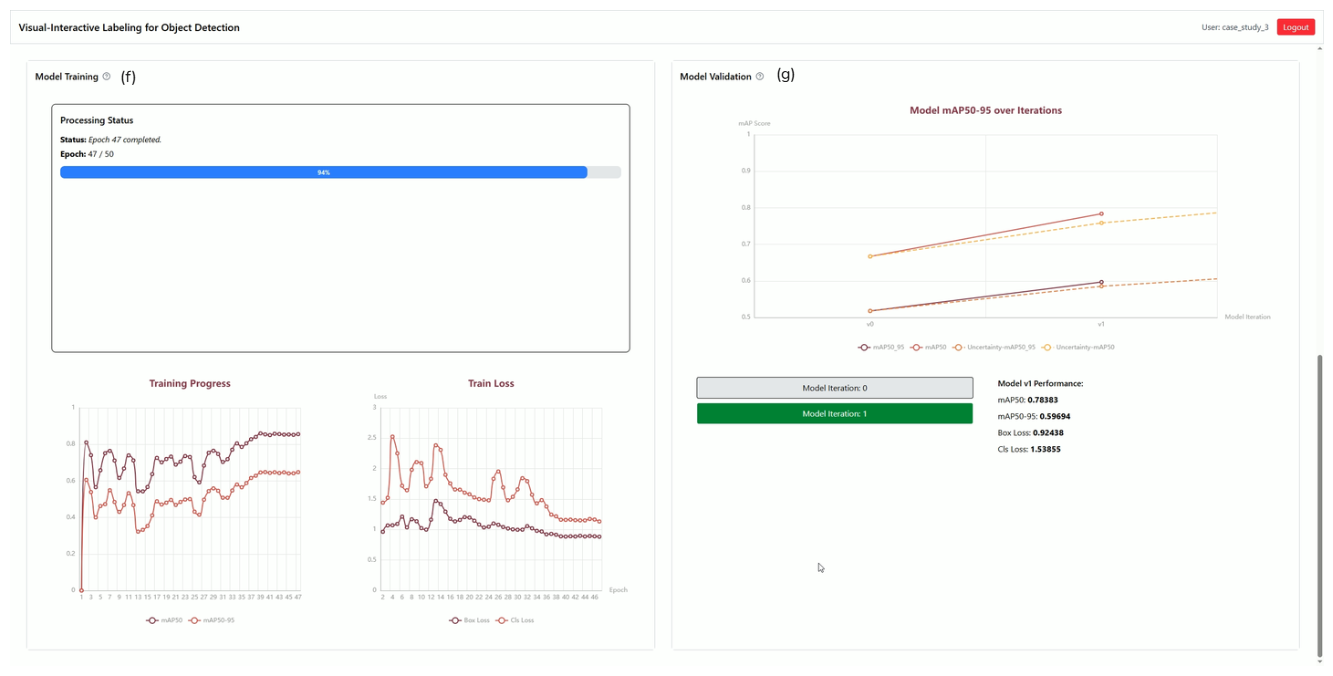}
\end{center}
\caption{Training page of VILOD. (f) Displays the Model Training View, enabling real-time training monitoring. (g) Shows the Model Validation view, displaying performance metrics of trained models.}
\label{trainingpage}
\end{figure}

The user interface is developed as a Next.js application leveraging TypeScript and React. This architecture choice enables a responsive single-page application experience with client-side state management and efficient component rendering. The frontend communicates with the backend through API calls and maintains real-time updates via WebSocket connections, particularly for monitoring training progress. Renderings and visualizations are created with the help of the Apache ECharts library. 
The different components of the front-end dashboard, as depicted in \ref{maindashboard} and \ref{trainingpage}, include:
\begin{enumerate}[label=(\alph*)]
    \item A Data View component featuring an interactive 2D scatter plot visualization of the African Wildlife image dataset using t-SNE dimensionality reduction. 
    \item A Model View visualization component that gives insight into the latest model iteration's prediction state and the state of the labeled data pool. 
    \item The Selected Images View that interacts with the lasso tool in the Data View, showing thumbnails of the selected images and their status.
    \item A Labeled Images view that keeps track of the annotated images and the progress toward the labeling budget.
    \item An interactive image labeling interface with bounding box drawing capabilities.
    \item A Model Training component allowing the user to monitor model training metrics in real-time.
    \item A Model Validation component featuring the performance trajectories of trained models, as evaluated by the validation set during training. 

\end{enumerate} 

The \textit{Data View} component is the main source of information in the system. 
It consists of a scatterplot visualizing the 2D datapoints of the images from the feature extraction and dimensionality reduction process described in the previous section. 
It provides a compact visualization of the entire training dataset. It includes several visual cues intended to help the user explore and understand both the original data and the currently trained model's state. 
One such visual cue is the \textit{AL Samples}, which are images that are selected by the employed active learning uncertainty algorithm. The AL samples are extra emphasized in the scatterplot by coloring them orange, using a triangle as a symbol, and rendering them larger than other points. Another feature of the Data View is the \textit{Uncertainty Heatmap}. 

\begin{figure}[ht!]
\begin{center}
\includegraphics*[width=0.6\columnwidth]{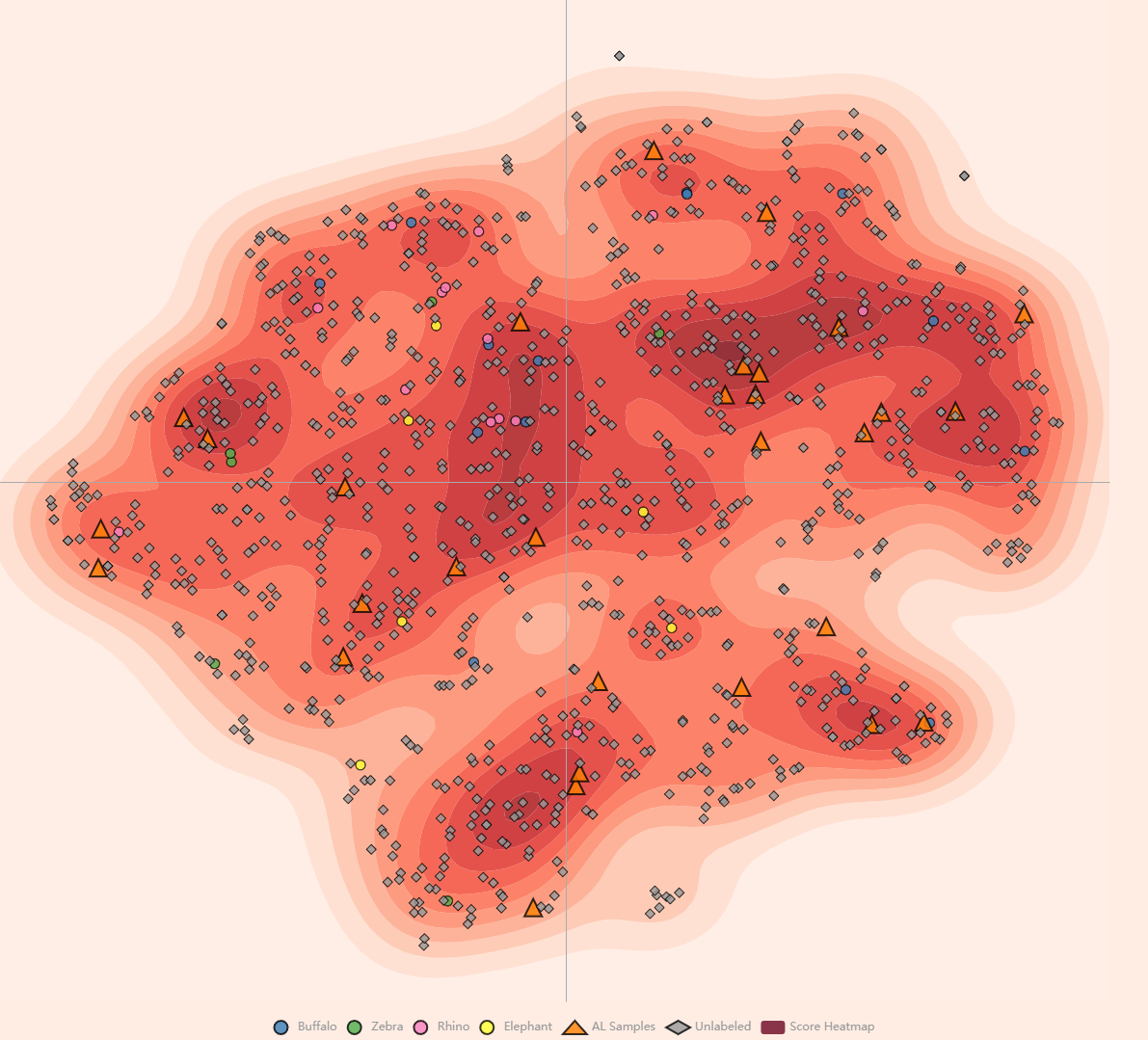}
\end{center}
\caption{The Scatterplot visualization in the Data View, initial state.}
\label{dataview}
\end{figure}

The uncertainty heatmap provides a visual overlay on the t-SNE scatterplot, designed to guide the user toward regions of high model uncertainty. The uncertainty heatmap is a Kernel Density Estimation (KDE) visualized as a filled contour plot.
The heatmap is overlaid transparently onto the scatterplot, allowing users to simultaneously perceive the data distribution and the model's uncertainty landscape, thereby facilitating the identification of potentially informative regions for labeling or to simply enable the user to make more informed decisions. 
Additionally, already labeled samples in the data view are colored by the corresponding class color. 
Since the detection task is multiclass by nature, class color is assigned by the majority class in each picture. 
Unlabeled points are distinctly visualized by using a diamond symbol and a gray color. 
The scatterplot, including these features, as seen from the initial state of the VILOD tool, is displayed in Figure \ref{dataview}. 

\begin{figure}[ht!]
\begin{center}
\includegraphics*[width=0.8\columnwidth]{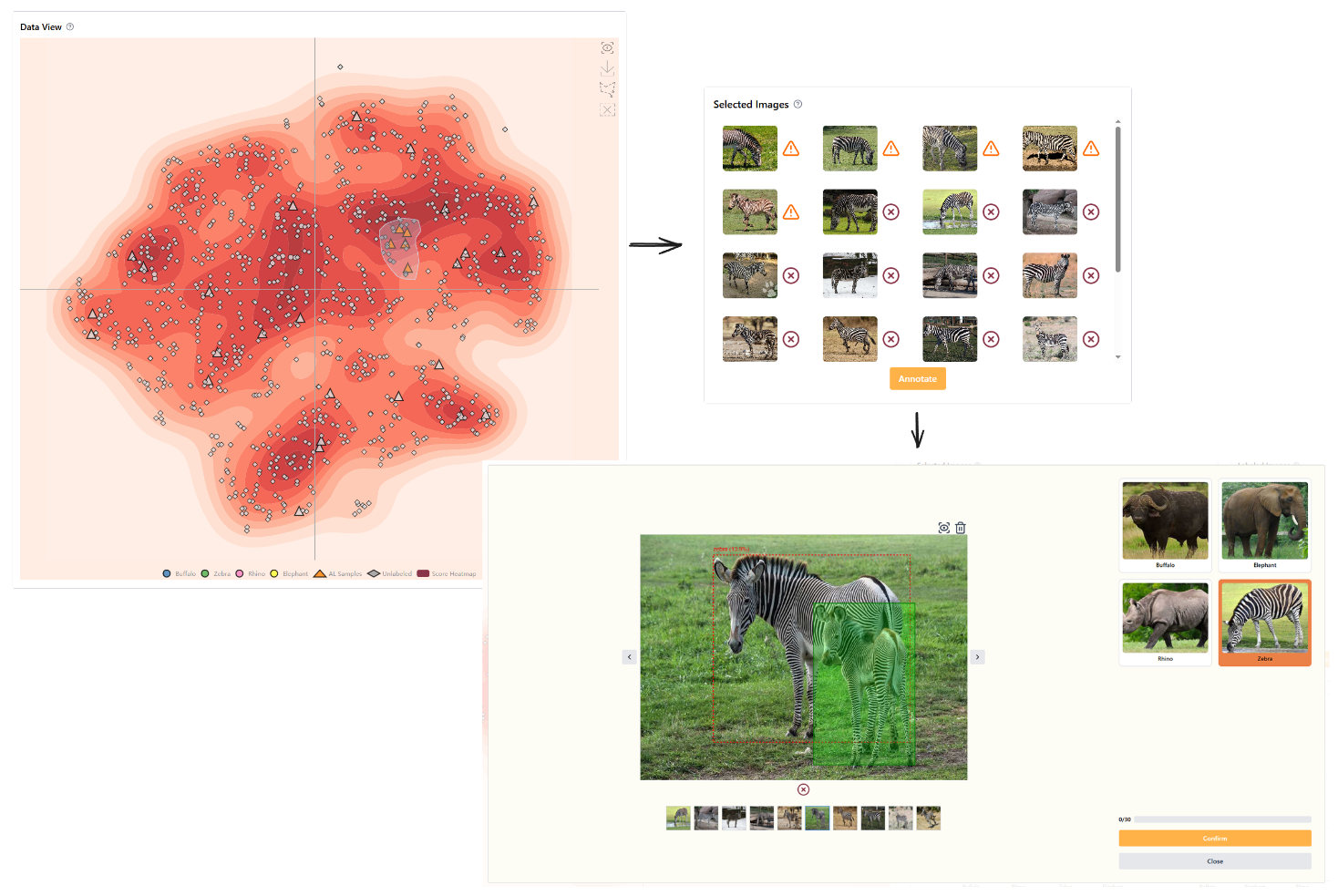}
\end{center}
\caption{Workflow showing how the linkage between different views is utilized in the tool.}
\label{linkage}
\end{figure}

Additionally, the scatter plot features a lasso selection tool, allowing the user to select a group of images for further exploration. 
The selection made in the Data View is linked to the \textit{Selected Images View}.
The selection view previews thumbnails of the selected images, together with a symbol indicating the status of the image. 
An orange triangle if the image is an AL suggested sample, a red cross if the image has yet to be labeled, or a green check mark if the image is already labeled. 
From the Selected Image view, the user can annotate the images by opening the \textit{Annotation Modal} through the \textit{Annotate} button. The workflow of selecting data points with the lasso tool, inspecting them in the Selected Images View, and finally annotating the images in the Annotation Modal is depicted in \ref{linkage}.

The \textit{Annotation Modal} is the interface for users to annotate the unlabeled images. 
The user can browse and annotate the selected images by selecting a pre-defined class label and drawing bounding boxes on the displayed image, encapsulating the objects. 
Additionally, the interface allows the user to see the predicted bounding boxes made by the latest model iteration, drawn as red bounding boxes with dotted lines. 
The user-made bounding boxes are drawn as green boxes. Some extra functionality includes the ability to toggle on-off model-predicted bounding boxes and to clear any existing user-drawn bounding boxes on the currently displayed image. 
It also includes a progress bar, helping the user keep track of the labeling progress for the current training iteration. 
Upon pressing the \textit{Confirm} button, the annotation modal will close, and all annotated images will move from the Selected Images view to the Labeled Images view.

\begin{figure}[ht!]
\begin{center}
\includegraphics*[width=0.8\columnwidth]{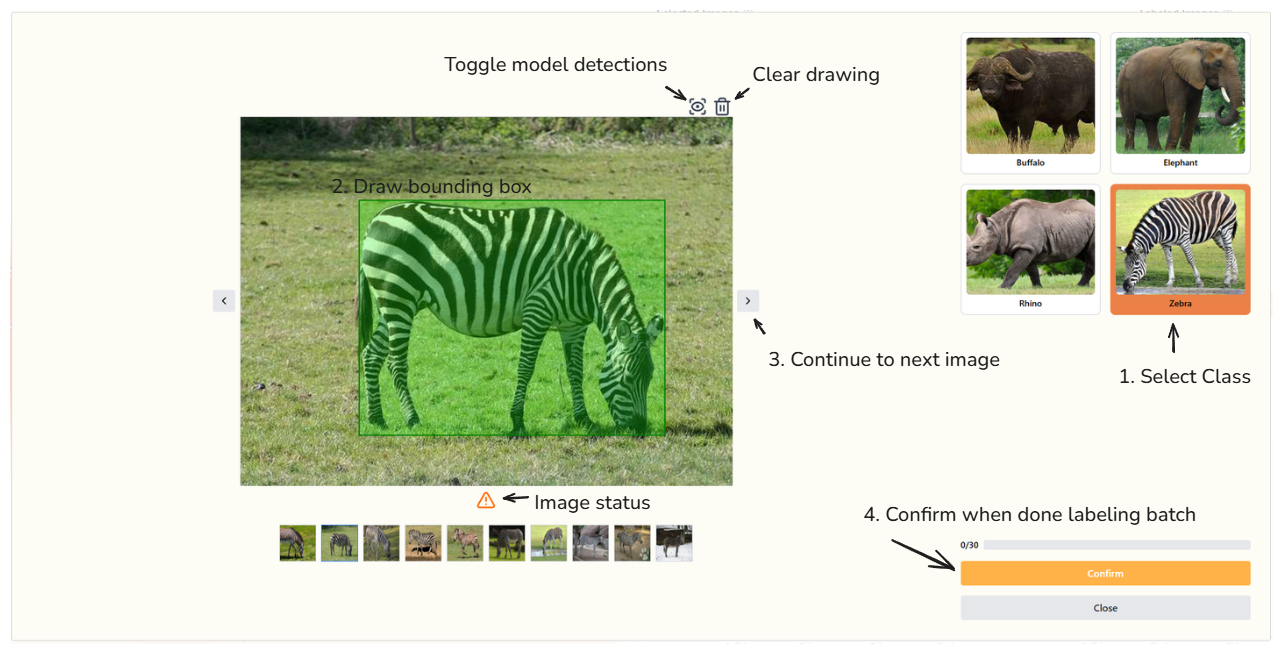}
\end{center}
\caption{The annotation interface of the VILOD system}.
\label{annotationworkflow}
\end{figure}

\begin{figure}[ht!]
\begin{center}
\includegraphics*[width=0.8\columnwidth]{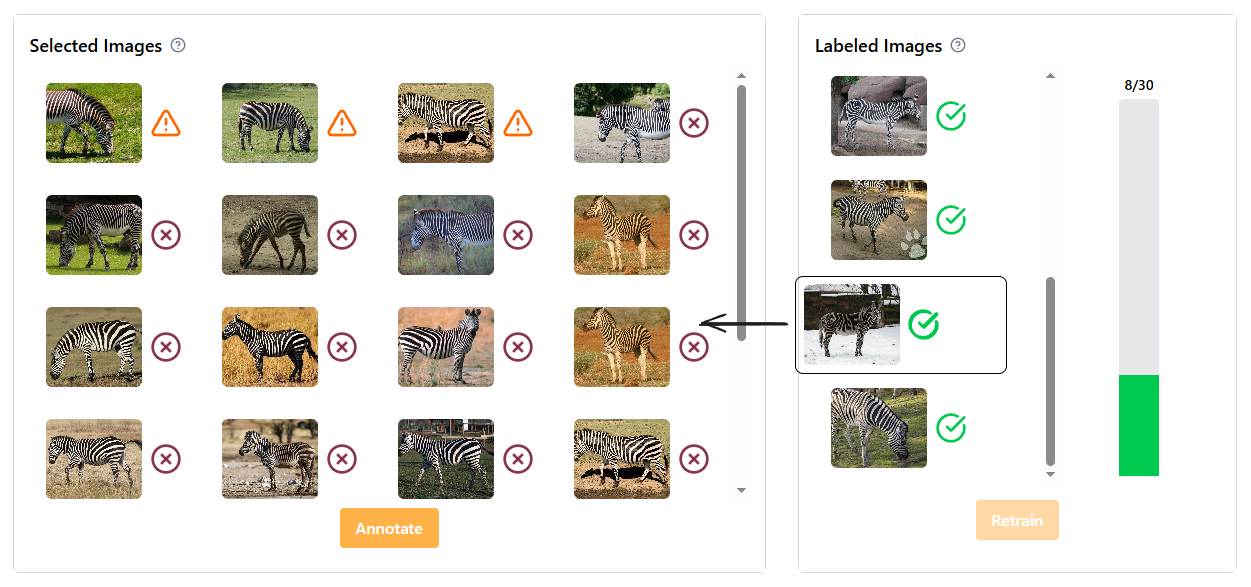}
\end{center}
\caption{The \textit{Labeled Images} view and the removal of an image from the labeled pool.}
\label{dragdrop}
\end{figure}

The \textit{Labeled Images} view helps the user keep track of which images have been labeled, as well as showing labeling budget progression for the training iteration. 
The user can undo an annotation by selecting the thumbnail of an image in the Labeled Image View and simply dragging and dropping it back to the Selected Image area, as seen in Figure \ref{dragdrop}. 
This component also features a progress bar, which, when filled, will unlock the \textit{Retrain} button, enabling the user to trigger a new round of training with these newly annotated images.

Furthermore, the \textit{Model View} component in Figure \ref{maindashboard} consists of two charts. One is a box plot visualizing the distribution of predictions of each class in the dataset. 
The purpose of the box plot is to give the user more granular insight into the model's performance on a class-to-class basis. 
For example, it allows a user to examine which class is performing better or worse, including the spread of the distribution and outliers. 
The other chart is a stacked bar chart that visualizes the class balance in the currently labeled train set. It will show the number of class instances that exist in the training set at the start of the training iteration in a darker class color and will stack newly labeled instances of a class in a lighter class color, as they are being labeled in the current iteration. This chart intends to inform the user of any class imbalance, hence allowing for more informed decisions on which samples to label next. 

Finally, VILOD's training page (Figure \ref{trainingpage}) features a \textit{Model Training} and \textit{Model Validation} component. 
The Model Training component displays real-time training progress. 
Showing the progression through the training epochs, both in terms of the model's mAP metrics and the training loss in terms of class and box loss. The Model Validation component allows the user to view and compare performance metrics of the model through training iterations, as evaluated on the validation set during training.

\subsubsection{System Architecture}
\label{SystemArchitecture}
\begin{figure}[ht!]
\begin{center}
\includegraphics*[width=0.8\columnwidth]{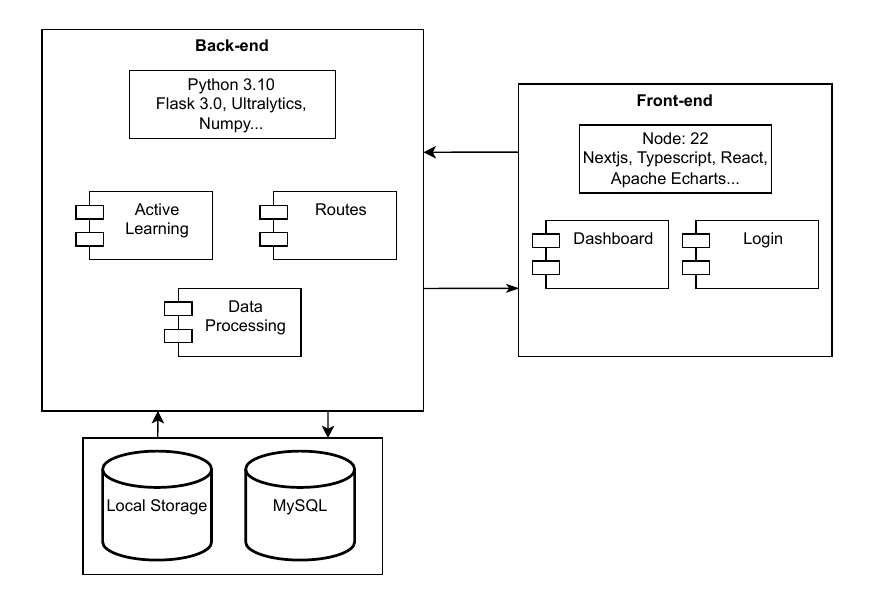}
\end{center}
\caption{An overview of the System Architecture of VILOD}
\label{sysarc-vilod}
\end{figure}

The Visual Interactive Labeling tool for Object Detection (VILOD) employs a containerized client-server architecture. The system is composed of three primary components. A front-end web application, a backend API, and a MySQL database. 
Each is encapsulated in a Docker container and orchestrated through Docker Compose. 
In addition, an nginx is used as a reverse proxy in the system, routing API requests and WebSocket connections as well as handling the SSL certification for the deployed application. Figure \ref{sysarc-vilod} describes the high-level architecture of the proposed VILOD system.\\

The back-end is implemented as a Flask application with a structured design. It includes a routing layer supported by the Flask library. 
This layer exposes API routes regarding data fetching and authentication for the front-end, as well as a WebSocket connection for triggering training and sending real-time progress updates. 
The architecture includes an AL layer, which utilizes the Ultralytics library for retraining and fine-tuning YOLO detection models, as well as for running inference and validation. It also contains a data processing layer that is responsible for writing to and reading from the MySQL database and preparing new annotation data received from the front-end.
This includes writing YOLO formatted annotation text files and copying image files to a user's training set folder, to be utilized in training new model iterations. 

The back-end component is responsible for calculating and suggesting the AL samples displayed in the front-end Data View. In practice, this involves querying the database for the image detections made by the latest model of a given user. 
Averaging the detection scores related to an image, sorting the image-score map on ascending values, and selecting the top n images. 
More formally, the process is defined by Algorithm \ref{alg:al_selection_from_scores}:

\begin{algorithm}[H]
\caption{Active Learning Sample Selection}
\label{alg:al_selection_from_scores}
\begin{algorithmic}[1]
\State \textbf{Inputs:}
\State \quad $ImageDetectionsMap$: A map where keys are image identifiers and values
\State \quad \quad are arrays of individual object detection scores for that image. 
\State \quad \quad An empty array signifies no detections.
\State \quad $D_{exclude}$: A set of image identifiers to be excluded, such as already labeled samples.
\State \quad $N_{budget}$: The predefined number of samples to be selected.

\Procedure{SelectALSamples}{$ImageDetectionsMap, D_{exclude}, N_{budget}$}
    \State $CandidateImageAvgScores \leftarrow \emptyset$
    \ForAll{image $i$ with $DetectionScores_i$ in $ImageDetectionsMap$}
        \If{$i \notin D_{exclude}$}
            \State $AvgConf_i \leftarrow 0.0$
            \If{$DetectionScores_i \text{ is not empty}$}
                \State $TotalScore_i \leftarrow 0$
                \ForAll{score $s \in DetectionScores_i$}
                    \State $TotalScore_i \leftarrow TotalScore_i + s$
                \EndFor
                \State $AvgConf_i \leftarrow TotalScore_i / \text{Length}(DetectionScores_i)$
            \EndIf
            \State Add $(i, AvgConf_i)$ to $CandidateImageAvgScores$
        \EndIf
    \EndFor

    \State $SortedCandidates \leftarrow \text{SortByAscendingValue}(CandidateImageAvgScores)$

    \State $D_{AL} \leftarrow \text{GetFirstNItems}(SortedCandidates, N_{budget})$
    \State \Return $D_{AL}$
\EndProcedure
\end{algorithmic}
\end{algorithm}

In addition, the back-end component calculates and generates the KDE heatmap overlaid in the Data View scatter plot, as depicted in \ref{dataview}. Specifically, a Gaussian KDE implementation provided by the SciPy library \cite{SciPyKDE}. It estimates the density of the uncertainty scores across the 2D projection of the data. The input for this heatmap consists of the x and y coordinates of each image in the scatterplot, along with their corresponding average confidence score, as generated by the latest model iteration. 
It uses the 'Reds' colormap, where more intense shades of red indicate regions with a higher aggregated uncertainty. 

An important aspect of the heatmap generation is the weighting mechanism: data points are weighted based on their uncertainty, calculated as:
\begin{displaymath}
    (1-avgConfidence)^2
\end{displaymath}

The quadratic weighting ensures that images with lower average confidence scores (and thus higher uncertainty) contribute more significantly to the density estimation. 
Consequently, areas in the scatterplot that contain a higher concentration of uncertain images will appear more prominent in the heatmap.

\begin{figure}[ht!]
\begin{center}
\includegraphics*[width=0.8\columnwidth]{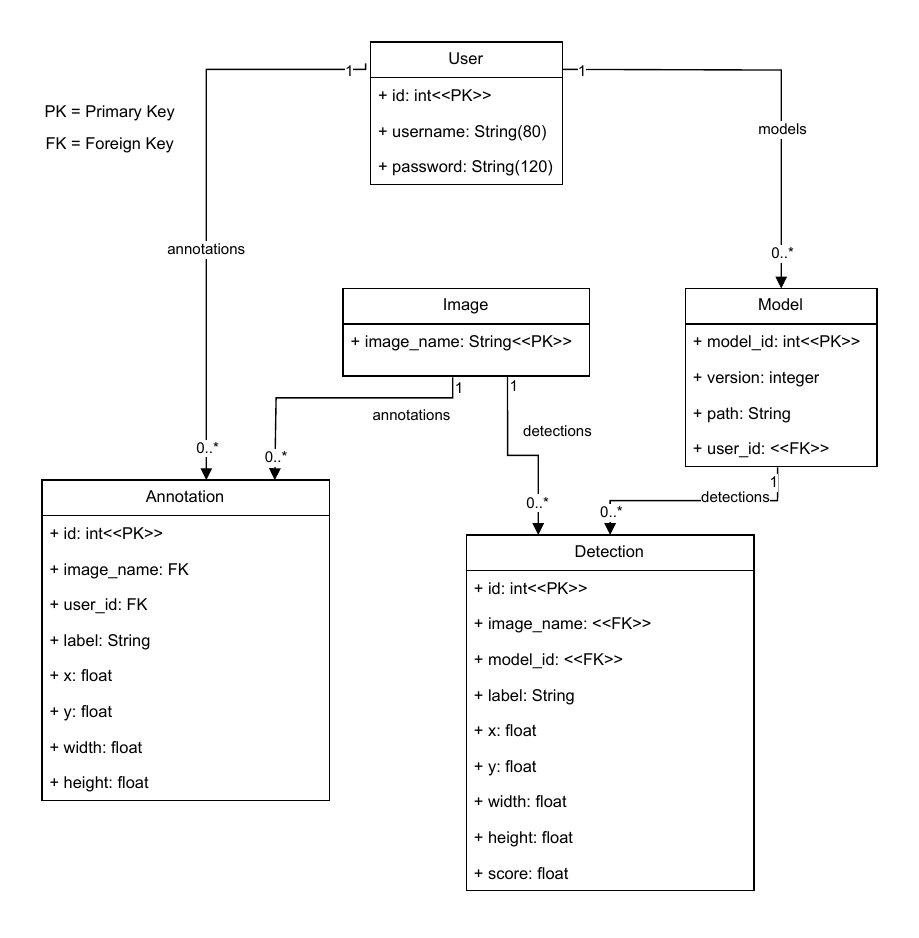}
\end{center}
\caption{The Database Schema of the mySQL service.}
\label{dbrelation}
\end{figure}

The database component consists of a MySQL database, allowing for persistent storage across users. The database includes five tables: User, Image, Model, Annotation, and Detection. Their relations and attributes can be seen in Figure \ref{dbrelation}. 
By using this relational schema, the process of tracking user-specific progress, such as annotation history, model versions, and detections for a given model, is simplified. Note that images and model tables in the database mainly exist as helpers for easier lookup and tracking; the actual image files and model weights are stored on file within the local file storage of the machine running the service.

\subsubsection{Uncertainty-based Active Learning Baseline Definition}
A computational simulation was conducted to establish an AL baseline using an uncertainty sampling strategy, serving as a comparative benchmark for RQ2, and that was integrated in VILOD. 
This simulation modeled an automated AL process without human interaction. 
The simulation commenced using the identical $M_0$ model prepared previously. 
It proceeded iteratively for 5 cycles. In each iteration t (from 1 to 5), the current model $M_{t-1}$ generated predictions for the currently unlabeled images $U_{current}$. 
An uncertainty sampling strategy ("Lowest Average Confidence") was applied, selecting 30 images from $U_{current}$ exhibiting the lowest average scores. 
These 30 selected images were considered "labeled" (using ground truth labels) and moved from $U_{current}$ to the growing labeled set $L_{current}$. 
The model was retrained starting from the previous iteration's best weights ($M_{t-1}$), using the updated $L_{current}$ and the fixed Validation Set, employing the same fixed 50-epoch training parameters (epochs=50 seed=42, imgsz=640) used for $M_0$. The best performing model weights (best.pt) for the iteration ($M_t$) were saved. 
The primary data recorded were the models $M_0$ through $M_5$ and their performance metrics evaluated on the held-out Test Set.

\subsubsection{Use Cases}
A multi-use case study forms an empirical investigation of this thesis. 
It involved the execution of the interactive active learning process three separate times by the author (e.g., acting as expert user), each adhering to a different predefined strategy for sample selection and model training using the VILOD system. 
Each case starts from the identical $M_0$ model and proceeds for 5 iterations with a budget of 30 samples per iteration. 
Retraining in each iteration used the consistent fixed 50-epoch parameters, but the best-performing epoch weights are saved and used as the new model iteration. 
The three use cases carried out were:

\paragraph{Use Case 1 - Exploration \& Structure Focus.} The selection process primarily utilized the t-SNE Data View to identify visually distinct clusters, sparse regions, or dense unlabeled areas for sampling, aiming for broad coverage of the feature space. 
    The \textit{Model View} was consulted to guide selections towards improving class balance or addressing classes with poor overall confidence. 
    Uncertainty information (heatmap intensity, specific AL suggestions) was deliberately de-emphasized as the primary driver for selection.

\paragraph{Use Case 2 - Uncertainty-Driven Focus.} The selection process primarily followed the signals provided by the uncertainty heatmap and the AL suggestions. Samples were chosen predominantly from the darkest red heatmap regions and included a large proportion of the AL-suggested images. The human role was primarily limited to basic quality control, filtering out only samples deemed entirely unusable (e.g., corrupt images), but otherwise trusting the model's uncertainty metric, even for visually "noisy" images. Less emphasis was placed on deliberate exploration or class balancing beyond what emerged from the uncertainty sampling.

\paragraph{Use Case 3 - Balanced Guidance Integration.} This scenario simulated the intended "best practice" use, requiring the synthesis of all available guidance. The process involved inspecting uncertainty (heatmap, AL suggestions), cross-referencing with data structure (t-SNE density, clusters), and consulting model feedback (Model View confidence/balance). Selections are made to address high-uncertainty areas, explore diverse unlabeled regions, improve class balance, and potentially correct specific observed model errors.

For each of the three scenarios, detailed qualitative data were recorded in a logbook along with screenshots. The sequence of selected images was logged. The final models M0-M5 generated from each scenario run were saved and subsequently evaluated on the held-out Test Set.

\subsection{Data Analysis}

The data analysis in this thesis will integrate quantitative performance evaluation with qualitative process analysis to thoroughly address the research questions. 
The primary quantitative aspect involves analyzing the performance of the object detection models generated throughout the various Use Case scenarios. 
To visualize progress and compare approaches, performance trajectories, plotting mAP against the model iterations, were generated for each use case and the baseline. 
This plotting and direct comparison of performance trajectories is specifically designed to answer RQ2, by showing how object detection performance differs when various visually-guided labeling strategies are employed relative to the baseline uncertainty sampling approach. 

Complementing the quantitative data, qualitative analysis will be performed on information extracted from the detailed logbook and screenshots compiled during the execution of the three use cases within the VILOD system. This qualitative data will be examined to understand how VILOD's visual guidance features, such as the t-SNE Data View, uncertainty heatmap, and Model View, were employed in each strategy and how they affected the performance of the model. It will also shed light on how these features impacted decision-making during sample selection. This qualitative analysis mainly addresses RQ1, concerning the enhancement of the user's capacity to interpret model state and implement informative labeling strategies.

The VILOD system itself is the main instrument for conducting experiments and gathering model performance data. Subsequent analysis, including the plotting of performance trajectories, was carried out using standard Python data analysis and visualization libraries such as matplotlib and SciPy.

\subsection{Reliability and Validity}

This section outlines the measures undertaken to enhance reliability and to address potential threats to the validity of the study. 

To promote reliability within this study, several measures were implemented. A standardized experimental setup was used for all experimental runs, including both the baseline simulation and the three use cases, which originated from an identical initial model. 
This model was trained on the same initial diverse set of 40 images, selected through K-Means clustering. 
Furthermore, consistent parameters were maintained throughout: the "African Wildlife" dataset splits (train, validation, test) remained fixed across all experiments, and model retraining in each iteration employed uniform parameters, including a set number of 50 epochs, a fixed image size (imgsz=640), and a constant random seed (seed=42) to ensure deterministic behavior where feasible. 
The Uncertainty Active Learning Baseline was executed as an automated simulation, thereby removing human variability from this critical reference point. 
Although the three expert-driven use cases were performed by the author, they adhered to predefined strategies for sample selection, guiding the interaction in a structured manner, with detailed logs and screenshots kept to document the process. 

The internal validity concerns of this thesis lie mainly in the fact that the author is acting as the single user in the use case studies, which introduces potential subjectivity and bias. 
This is mitigated by the use of clearly predefined strategies for each use case, aiming to standardize interaction with the VILOD tool, and by detailed logging to ensure transparency. 
The comparison against an automated baseline also provides a non-subjective benchmark. 

Regarding external validity, the findings of this thesis might be constrained by the use the use of a single dataset ("African Wildlife"), a particular object detection architecture (YOLOv11n), and a specific Active Learning suggestion mechanism (uncertainty sampling). 
Moreover, the author's familiarity with the system might not be representative of other potential users. While the chosen dataset and model are relevant in the object detection field and VILOD's principles are designed for adaptability, generalization of specific performance outcomes, warrants further research. 

\subsection{Ethical Considerations}

The research conducted in this thesis adheres to ethical conduct, especially concerning data usage. Since the use case studies described were conducted by the author acting as the expert user, no external human participants were involved in the data collection for VILOD's evaluation, thus negating the need for a formal informed consent process for this phase. 
The study utilizes the "African Wildlife" dataset, which is publicly accessible and does not include any personal data. 

Potential biases are an important consideration. The "African Wildlife" dataset may have inherent biases in terms of class distribution or image characteristics. 
While this study does not aim to mitigate these pre-existing dataset biases, their potential influence on model performance is acknowledged; VILOD's Model View, by displaying class balance, can help users become aware of such issues in their labeled set. 
Similarly, the underlying YOLOv11n model and t-SNE algorithm might possess their own biases. Although exploring these is beyond this thesis's scope, VILOD's interactive nature is designed to provide users with tools to understand and potentially counteract undesirable model behaviors. 

The VILOD tool is intended to make the development of object detection models more efficient, transparent, and strategically manageable, thereby contributing positively to AI development. No direct negative societal impacts are anticipated from this research prototype.  \\

\subsection{Summary}

This chapter has comprehensively detailed the methodology underpinning the development and evaluation of the VILOD system. The adopted research approach is a comparative analysis of use cases. In these, the author, simulating an expert user, leverages the VILOD tool to conduct object detection annotation according to several predefined strategies. 

The data collection process encompassed several stages: the initial preparation of the "African Wildlife" dataset, the development of an initial YOLOv11n model ($M_0$), the extraction of image features and their subsequent dimensionality reduction to 2D using t-SNE for effective visualization, and importantly, the design and implementation of the VILOD tool itself. An automated uncertainty-based active learning baseline was simulated to provide a benchmark against three distinct use case scenarios: \textit{Exploration \& Structure Focus}, \textit{Uncertainty-Driven Focus}, and \textit{Balanced Guidance Integration}, executed with VILOD, were compared. 

The subsequent data analysis merges quantitative metrics, mainly model performance (mAP) trajectories, with qualitative analysis derived from interaction logs. This approach aims to answer how VILOD facilitates different user strategies. Throughout the process, careful attention has been paid to reliability, ensured through standardized setups and parameters. External and internal validity have also been considered, with an acknowledgment of limitations such as the single-user evaluation and dataset specificity. Ethical considerations have primarily revolved around the use of public data and the author-centric nature of the experiments, leading to the conclusion that no significant negative ethical implications arise from this specific research. This methodology lays the groundwork for generating and analyzing the results that will be presented and interpreted in the chapters that follow.

\newpage

\section{Results and Analysis}
\label{ResultsAnalysis}

This chapter presents the findings obtained from the empirical investigation conducted to evaluate the VILOD system and address the research questions outlined in Section \ref{problemFormulation}. As detailed in Chapter 3, the methodology involved a comparative analysis of three distinct use case scenarios performed by the author using the VILOD tool, alongside an automated uncertainty-based AL baseline simulation. 
The primary goal was to investigate how VILOD's visual components facilitate the implementation of distinct, expert-driven labeling strategies (RQ1) and to compare the resulting performance trajectories of the object detection model against a baseline (RQ2).

The chapter is structured as follows: Sections \ref{use_case_stuctural}, \ref{use_case_uncertainty}, and \ref{use_case_balanced} detail the narrative and specific observations from each of the three executed use cases, "Exploration \& Structure Focus," "Uncertainty-Driven Focus," and \textit{Balanced Guidance Integration}, respectively. 
Each of these sections will describe and guide the reader through the labeling process, showcasing the different functionalities and possibilities enabled by the system. Section \ref{analysis} provides a comparative analysis of these strategies against each other and the automated baseline, highlighting key differences in performance trajectories and process characteristics. 
Finally, Section \ref{summaryoffindings} summarizes the principal findings of this empirical study. 

\subsection{Use Case 1: Exploration \& Structure Focus}
\label{use_case_stuctural}

In this use case, the primary strategy for selecting images to annotate was driven by an exploration of the dataset's structure as visualized in the VILOD Data View (the t-SNE scatterplot). The objective was to achieve broad coverage of the feature space by identifying and sampling from visually distinct clusters, sparse regions, or dense unlabeled areas. While model uncertainty information (heatmap intensity, AL suggestions) was available, it was deliberately de-emphasized as the primary selection driver. The Model View was consulted periodically to guide selections towards improving class balance or addressing classes with observably poor overall confidence distributions.

\subsubsection{Iteration 1}
\label{use_case_stuctural_Iteration_1}
\begin{figure}[ht!]
    \centering
    \includegraphics[width=0.4\textwidth]{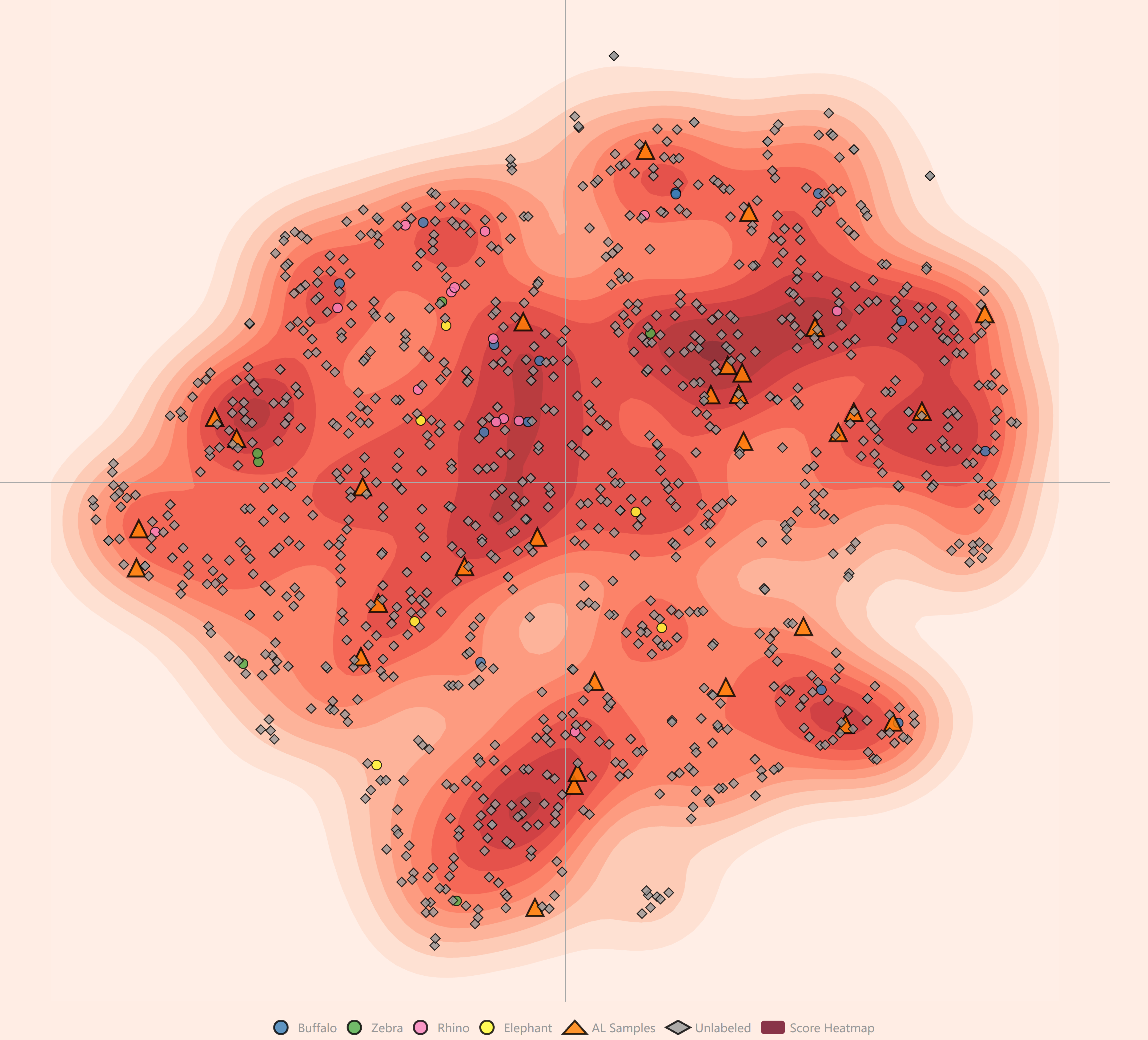}
    \caption{Data View at the start of iteration 1.}
    \label{c3it1_dataview}
\end{figure}

\begin{figure}[ht!]
  \centering
  \begin{minipage}[t]{0.4\textwidth}
    \centering
    \includegraphics[width=\linewidth]{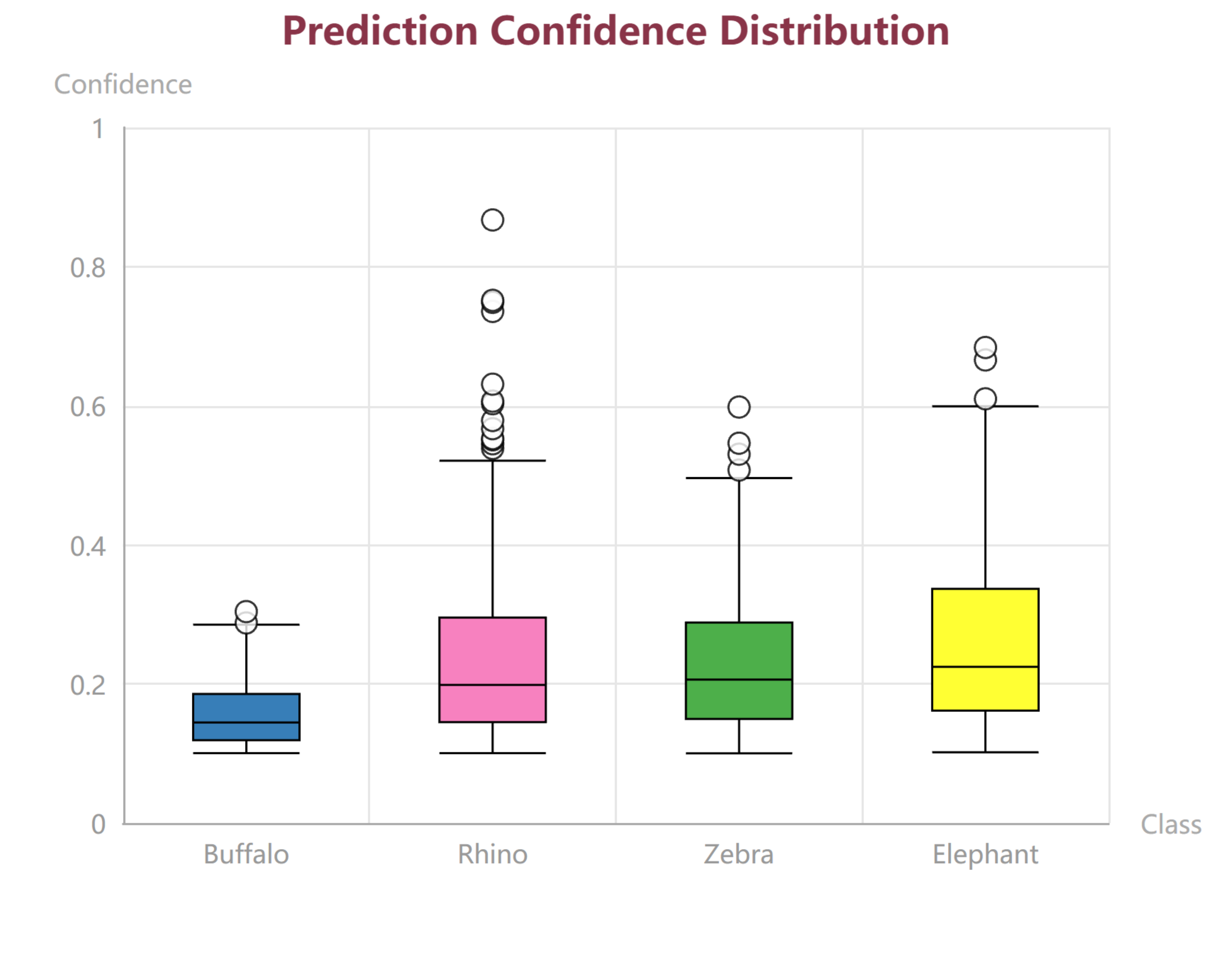}
  \end{minipage}
  \begin{minipage}[t]{0.4\textwidth}
    \centering
    \includegraphics[width=\linewidth]{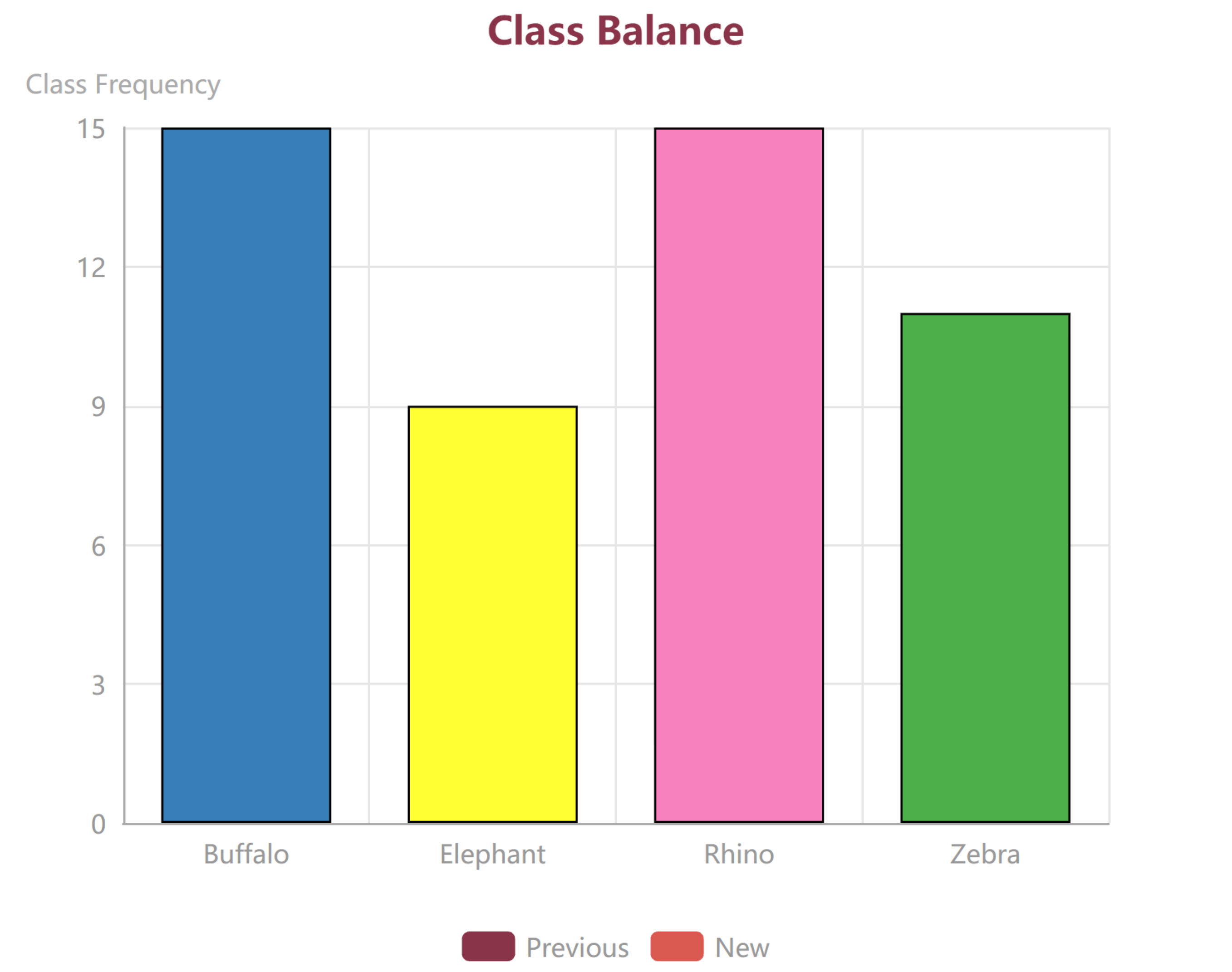}
  \end{minipage}
  \caption{Model View at the start of training iteration 1.}
  \label{c3it1_modelview}
\end{figure}

The first iteration starts with the Data View (Figure \ref{c3it1_dataview}) visualization and Model View (Figure \ref{c3it1_modelview}) as generated based on the initial model $m_0$. In this use case, the Data View is interpreted from a structural standpoint to identify potential patterns, clusters, or general features of the original data space from which samples can be selected.

\begin{figure}[ht!]
    \centering
    \includegraphics[width=0.5\textwidth]{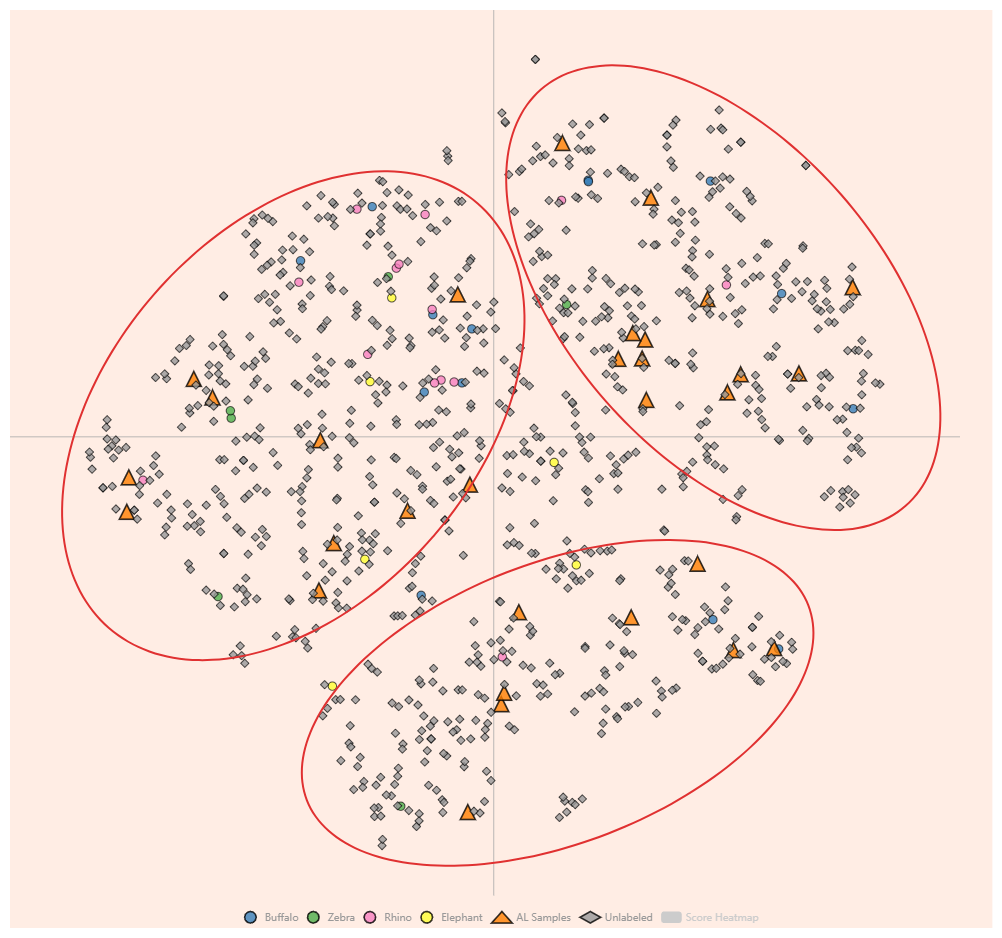}
    \caption{Three identified blobs within the Data View scatter plot.}
    \label{scatterplot_blobs}
\end{figure}

In broad terms, most of the data points in the scatterplot are divided into three "blobs," loosely separated by more sparse regions. For a better reference, these blobs have been highlighted in Figure \ref{scatterplot_blobs}. There is arguably another region of points belonging together, towards the middle of the visualization, positioned between the other blobs. Within these larger blobs, smaller regions of points clustered more tightly are visible. It is worth keeping in mind that this interpretation is based only on what can be visually discerned in this reduced 2D view of the original high-dimensional images. Structural patterns visible here are not guaranteed to be perfectly representative of the original feature space, but serve as the primary guide for this exploration-focused strategy

\begin{figure}[ht!]
  \centering
  \begin{minipage}[c]{0.49\textwidth}
    \centering
    \includegraphics[width=\linewidth]{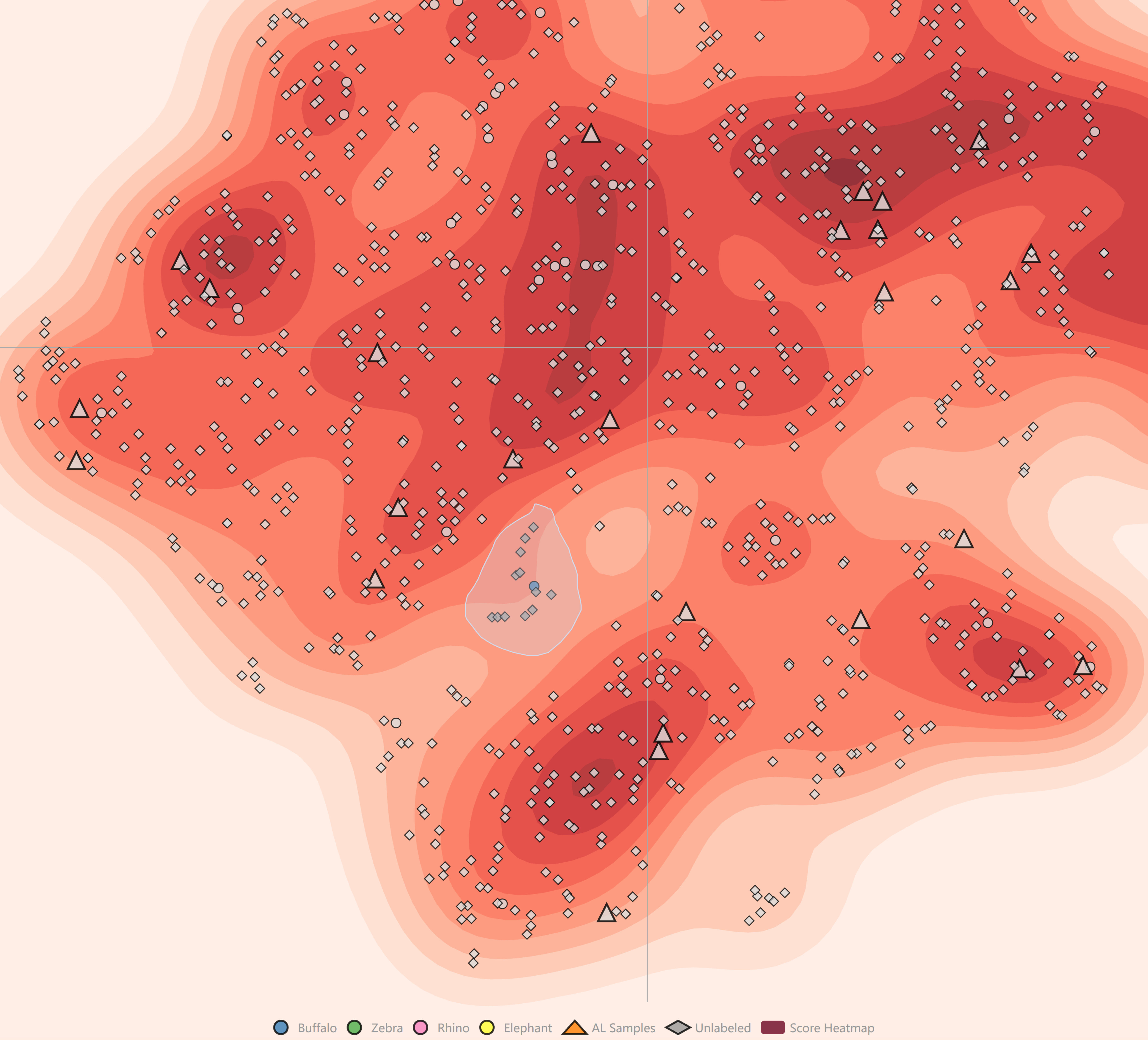}
  \end{minipage}
  \hfill
  \begin{minipage}[c]{0.49\textwidth}
    \centering
    \includegraphics[width=\linewidth]{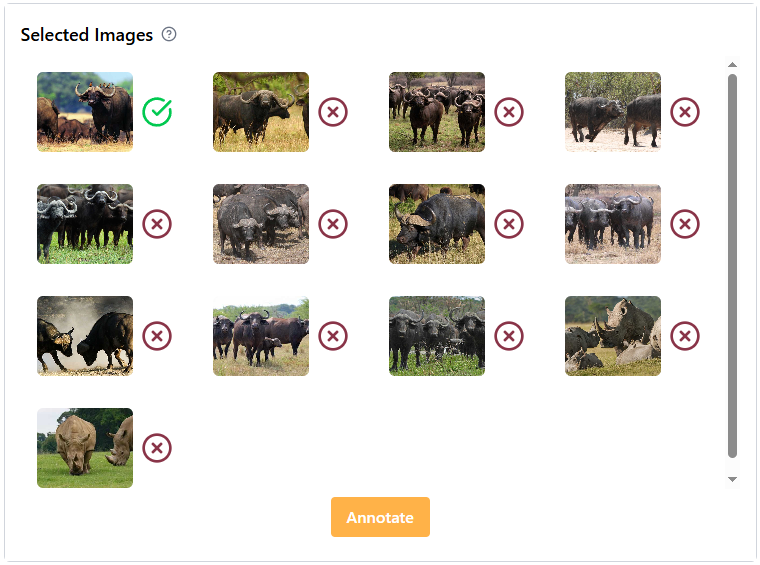}
  \end{minipage}
  \caption{First sample selection of iteration 1}
  \label{it1s1_dataview}
\end{figure}

\noindent The first selection is made in the third quadrant. Here lies a region of 13 images that exhibit some visual separation from the bulk of the other points, suggesting a potentially distinct subgroup. The lasso tool is utilized to draw a region around these samples for further inspection (Figure \ref{it1s1_dataview}). \\

\begin{figure}[ht!]
    \centering
    \includegraphics[width=0.5\textwidth]{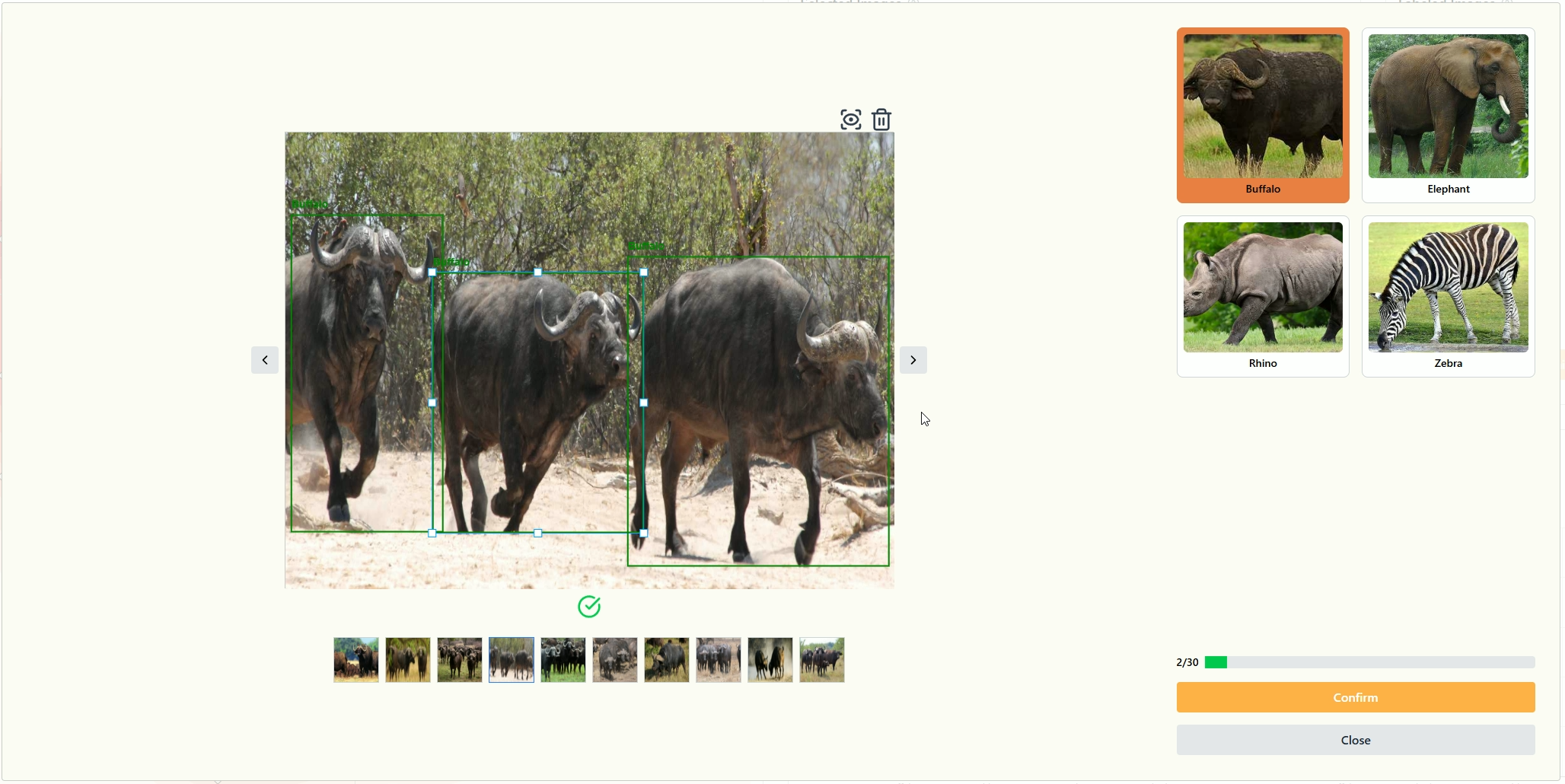}
    \caption{Annotating the first sample batch.}
    \label{i1s1_draw}
\end{figure}

The selected images seem to largely contain samples of buffalos. From here, these samples are opened in the drawing interface by pressing the "Annotate" Button. Since the aim for this initial iteration, and for the use case strategy as a whole, is to achieve broad coverage across the feature space rather than deeply sampling any single area, only three of these samples are annotated (Figure \ref{i1s1_draw}).

\noindent Continuing the exploration of the Data View, another region of samples is found at the bottom of the fourth quadrant (Figure \ref{it1s2}) that is also clearly visually separated from other points, making it a candidate for capturing further diversity. Three samples are annotated here, making the total budget progression 6/30 at this point.

\begin{figure}[ht!]
  \centering
  \begin{minipage}[c]{0.49\textwidth}
    \centering
    \includegraphics[width=\linewidth]{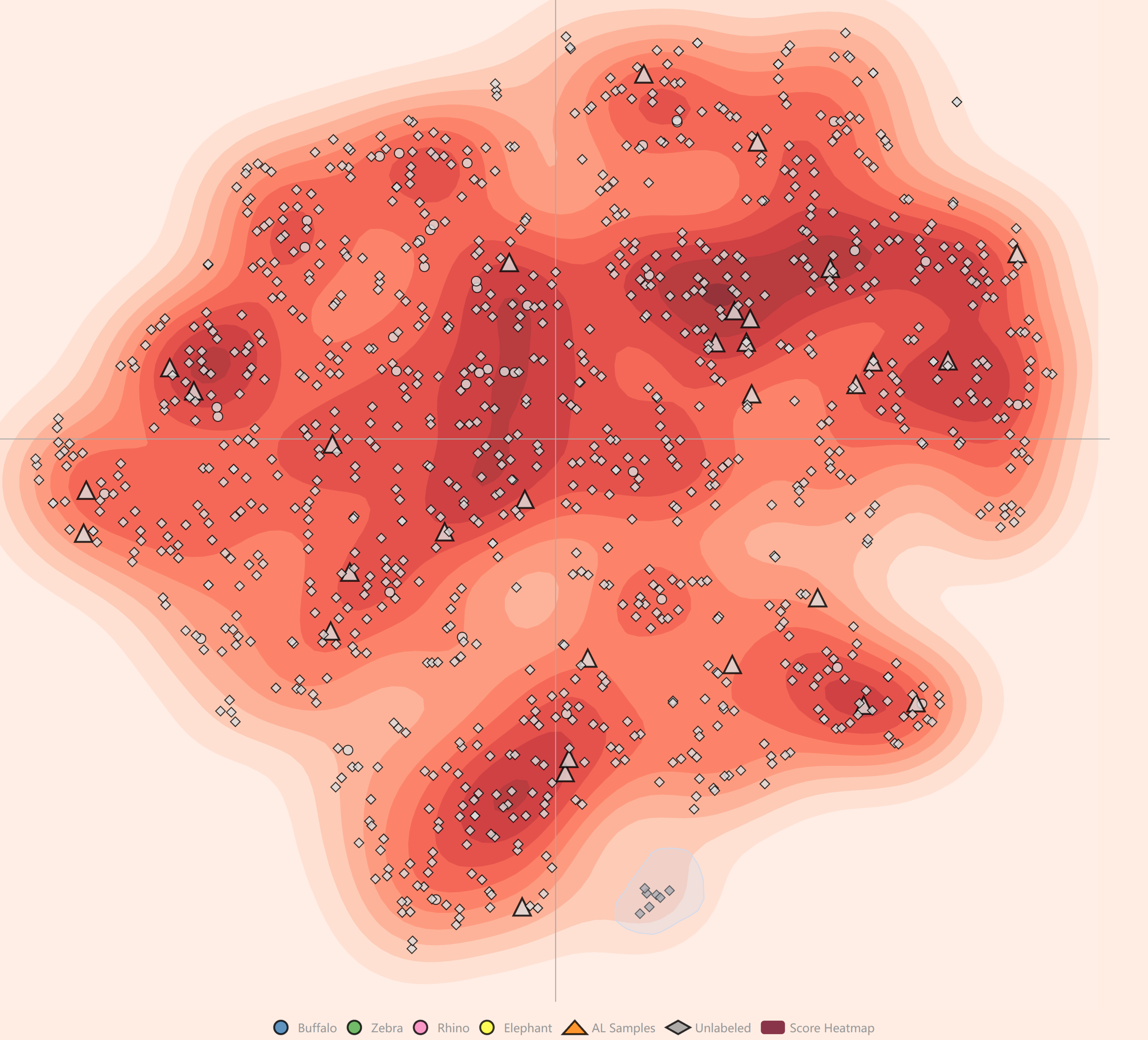}
  \end{minipage}
  \hfill
  \begin{minipage}[c]{0.49\textwidth}
    \centering
    \includegraphics[width=\linewidth]{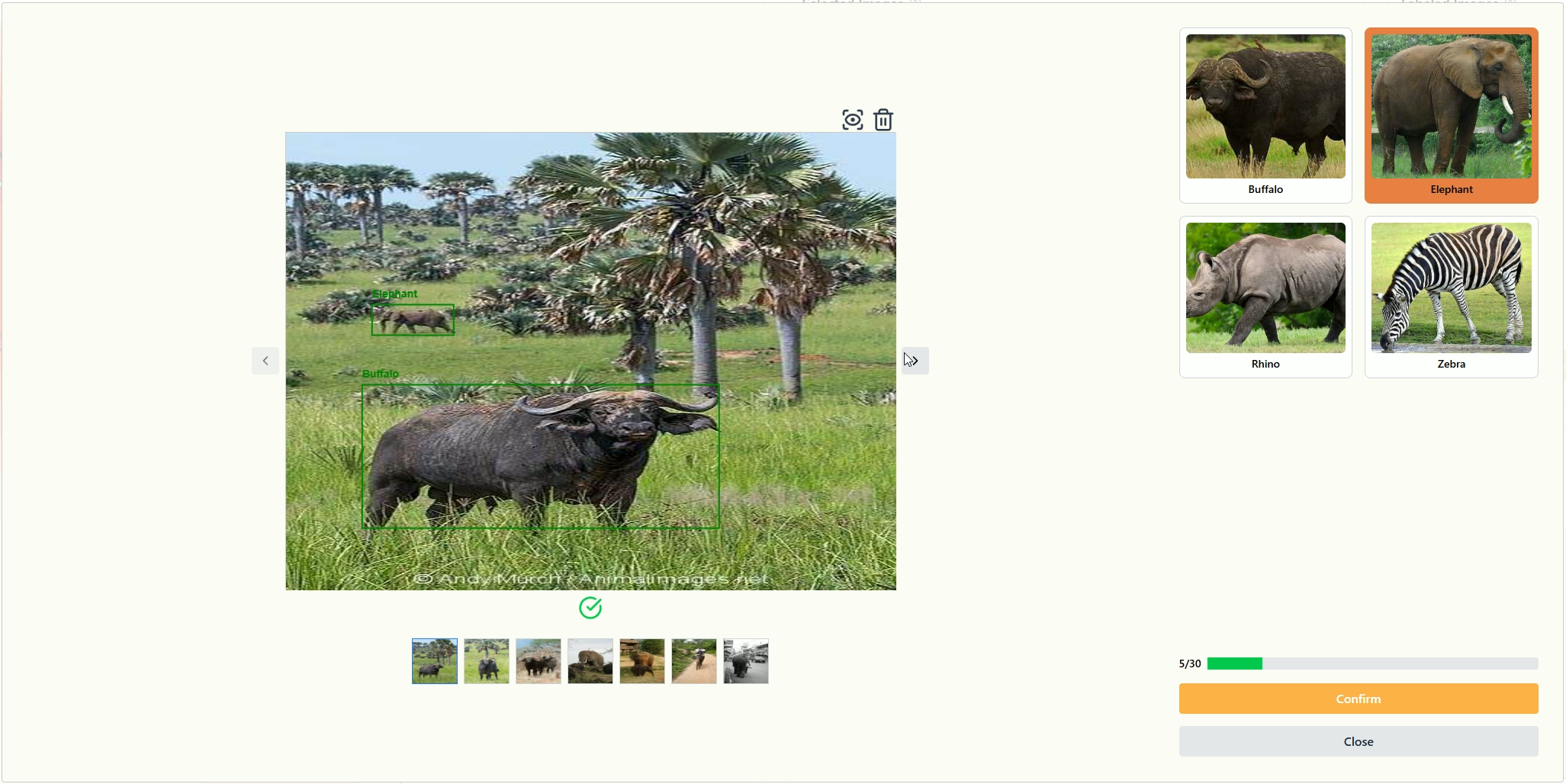}
  \end{minipage}
  \caption{Second sample selection and annotation.}
  \label{it1s2}
\end{figure}

\begin{figure}[ht!]
    \centering
    \includegraphics[width=0.5\textwidth]{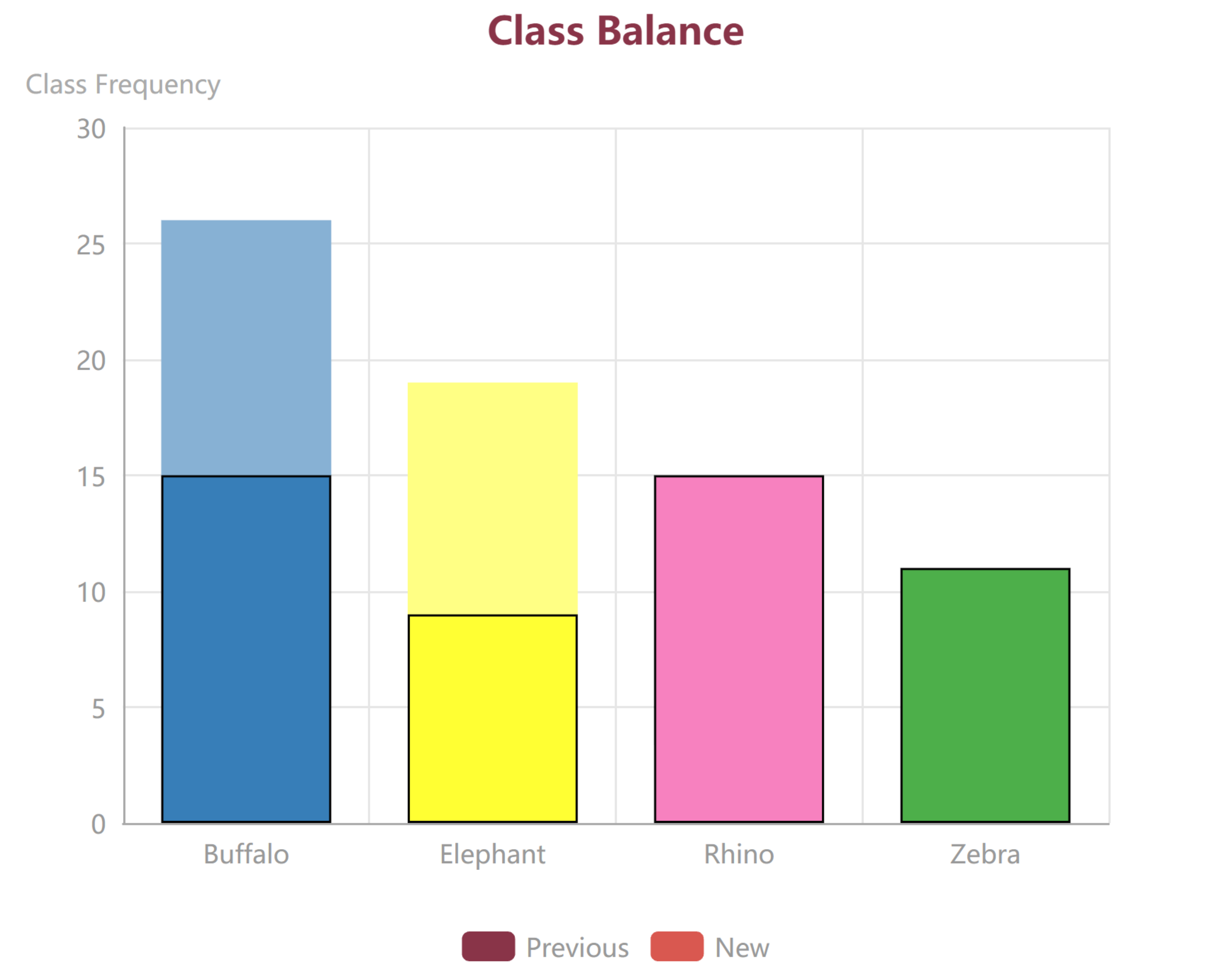}
    \caption{Class Balance after almost half the labeling budget is spent.}
    \label{i1_cb2}
\end{figure}

\newpage
\noindent After a few more selections guided by identifying distinct visual groupings, and with approximately half the labeling budget spent, the class balance chart in the Model View is consulted (Figure \ref{i1_cb2}). It's observed that only buffalo and elephant instances have been labeled thus far. To counteract this imbalance and still adhere to the goal of diverse coverage, for the second half of the labeling budget, an effort will be made to specifically target rhino and zebra samples by seeking unexplored structural regions that might contain these classes.

\noindent By exploring the Data View, now with an awareness of the needed classes, and hovering over data points to preview thumbnails, a larger dense region containing zebras is discovered in the first quadrant, and a smaller, sparser region suggesting rhinos is found in the third quadrant (Figure \ref{it1s2}). These structurally distinct areas, now also targeted for their potential class content, are sampled.

\begin{figure}[ht!]
  \centering
  \begin{minipage}[c]{0.49\textwidth}
    \centering
    \includegraphics[width=\linewidth]{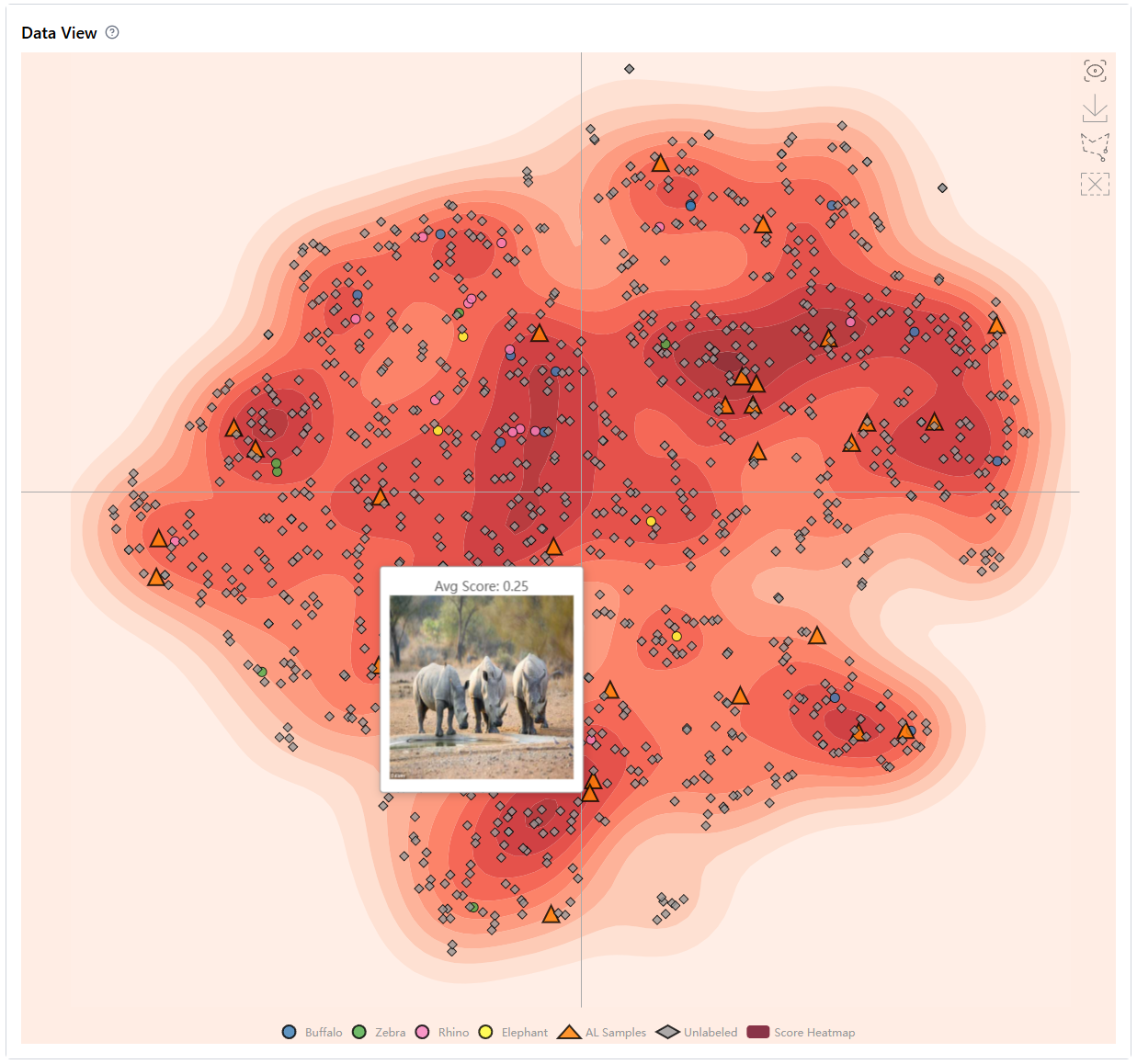}
  \end{minipage}
  \hfill
  \begin{minipage}[c]{0.49\textwidth}
    \centering
    \includegraphics[width=\linewidth]{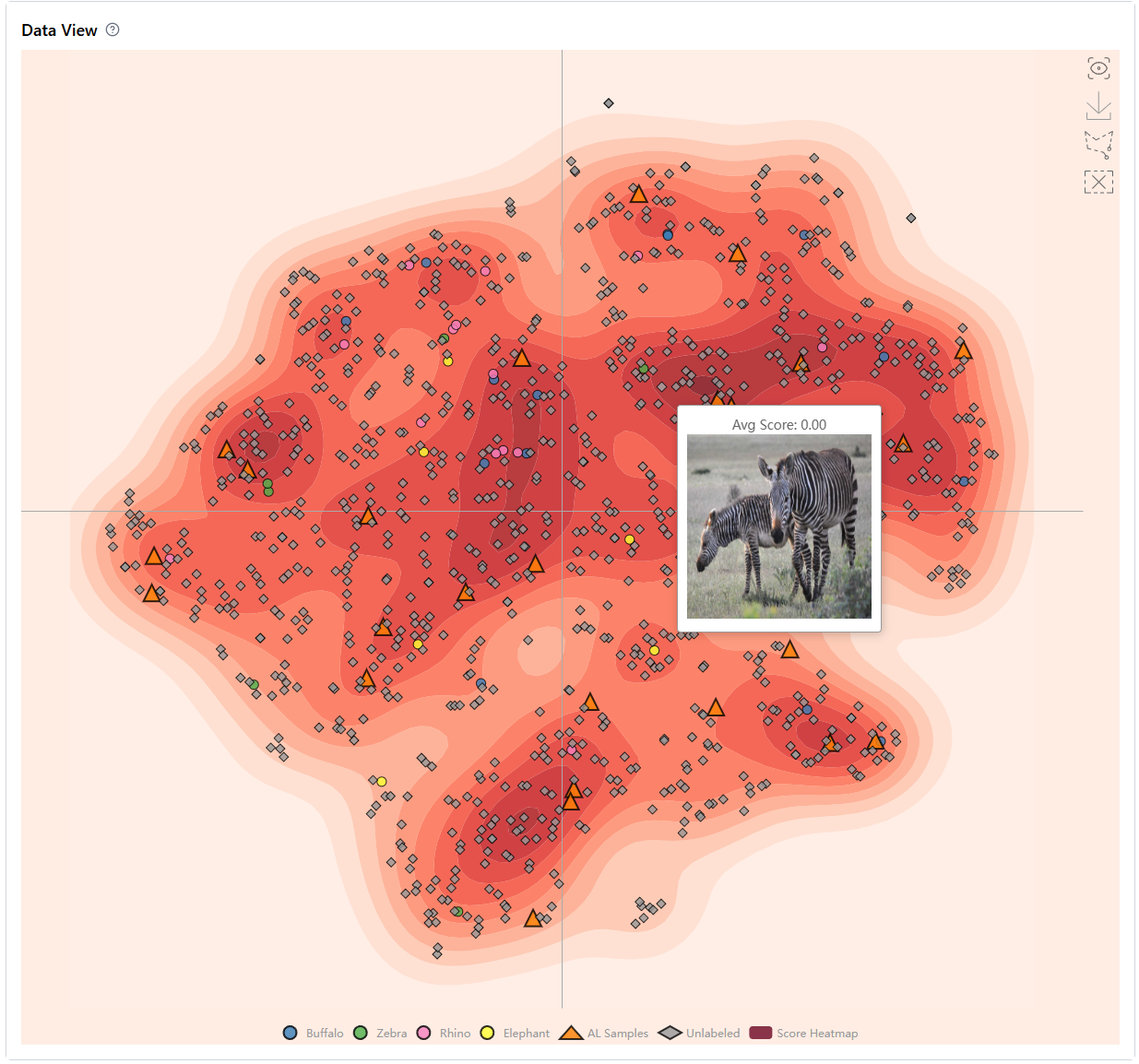}
  \end{minipage}
  \caption{Exploring the Dataview, finding instances of rhinos and zebras.}
  \label{it1s2}
\end{figure}

\begin{figure}[ht!]
  \centering
  \begin{minipage}[c]{0.4\textwidth}
    \centering
    \includegraphics[width=\linewidth]{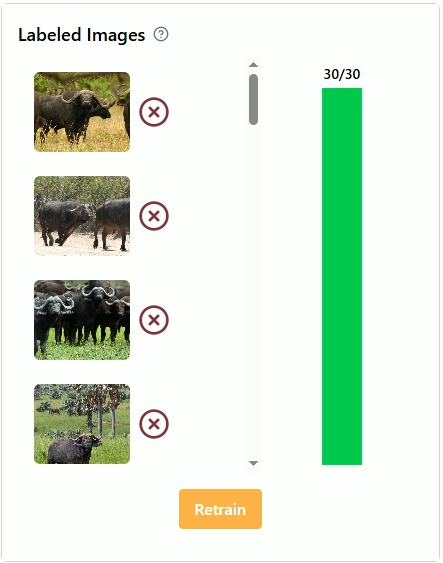}
  \end{minipage}
  \begin{minipage}[c]{0.4\textwidth}
    \centering
    \includegraphics[width=\linewidth]{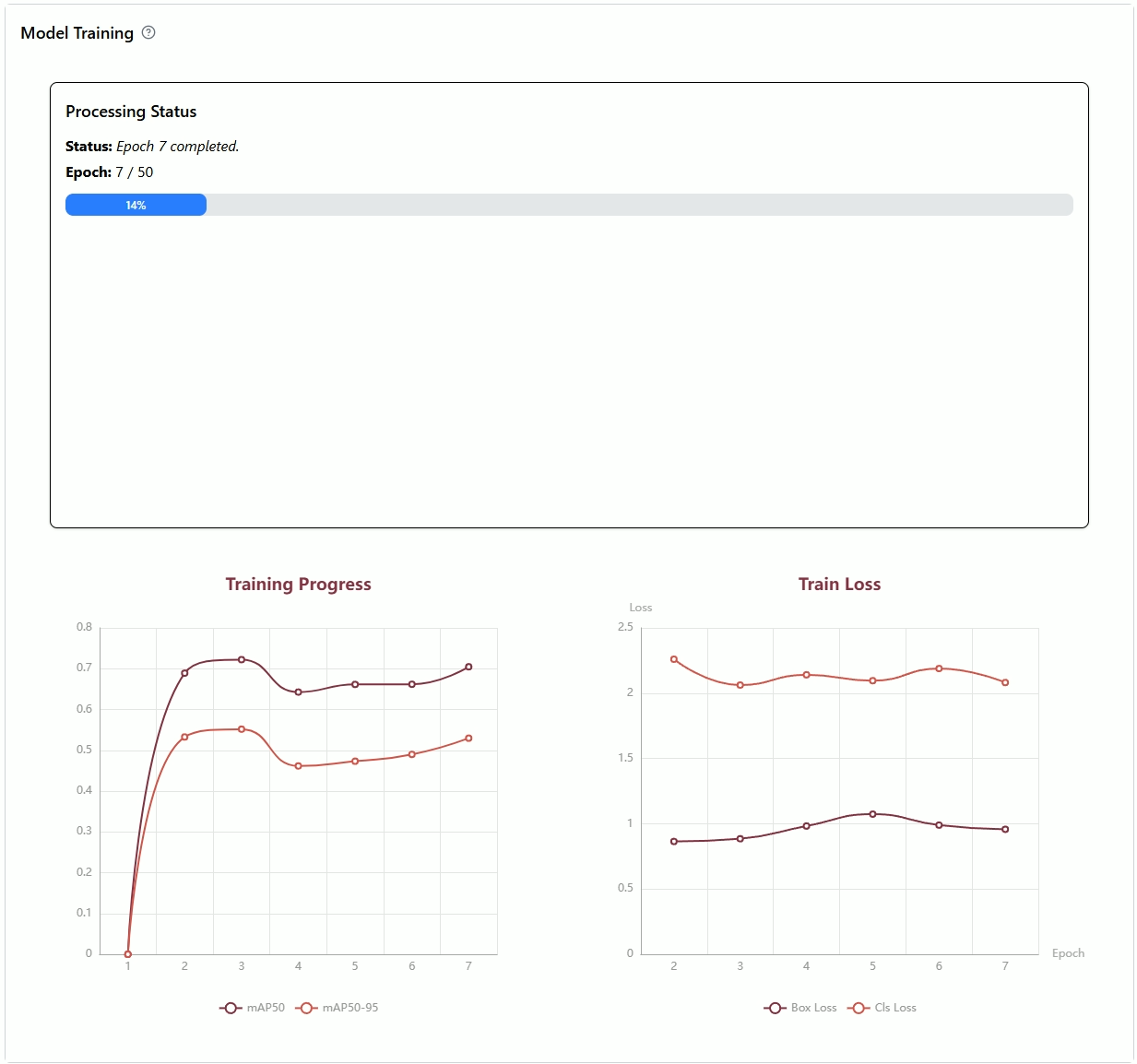}
  \end{minipage}
  \caption{Starting retraining and monitoring the training process.}
  \label{it2c1_retrain}
\end{figure}

\newpage
\noindent This process of identifying structurally unique areas and then considering class balance continues until the labeling budget of 30 images is met. Retraining of the model is initialized, and the training process is monitored (Figure \ref{it2c1_retrain}) before moving on to iteration 2 with updated states.

\subsubsection{Iteration 2}
\begin{figure}[ht!]
    \centering
    \includegraphics[width=0.5\textwidth]{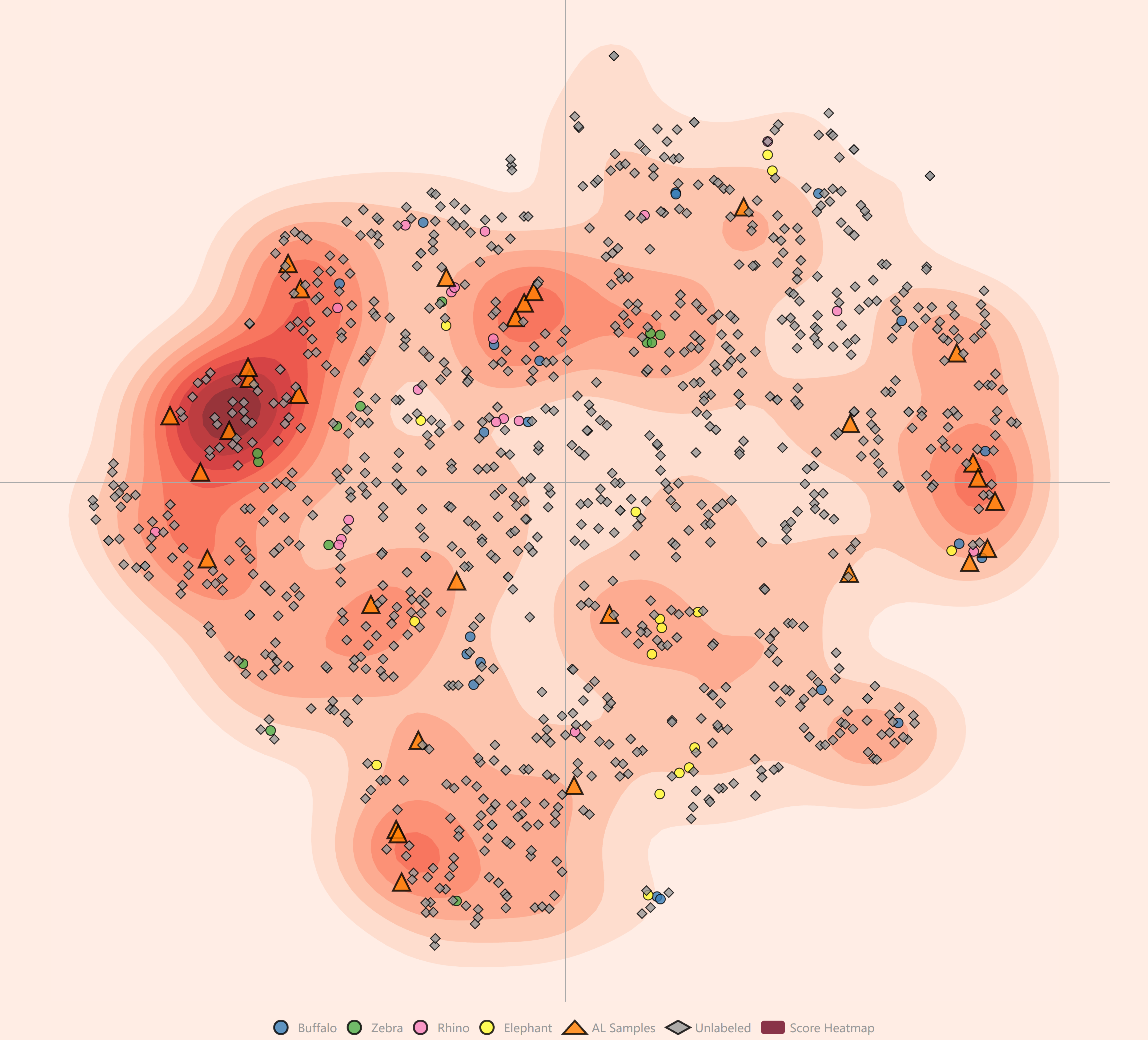}
    \caption{Data View in the start of iteration 2.}
    \label{c2_iteration2_dv}
\end{figure}

\begin{figure}[ht!]
  \centering
  \begin{minipage}[c]{0.49\textwidth}
    \centering
    \includegraphics[width=\linewidth]{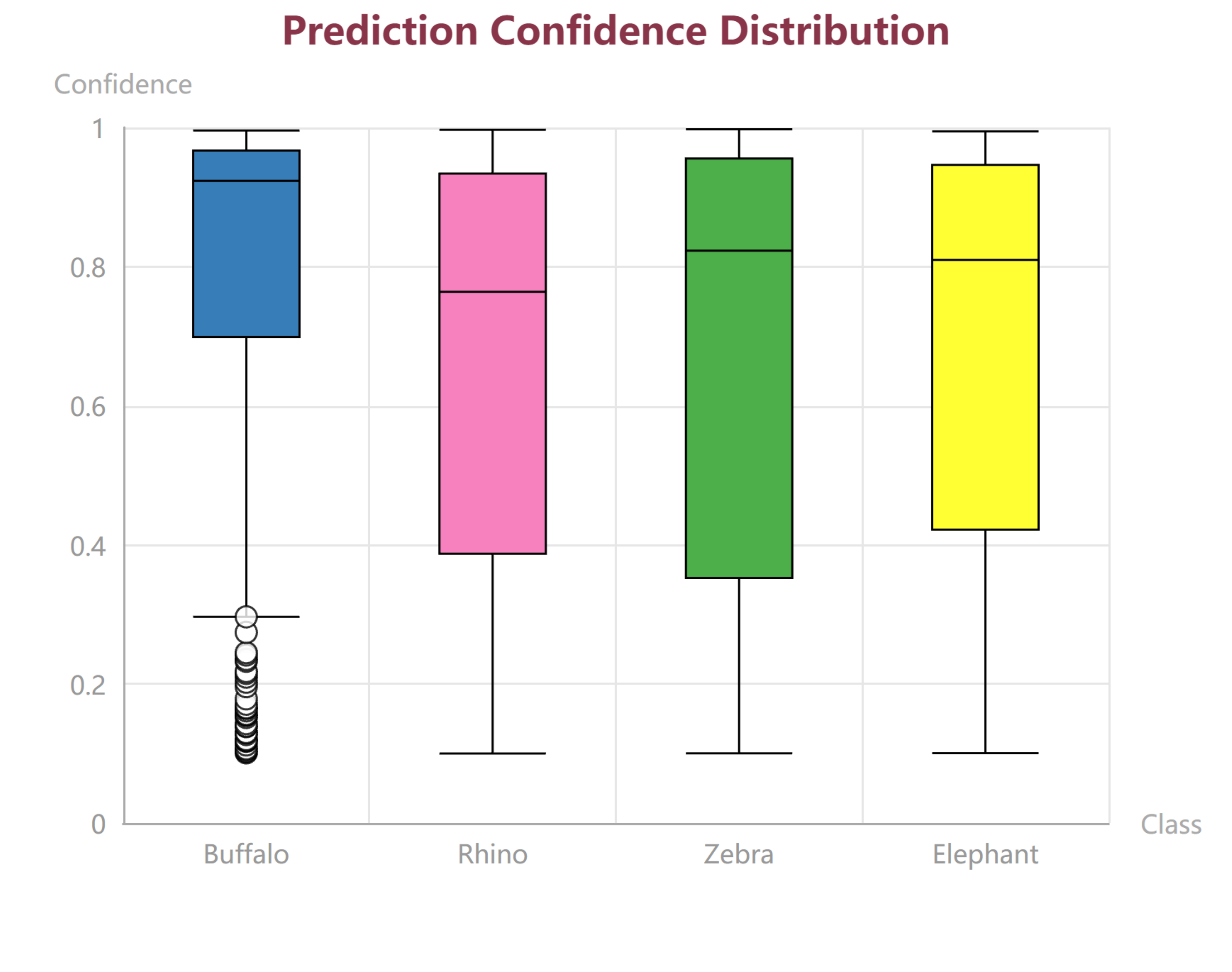}
  \end{minipage}
  \hfill
  \begin{minipage}[c]{0.49\textwidth}
    \centering
    \includegraphics[width=\linewidth]{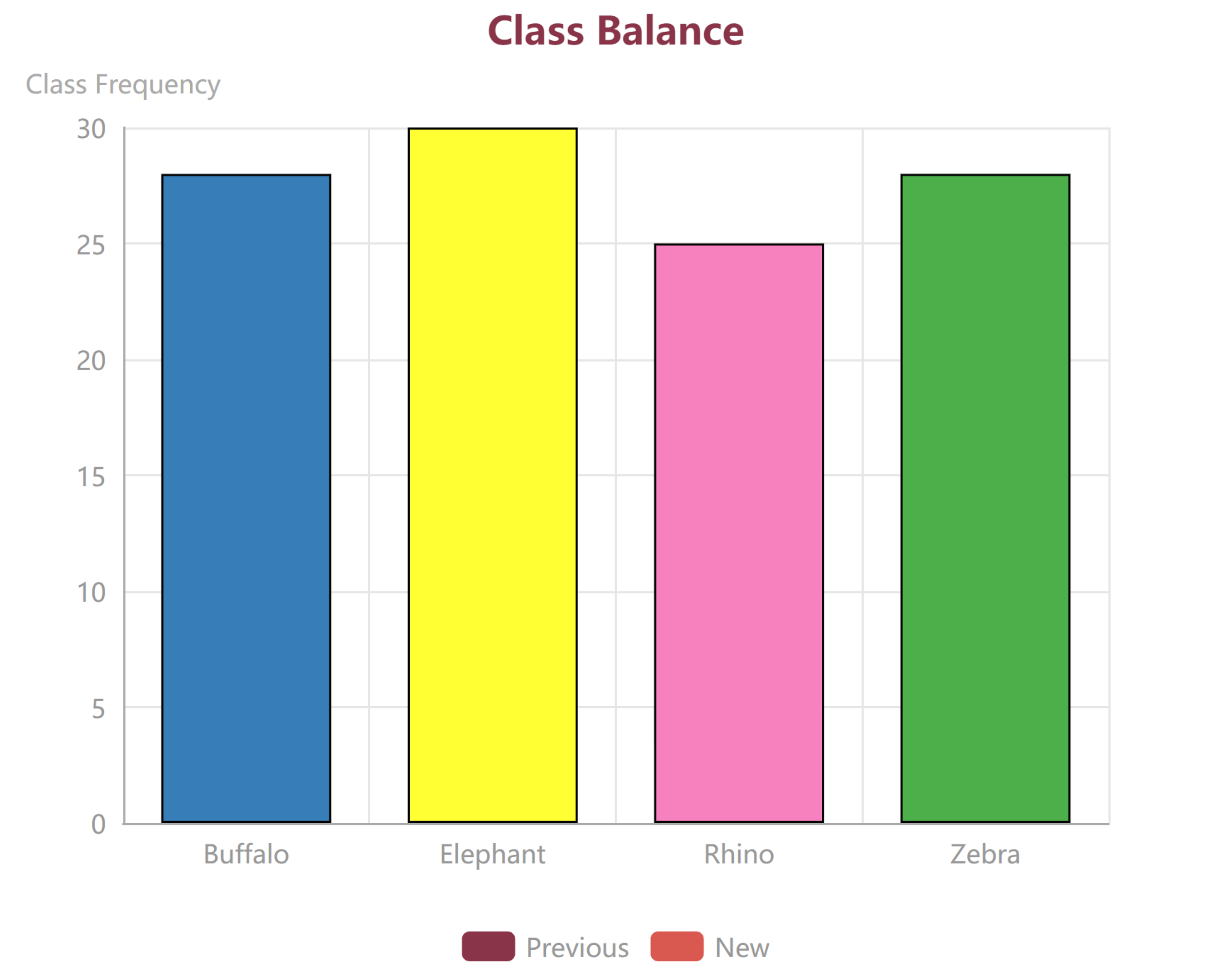}
  \end{minipage}
  \caption{Model View in the start of training iteration 2.}
  \label{fig:iteration2-comparisons1i2}
\end{figure}

Following the first iteration of exploration-driven labeling, the Data View \ref{c2_iteration2_dv} now highlights the initially selected structural regions with their respective class colors. This visual confirmation of labeled areas allows the subsequent exploration to focus on new, still unlabeled regions of the scatterplot. For example, the smaller cluster selected in the fourth quadrant now has three colored points. Namely, two samples of buffalos and one sample of elephants. The predictions confidence distribution is updated with predictions made by the newly trained model on the unlabeled training set, and finally, the class balance is updated to reflect the current number of class instances in the labeled training set, as seen in Figure \ref{fig:iteration2-comparisons1i2}. \\

\noindent The median confidence of the predictions is now above 0.7 for all the classes. However, there is a large spread of prediction confidences within the class distributions, except for the Buffalo class, which seems to perform well already. The class balance (Figure \ref{fig:iteration2-comparisons1i2}) is, as intended, more even than midway through Iteration 1, with classes now mostly represented equally in the labeled training set. An attempt will still be made to maintain a decent class balance however, due to the observation that the model is already quite confident at predicting buffalos in the unlabeled training set, this could indicate that the model has learned this class faster and might not need as many training samples as the other classes.

\begin{figure}[ht!]
    \centering
    \includegraphics[width=0.5\textwidth]{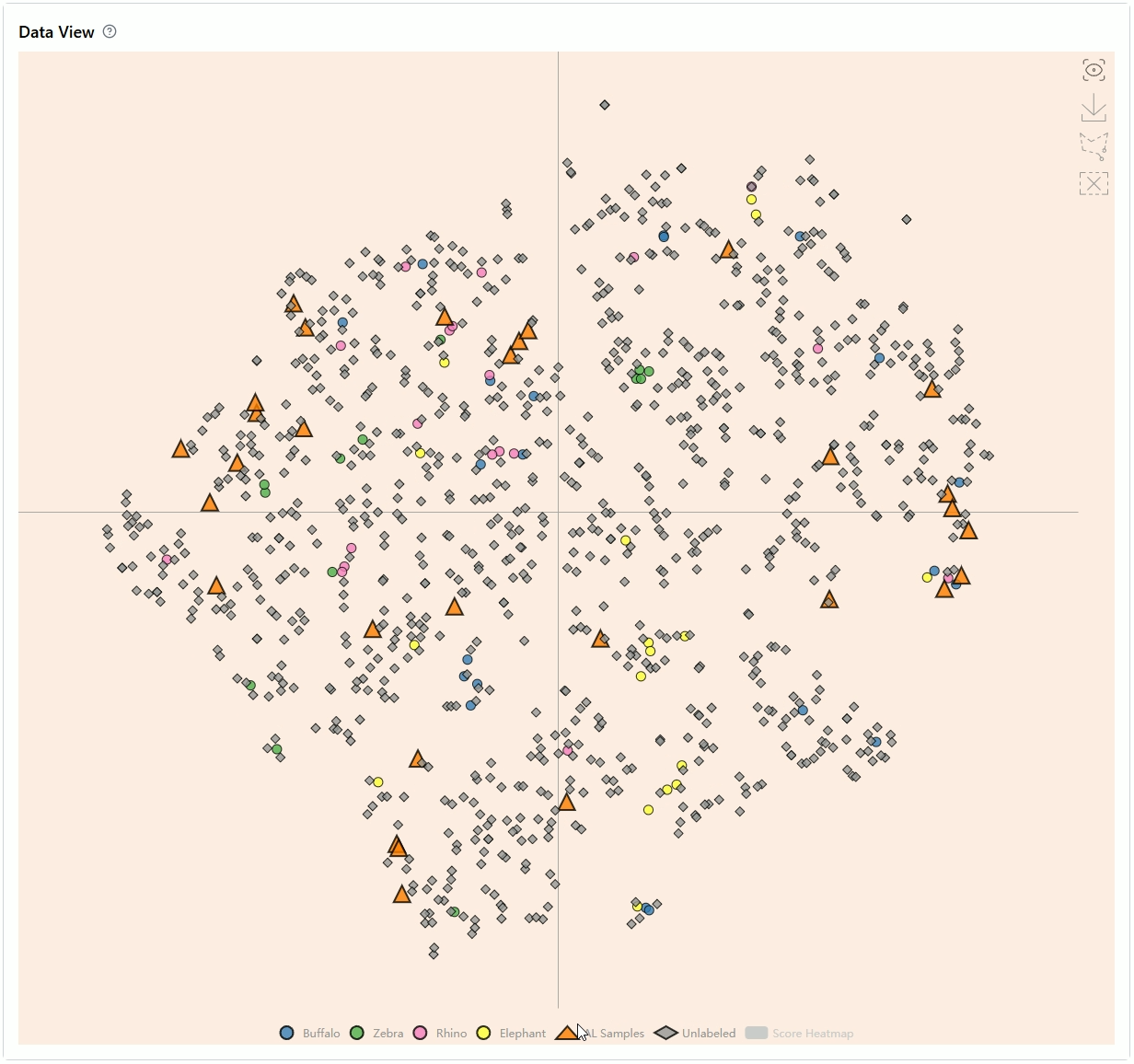}
    \caption{Data View of Iteration 2 with disabled heatmap.}
    \label{it2_dataview_no_map}
\end{figure}

To get a clearer view of the data distribution, the score heatmap is disabled by pressing its legend in the Data View, as seen in Figure \ref{it2_dataview_no_map}. This allows for easier distinction of regions that were sampled in the previous iteration, and can help guide towards either finding new regions still not represented in the labeled set, or adding more samples from these regions.

\begin{figure}[ht!]
  \centering
  \begin{minipage}[c]{0.49\textwidth}
    \centering
    \includegraphics[width=\linewidth]{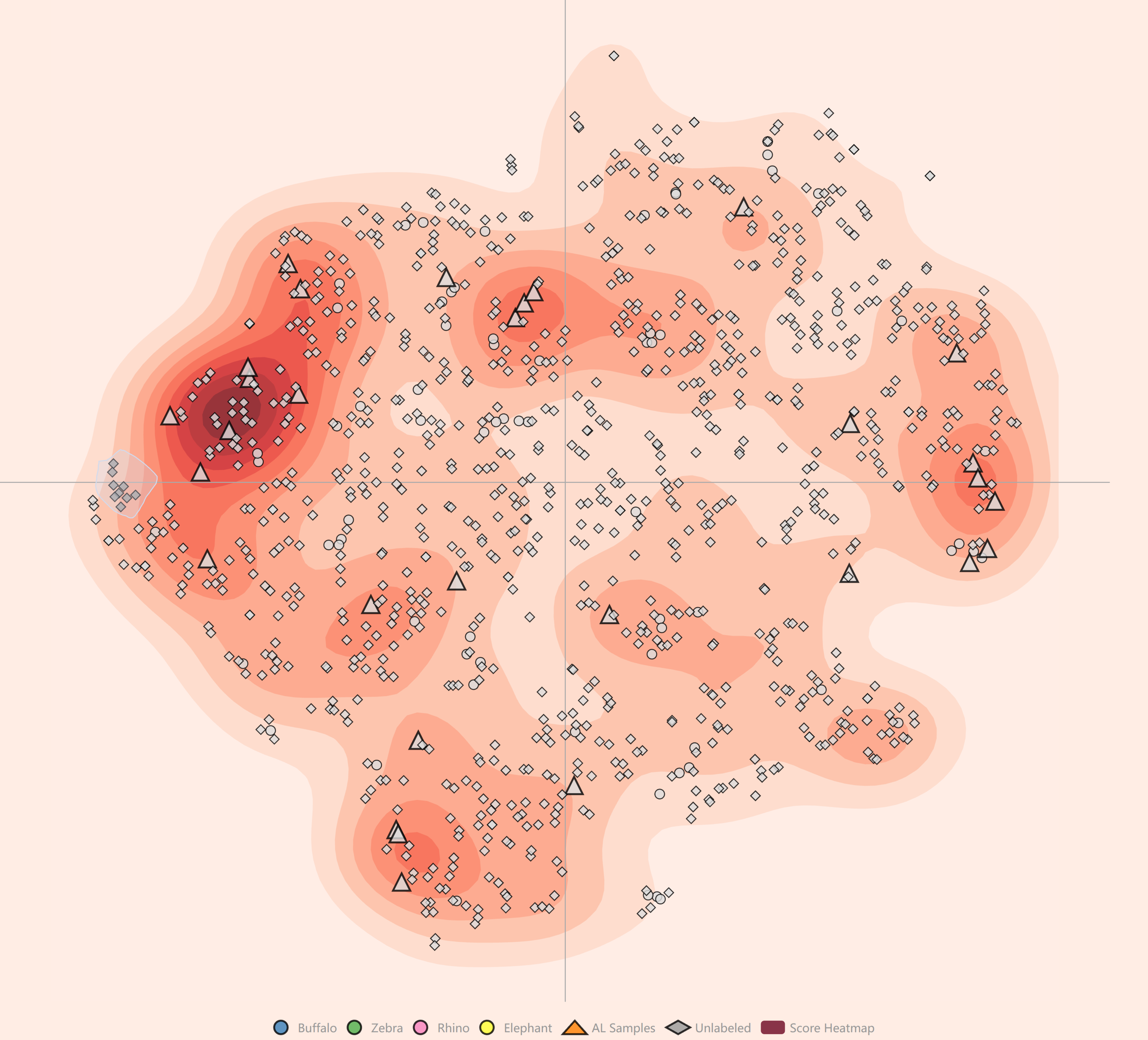}
  \end{minipage}
  \hfill
  \begin{minipage}[c]{0.49\textwidth}
    \centering
    \includegraphics[width=\linewidth]{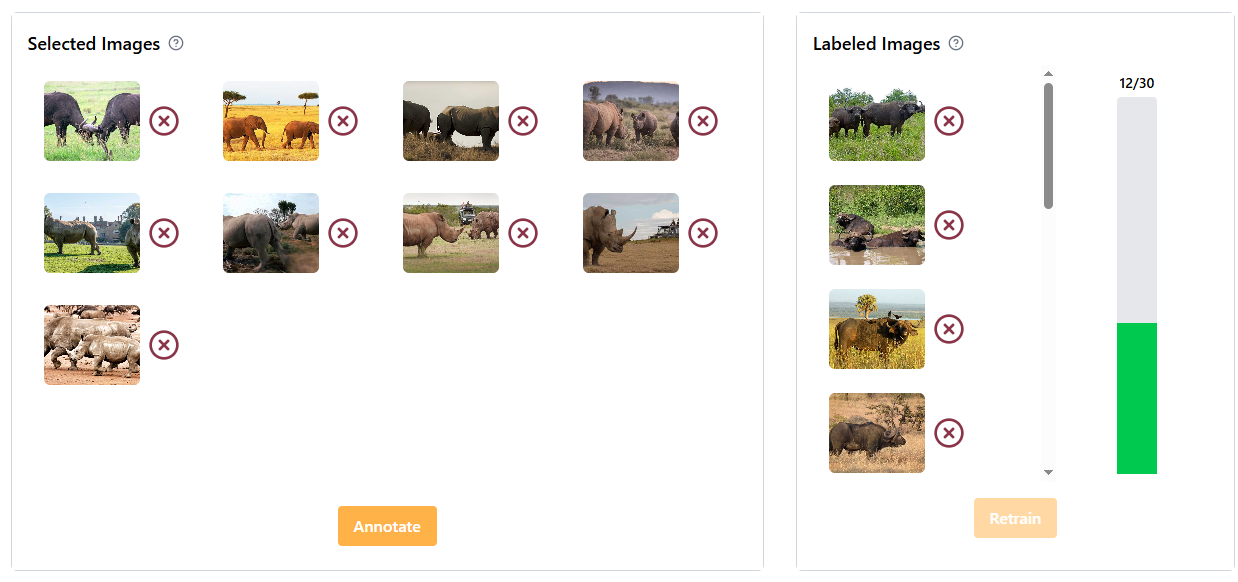}
  \end{minipage}
  \caption{A selection of a smaller region made in iteration 2.}
  \label{c1i2s1}
\end{figure}

After disabling the heatmap to get a clearer view of the raw data distribution, a small, somewhat isolated cluster of points was identified on the far negative x-axis (\ref{c1i2s1}). This region is selected as it represents a previously unsampled area of the feature space, potentially holding new visual characteristics. Four of these samples are annotated and added to the labeled dataset.

\begin{figure}[ht!]
  \centering
  \begin{minipage}[c]{0.3\textwidth}
    \centering
    \includegraphics[width=\linewidth]{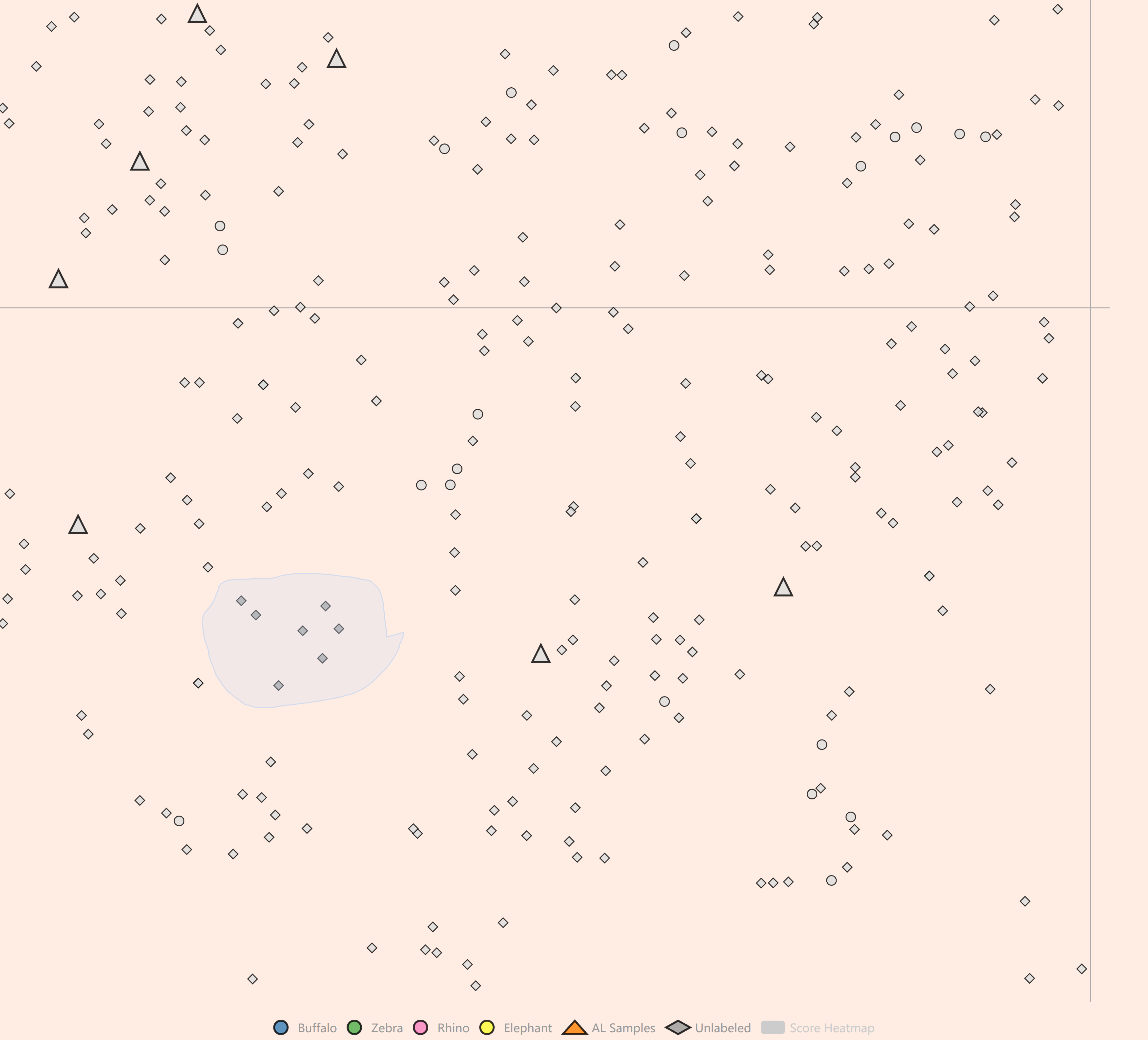}
  \end{minipage}
  \begin{minipage}[c]{0.3\textwidth}
    \centering
    \includegraphics[width=\linewidth]{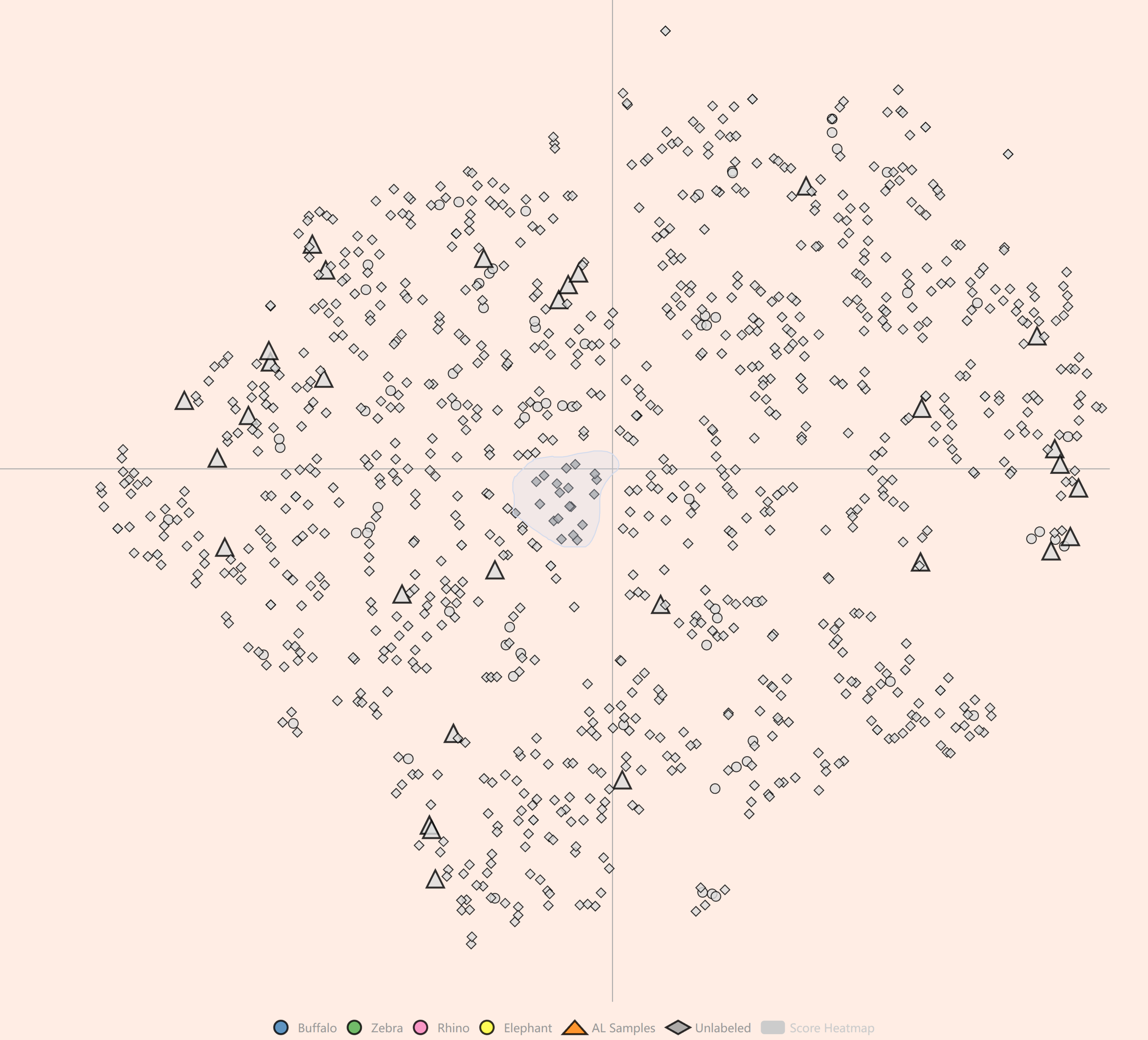}
  \end{minipage}
  \vspace{0.2em} \\
  \begin{minipage}[c]{0.3\textwidth}
    \centering
    \includegraphics[width=\linewidth]{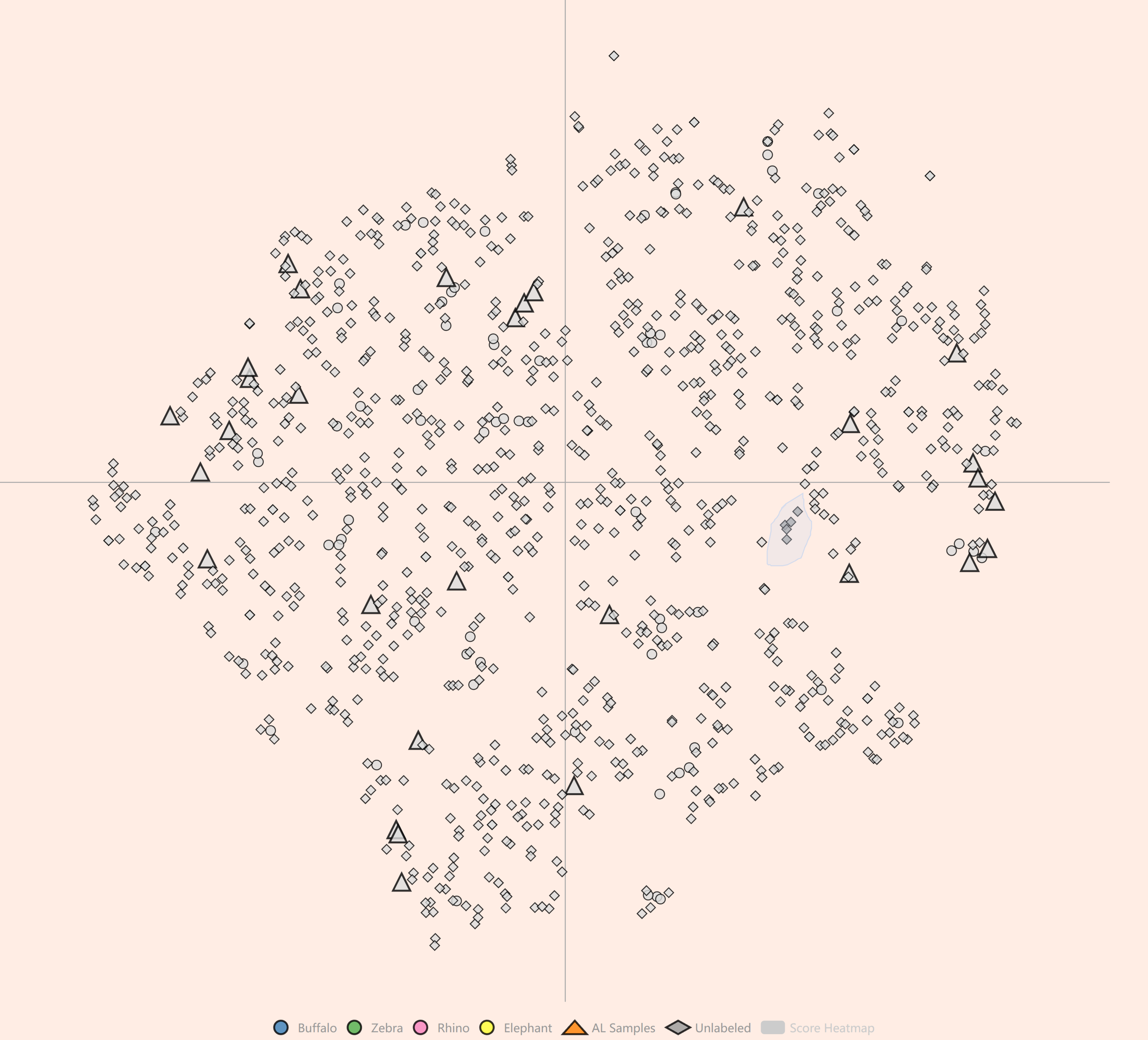}
  \end{minipage}
  \begin{minipage}[c]{0.3\textwidth}
    \centering
    \includegraphics[width=\linewidth]{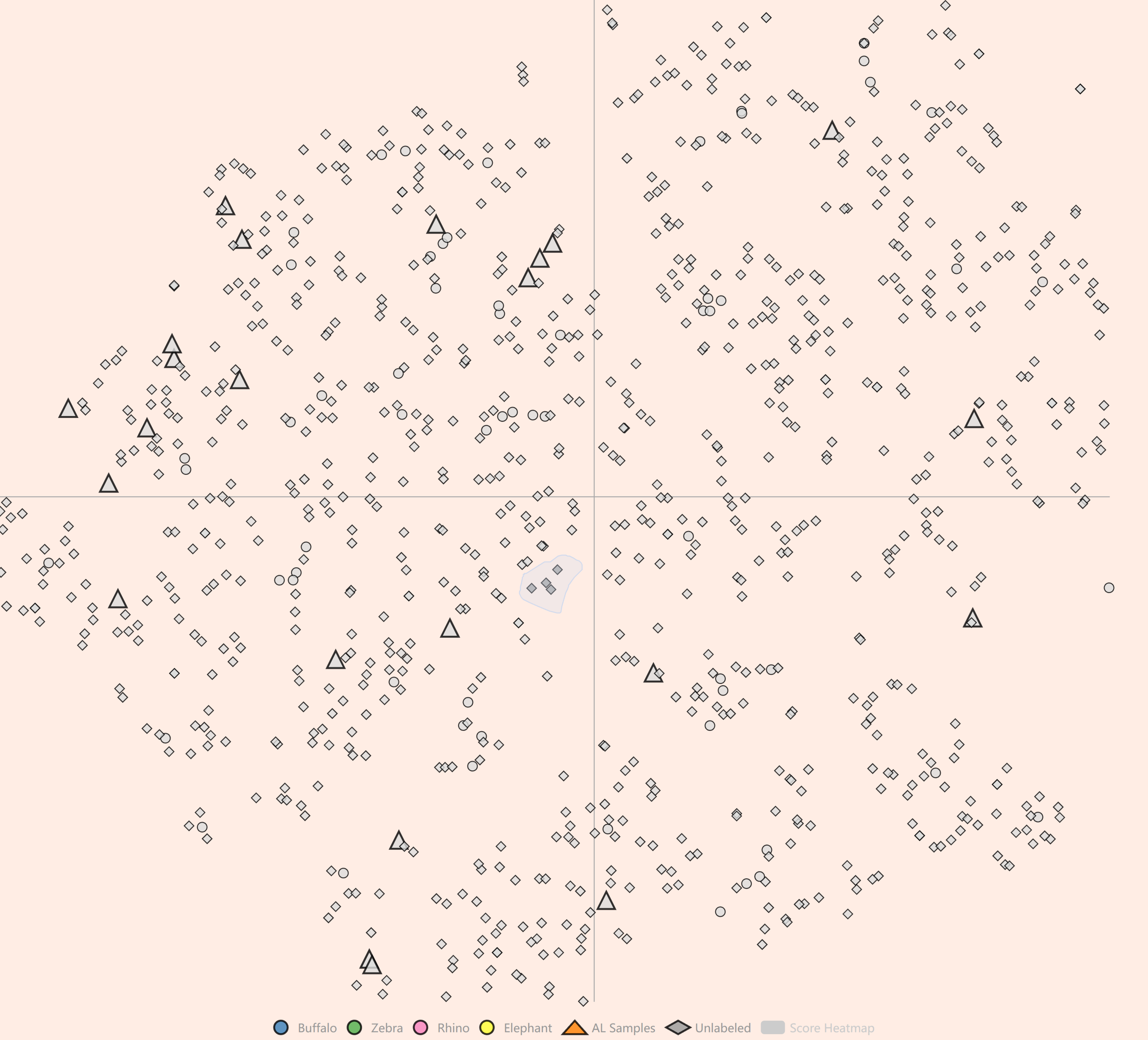}
  \end{minipage}
  \caption{Four selections made in iteration 2. Targeting areas with no previously labeled samples.}
  \label{iteration_2 selection}
\end{figure}

The exploration continues throughout the rest of iteration 2. Four additional selections, each targeting distinct areas with no previously labeled instances or very sparse labeling, can be seen in Figure \ref{iteration_2 selection}. This demonstrates the ongoing effort to broaden the labeled sets coverage of the feature space

\subsubsection{Iteration 3}
\begin{figure}[ht!]
    \centering
    \includegraphics[width=0.4\textwidth]{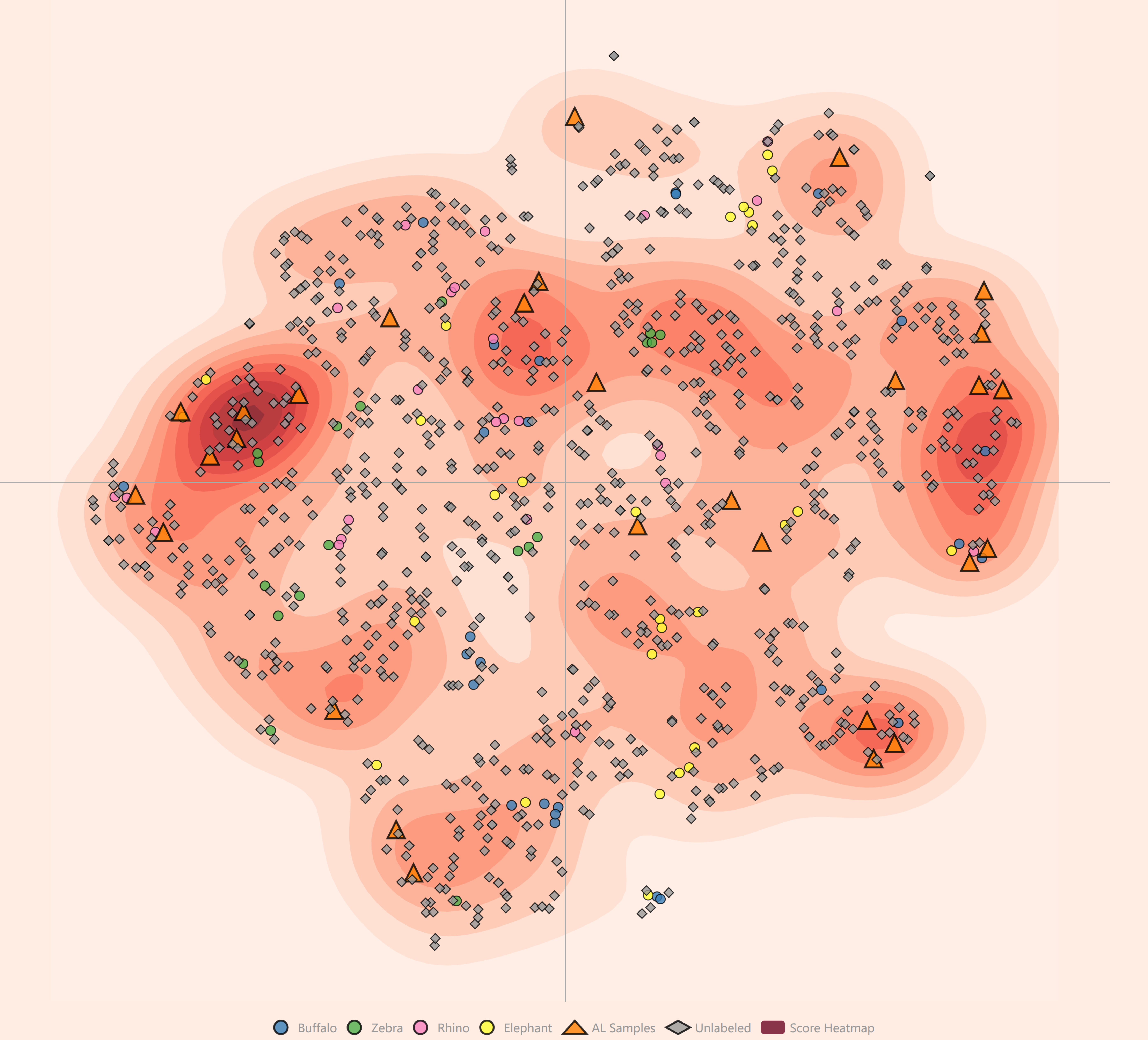}
    \caption{Data View in the start of iteration 3.}
    \label{it3_dataview}
\end{figure}

\begin{figure}[ht!]
  \centering
  \begin{minipage}[c]{0.4\textwidth}
    \centering
    \includegraphics[width=\linewidth]{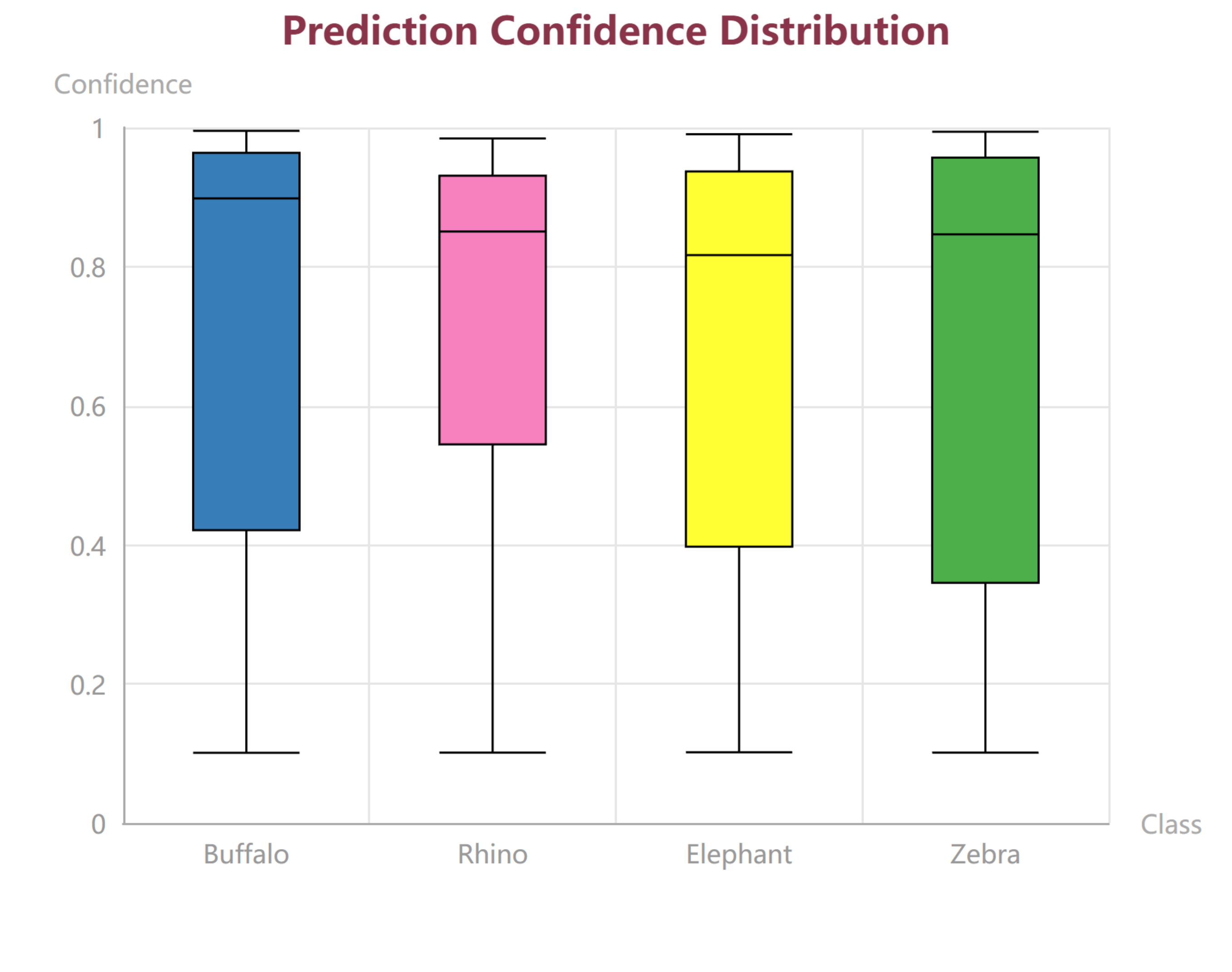}
  \end{minipage}
  \begin{minipage}[c]{0.4\textwidth}
    \centering
    \includegraphics[width=\linewidth]{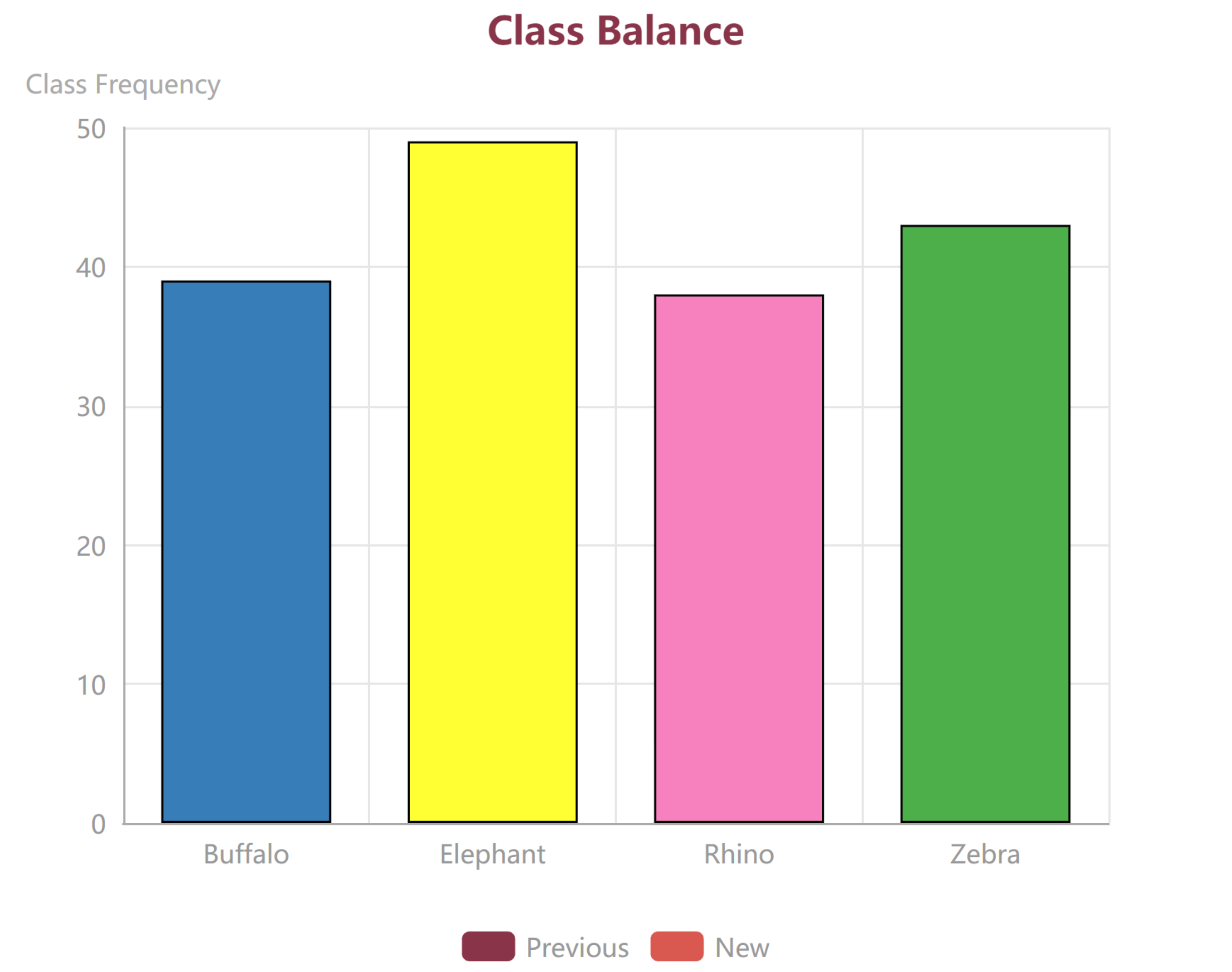}
  \end{minipage}
  \caption{Model View at the start of training iteration 3.}
  \label{fig:iteration3-comparison}
\end{figure}

In Figure \ref{it3_dataview}, the updated Data View for iteration 3 can be viewed. In this use case, the updated view is mainly interesting to examine to get an overview of where previous samples have been labeled, since the structure of the points does not change over iterations. More colored points are starting to appear all over the scatterplot, which has been the target so far. It is becoming harder to identify entirely unlabeled regions. The strategy will remain the same for this iteration, but if it is considered that a diverse space of the data has been covered, expansion of the labeled regions by labeling neighboring points can also begin. There is still a decently balanced representation of the classes in the labeled training set, but elephants are slightly more represented than the other classes, as seen in Figure \ref{fig:iteration3-comparison}. Notable in the Prediction Confidence Distribution is that the Buffalo class is now having a wider distribution again, where most of the predictions lie between 0.4-0.95 confidence, in comparison to the previous iteration, which had a generally higher confidence distribution.

\begin{figure}[ht!]
  \centering
  \begin{minipage}[c]{0.3\textwidth}
    \centering
    \includegraphics[width=\linewidth]{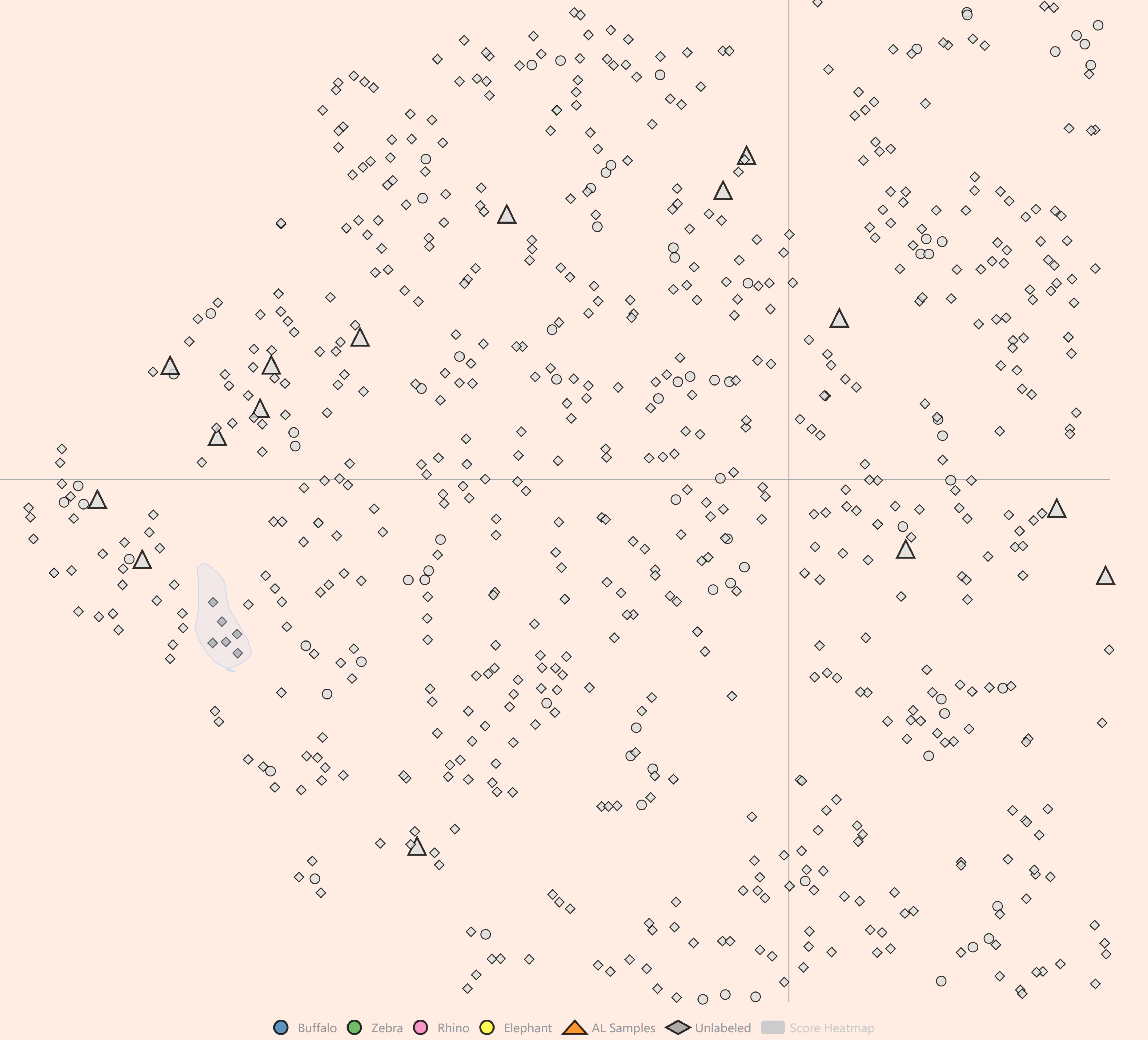}
  \end{minipage}
  \begin{minipage}[c]{0.3\textwidth}
    \centering
    \includegraphics[width=\linewidth]{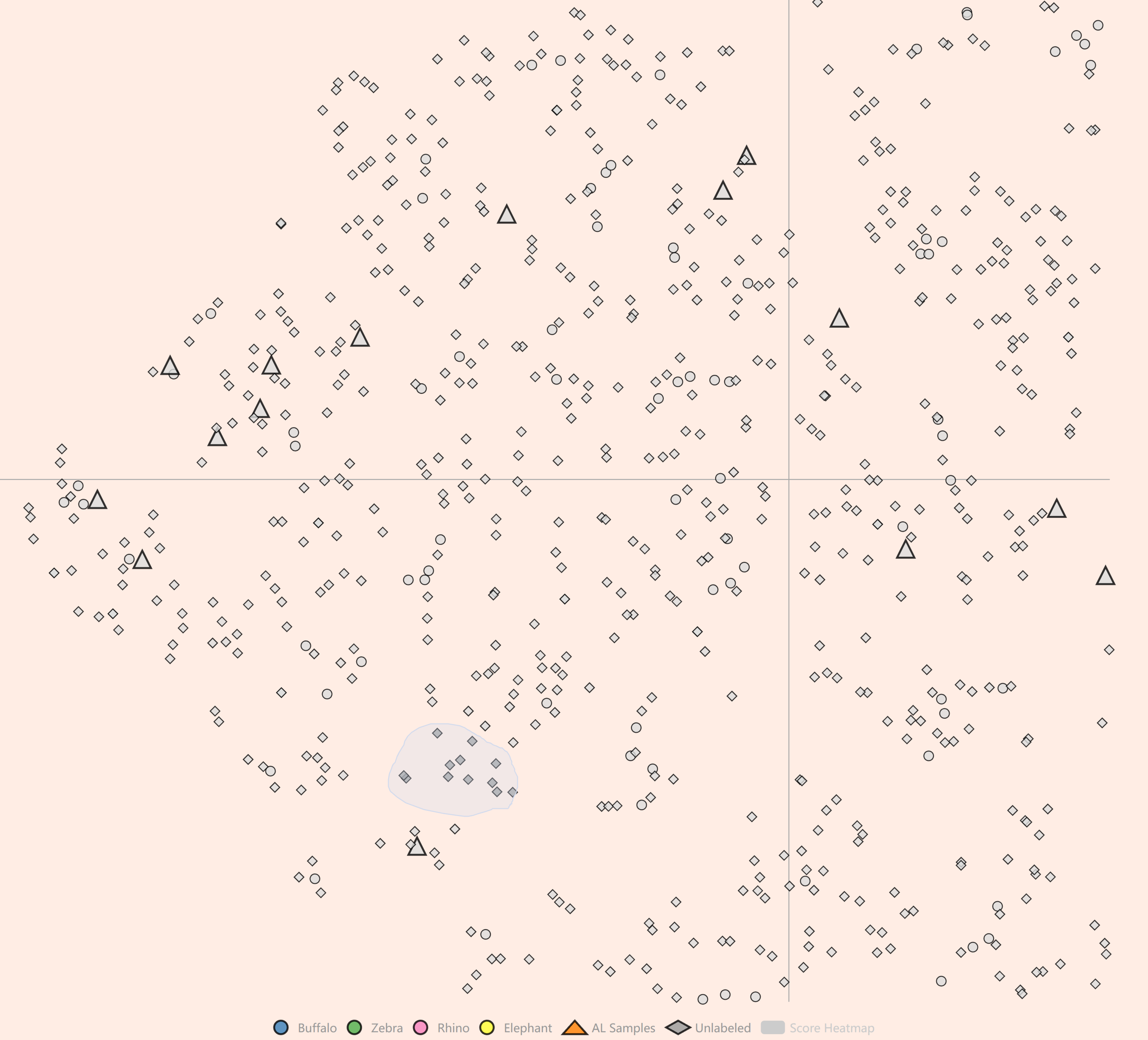}
  \end{minipage}
  \vspace{0.2em} \\
  \begin{minipage}[c]{0.3\textwidth}
    \centering
    \includegraphics[width=\linewidth]{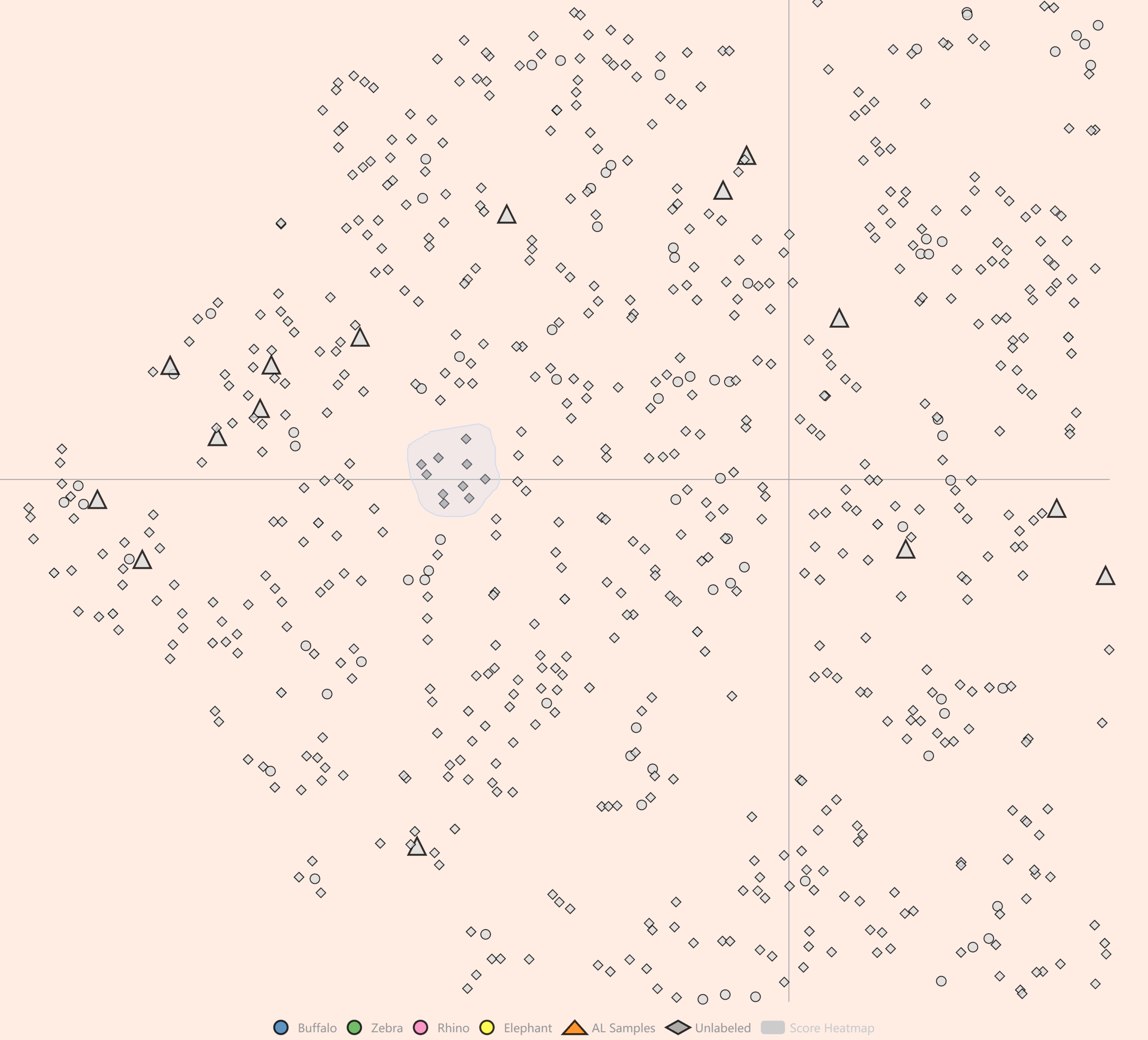}
  \end{minipage}
  \begin{minipage}[c]{0.3\textwidth}
    \centering
    \includegraphics[width=\linewidth]{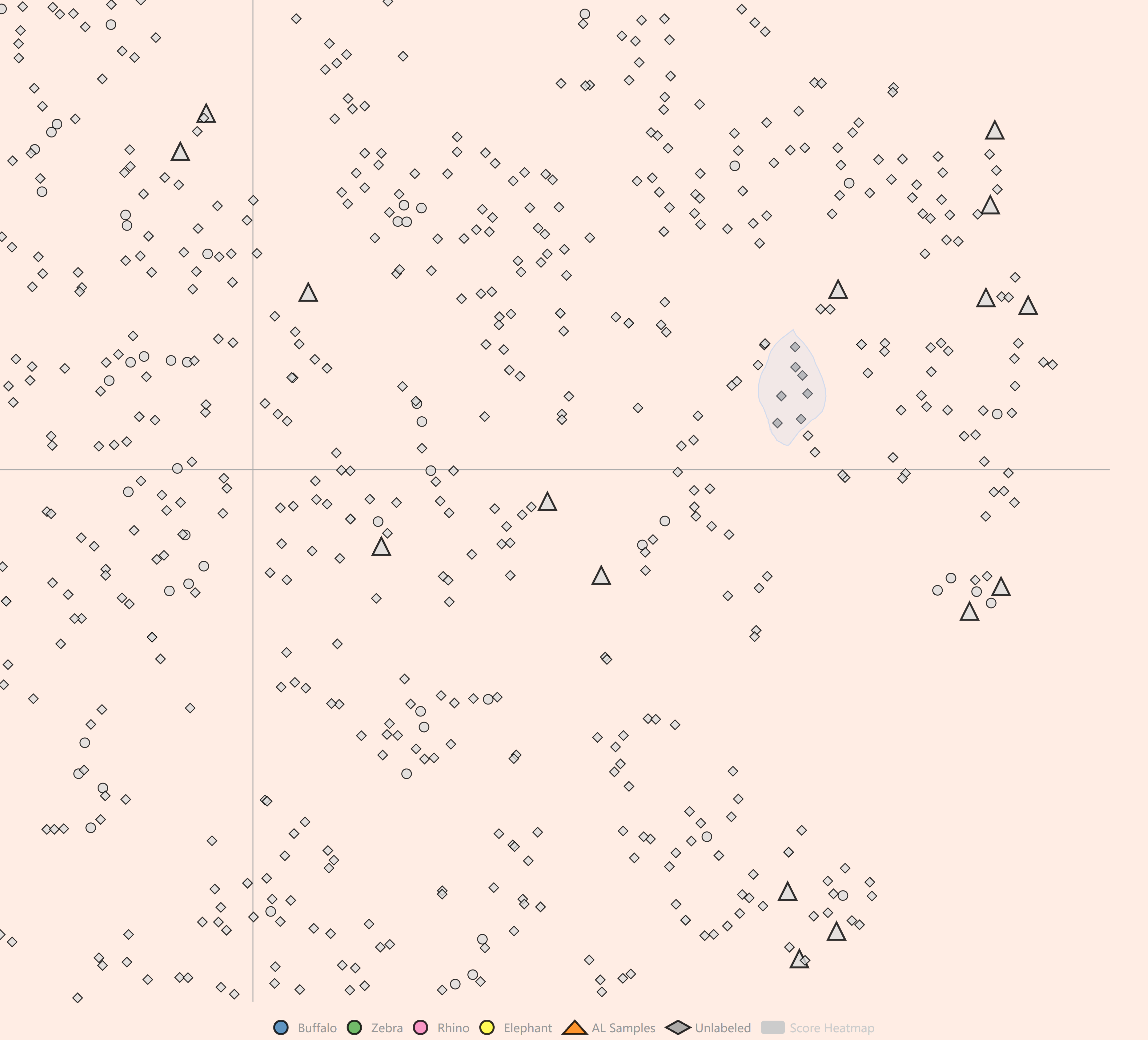}
  \end{minipage}
  \caption{Four selections made in iteration 2. Targeting areas with no previously labeled samples.}
  \label{c1it3s}
\end{figure}

\newpage
A sample of four different selections made in iteration 3, continuing to target previously unsampled or sparsely sampled structural areas, is presented in \ref{c1it3s}.

\subsubsection{Iteration 4}
\begin{figure}[ht!]
    \centering
    \includegraphics[width=0.4\textwidth]{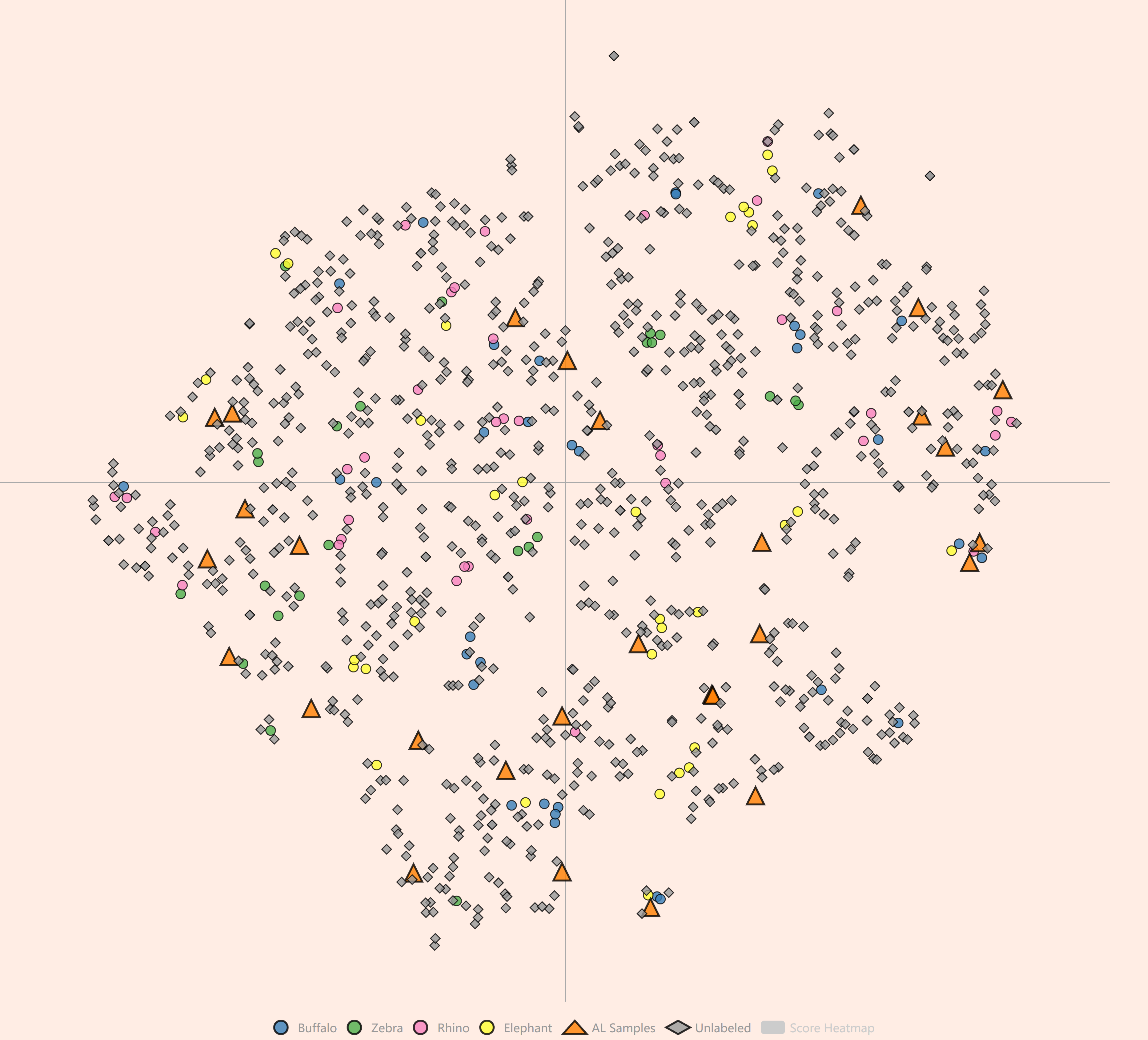}
    \caption{Data View at the start of iteration 4, without heatmap}
    \label{it4c1_dataview}
\end{figure}

Beginning the fourth iteration, 60 samples remain to be labeled to reach the total budget of 150. The Data View  \ref{it4c1_dataview} is consulted again. The feature space, as represented by the t-SNE plot, appears largely covered at this point. Instances of all four classes are represented within the main "blobs" identified in Iteration 1.

\begin{figure}[ht!]
  \centering
  \begin{minipage}[c]{0.4\textwidth}
    \centering
    \includegraphics[width=\linewidth]{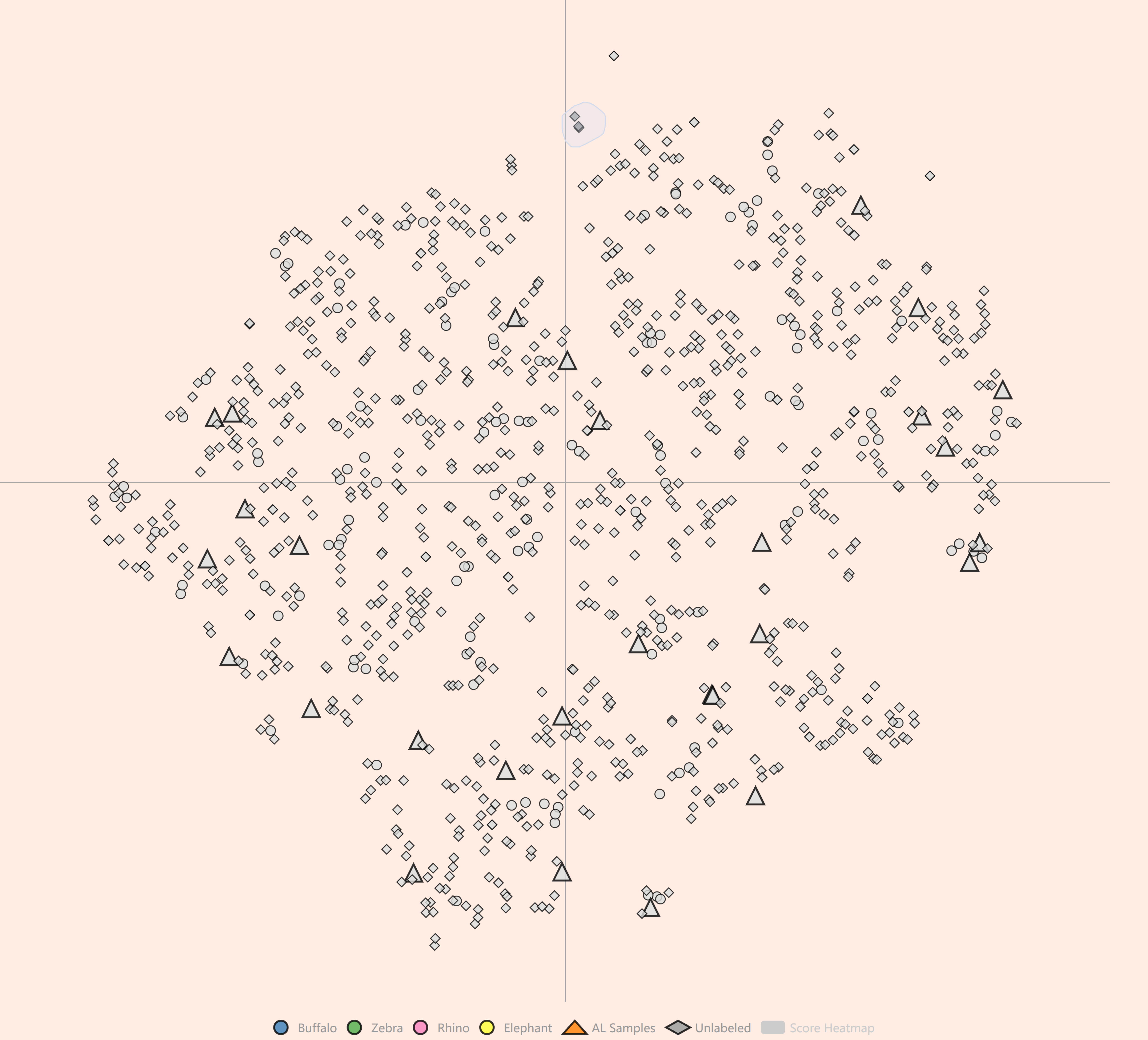}
  \end{minipage}
  \begin{minipage}[c]{0.4\textwidth}
    \centering
    \includegraphics[width=\linewidth]{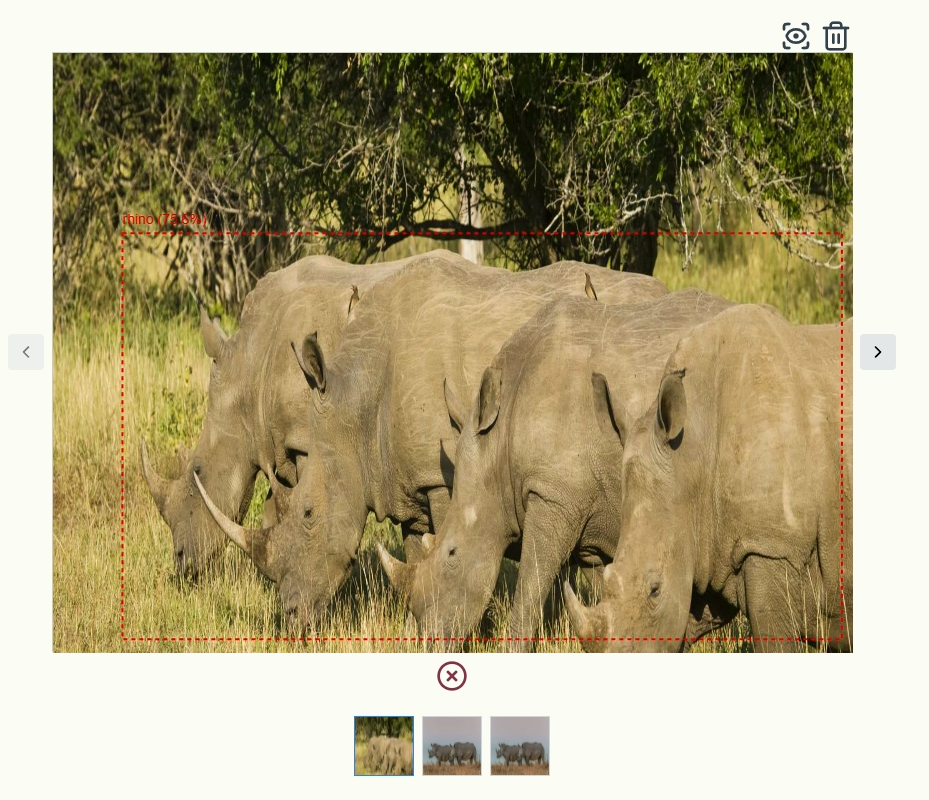}
  \end{minipage}
  \caption{A selection of three points closely coupled together but separated from the rest of the points}
  \label{edge_points}
\end{figure}

What remains unlabeled are largely more subtle structural features or what might be considered edge cases. For example, a few very small, tightly coupled clusters of points that were previously overlooked or deemed lower priority than larger structures. An example of such a selection is presented in Figure \ref{edge_points}. The nature of these points is very similar. They depict rhinos stacked together as seen from the side. Two of the samples even look to be identical, highlighting potential near-duplicates or very similar scenes within the dataset. The model-made prediction on one of the images reveals that it is having trouble discerning the individual rhinos when stacked, as it groups all four rhinos into a single bounding box. This is a valuable qualitative insight stemming from exploring these tight structural clusters. The sampling process throughout iteration 4 continues in this manner, focusing on these smaller, distinct remaining groups and filling the budget with such selections.

\subsubsection{Iteration 5}
\begin{figure}[ht!]
  \centering
  \begin{minipage}[c]{0.4\textwidth}
    \centering
    \includegraphics[width=\linewidth]{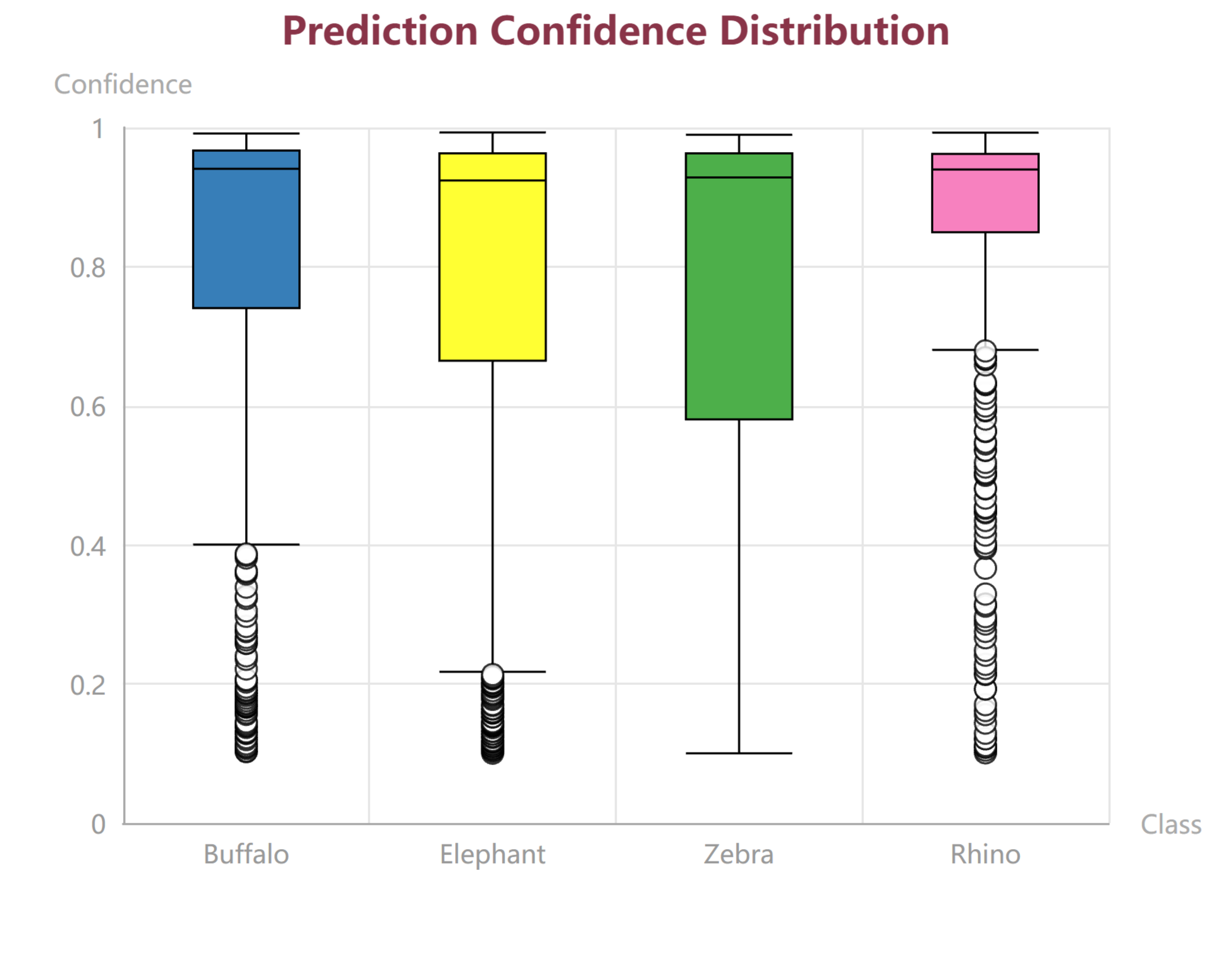}
  \end{minipage}
  \begin{minipage}[c]{0.4\textwidth}
    \centering
    \includegraphics[width=\linewidth]{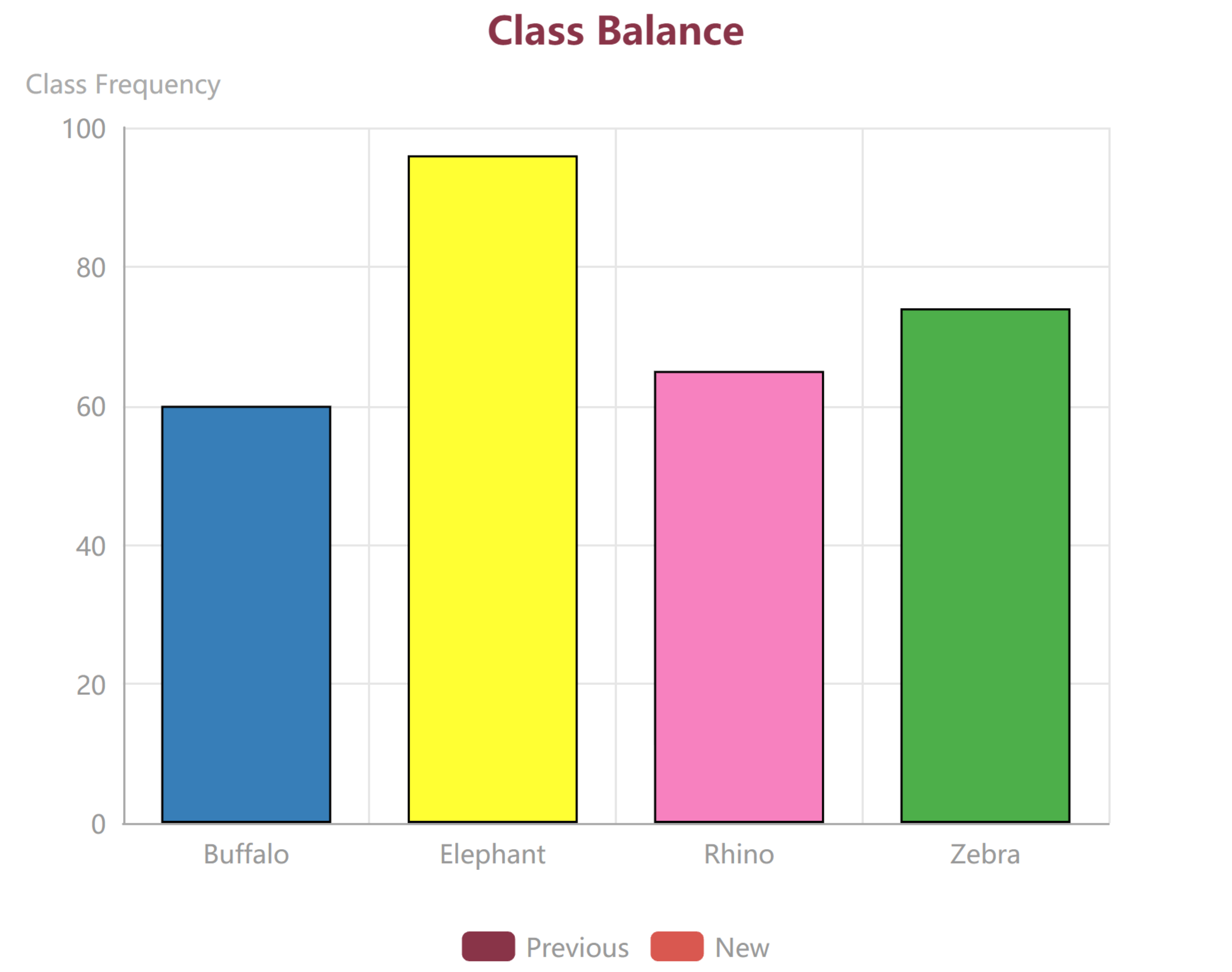}
  \end{minipage}
  \caption{Model View at the start of training iteration 5.}
  \label{fig:iteration5-comparison}
\end{figure}

This iteration will conclude the "Exploration \& Structure Focus" use case. Thirty samples are left to annotate. The Model View \ref{fig:iteration5-comparison} reveals an overall good Prediction Confidence Distribution. The median confidence for all classes is now above 0.9. The defined sampling will be maintained for this iteration. There are still a few very small, isolated regions of unlabeled datapoints, often on the periphery of the main distribution. These are prioritized first. The rest of the budget will then be spent on carefully selecting neighboring points of already labeled samples in sparser, previously under-sampled regions, to ensure even these less prominent areas of the feature space are represented.

\begin{figure}[ht!]
  \centering
  \begin{minipage}[c]{0.3\textwidth}
    \centering
    \includegraphics[width=\linewidth]{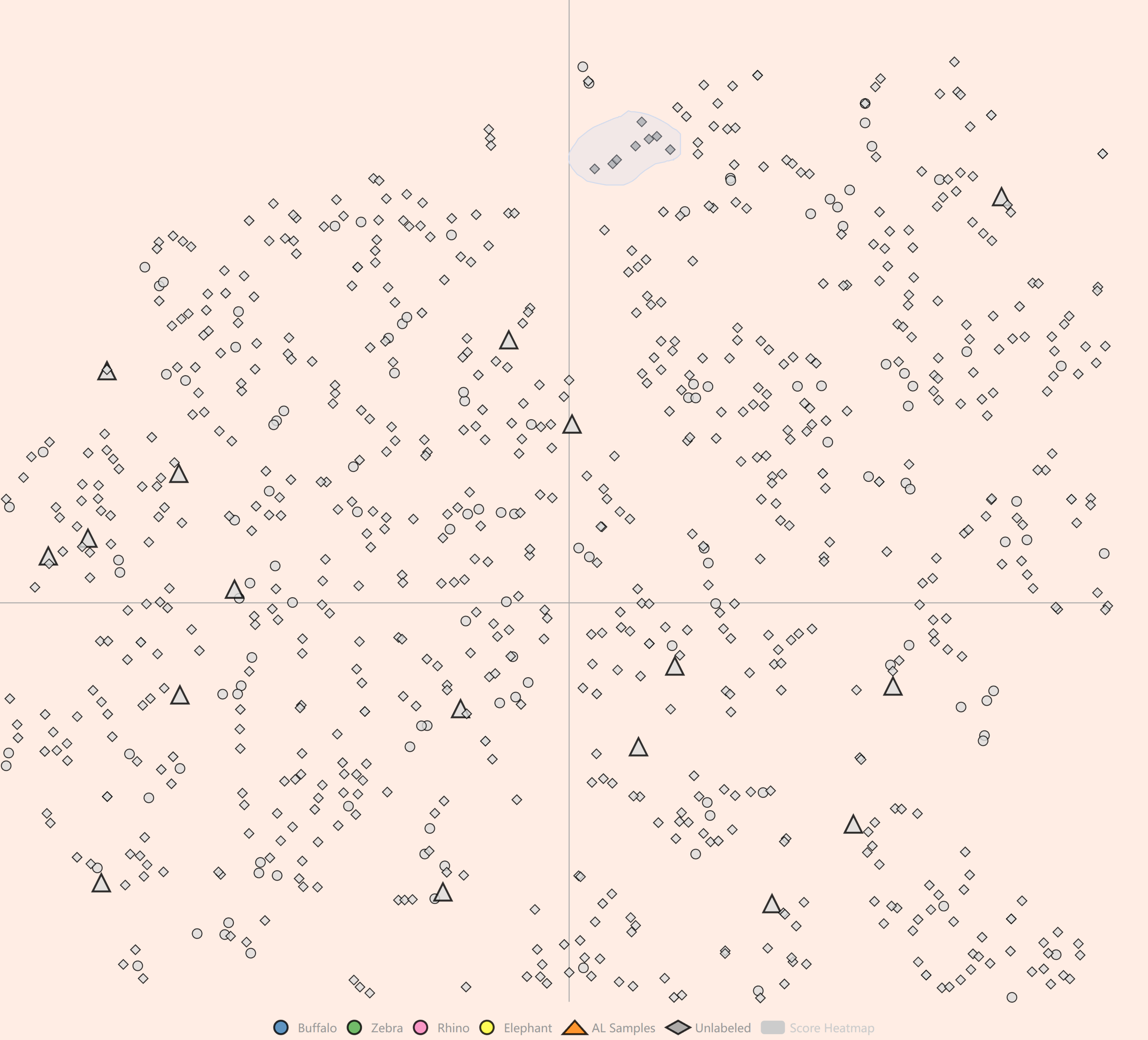}
  \end{minipage}
  \begin{minipage}[c]{0.3\textwidth}
    \centering
    \includegraphics[width=\linewidth]{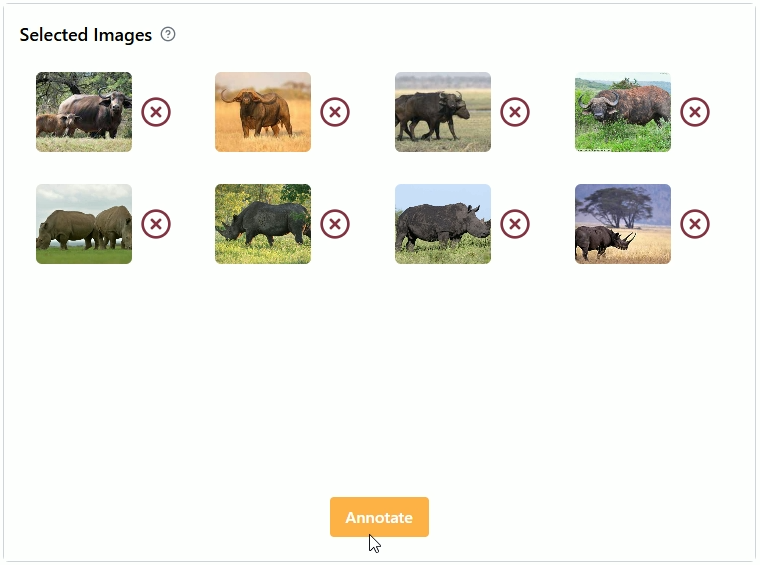}
  \end{minipage}
  \vspace{0.2em} \\
  \begin{minipage}[c]{0.3\textwidth}
    \centering
    \includegraphics[width=\linewidth]{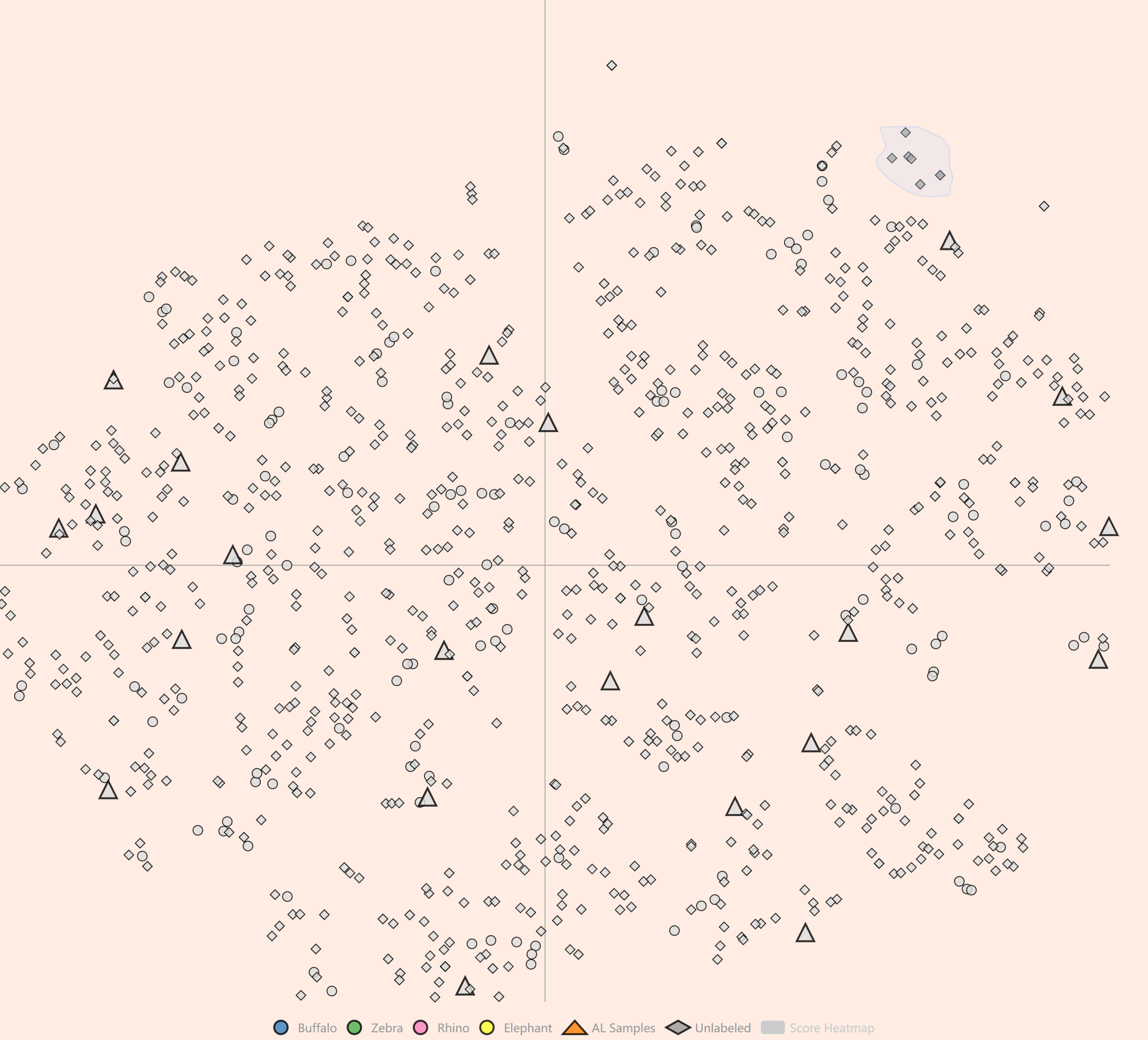}
  \end{minipage}
  \begin{minipage}[c]{0.3\textwidth}
    \centering
    \includegraphics[width=\linewidth]{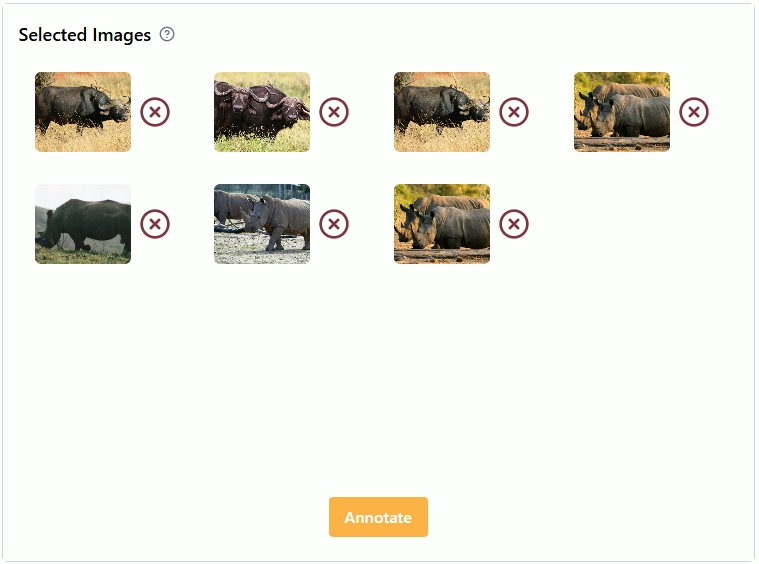}
  \end{minipage}
  \caption{Two selections made in iteration 5. Targeting areas with no previously labeled samples.}
  \label{c1it3s_two_samples}
\end{figure}

Figure \ref{c1it3s_two_samples} displays an example of two such prioritized regions, targeting some of the final remaining small pockets of unlabeled data points. These are annotated, and the final model training for this use case is initiated. 

This exploration-driven approach, by consistently prioritizing broad structural coverage first and also considering class balance, aimed to build a diverse labeled dataset reflecting the varied nature of the visual feature space. This strategy, facilitated by VILOD's depiction of the data's feature space, allowed for a systematic discovery of the dataset's intrinsic structures, potentially uncovering visual groups or outlier characteristics independently of the model's biases. A full list of annotated samples from this use case is provided in Appendix \ref{appendixA}. The quantitative results and a comparative analysis against other strategies will be presented in Section \ref{analysis}.

\subsection{Use Case 2: Uncertainty-Driven Focus}
\label{use_case_uncertainty}

The second use case shifted the sample selection strategy to be primarily guided by model uncertainty, as presented within the VILOD interface. The main objective here was to investigate the impact of closely following the signals provided by the uncertainty heatmap overlay on the t-SNE Data View and the explicit AL sample suggestions, also showcasing how the VILOD tool can facilitate a more algorithmic-like approach if the user chooses to. Selections were predominantly made from the most intense (darkest red) regions of the uncertainty heatmap and included a significant proportion of the AL-suggested images, which are highlighted in the Data View. The human's role in this use case is largely to adhere to the uncertainty metrics, with intervention limited to basic quality control, such as filtering out any samples that might be entirely unusable. Less emphasis was placed on deliberate exploration of the feature space or proactive class balancing beyond what naturally emerged from the uncertainty-driven sampling process. As with the previous use case, this scenario commenced with the identical initial model and involved five iterations, adding 30 newly annotated images to the training set in each iteration.

\subsubsection{Iteration 1}
\begin{figure}[ht!]
  \centering
  \begin{minipage}[c]{0.49\textwidth}
    \centering
    \includegraphics[width=\linewidth]{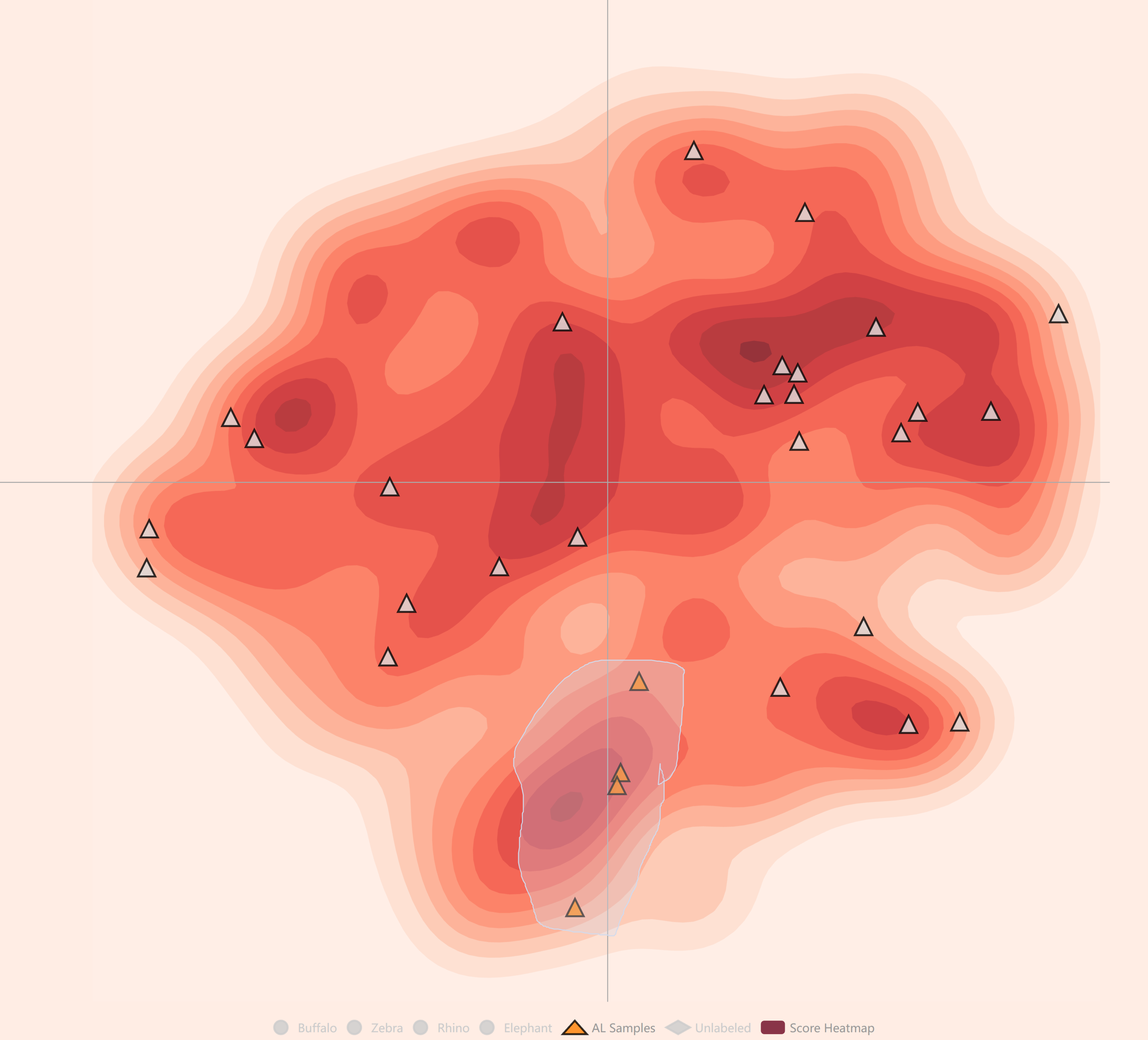}
  \end{minipage}
  \hfill
  \begin{minipage}[c]{0.49\textwidth}
    \centering
    \includegraphics[width=\linewidth]{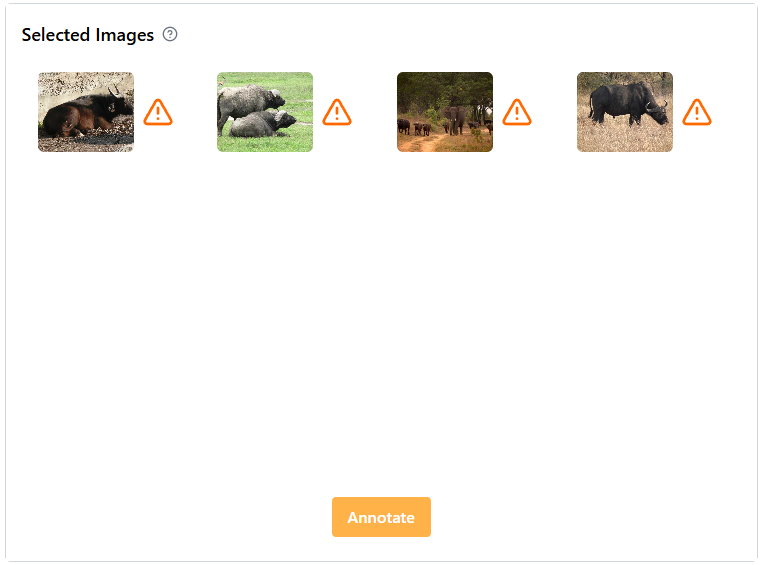}
  \end{minipage}
  \caption{Isolated AL samples and Uncertainty heatmap and the first sample selection of iteration 1.}
  \label{c2it1_AL}
\end{figure}

This use case also commenced with the identical initial model $M_0$. Therefore, the foundational Data View \ref{c3it1_dataview} and Model View \ref{c3it1_modelview}, as depicted in Section \ref{use_case_stuctural_Iteration_1}, represent the starting visual state before any strategy-specific interactions in this iteration. To begin the labeling process for the uncertainty-driven sample selection, the first action is to adapt the Data View by isolating the AL samples and the uncertainty heatmap. This is achieved by turning off the other points by pressing their respective legends in the interface. The lasso tool can then be utilized to start inspecting these samples, an example of which is shown in Figure \ref{c2it1_AL}. The quality check is performed by progressively moving the lasso selection and visually inspecting the samples suggested by the AL algorithm. Samples considered acceptable in quality are annotated; samples that have very poor resolution or very obscured objects are not annotated. In the first iteration, all the AL samples pass this basic quality check.

\subsubsection{Iteration 2}
\begin{figure}[ht!]
    \centering
    \includegraphics[width=0.5\textwidth]{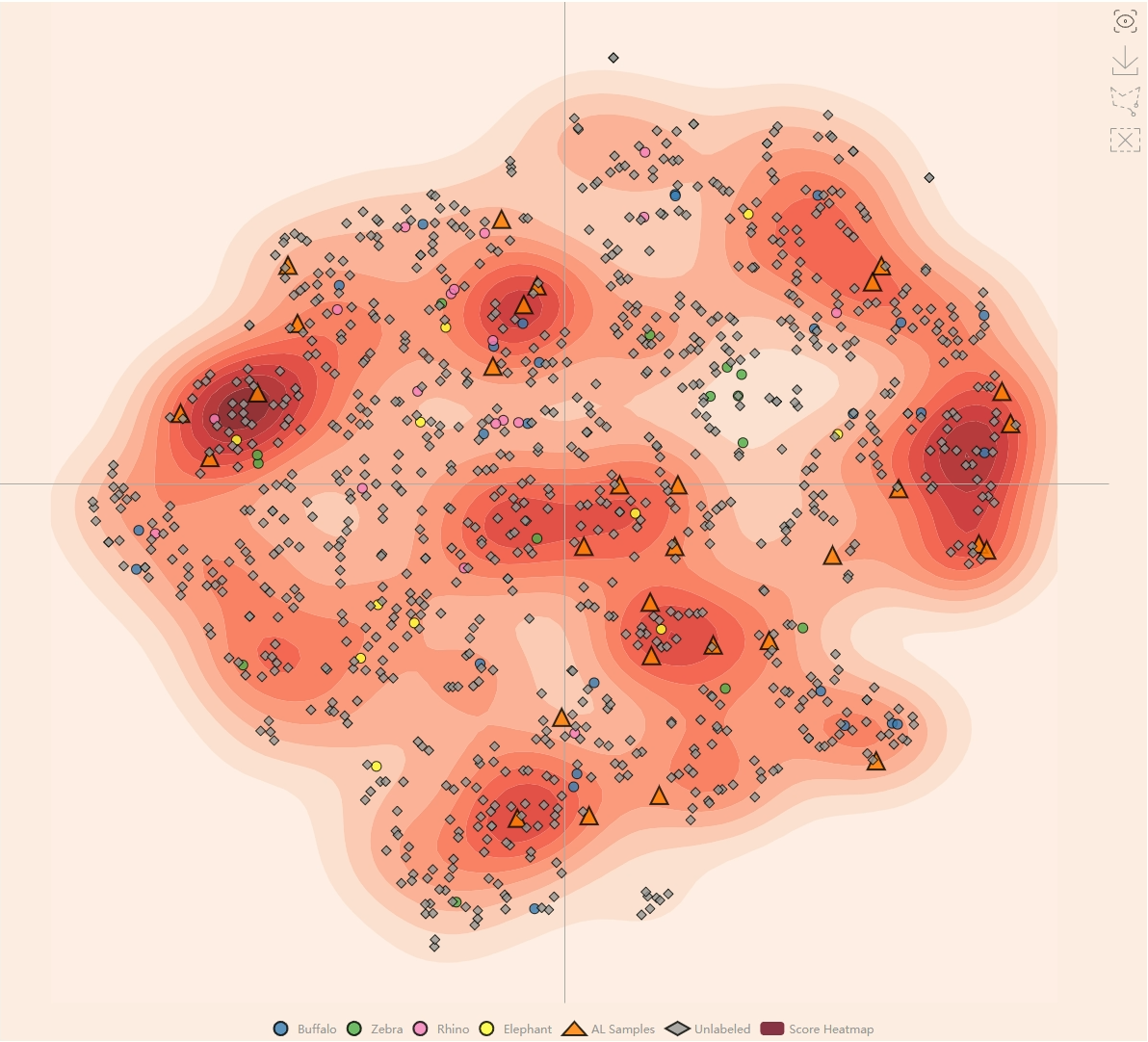}
    \caption{Data View iteration 2.}
    \label{it2c2_dataview}
\end{figure}

\begin{figure}[ht!]
  \centering
  \begin{minipage}[c]{0.49\textwidth}
    \centering
    \includegraphics[width=\linewidth]{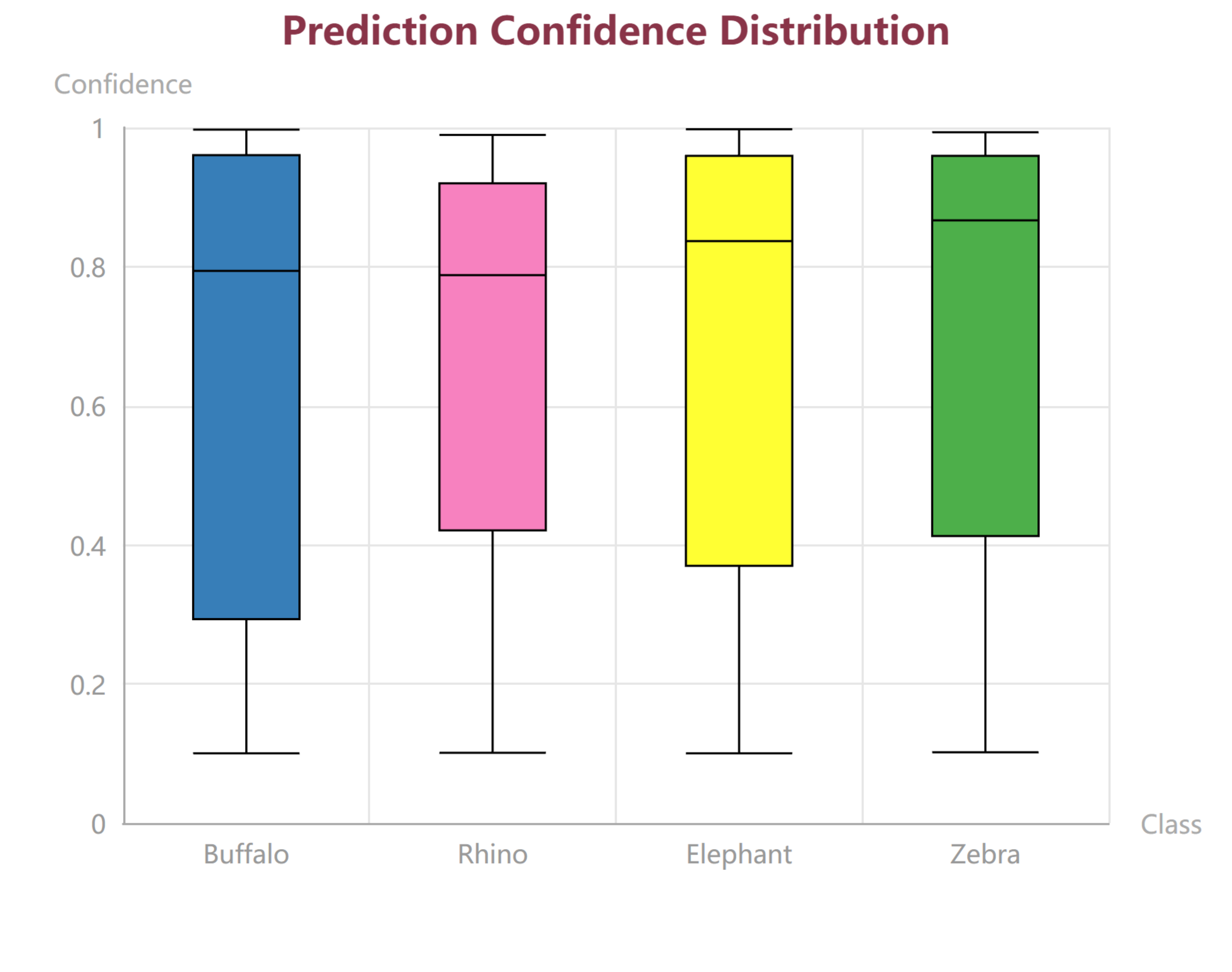}
  \end{minipage}
  \hfill
  \begin{minipage}[c]{0.49\textwidth}
    \centering
    \includegraphics[width=\linewidth]{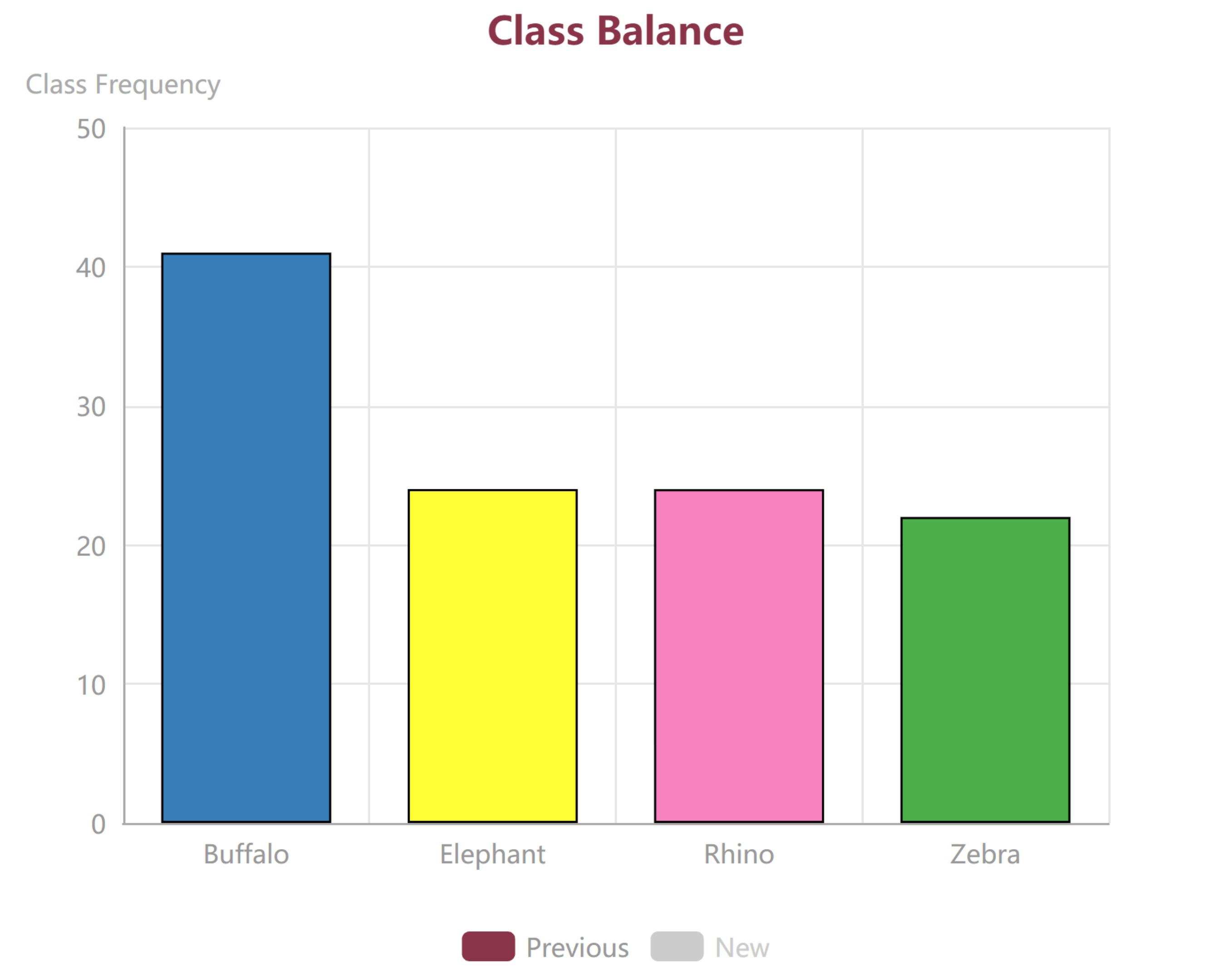}
  \end{minipage}
  \caption{Model View in the start of training iteration 2.}
  \label{fig:c2iteration3-comparison}
\end{figure}

In iteration 2, the uncertainty landscape has shifted slightly (Figure \ref{it2c2_dataview}). The most prominent red regions from iteration 1 are still visible but are now more dense and distinct, with regions of low uncertainty surrounding them. There is a large dark red region towards the far right middle, at the intersection of quadrant one and four; another is towards the far left in the second quadrant, and an additional three to four more regions are spread across the Data View. The AL-suggested samples are mostly clustered around these regions. It is also noted that the AL samples annotated in the previous iteration skewed the class balance (Figure \ref{fig:c2iteration3-comparison}) towards the Buffalo class, which is almost two times more represented than the other three classes. This, however, does not appear to have had any greater impact on the performance of each class. The distribution of the prediction confidences for each class is largely the same, though Buffalo is arguably the worst-performing, with the lowest median and largest spread. Despite this, the strategy of inspecting and annotating mostly AL samples will be maintained.

\begin{figure}[ht!]
    \centering
    \includegraphics[width=0.9\textwidth]{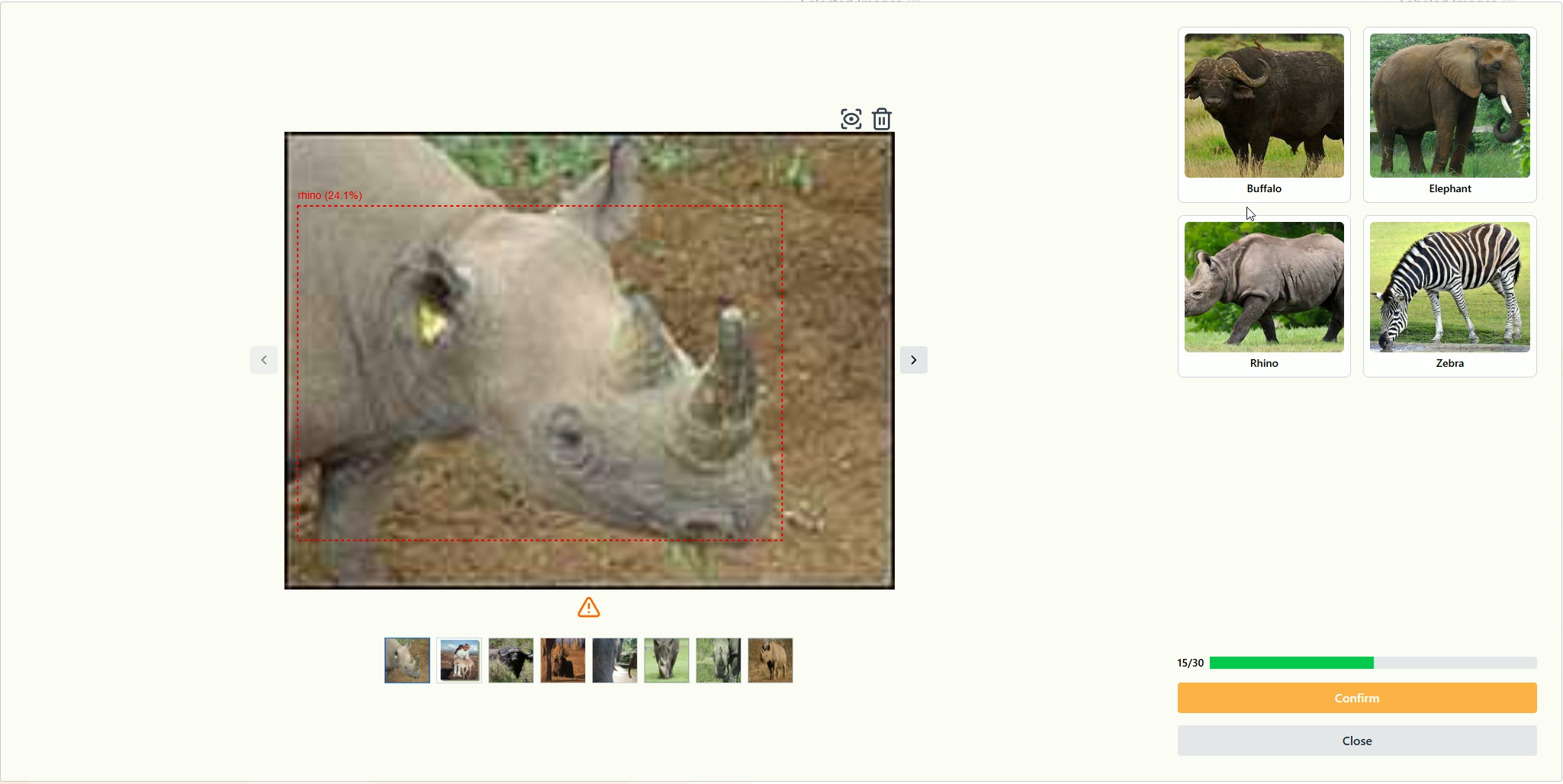}
    \caption{Example of a sample suggested by the Active Learning algorithm.}
    \label{c2i2_sel_al}
\end{figure}

\newpage
Upon inspecting the AL samples, it is noticeable that the characteristics of the samples suggested by the algorithm are quite different from the first iteration. In the first iteration, the model is not very confident in any of its predictions, leading to a wider spread of samples suggested by the algorithm. Perhaps even samples that are easy to classify, since the "hard-to-classify" samples might not have predictions yet. In this iteration, the majority of the AL samples appear to be these harder-to-classify samples, in the sense that they are often low-resolution images or close-up images showing only half of an object. Understandably, the model has a harder time predicting these. However, in most of them, the animals are still clearly visible, and to build a robust model, it is still desirable to include samples like these. An example of a selection made in a darker red region, which includes two AL samples, can be seen in Figure \ref{c2i2_sel_al}. In one sample, it is also noted that the model correctly classifies the animal in the image as a rhino. However, it is struggling with localization.

\newpage
\subsubsection{Iteration 3}
\begin{figure}[ht!]
    \centering
    \includegraphics[width=0.4\textwidth]{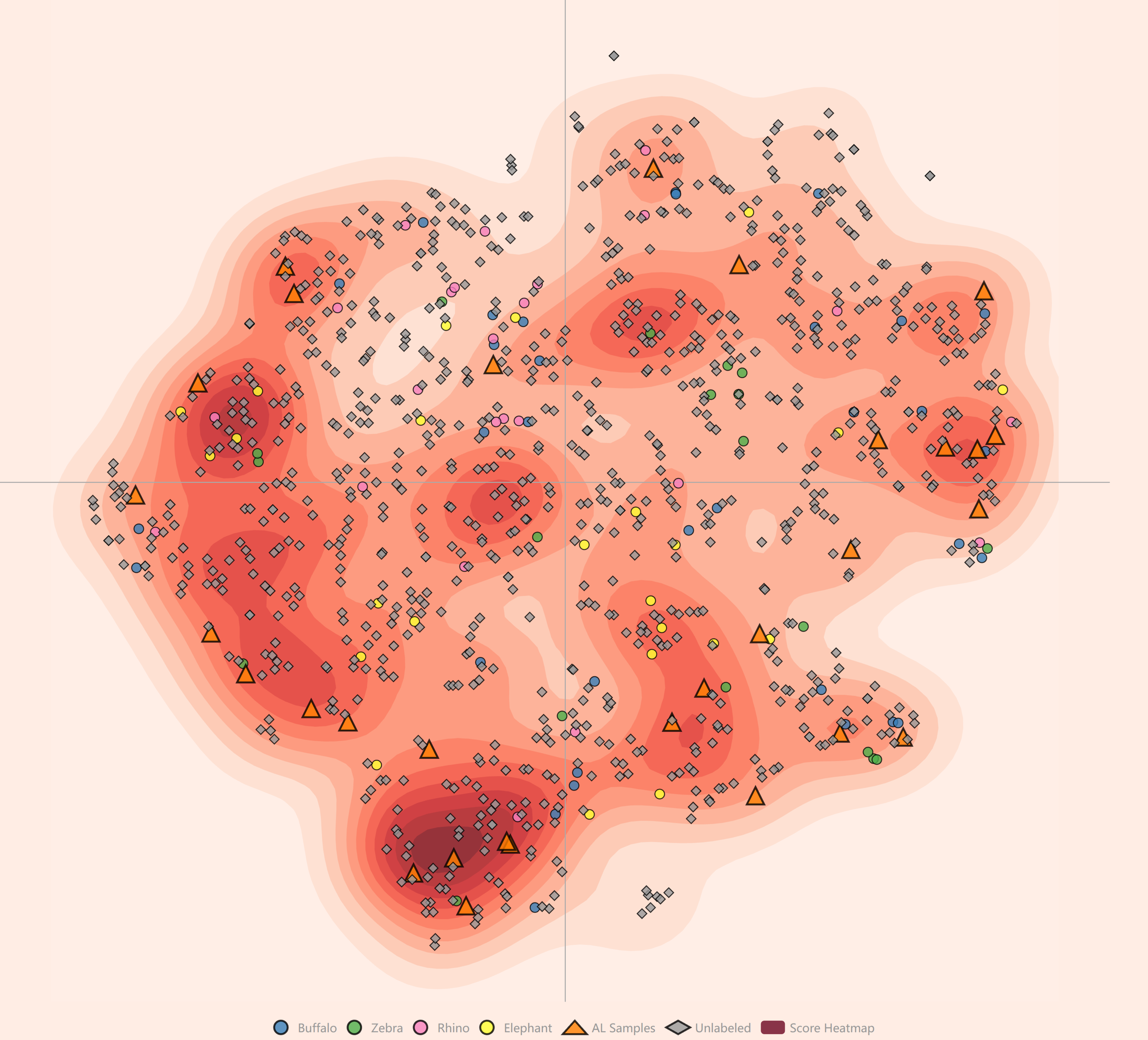}
    \caption{Data View iteration 3}
    \label{c2it3_dataview}
\end{figure}

\begin{figure}[ht!]
  \centering
  \begin{minipage}[c]{0.4\textwidth}
    \centering
    \includegraphics[width=\linewidth]{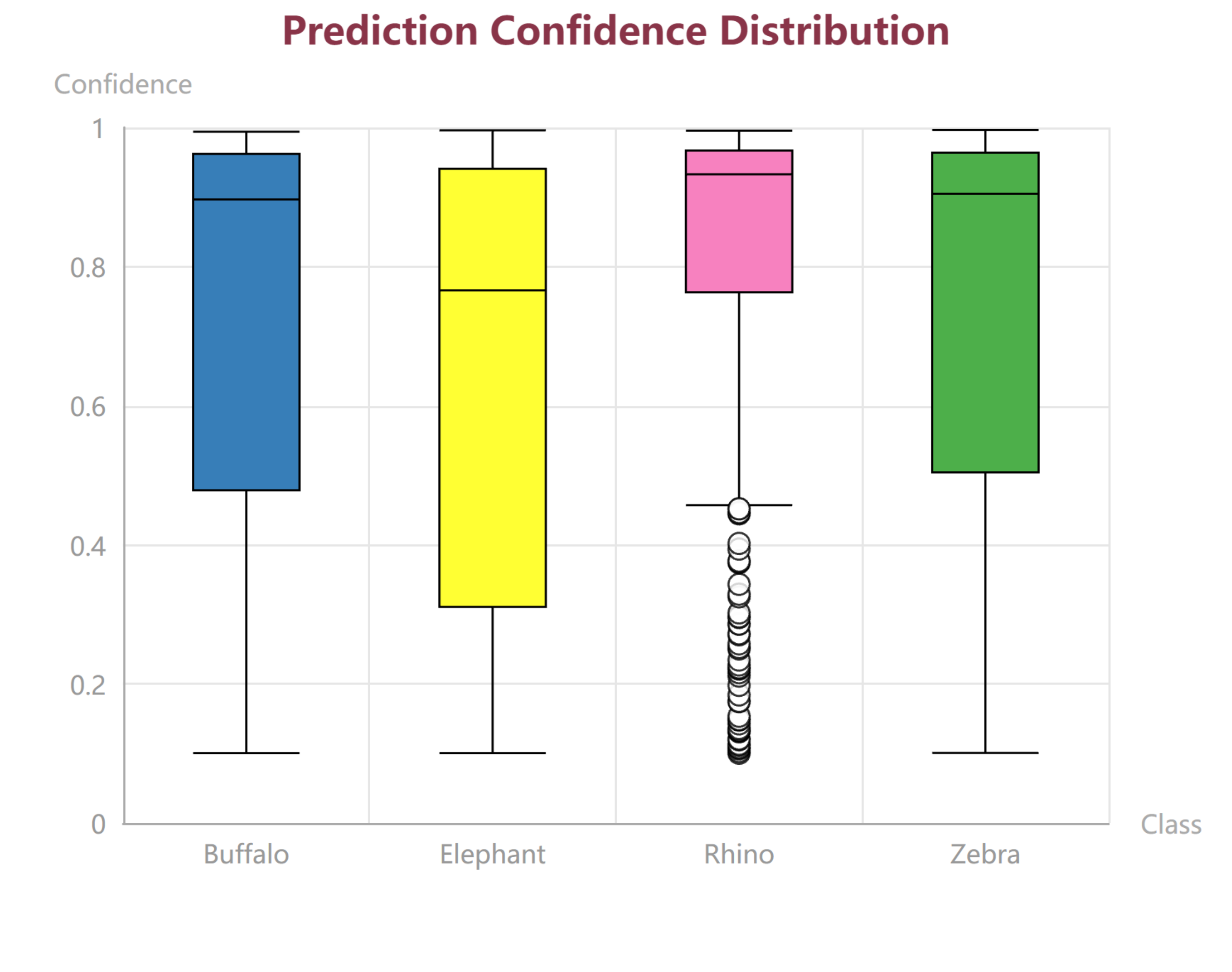}
  \end{minipage}
  \begin{minipage}[c]{0.4\textwidth}
    \centering
    \includegraphics[width=\linewidth]{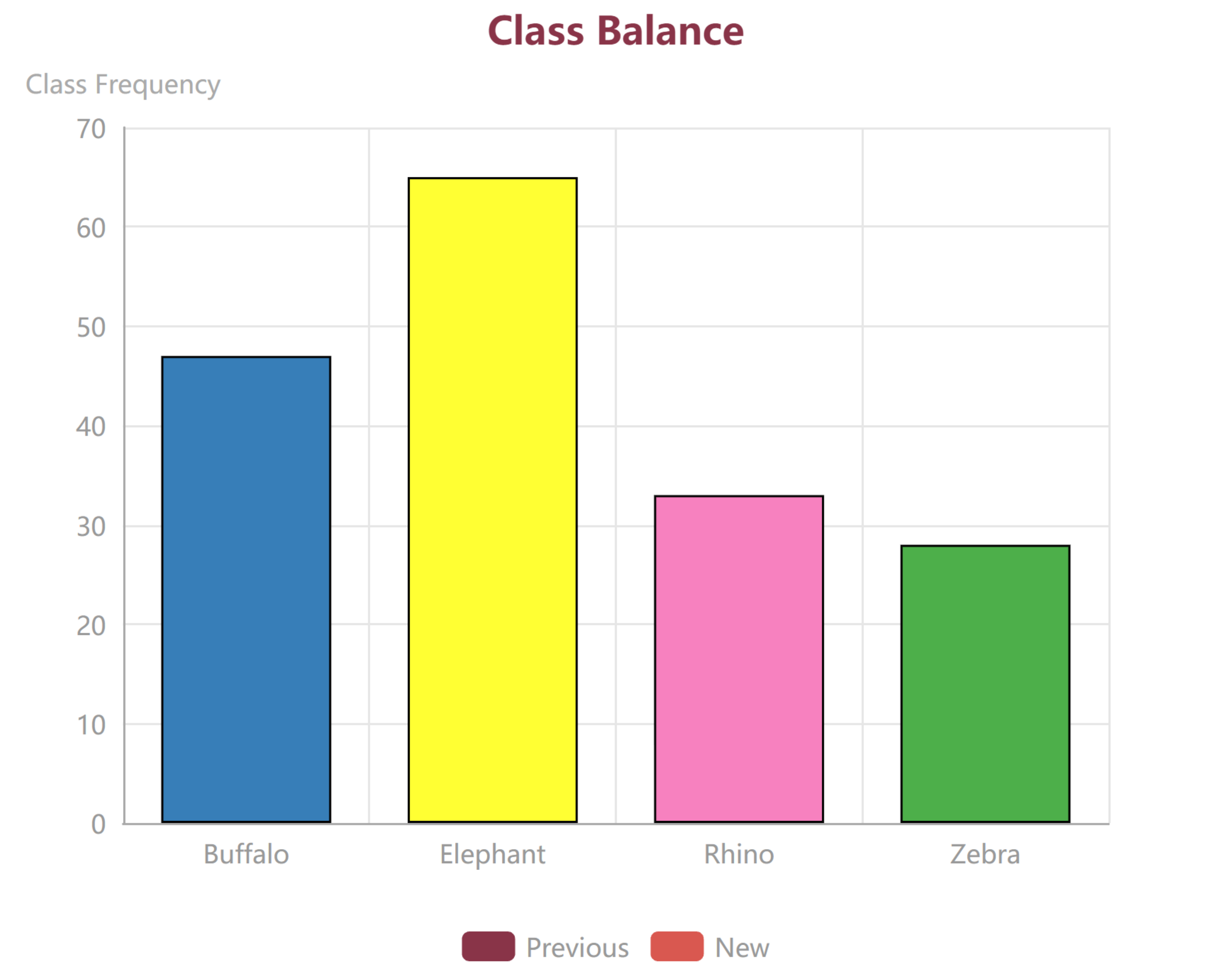}
  \end{minipage}
  \caption{Model View at the start of training iteration 2.}
  \label{mvc2i3}
\end{figure}

In this iteration, the larger red area at the intersection of quadrant one and four from the previous iteration is now not as prominent, although many of the AL-suggested samples are still located in and around that region. Instead, the uncertainty landscape seems to have moved more towards the second and third quadrants of the Data View (Figure \ref{c2it3_dataview}). The class balance is more skewed towards elephants, which have about 65 instances present in the labeled training set (Figure \ref{mvc2i3}). Zebras and rhinos are the most underrepresented classes. Although there still does not seem to be much correlation between the number of class instances present in the labeled training set and the model's prediction confidences for those classes. Rhino is the most confident class, and confidence scores between 0.1-0.45 are considered outliers in the box plot, which is a good sign. Elephants, which are most represented in the training set, have the largest spread of confidence and the lowest median.\\

\begin{figure}[ht!]
  \centering
  \begin{minipage}[c]{0.49\textwidth}
    \centering
    \includegraphics[width=\linewidth]{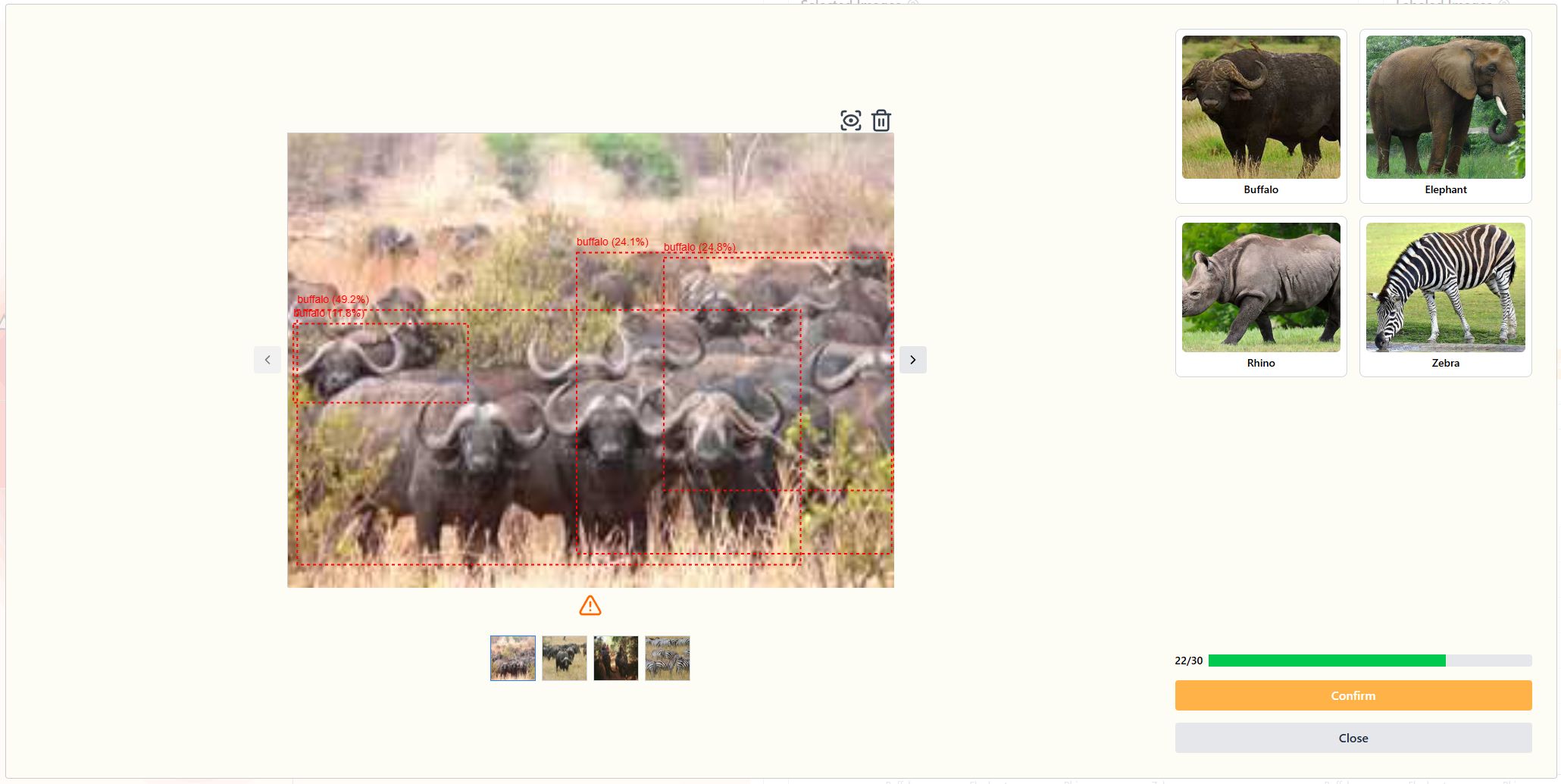}
    \caption{Example of a noisy sample with many partially hidden instances of the Buffalo class.}
    \label{noisy_buffalo}
  \end{minipage}
  \hfill
  \begin{minipage}[c]{0.49\textwidth}
    \centering
    \includegraphics[width=\linewidth]{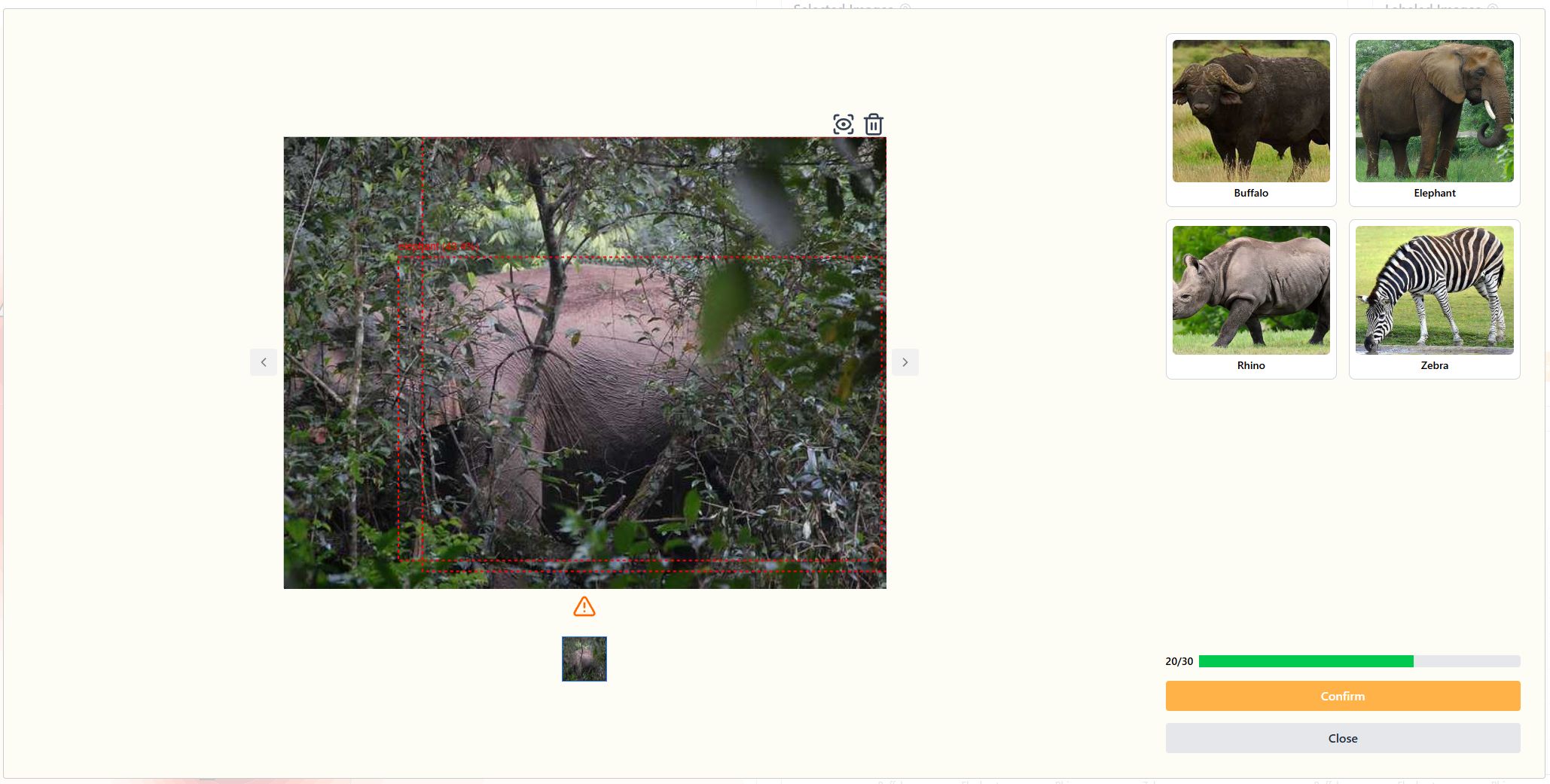}
    \caption{Example of a noisy sample with a largely obscured view of an elephant}
    \label{noisy_elepahnt}
  \end{minipage}
\end{figure}

A common feature of the AL samples this round is that they contain many instances of animals, often times also obscured or far in the background. One such example can be seen in Figure \ref{noisy_buffalo}. This sample is not only very tough to find all present buffalos, since many are partly hidden in the background, but it is also a low-resolution sample. For that reason, it was discarded as it was deemed it would introduce too much noise for the model and not add much value. Similarly, Figure \ref{noisy_elepahnt} shows an AL-suggested sample of what looks to be an elephant largely hidden behind trees and bushes. Both of these samples were discarded and replaced with other samples from high-uncertainty regions.

\subsubsection{Iteration 4}
\begin{figure}[ht!]
    \centering
    \includegraphics[width=0.5\textwidth]{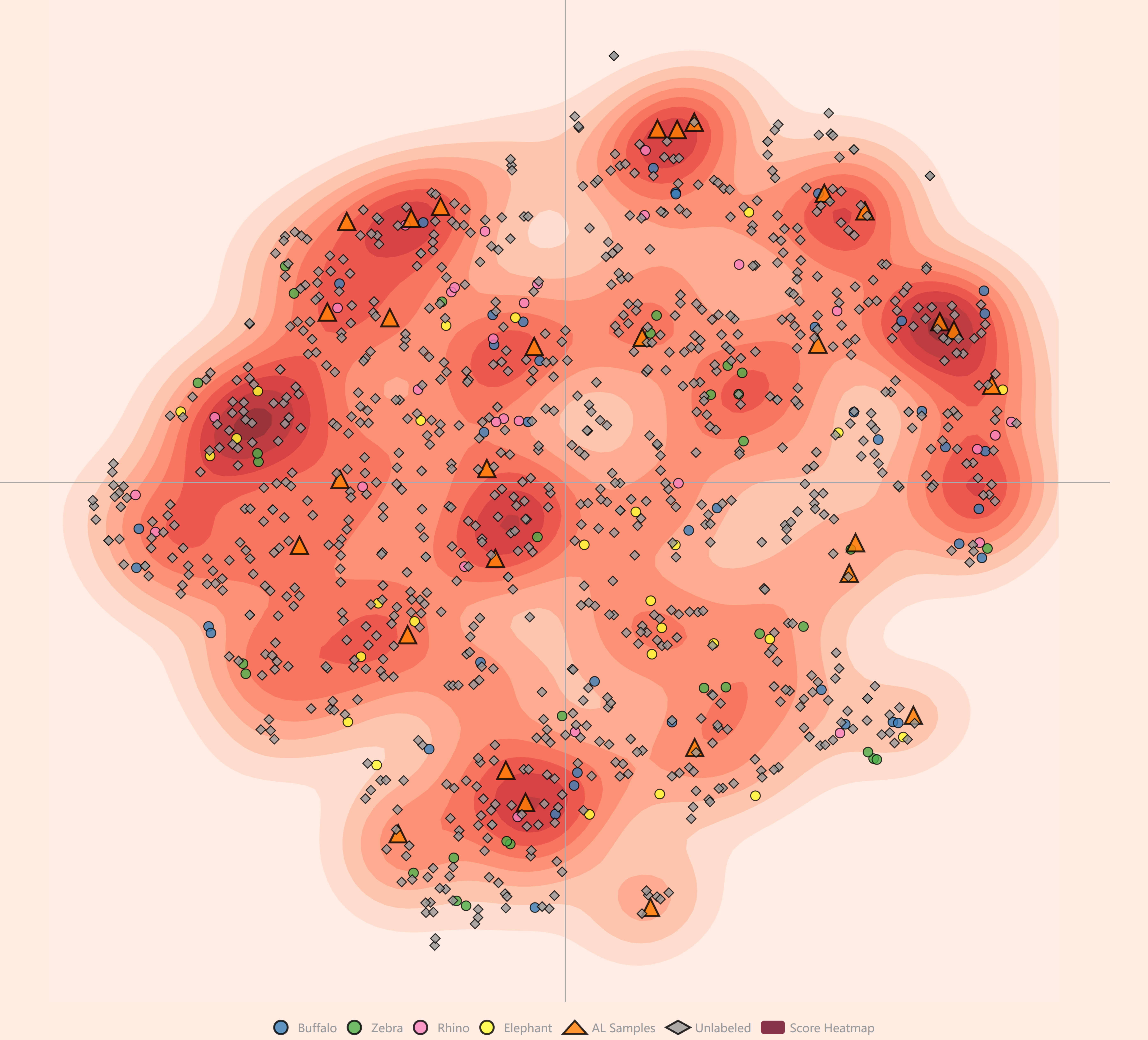}
    \caption{Data View iteration 4}
    \label{it4_dataview}
\end{figure}

\begin{figure}[ht!]
  \centering
  \begin{minipage}[c]{0.49\textwidth}
    \centering
    \includegraphics[width=\linewidth]{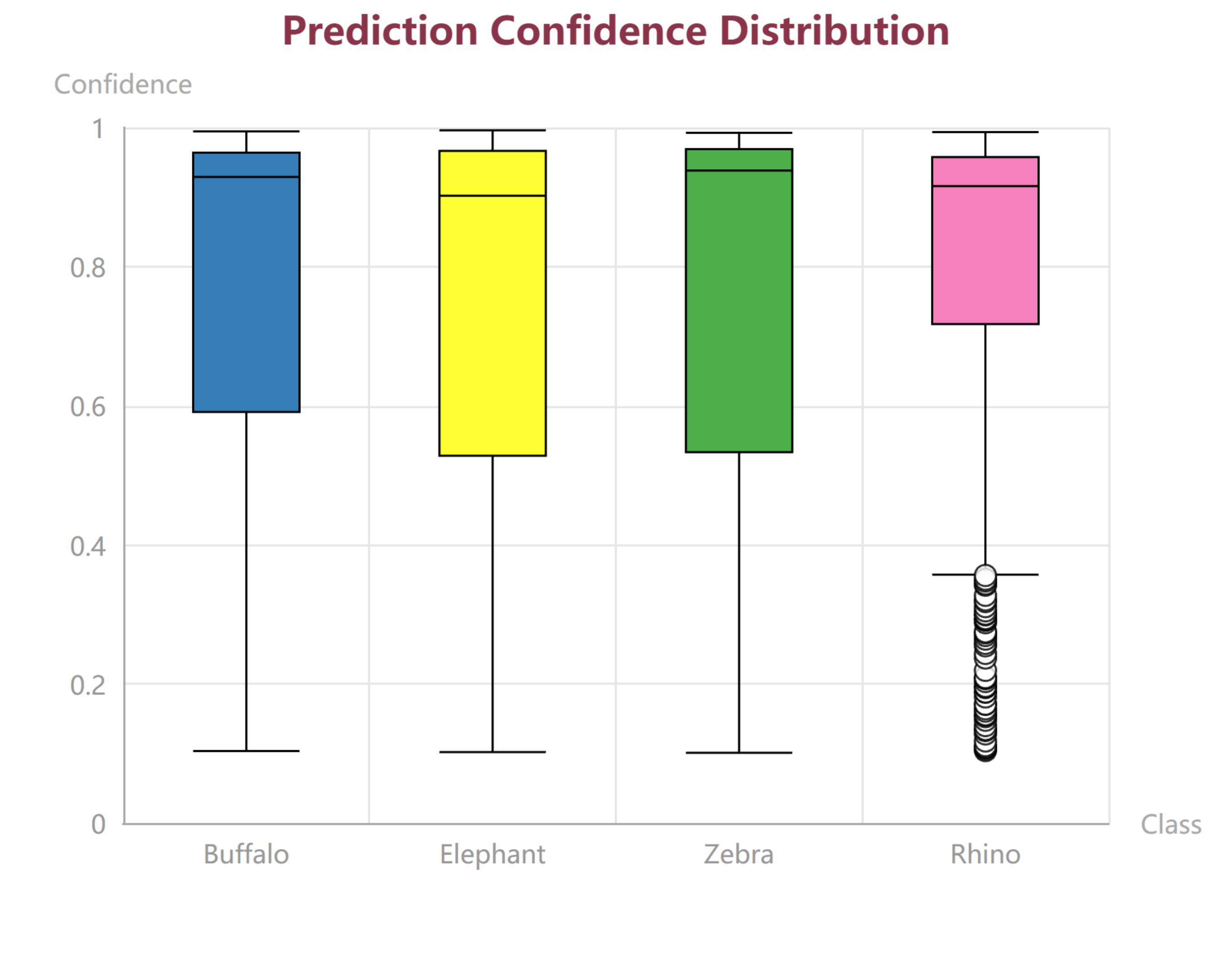}
  \end{minipage}
  \hfill
  \begin{minipage}[c]{0.49\textwidth}
    \centering
    \includegraphics[width=\linewidth]{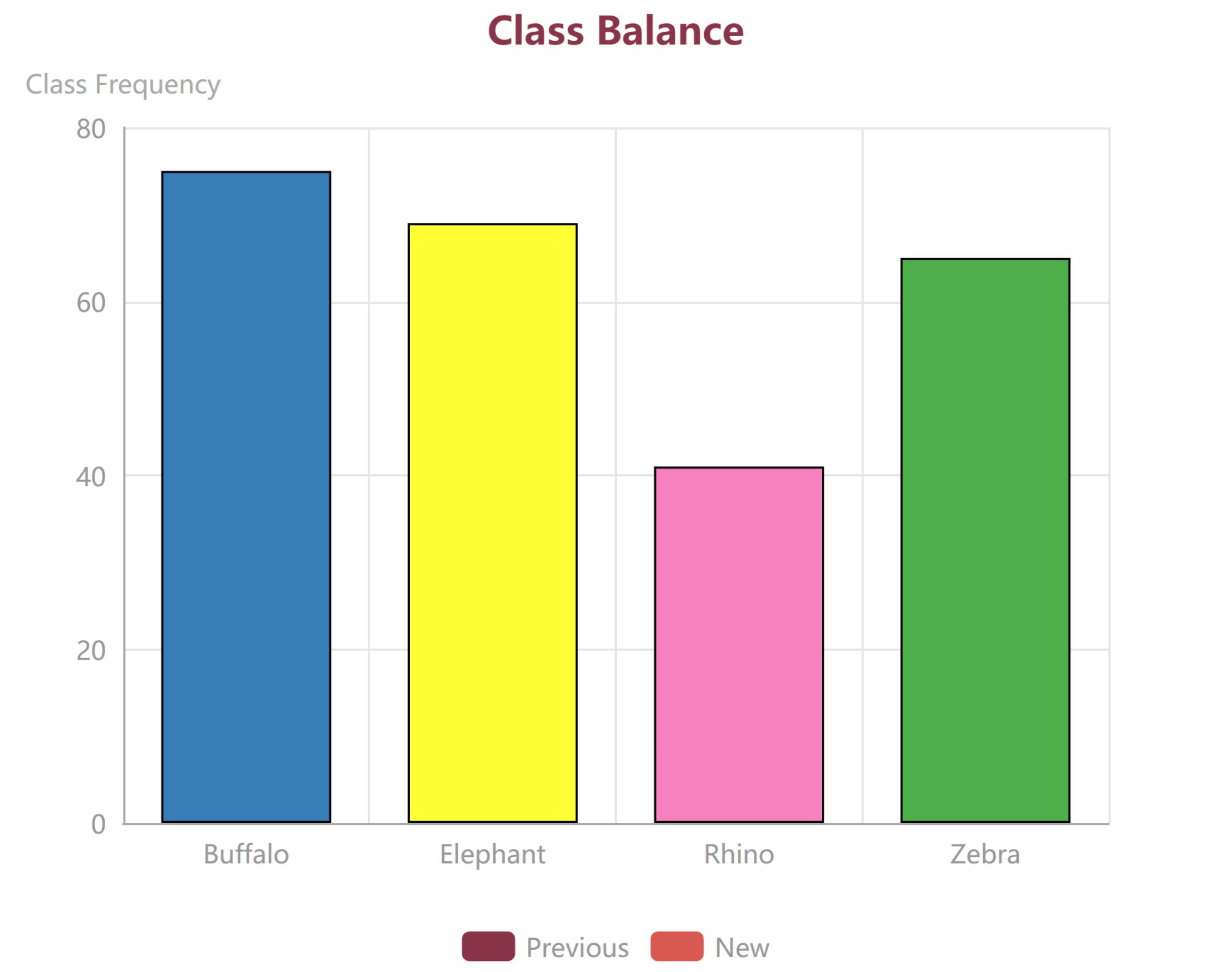}
  \end{minipage}
  \caption{Model View at the start of training iteration 4.}
  \label{it4_modelview}
\end{figure}

After another round of retraining was completed, and a total of 90 images were added to the labeled training set, the model still seems to improve with regard to the Prediction Confidence Distribution, as seen in Figure \ref{it4_modelview} and compared to the previous iteration. However, the heatmap in the Data View in Figure \ref{it4_dataview} shows more dense regions of high uncertainty than earlier. Some regions that were previously considered low uncertainty regions are now high-uncertainty regions. This could be concerning, as it points to the model perhaps forgetting important "easier" structures that it learned very early. The class balance is slightly more balanced now. However, the Rhino class is mostly falling behind the other three classes

\begin{figure}[ht!]
    \centering
    \includegraphics[width=0.7\textwidth]{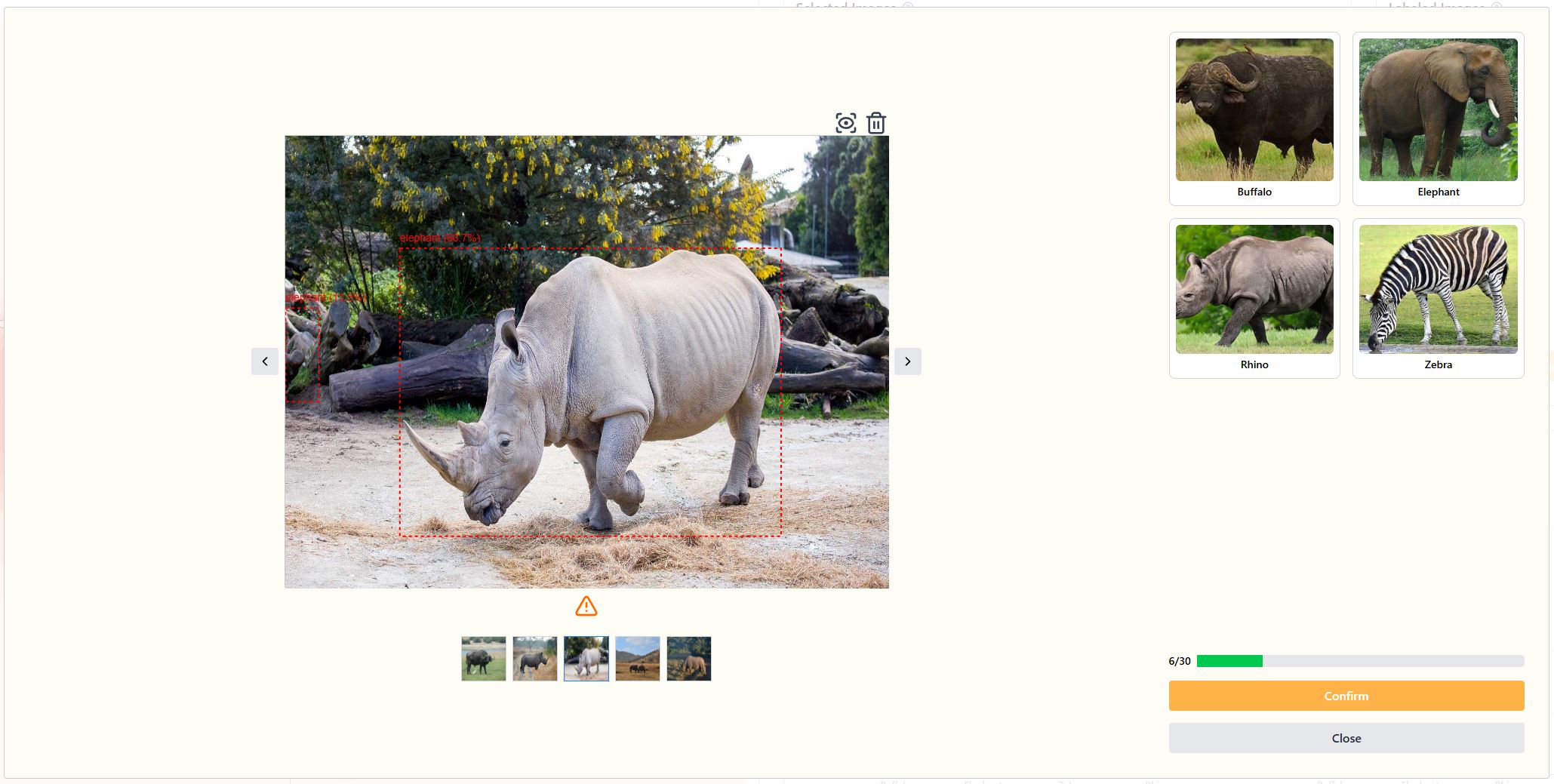}
    \caption{A confidently misclassified sample.}
    \label{rhino_not_ele}
\end{figure}

On closer inspection of samples suggested by the AL algorithm in this iteration, it is noticed that the model sometimes, quite confidently, misclassifies rhinos as elephants. This occurs even for samples that could be considered easier, with a clear view of the object, good lighting, and high resolution. As seen in Figure \ref{rhino_not_ele}, the model has mistakenly predicted a rhino as an elephant with 60\% confidence.

\newpage
Although the Prediction Confidence Distribution in the Model View points to rhinos being the most confidently predicted class, the interplay between the different visualizations and inspecting AL suggestions tells a more nuanced story. Labeling of the AL-suggested samples continues, hoping that this inconsistency will resolve itself in this or the next training round.

\newpage
\subsubsection{Iteration 5}
\begin{figure}[ht!]
    \centering
    \includegraphics[width=0.5\textwidth]{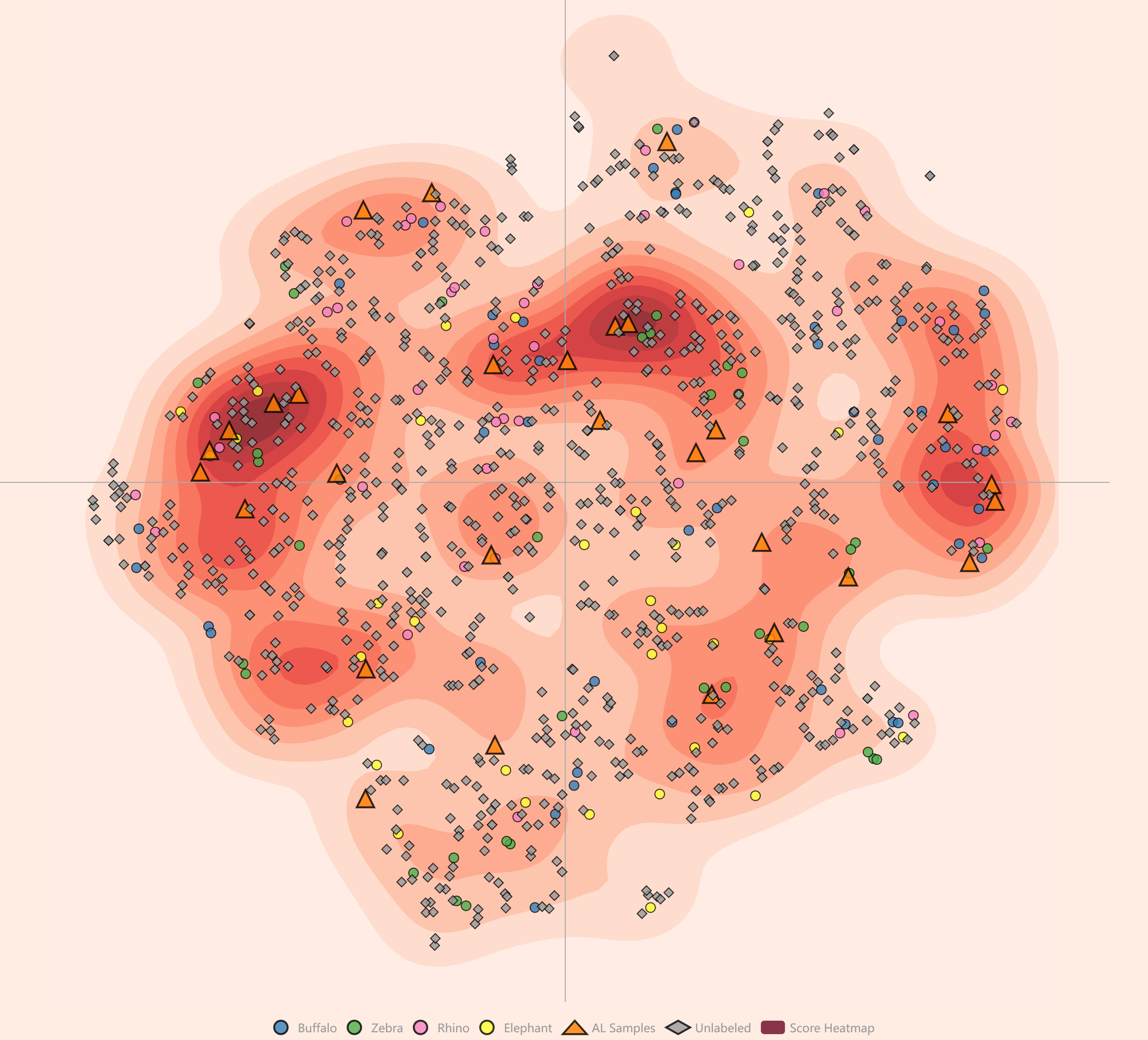}
    \caption{Data View iteration 5}
    \label{it5_dataview}
\end{figure}

\begin{figure}[ht!]
  \centering
  \begin{minipage}[c]{0.49\textwidth}
    \centering
    \includegraphics[width=\linewidth]{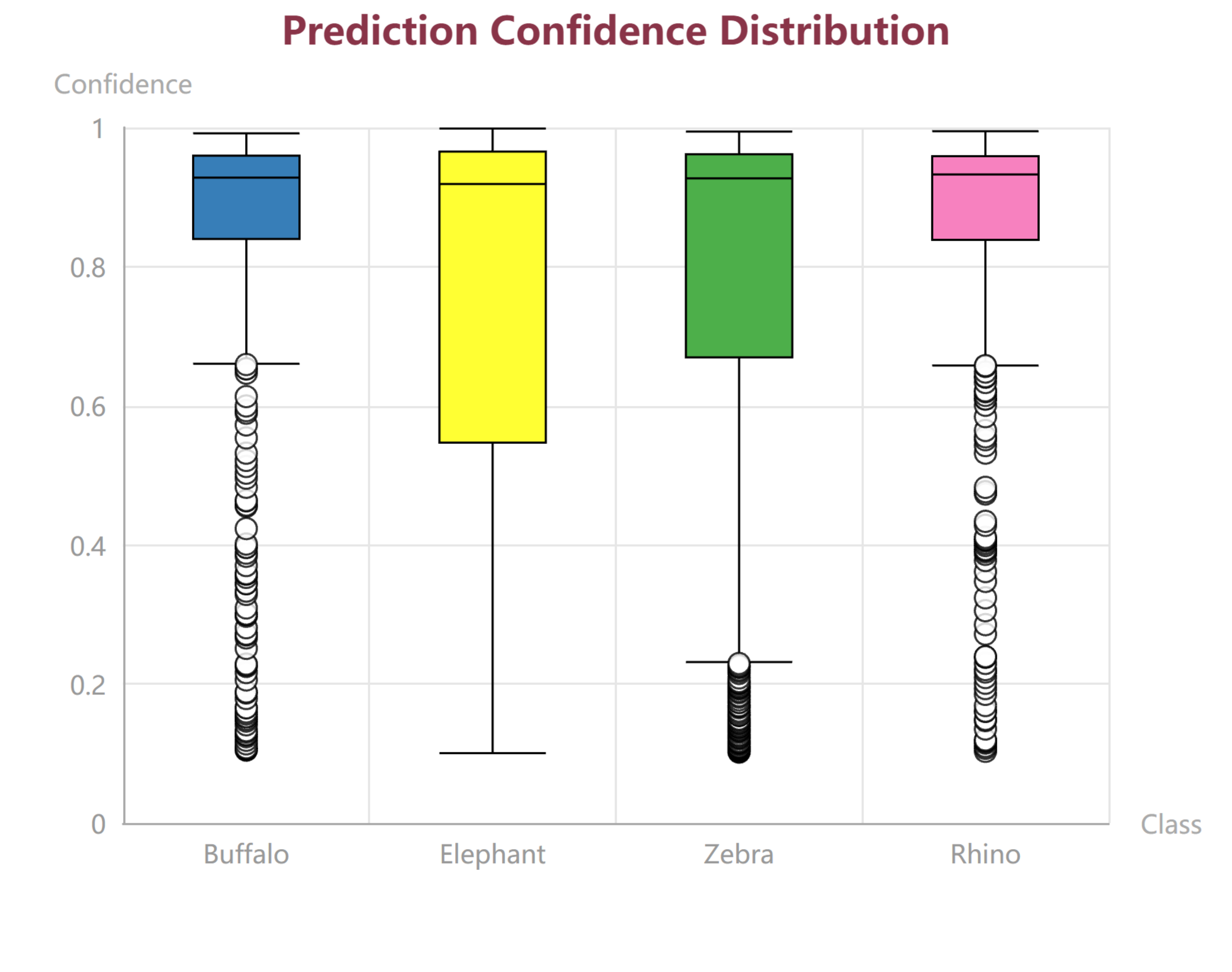}
  \end{minipage}
  \hfill
  \begin{minipage}[c]{0.49\textwidth}
    \centering
    \includegraphics[width=\linewidth]{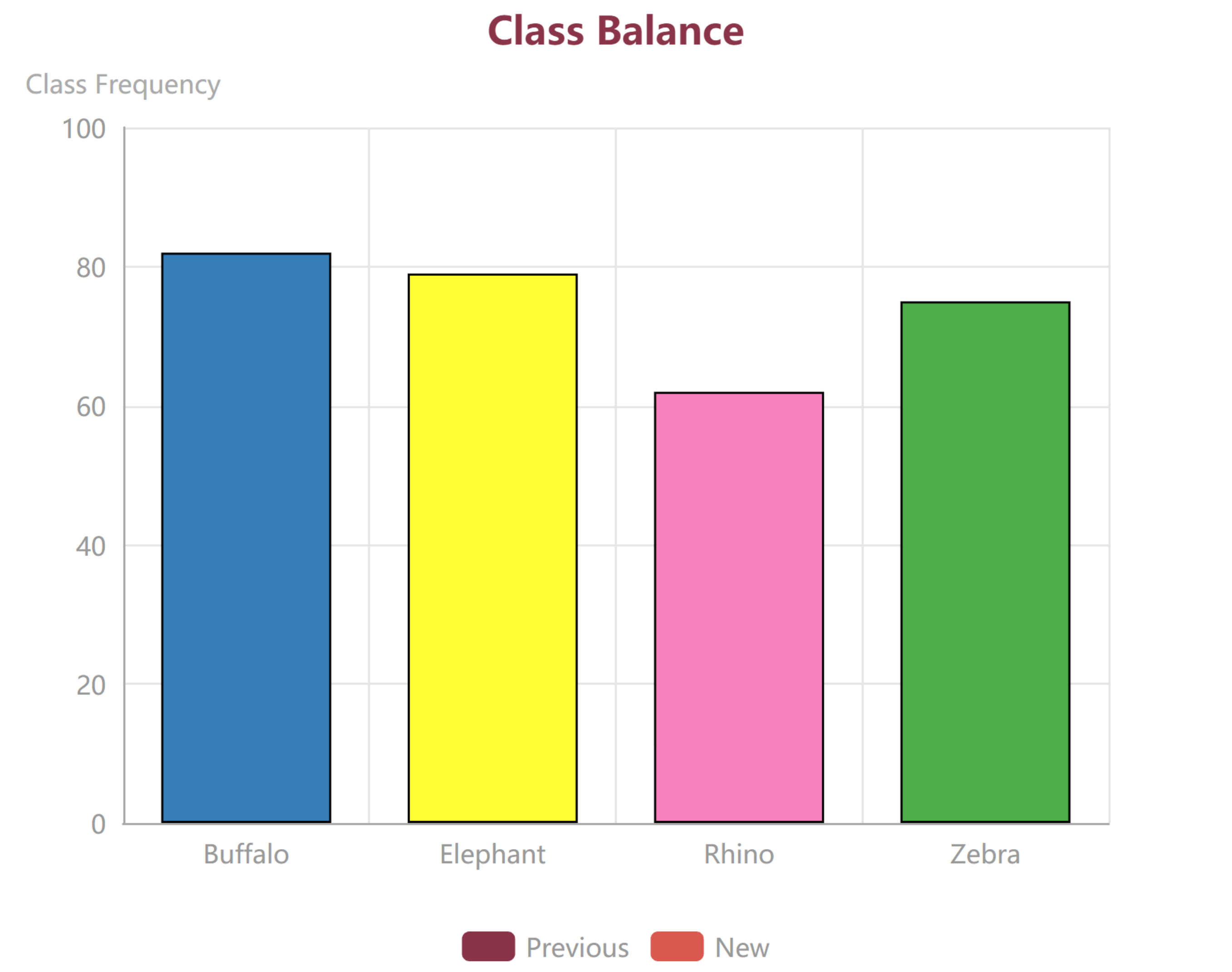}
  \end{minipage}
  \caption{Model View at the start of training iteration 5.}
  \label{it5_modelview}
\end{figure}

Starting the final training round, it looks like iteration 4 was an improvement in terms of the uncertainty landscape. There is a lot less dark red; instead, larger regions of low uncertainty are seen, as depicted in Figure \ref{it5_dataview}. This is also confirmed by the box plot in the Model View, as seen in Figure \ref{it5_modelview}. The boxes are tighter, and the median confidence for each class is above 0.9. Most of the predictions lie in the 0.6-0.95 range, and many low predictions are now considered outliers. Elephants still have a larger spread of prediction scores than the other classes, but are still performing well. The representation of the classes in the labeled training set is surprisingly balanced.

\begin{figure}[ht!]
  \centering
  \begin{minipage}[c]{0.49\textwidth}
    \centering
    \includegraphics[width=\linewidth]{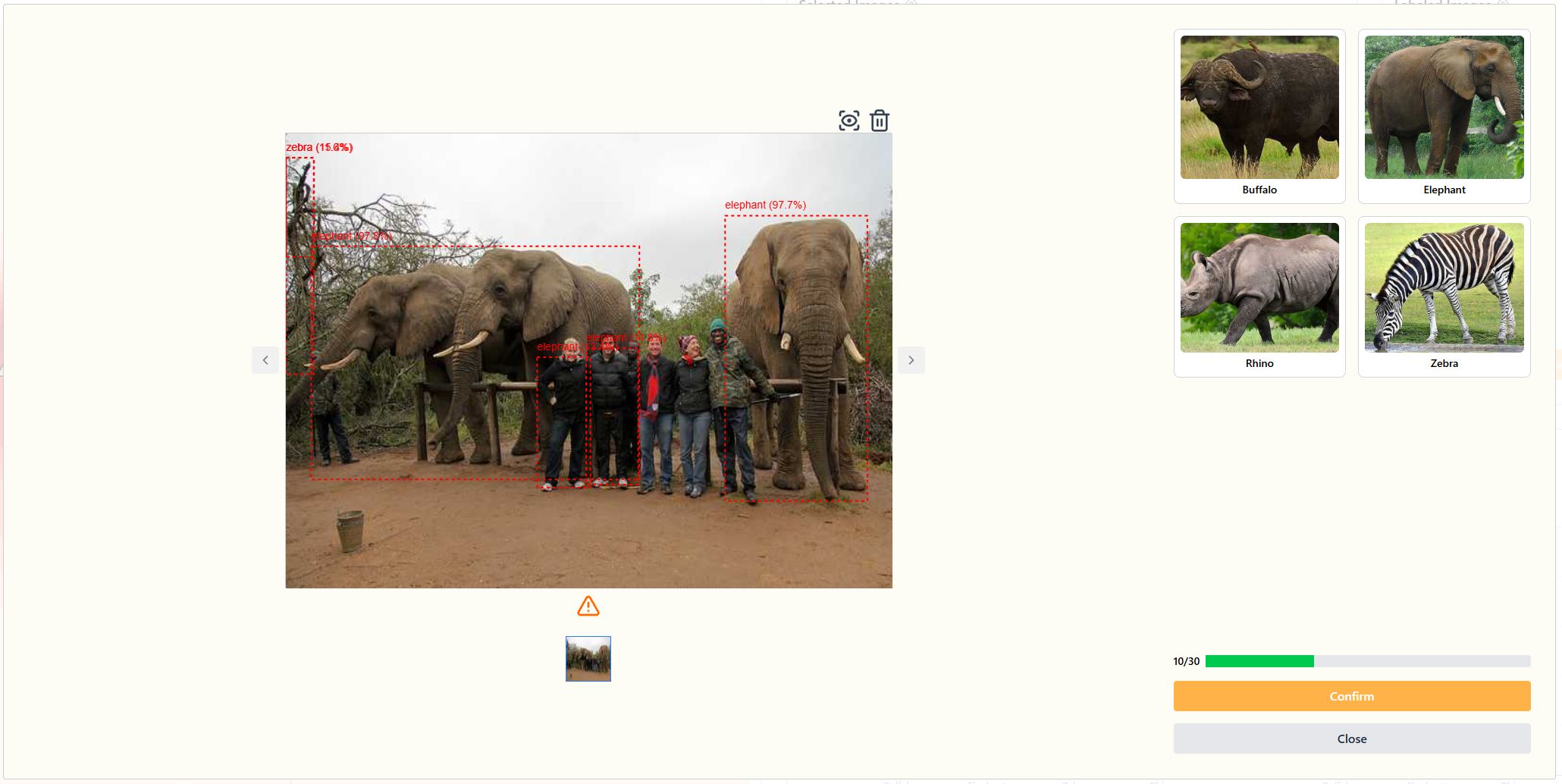}
  \end{minipage}
  \hfill
  \begin{minipage}[c]{0.49\textwidth}
    \centering
    \includegraphics[width=\linewidth]{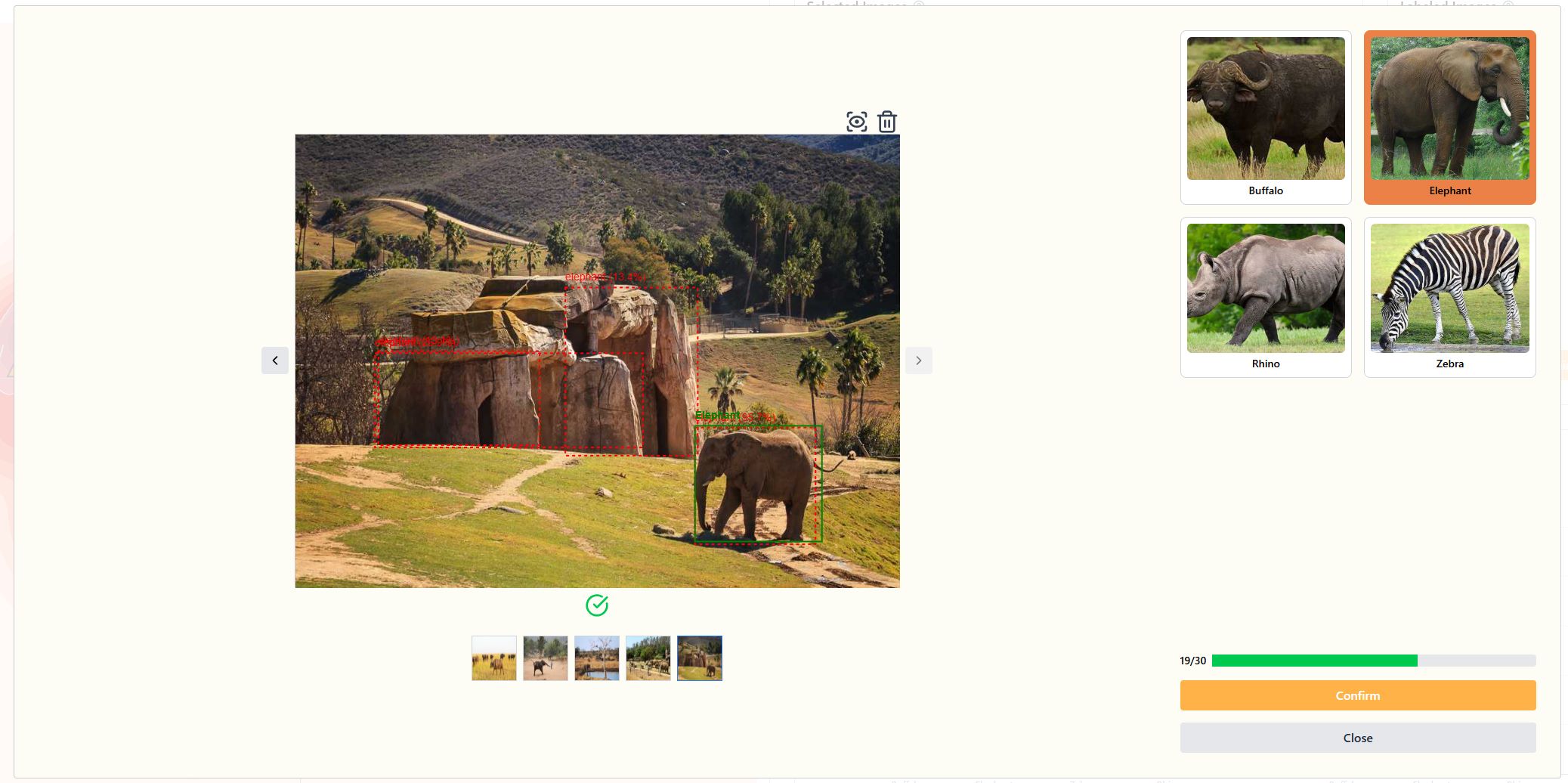}
  \end{minipage}
  \caption{AL samples in iteration 5.}
  \label{it5_AL_samples}
\end{figure}

As indicated in the Model View and confirmed when inspecting the AL suggestions, the model is still struggling with predicting some instances of elephants, especially samples with multiple instances of them, as seen in Figure \ref{it5_AL_samples}. In the left image, the model makes a few low-confidence, or "noisy" predictions, encapsulating humans as elephants. However, it fails to detect one visible elephant, which could be more problematic. In the right image, it properly detects an elephant in the foreground with high confidence, but it makes a few low-confidence predictions of stone formations in the background \\

At this point in the training process, it seems like the AL suggestions mainly depict minor errors. It is clear that the model is already performing well in general, and the focus is now more on edge cases. The low-confidence detections that are made in Figure \ref{it5_AL_samples} might not necessarily be a problem. Detections like these would be filtered out in a deployment setting by adjusting the allowed prediction confidence to achieve the highest amount of true positives.

Labeling this final round of AL samples and training the last iteration of the model with a total of 190 labeled images in the training set concludes this use case, which focused on uncertainty sampling as facilitated by the VILOD tool. This uncertainty-driven approach highlighted the dynamic nature of AL suggestions and underscores the importance of human-in-the-loop quality control (as seen in Iteration 3) to filter out genuinely uninformative or noisy samples that a purely automated system might select. Final evaluation of the trained model is presented in the following Analysis \ref{analysis} section, and the full selection of labeled samples for each iteration is provided in Appendix \ref{appendixA}.

\subsection{Use Case 3: Balanced Guidance Integration}
\label{use_case_balanced}

This use case is designed to simulate the intended \textit{best practice} application of the VILOD system. 
This approach involves a holistic synthesis of all available visual guidance and model feedback to inform sample selection. 
The process included inspecting model uncertainty through the heatmap and AL suggestions, cross-referencing these with the dataset's structure as revealed by the t-SNE Data View (identifying clusters and data density), and actively consulting the Model View for insights into prediction confidence distributions and class balance. 
Selections were strategically made with multiple objectives in mind: addressing high-uncertainty areas, exploring diverse unlabeled regions of the feature space, actively working to improve class balance in the labeled set, and, where possible, attempting to correct specific, observably problematic model behaviors. 
The following sections will provide a narrative through five training iterations using the VILOD tool. Consistent with the overall methodology for all use cases defined in Chapter 3, this scenario commenced from the identical initial model ($M_0$) and proceeded through five iterations, with 30 new images selected and annotated in each iteration to incrementally expand the training dataset.

\subsubsection{Iteration 1}

In the first iteration, the visualizations are based on the generated base model $m_0$ as depicted in \ref{c3it1_dataview} and \ref{c3it1_modelview}.

On first glance, it is noted that the heatmap has many dark red regions, indicating a high degree of uncertainty in the model predictions. This is further substantiated by the Prediction Confidence Distribution chart. Most of the predictions for all classes fall within the range of 0.1 to 0.35 confidence. The representation of the classes in the initial training set is not perfectly balanced. The buffalo and rhino classes are represented by 15 instances each, whereas elephants and zebras have 9 and 11, respectively. In this early training iteration, the selection strategy aims to balance the instances of each class in the labeled training set by selecting a diverse set of samples that cover the entire feature space. This could also include samples suggested by the AL algorithm, but it is not the main focus. Images considered hard or edge cases will be avoided, as the aim is not to introduce too much noise this early\\

\begin{figure}[ht!]
  \centering
  \begin{minipage}[c]{0.4\textwidth}
    \centering
    \includegraphics[width=\linewidth]{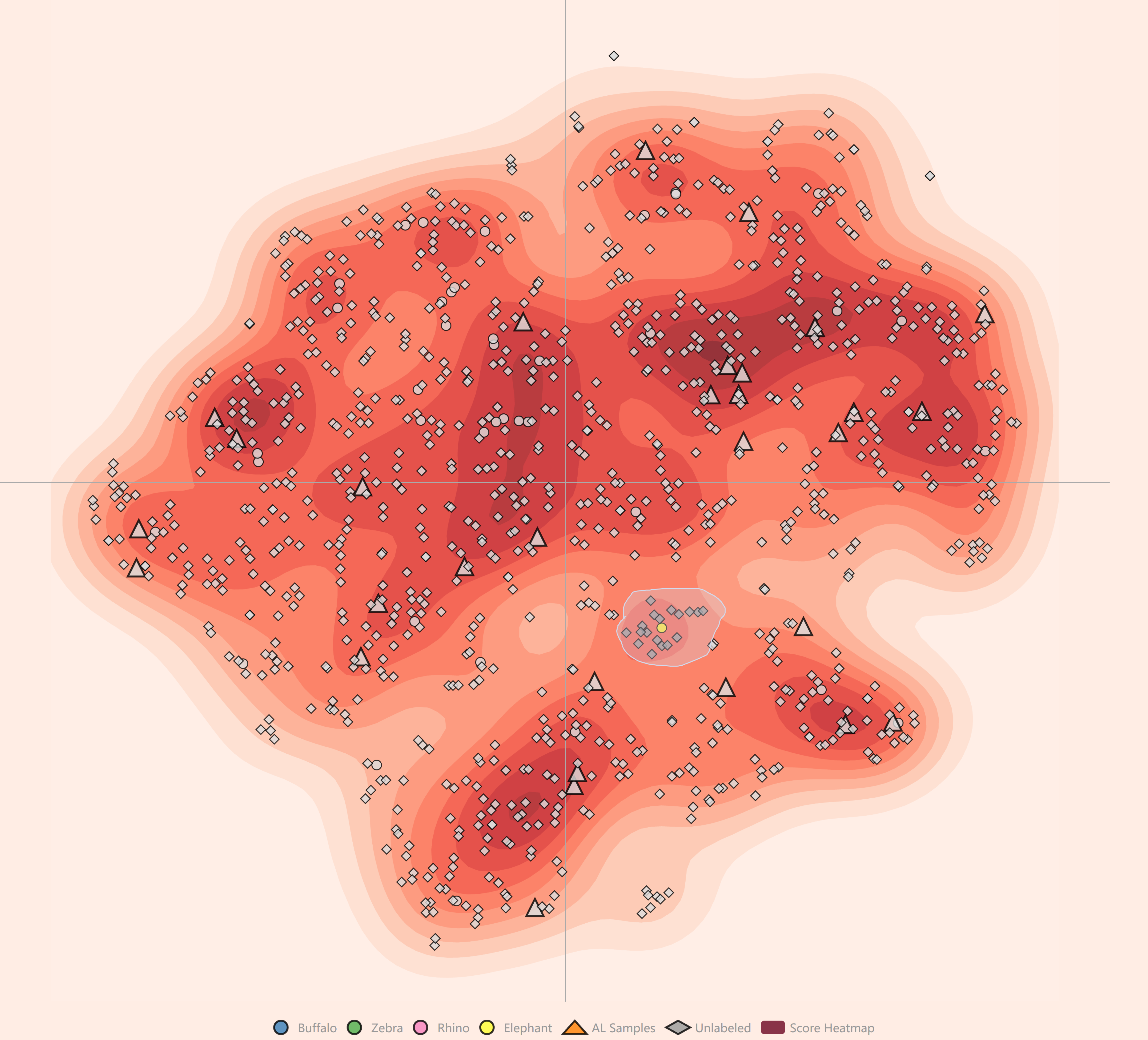}
  \end{minipage}
  \begin{minipage}[c]{0.4\textwidth}
    \centering
    \includegraphics[width=\linewidth]{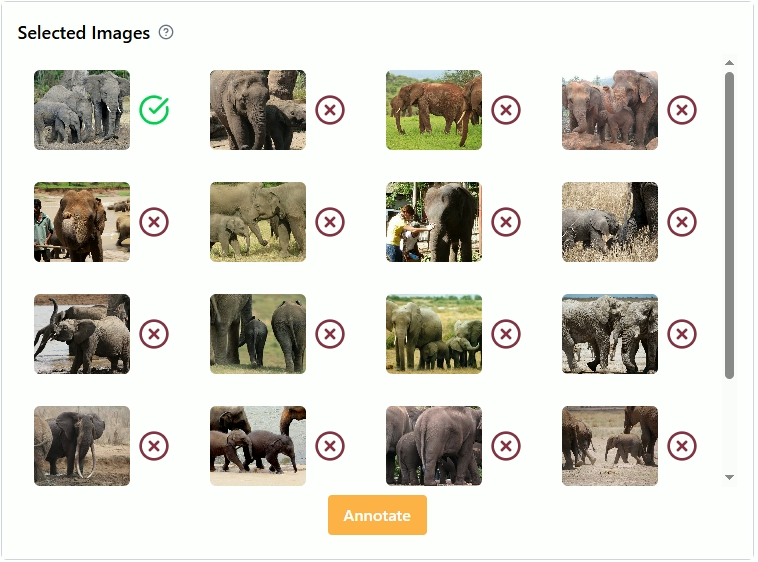}
  \end{minipage}
  \caption{First selection of samples in iteration 1.}
  \label{c3s1}
\end{figure}

The first selection is made by using the lasso tool to encapsulate a visibly separated cluster of points in the fourth quadrant of the Data View, as seen in Figure \ref{c3s1}. This selection largely contains images with instances of elephants, which are desired to balance the training set. Since the goal is to cover a large area of the feature space, only three of the selected images in this region are annotated and added to the training set.

Attention then moves to the first quadrant, highlighting another set of points closely coupled together. This selection predominantly consists of samples depicting rhinos but also includes some elephants, often with a side-view perspective of the animal, as seen in Figure \ref{c3s2}.

\begin{figure}[ht!]
  \centering
  \begin{minipage}[c]{0.4\textwidth}
    \centering
    \includegraphics[width=\linewidth]{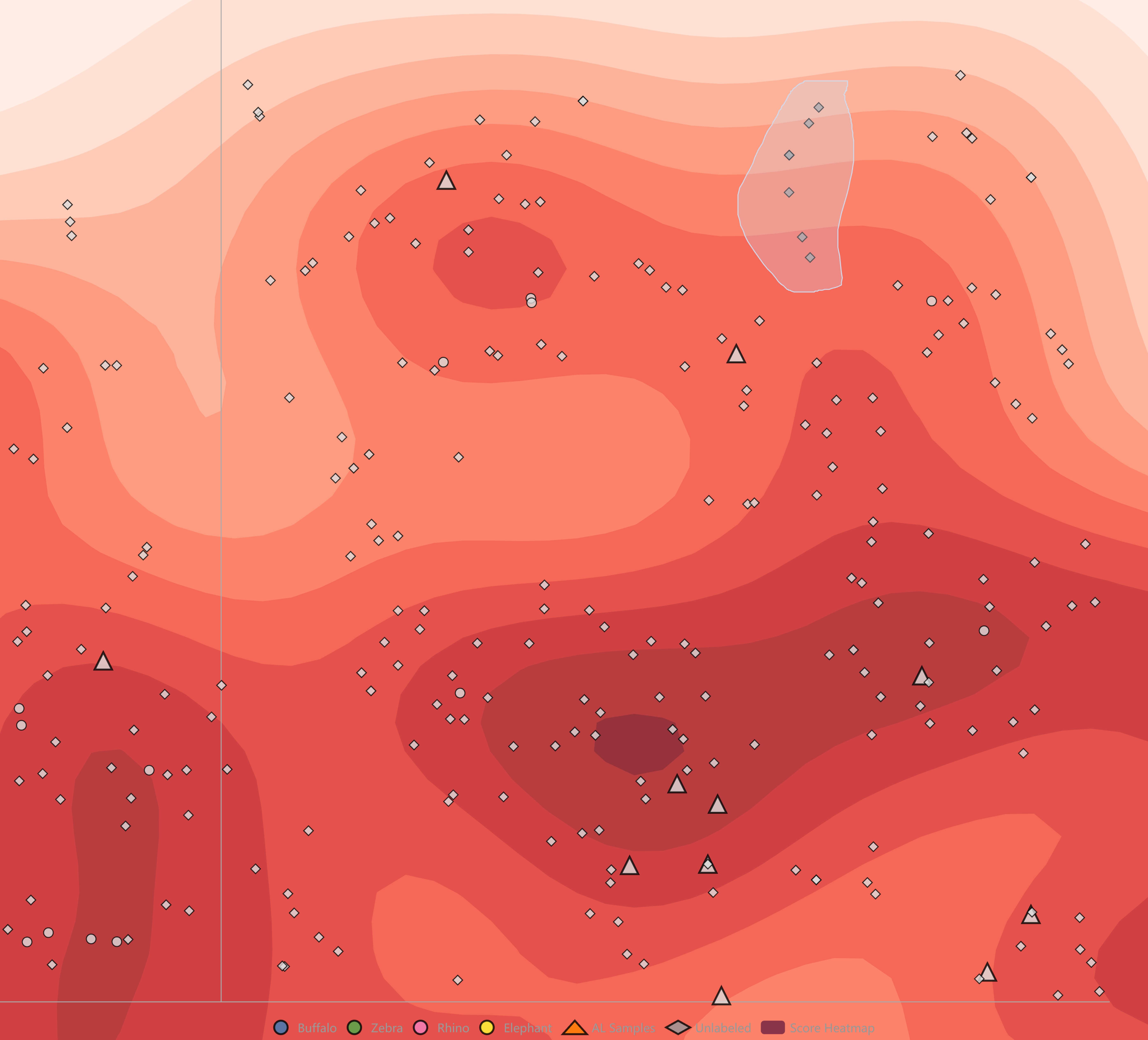}
  \end{minipage}
  \begin{minipage}[c]{0.4\textwidth}
    \centering
    \includegraphics[width=\linewidth]{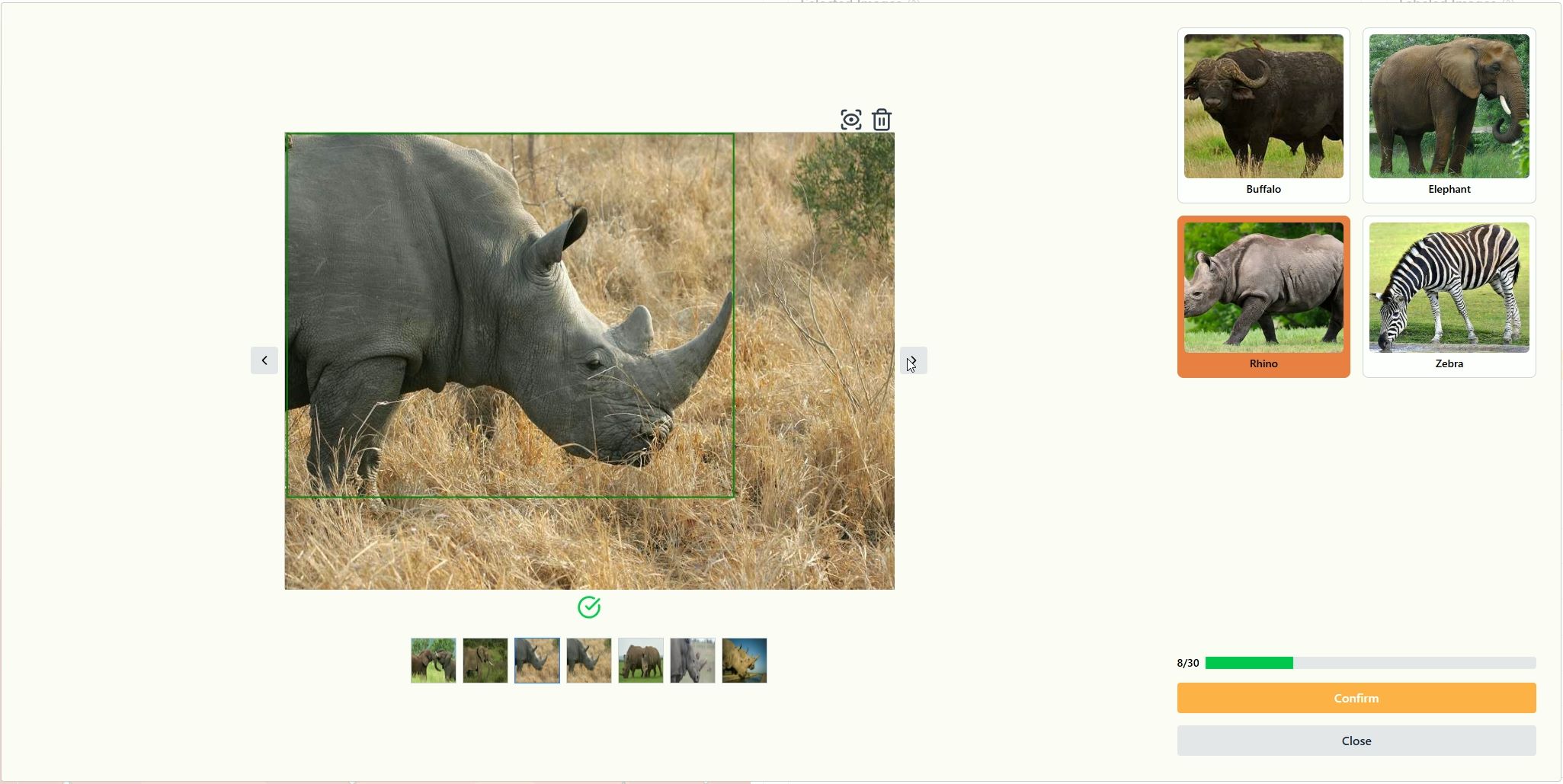}
  \end{minipage}
  \caption{Selection of samples depicting rhinos from a side view perspective.}
  \label{c3s2}
\end{figure}

\begin{figure}[ht!]
    \centering
    \includegraphics[width=0.4\textwidth]{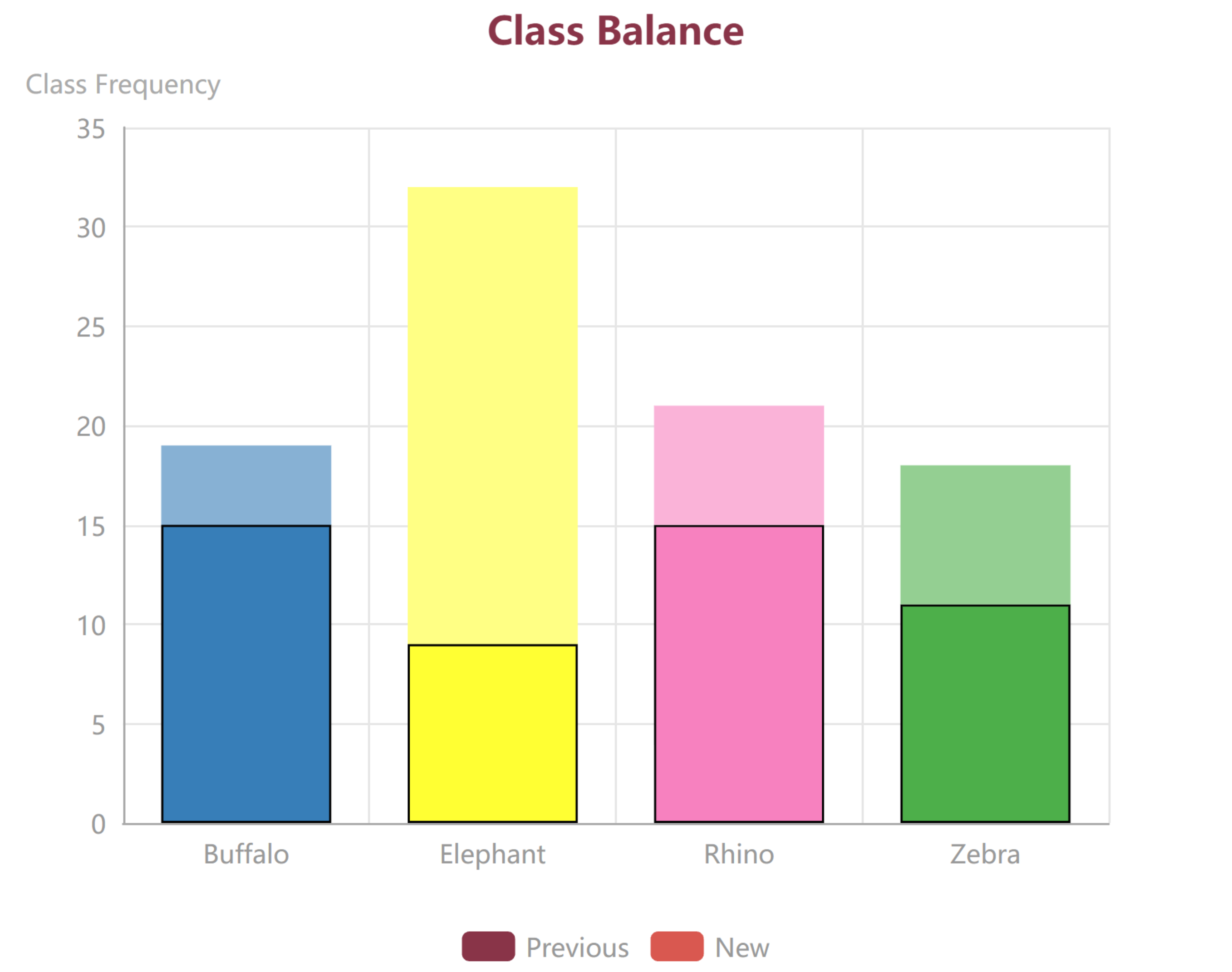}
    \caption{Class Balance after half the labeling budget is spent.}
    \label{c3it1_cbhalf}
\end{figure}

After a few more selections and with almost half the labeling budget spent, the Model View is consulted to assess the class balance (Figure \ref{c3it1_cbhalf}). Buffalo, rhino, and zebras are lagging behind elephants. Hence, subsequent selections made in this iteration will mostly exclude elephants.

\begin{figure}[ht!]
  \centering
  \begin{minipage}[c]{0.40\textwidth}
    \centering
    \includegraphics[width=\linewidth]{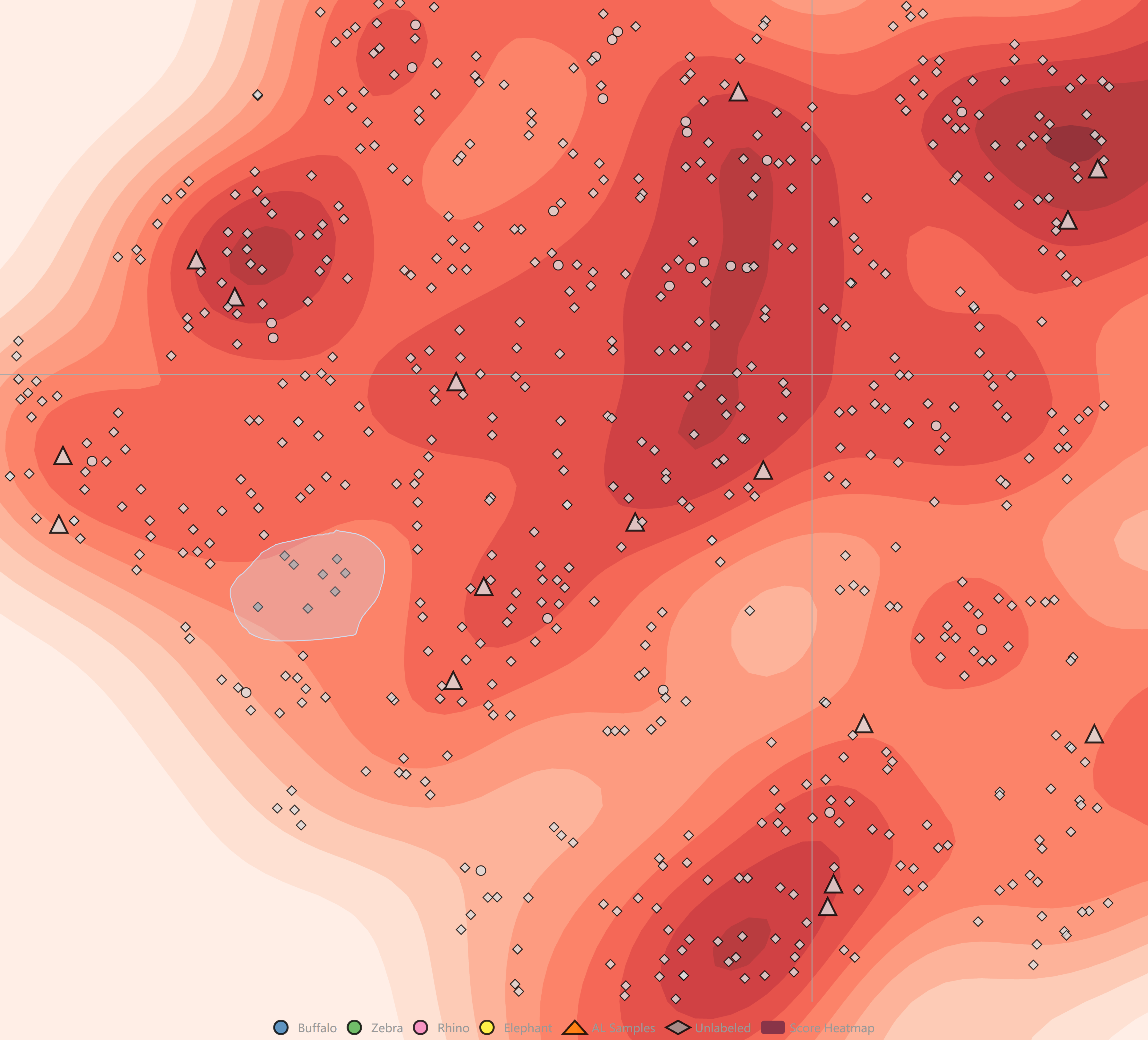}
  \end{minipage}
  \begin{minipage}[c]{0.4\textwidth}
    \centering
    \includegraphics[width=\linewidth]{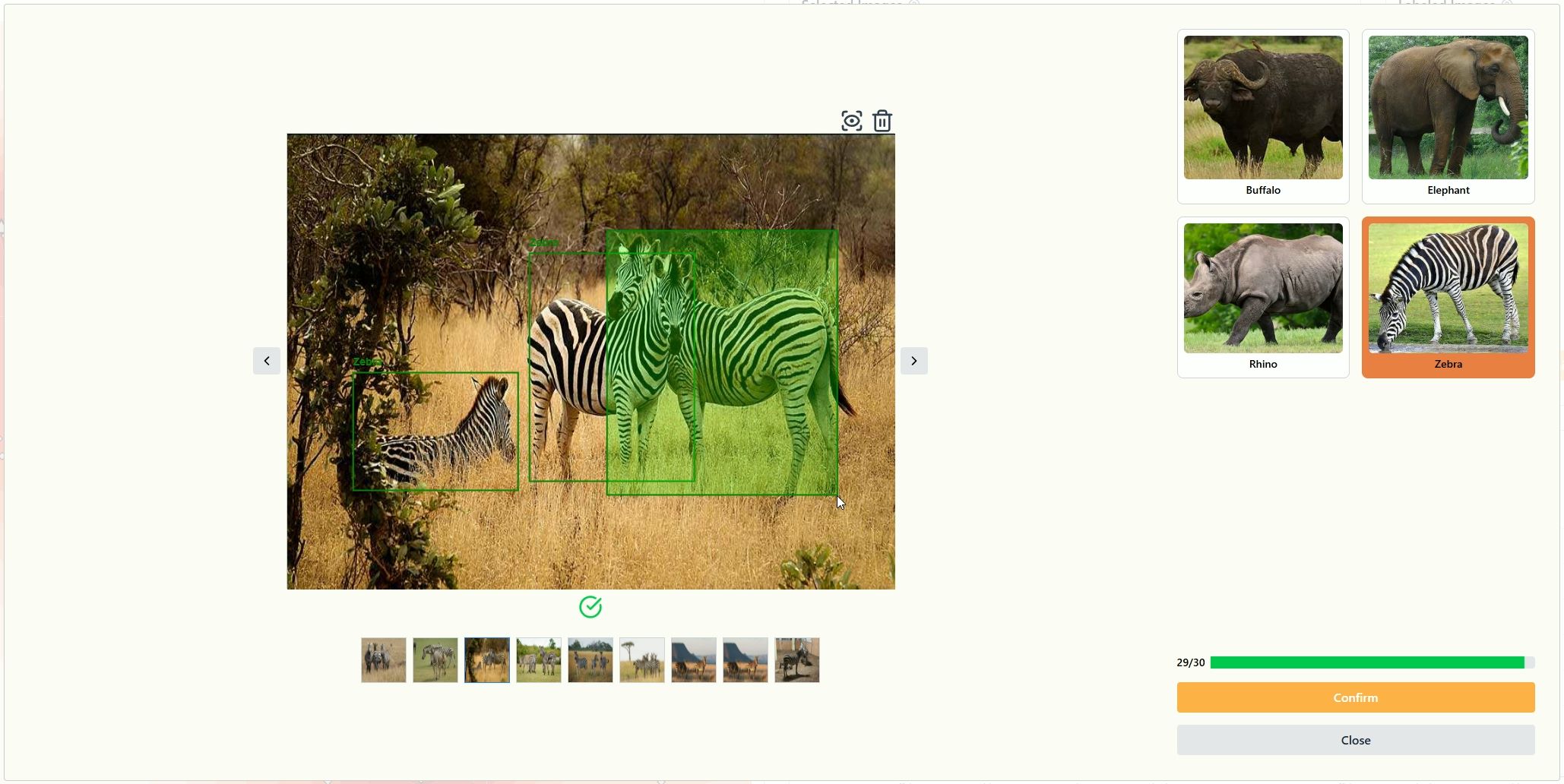}
  \end{minipage}
  \caption{Final selection and annotation for iteration 1.}
  \label{c3s3}
\end{figure}

For the final sampling of this iteration, more instances of zebras are sought. Exploring the Data View reveals a region of points in the third quadrant (see Figure \ref{c3s3}). These are clear images, containing multiple instances of zebras. Annotating these samples concludes this iteration, and the new model is trained by pressing the retrain button in the Labeled Images view. 

\subsubsection{Iteration 2}
\begin{figure}[ht!]
    \centering
    \includegraphics[width=0.4\textwidth]{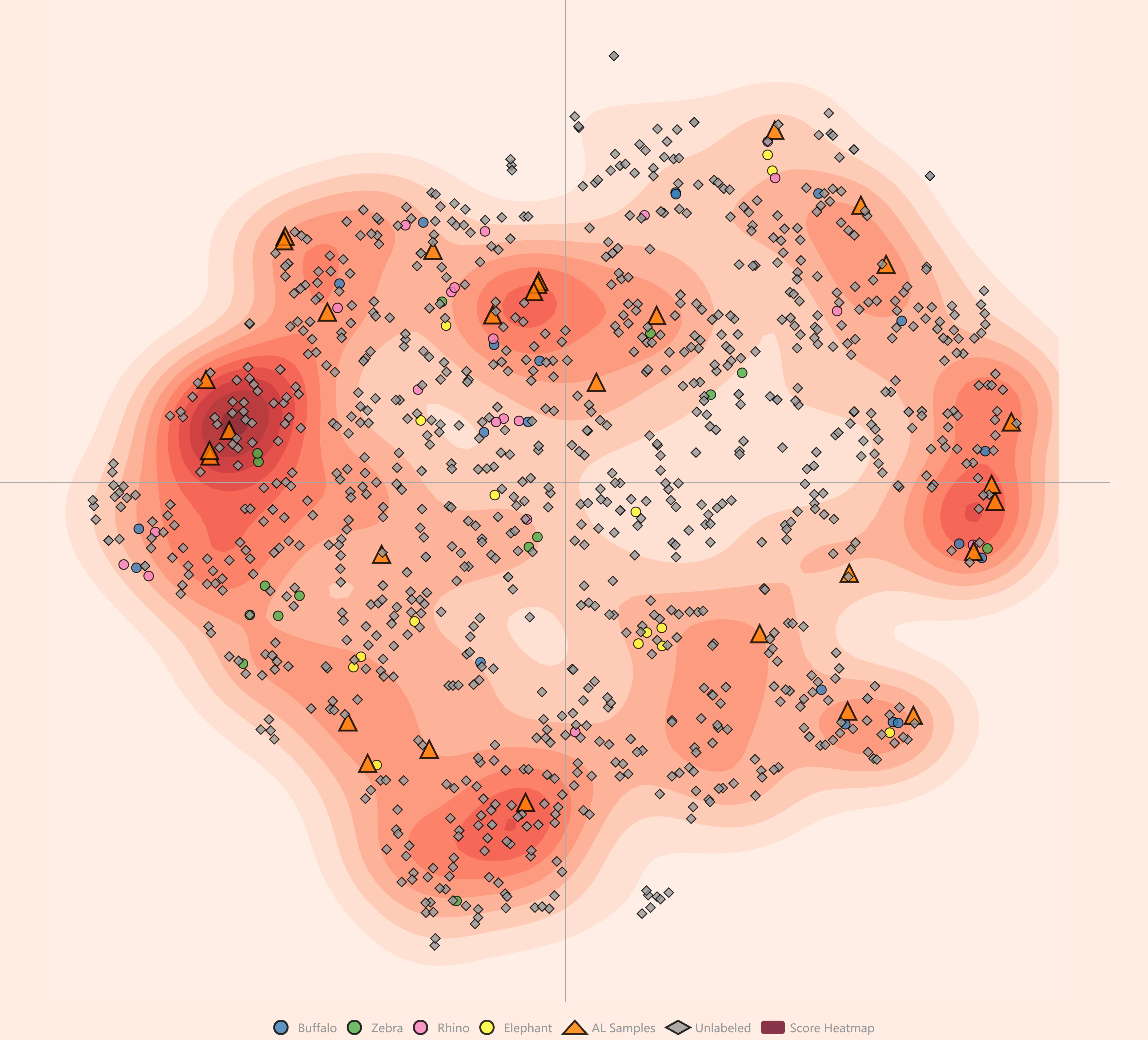}
    \caption{Data view in iteration 2.}
    \label{c3it2_dataview}
\end{figure}

\begin{figure}[ht!]
  \centering
  \begin{minipage}[t]{0.40\textwidth}
    \centering
    \includegraphics[width=\linewidth]{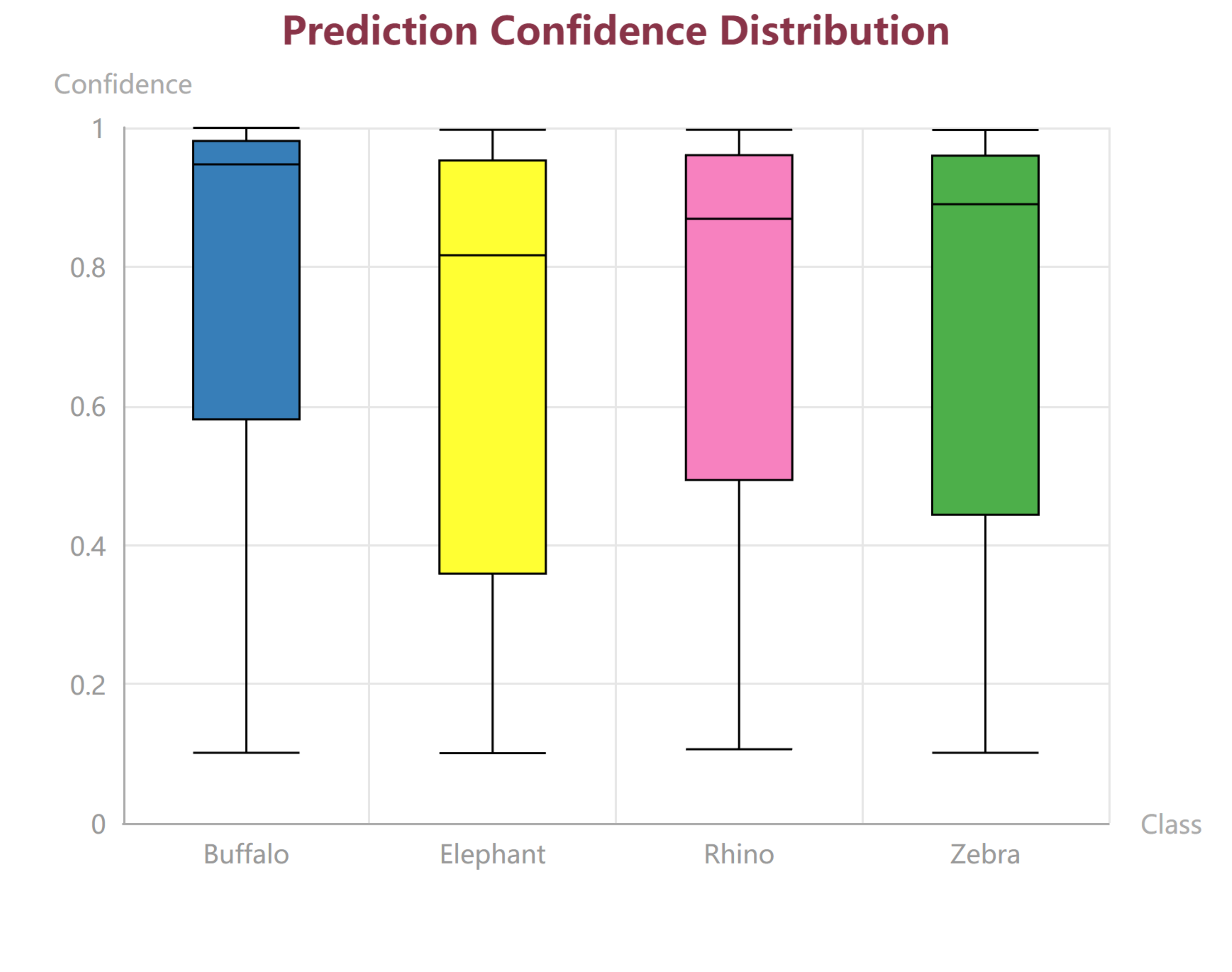}
  \end{minipage}
  \begin{minipage}[t]{0.40\textwidth}
    \centering
    \includegraphics[width=\linewidth]{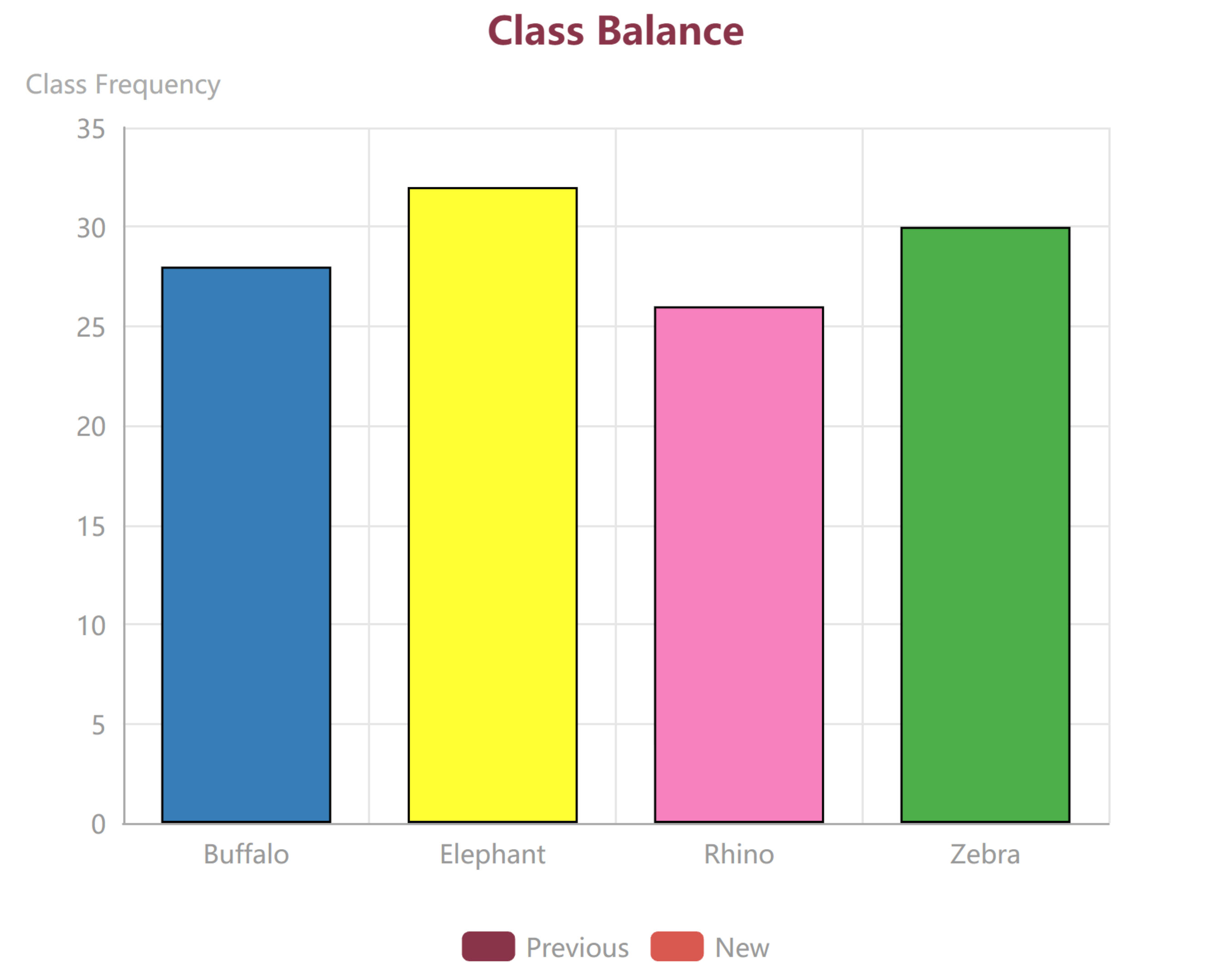}
  \end{minipage}
  \caption{Model View at the start of iteration 2.}
  \label{c3it2_modelview}
\end{figure}

The updated states in iteration 2 reveal an overall much higher prediction confidence, as seen in the Prediction Confidence Distribution chart of the model Model View in Figure \ref{c3it2_modelview}. All four classes are now similarly represented in the labeled training set. The uncertainty map in Figure \ref{c3it2_dataview} also confirms a general higher confidence, with mainly two to three distinct dark red regions indicating high uncertainty. The strategy in this iteration will be similar to iteration 1. However, with a general higher confidence in the predictions, more points from the high uncertainty regions and AL suggested samples can be considered.

To begin this training iteration, the high uncertainty region in the second quadrant is selected to inspect the samples further. A common feature in this region seems to be that the images most often include more than one instance of a class. There is also a sample of mixed classes. It is interesting to note that many very similar-looking images are clustered together. For example, there are two to three different images of elephants drinking from a pool of water grouped together. Additionally, a few similar-looking samples of elephants are in some kind of enclosed space, with man-made structures in the background. This observation highlights the benefit of using a pre-trained detection model to extract image embeddings and visualize them with dimensionality reduction. This selection and the preview of the images can be seen below in  Figure \ref{c3it2_selections}.

\begin{figure}[ht!]
  \centering
  \begin{minipage}[c]{0.4\textwidth}
    \centering
    \includegraphics[width=\linewidth]{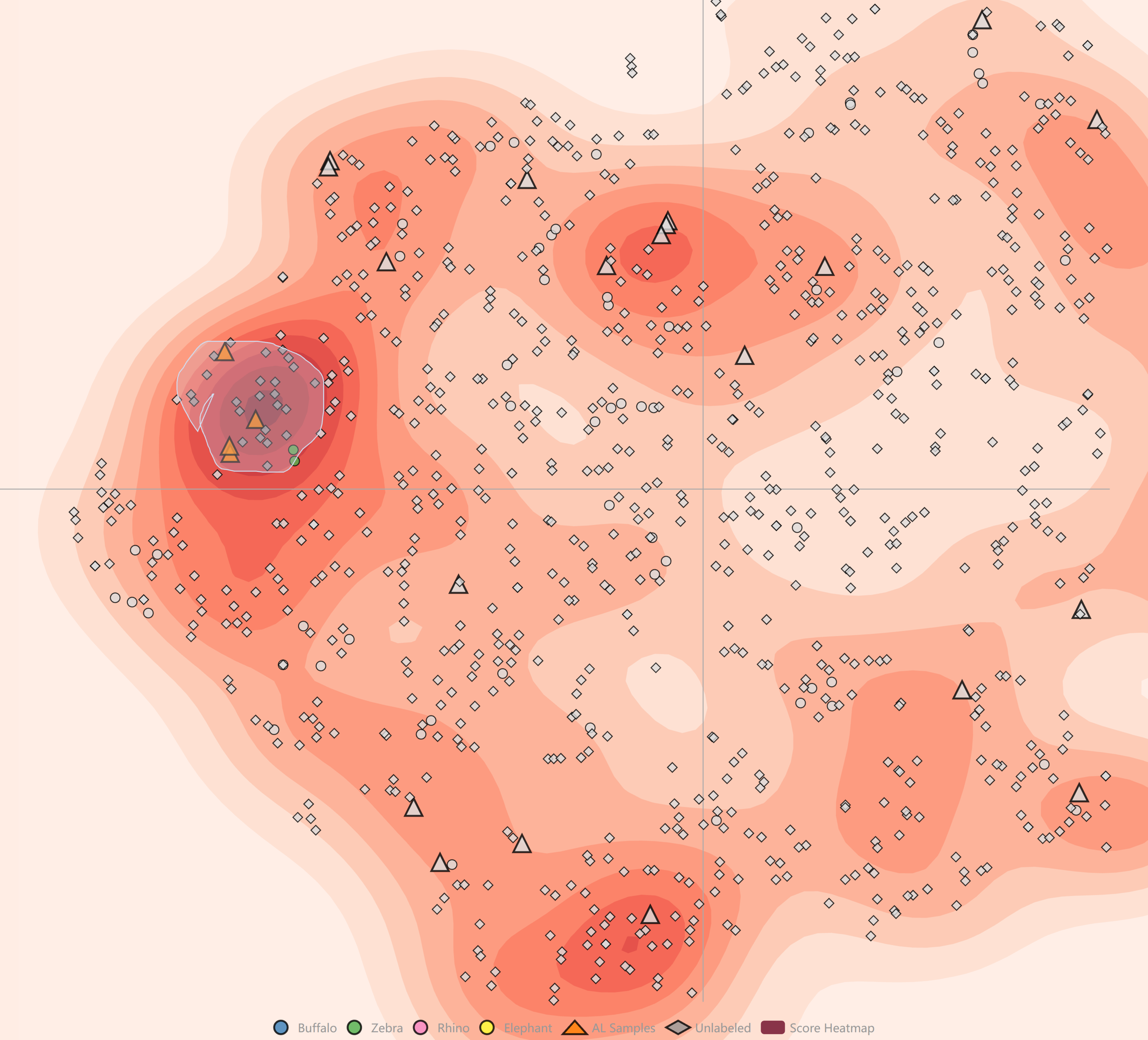}
  \end{minipage}
  \begin{minipage}[c]{0.4\textwidth}
    \centering
    \includegraphics[width=\linewidth]{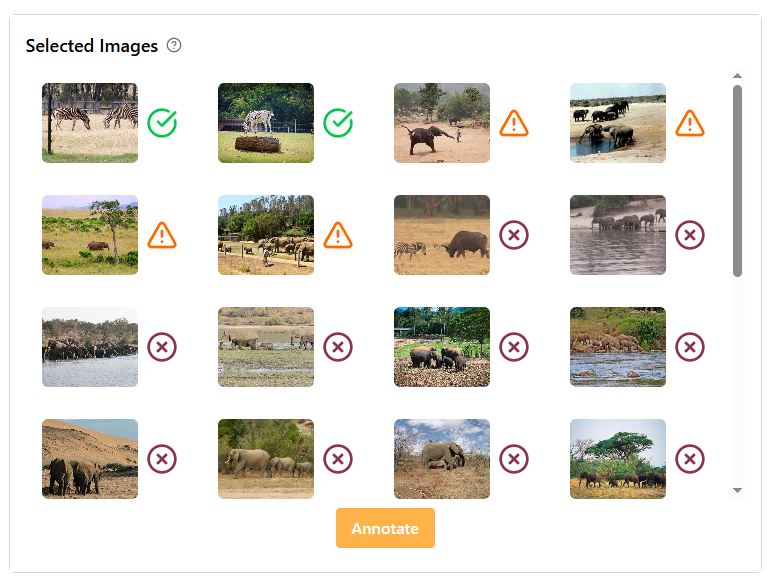}
  \end{minipage}
  \caption{Selection of a high uncertainty region in the first quadrant.}
  \label{c3it2_selections}
\end{figure}

\begin{figure}[ht!]
    \centering
    \includegraphics[width=0.6\textwidth]{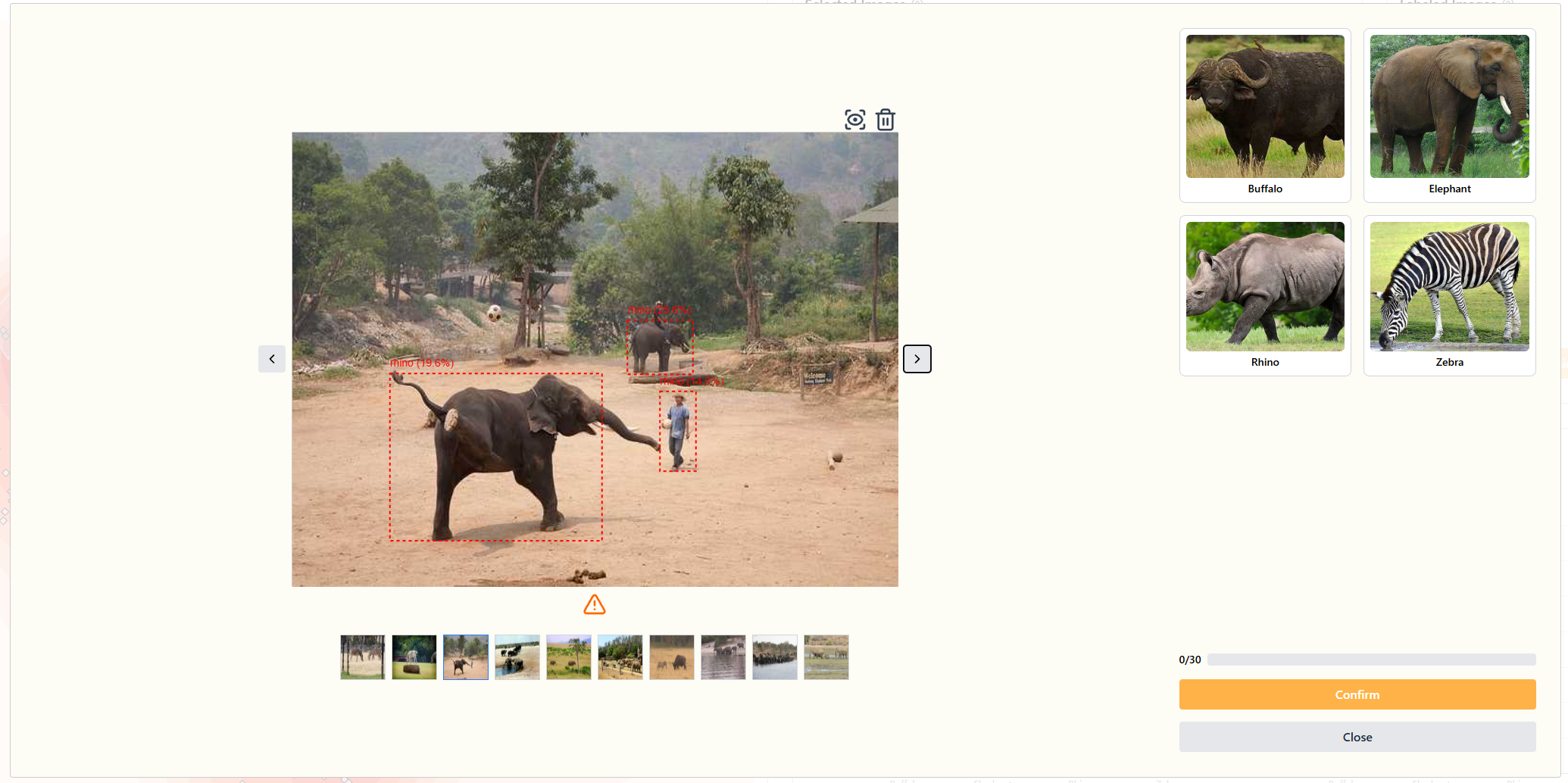}
    \caption{AL suggested sample with misclassified predictions made by the model.}
    \label{c3it2_wrong}
\end{figure}

\newpage
Inspecting the model-made bounding boxes on the individual images also gives insight into the uncertainty in this region. The model has not yet learned to distinguish elephants and rhinos in this type of setting, as seen in Figure \ref{c3it2_wrong}. Eight images are annotated in this region, hoping it will be enough to increase the overall confidence in this area and, consequently, the performance of the overall model.

\begin{figure}[ht!]
  \centering
  \begin{minipage}[c]{0.4\textwidth}
    \centering
    \includegraphics[width=\linewidth]{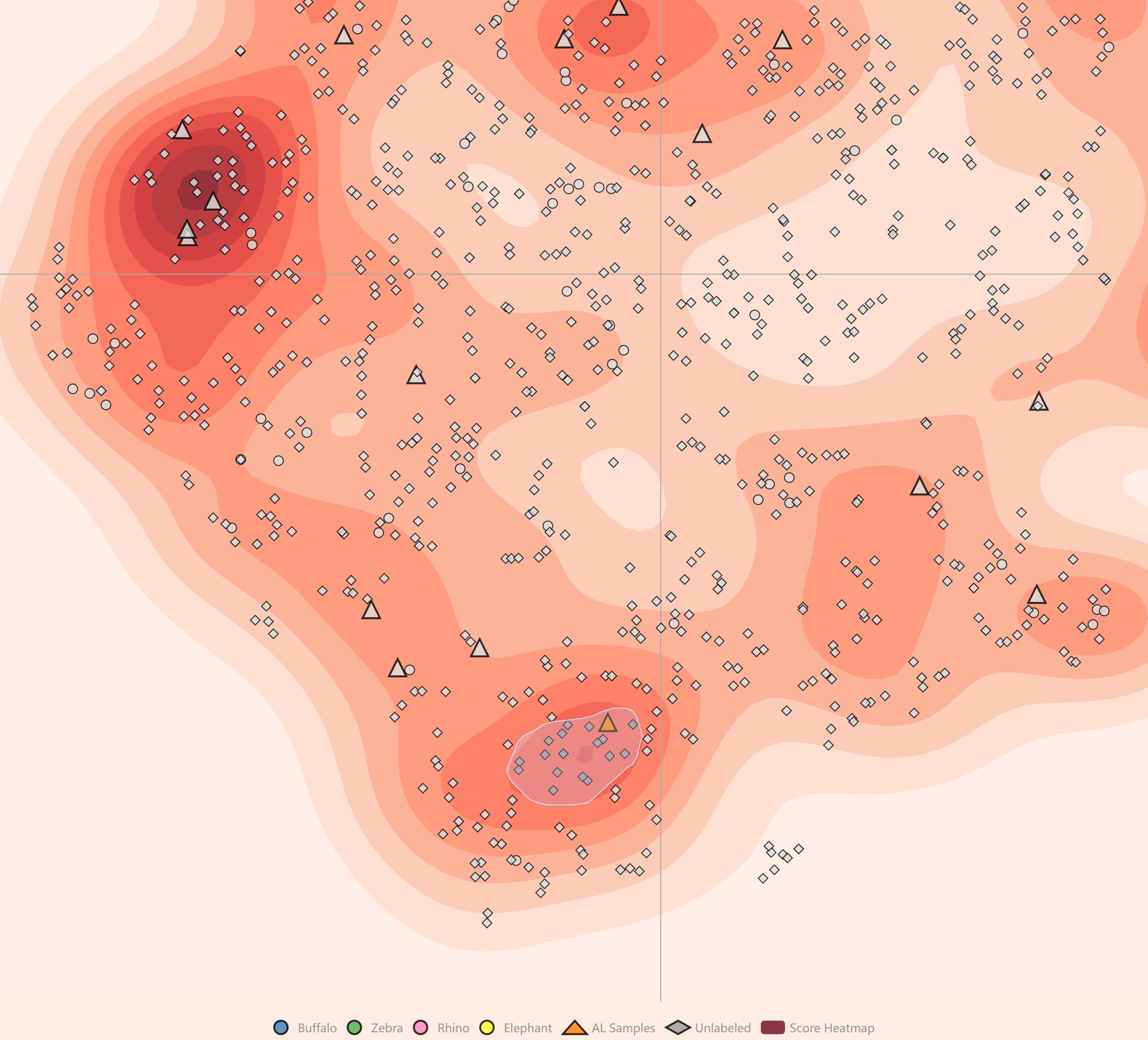}
  \end{minipage}
  \begin{minipage}[c]{0.4\textwidth}
    \centering
    \includegraphics[width=\linewidth]{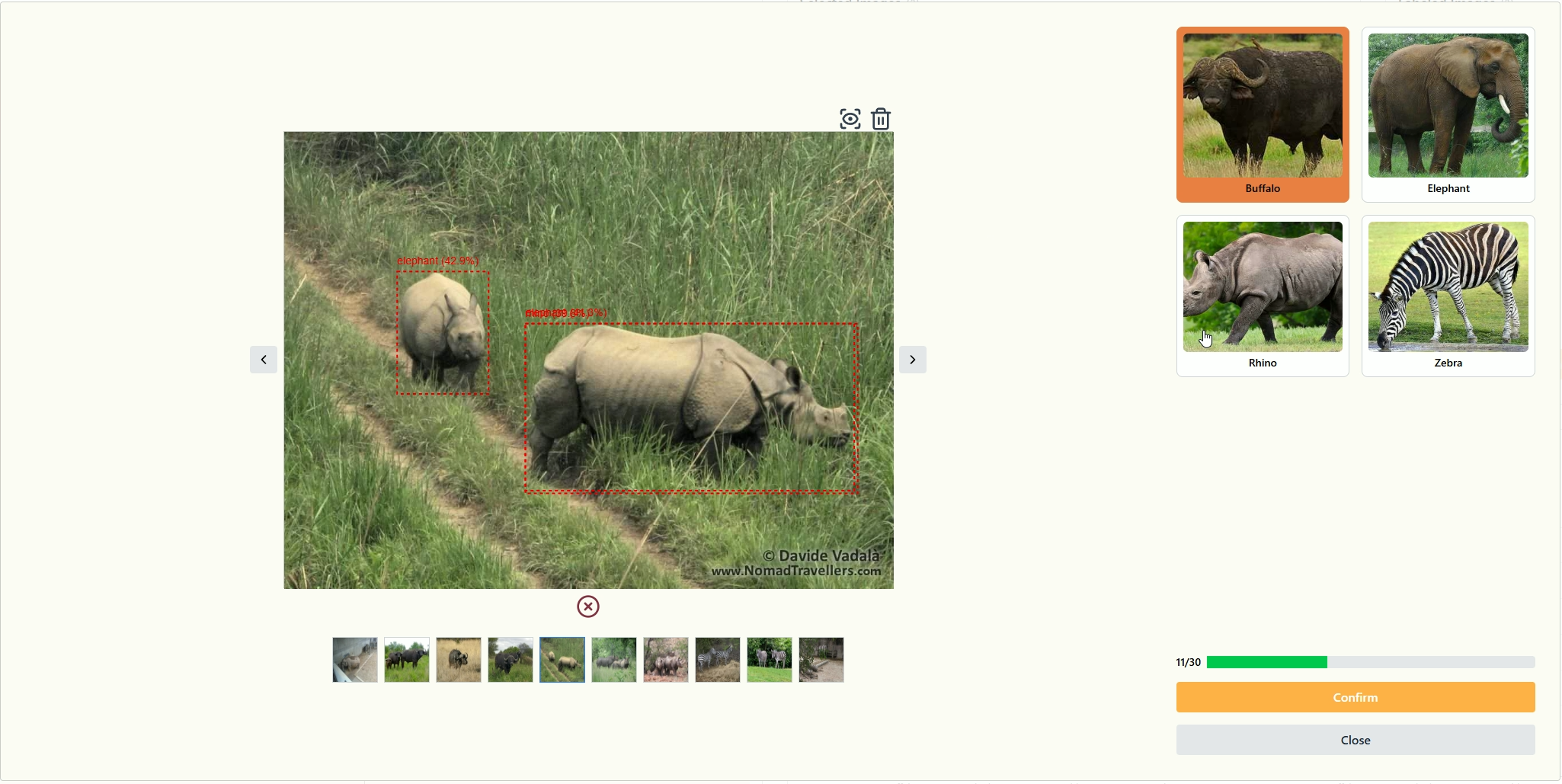}
  \end{minipage}
  \caption{Second selection and another misclassified sample.}
  \label{c3i2_wrong2}
\end{figure}

Moving on, a new selection is made in the bottom middle of the Data View visualization. This is a region of high uncertainty. It includes one AL sample, and additionally, no images have been labeled in or around this region previously. This selection contains mostly rhinos but also samples of buffalos. The same type of misclassifications is observed in this region as well (Figure \ref{c3i2_wrong2}), where the model is predicting rhinos as elephants. The rest of iteration 2 continues in the same manner, targeting unlabeled regions and uncertain samples. 

\newpage
\subsubsection{Iteration 3}
\begin{figure}[ht!]
    \centering
    \includegraphics[width=0.4\textwidth]{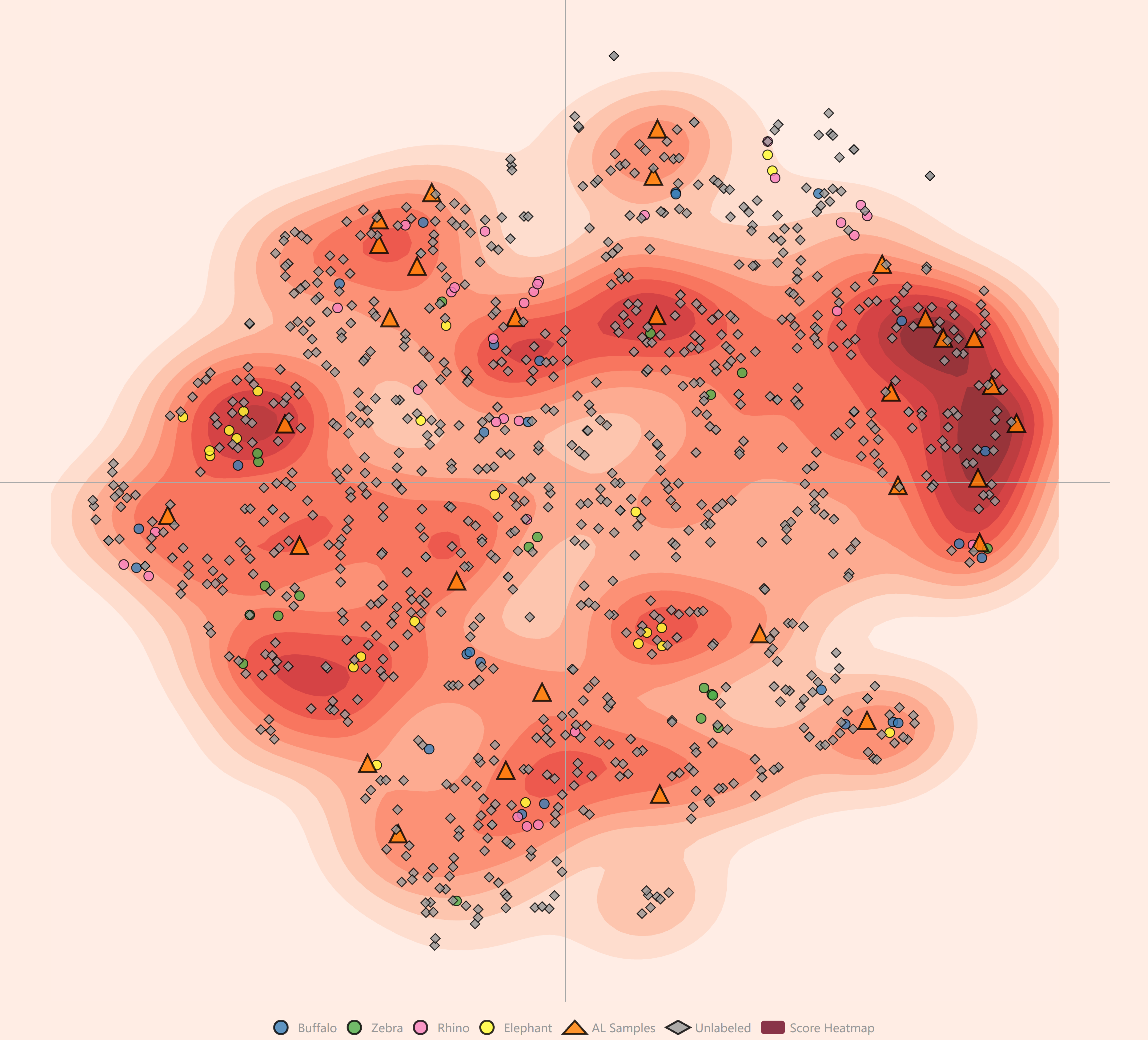}
    \caption{Data View start of iteration 3.}
    \label{c3it3_dataview}
\end{figure}

\begin{figure}[ht!]
  \centering
  \begin{minipage}[c]{0.4\textwidth}
    \centering
    \includegraphics[width=\linewidth]{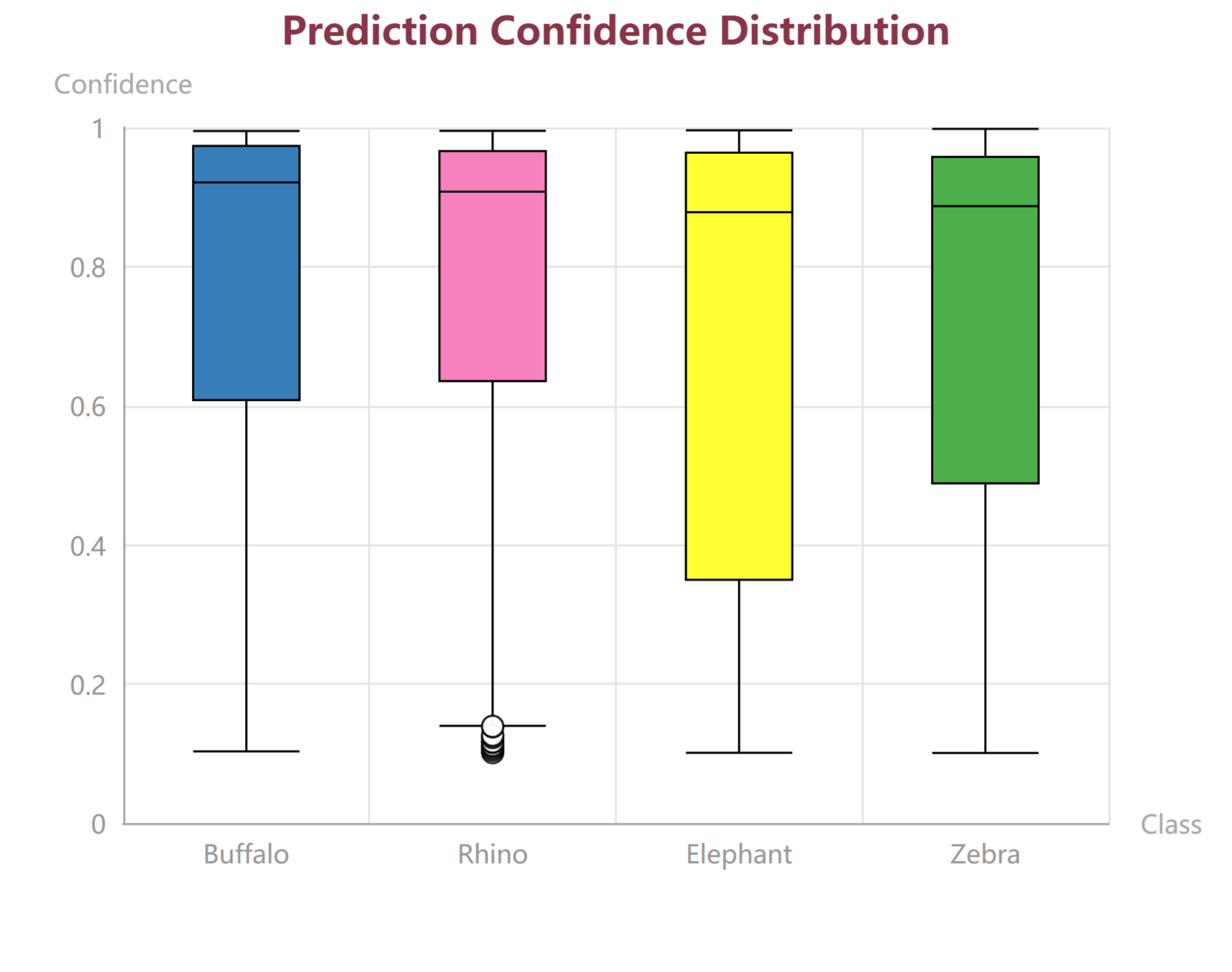}
  \end{minipage}
  \begin{minipage}[c]{0.4\textwidth}
    \centering
    \includegraphics[width=\linewidth]{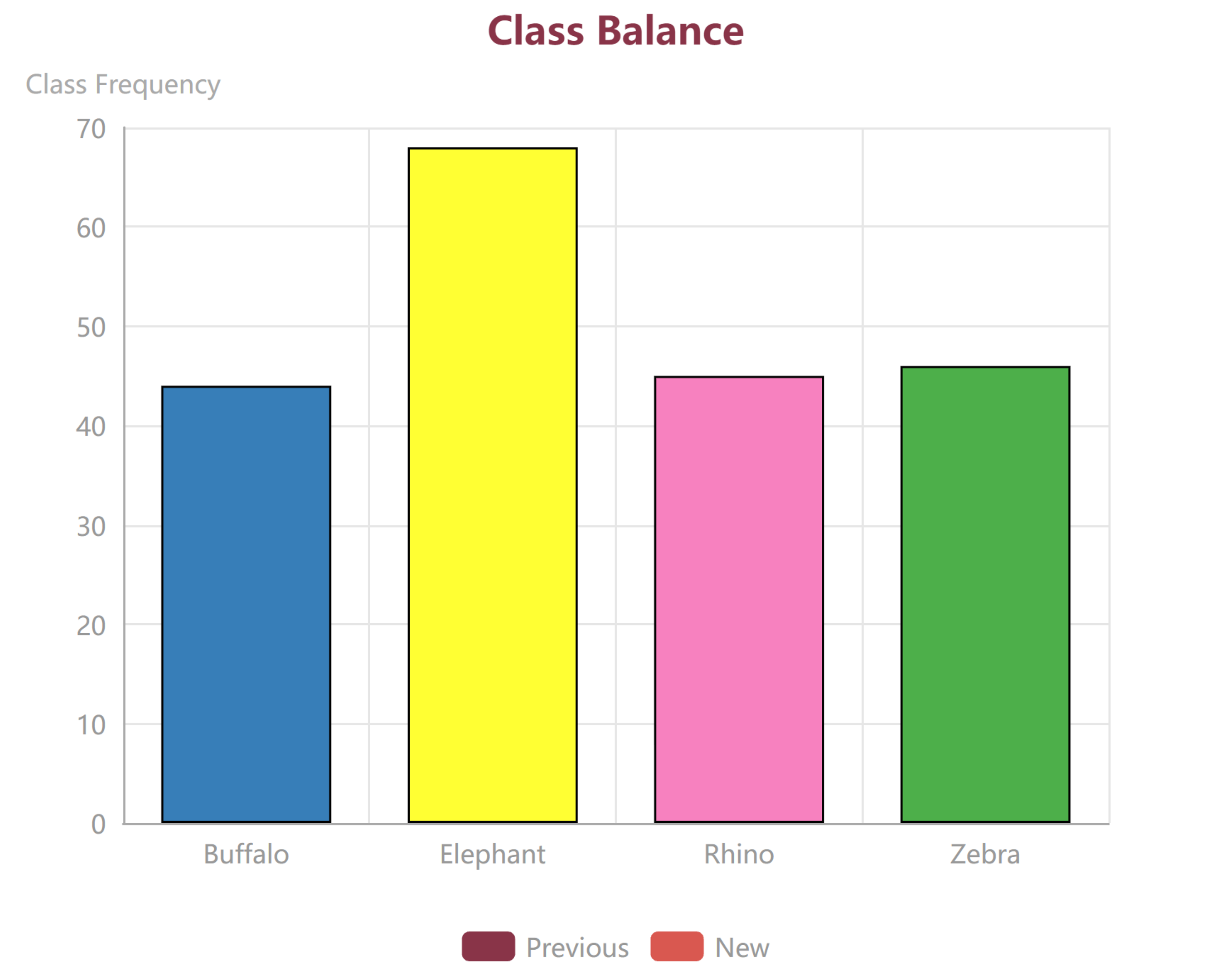}
  \end{minipage}
  \caption{Model View start of iteration 3.}
  \label{c3it3_modelview}
\end{figure}

Looking at the updated Data View in Figure \ref{c3it3_dataview}, the labeling efforts made in the red region of quadrant two do not seem to have had a great impact, at least not concerning the heatmap uncertainty, as a dark red region is still visible here. Overall, there seems to be slightly more uncertainty than previously. It is worth noting that this does not necessarily mean that the model's performance is worse, as several aspects have to be taken into consideration. For example, the model might now be correctly detecting more of the objects, but it is still not confident. Additionally, in the previous iteration, some false predictions were observed that could now have been corrected. The Prediction Confidence box plot in Figure \ref{c3it3_modelview} shows an increasing confidence compared to the previous iteration. For the Buffalo and Rhino classes, most of the confidence distribution lies over 0.6. The training set is not perfectly balanced, as many instances of elephants were labeled in the previous iteration.

\begin{figure}[ht!]
  \centering
  \begin{minipage}[c]{0.4\textwidth}
    \centering
    \includegraphics[width=\linewidth]{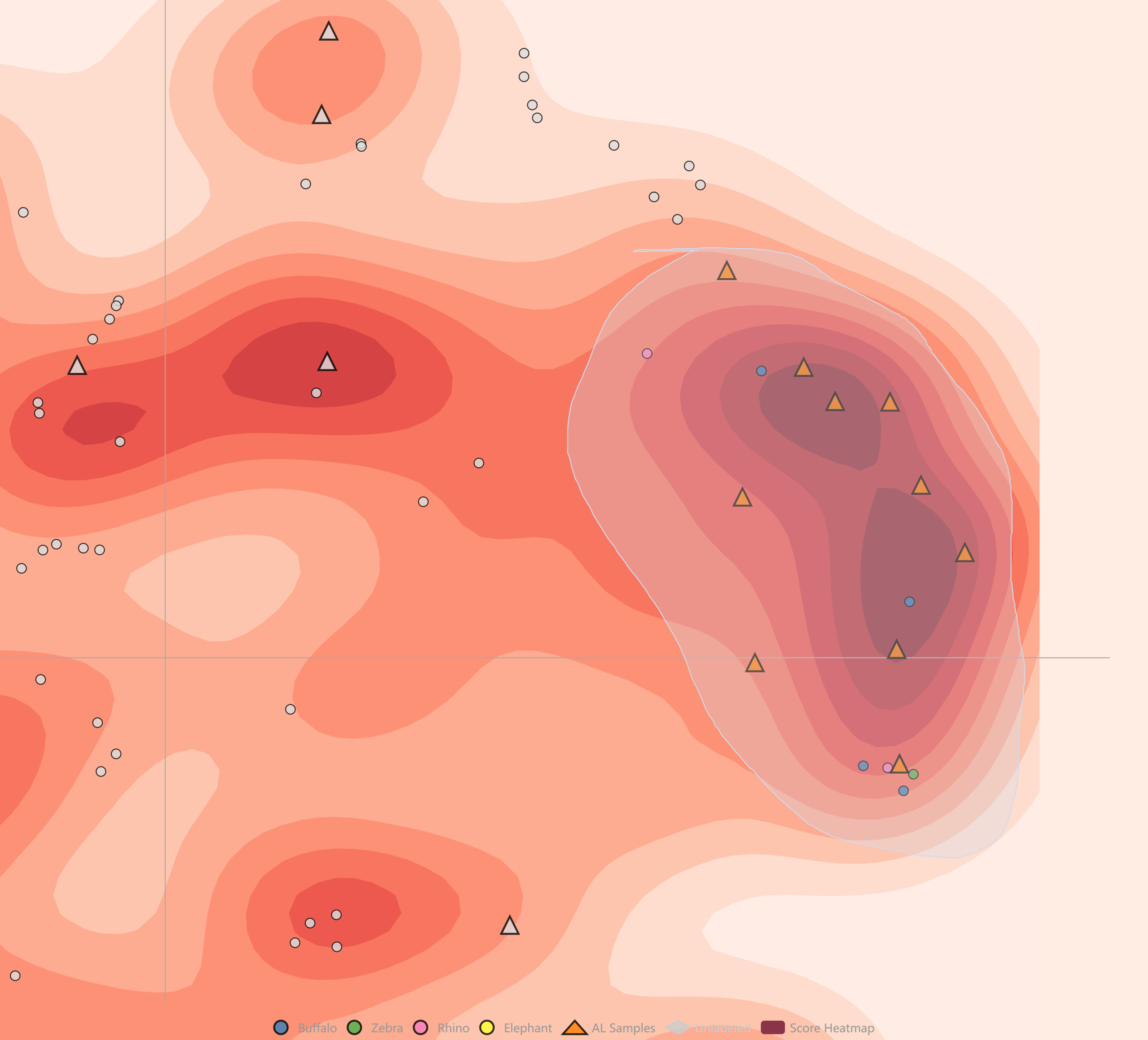}
  \end{minipage}
  \begin{minipage}[c]{0.4\textwidth}
    \centering
    \includegraphics[width=\linewidth]{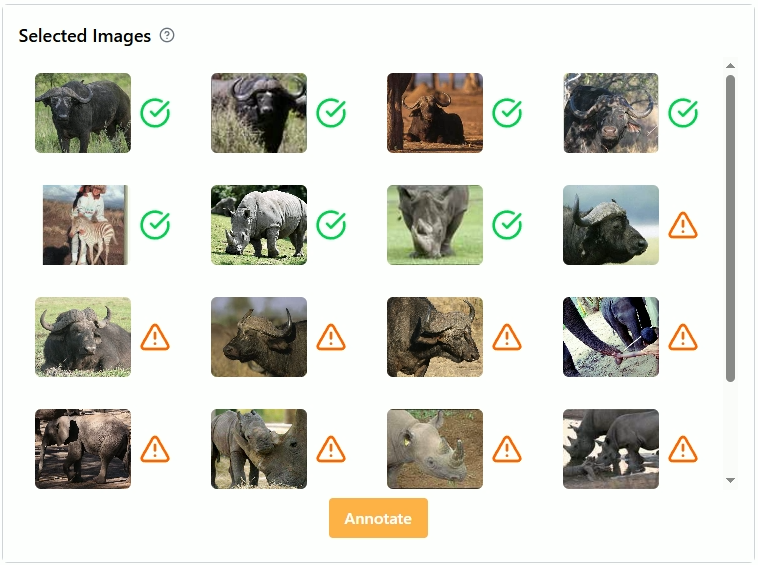}
  \end{minipage}
  \caption{Selecting AL samples in a high uncertainty region.}
  \label{c3it_s1}
\end{figure}

\newpage
The selection process this iteration will focus more on the AL samples and on trying to find samples considered hard to predict, for example, samples containing many instances of a class or multi-class samples. The process starts by highlighting the samples in and around the dark red area in the far right of the Data View, as seen in Figure \ref{c3it_s1}. It's a large selection containing a mix of samples. Eight of these samples are annotated and added to the labeled training set.

\begin{figure}[ht!]
  \centering
  \begin{minipage}[c]{0.49\textwidth}
    \centering
    \includegraphics[width=\linewidth]{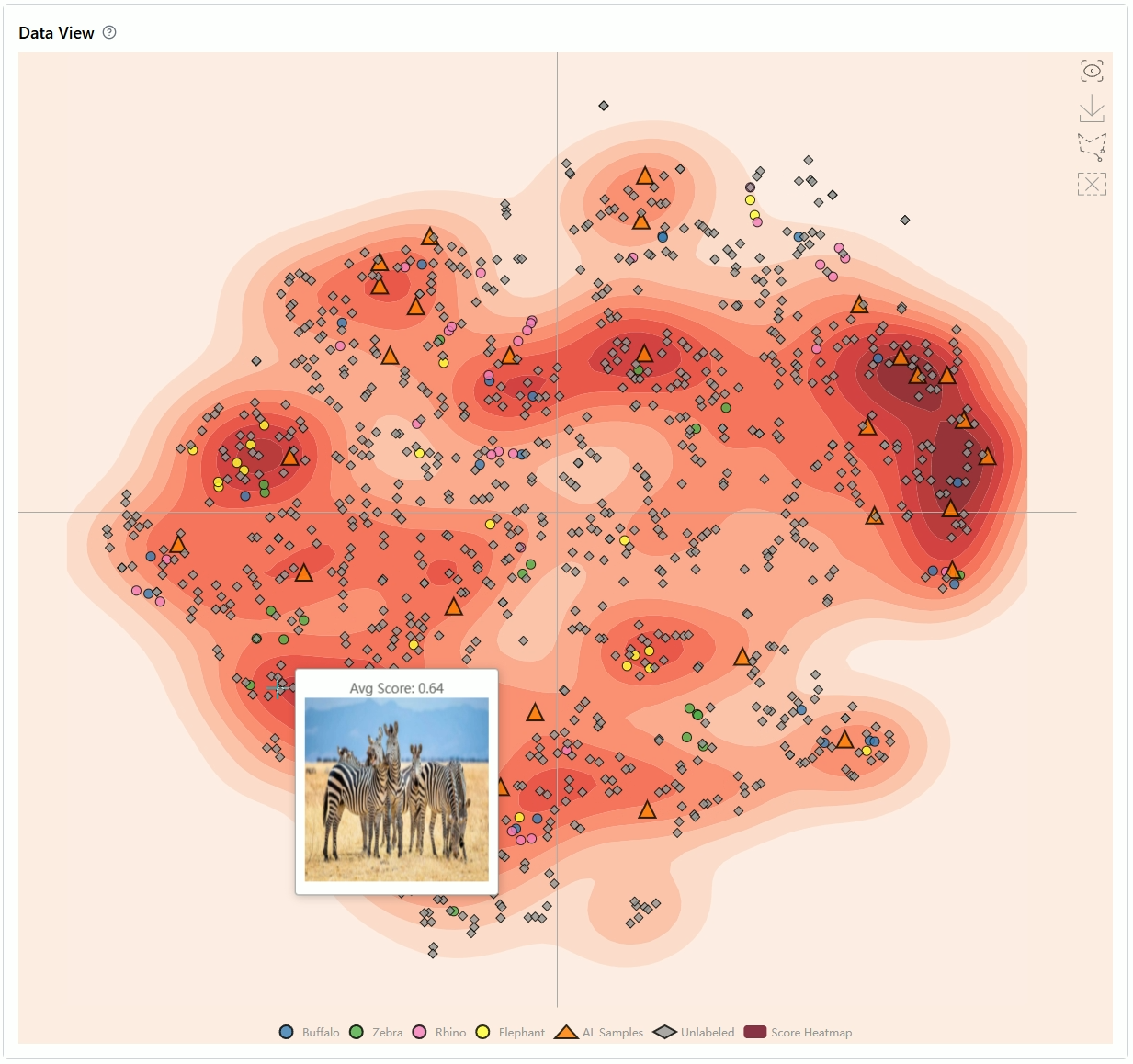}
    \caption{Exploring the Data View looking for samples of zebras.}
    \label{zebra}
  \end{minipage}
  \hfill
  \begin{minipage}[c]{0.49\textwidth}
    \centering
    \includegraphics[width=\linewidth]{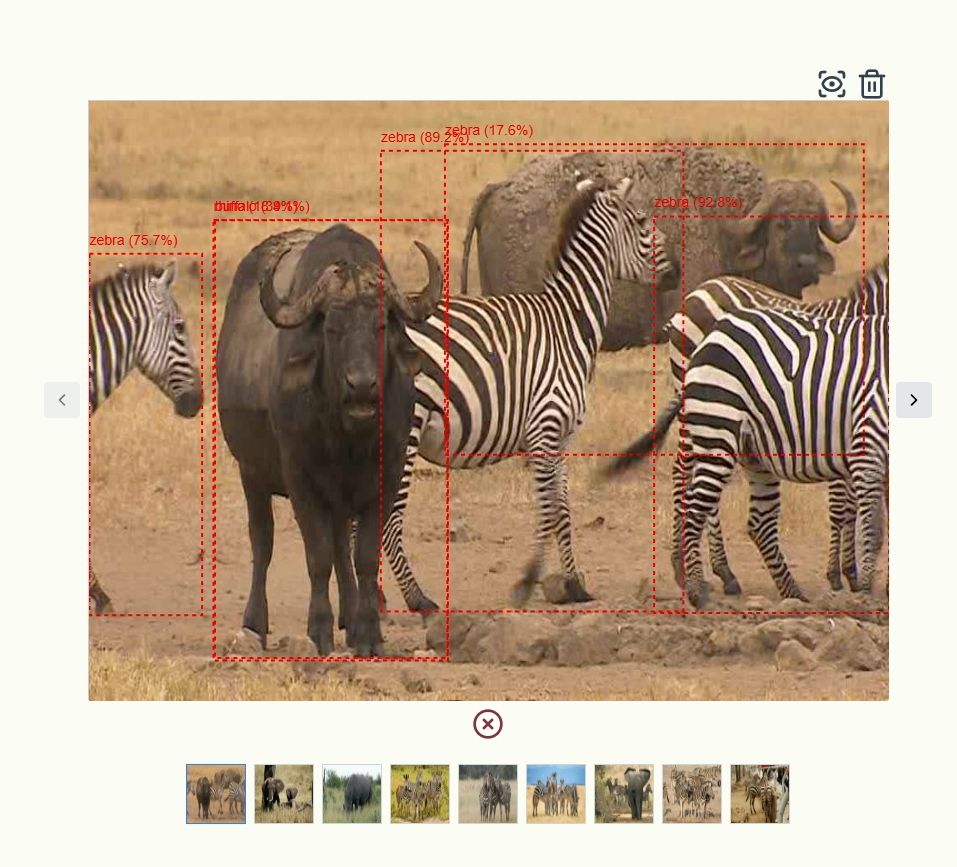}
    \caption{A sample containing both buffalos and zebras}
    \label{zebra_buffalo}
  \end{minipage}
\end{figure}

Following this, a search for samples of zebras that might exhibit higher uncertainty is conducted. By hovering over the data points, a region containing mostly zebras is found in the third quadrant. These are inspected more closely, as seen in Figure \ref{zebra}. Two of the samples in this region are multi-class instances, which could be important to label to build a robust model. One of them contains both zebras and buffalos, as seen in Figure \ref{zebra_buffalo}. The model does a decent job of predicting the classes visible in this picture, but it is not perfect. It has made three detections of zebras and one buffalo with higher confidence. However, there are four zebras and two buffalos present in the image. Additionally, the localization of the detections is not great, leaving room for improvement.

\begin{figure}[ht!]
    \centering
    \includegraphics[width=0.5\textwidth]{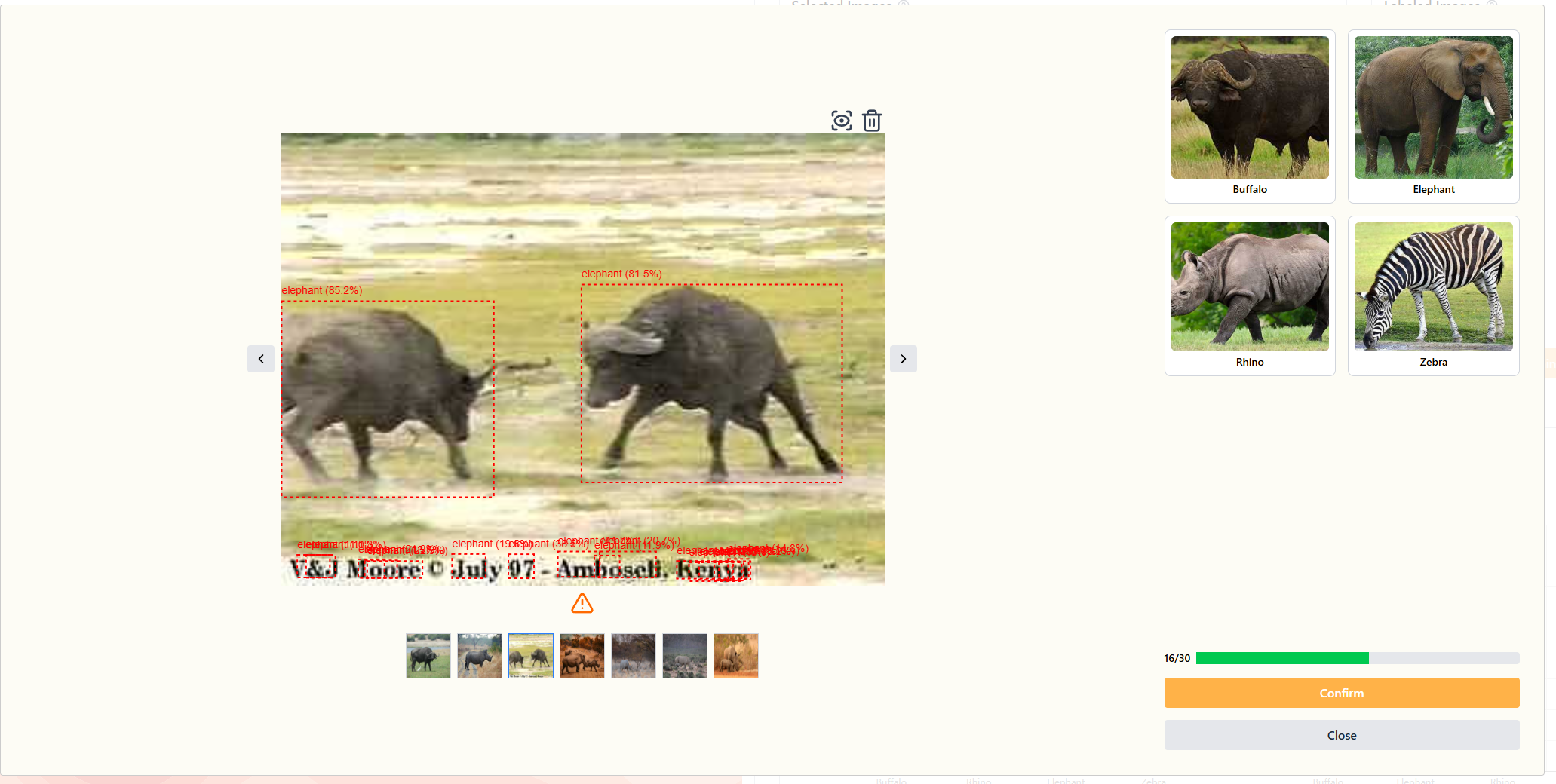}
    \caption{A discarded AL sample with noisy features.}
    \label{watermark_buffalo}
\end{figure}

\newpage
Furthermore, the exploration and annotation of the AL samples continues. One of the AL samples is a very noisy sample (Figure \ref{watermark_buffalo}). It is a low-resolution image, also containing watermark text, which seems to confuse the model significantly. Samples like these are not labeled, as there is no desire to introduce these kinds of structures to the model. In purely automatic uncertainty AL sampling, images like these could be added to the labeled training set, which might deteriorate the performance of the model.

\subsubsection{Iteration 4}
\begin{figure}[ht!]
    \centering
    \includegraphics[width=0.5\textwidth]{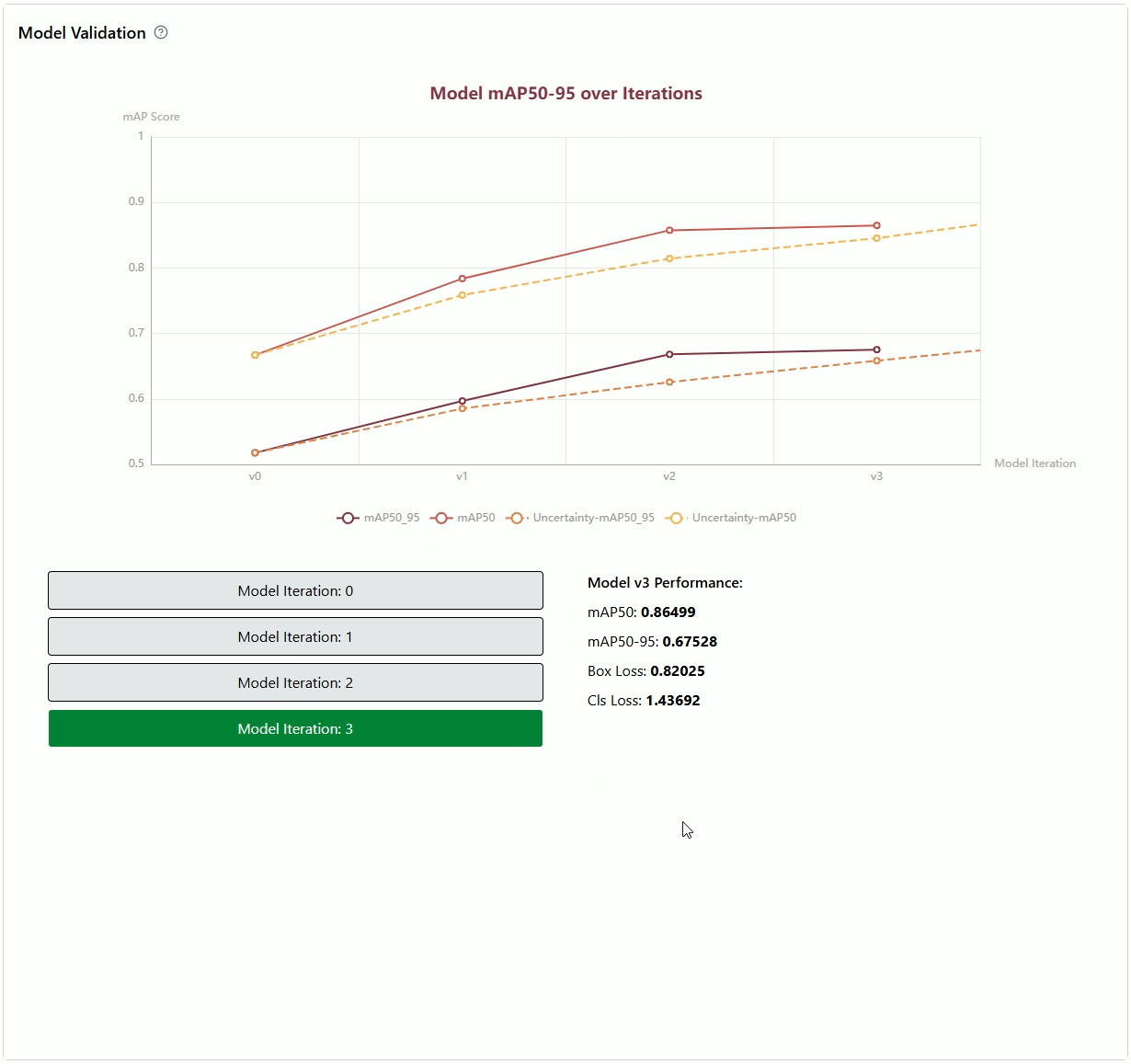}
    \caption{The Model Validation view after performing three rounds of iterative training.}
    \label{model_val}
\end{figure}

At the start of iteration 4, a total of 90 samples have been added to the training set. To get a glimpse of how the overall model performance has progressed over the iterations, the Model Validation View in Figure \ref{model_val}. Here, the mAP metric of the model as evaluated during training on the validation set can be viewed. The trained model is represented by the filled line segments in the chart, and the dotted lines represent the automatic AL simulation. The trained model had a very good performance trajectory in the first two training iterations, but the improvement slowed down in the third iteration. Perhaps too much consideration was put on the AL samples that round. Thus, an attempt will be made to reverse this trend by moving back to a more balanced approach in this round, exploring the Data View to see if there are any regions still having no labeled samples.

\begin{figure}[ht!]
    \centering
    \includegraphics[width=0.4\textwidth]{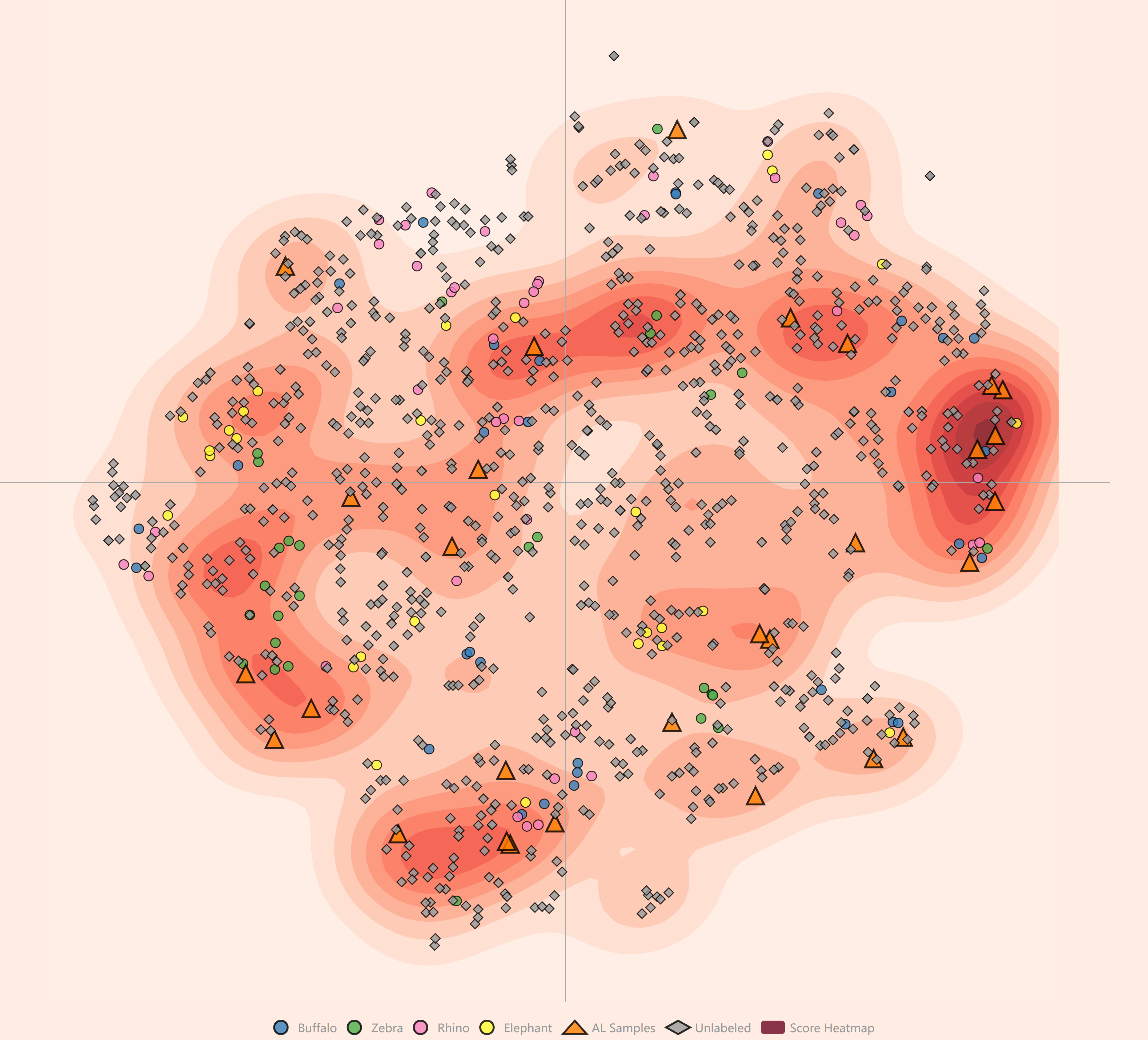}
    \caption{Data View start of iteration 4.}
    \label{c3it4_dataview}
\end{figure}

\begin{figure}[ht!]
  \centering
  \begin{minipage}[c]{0.4\textwidth}
    \centering
    \includegraphics[width=\linewidth]{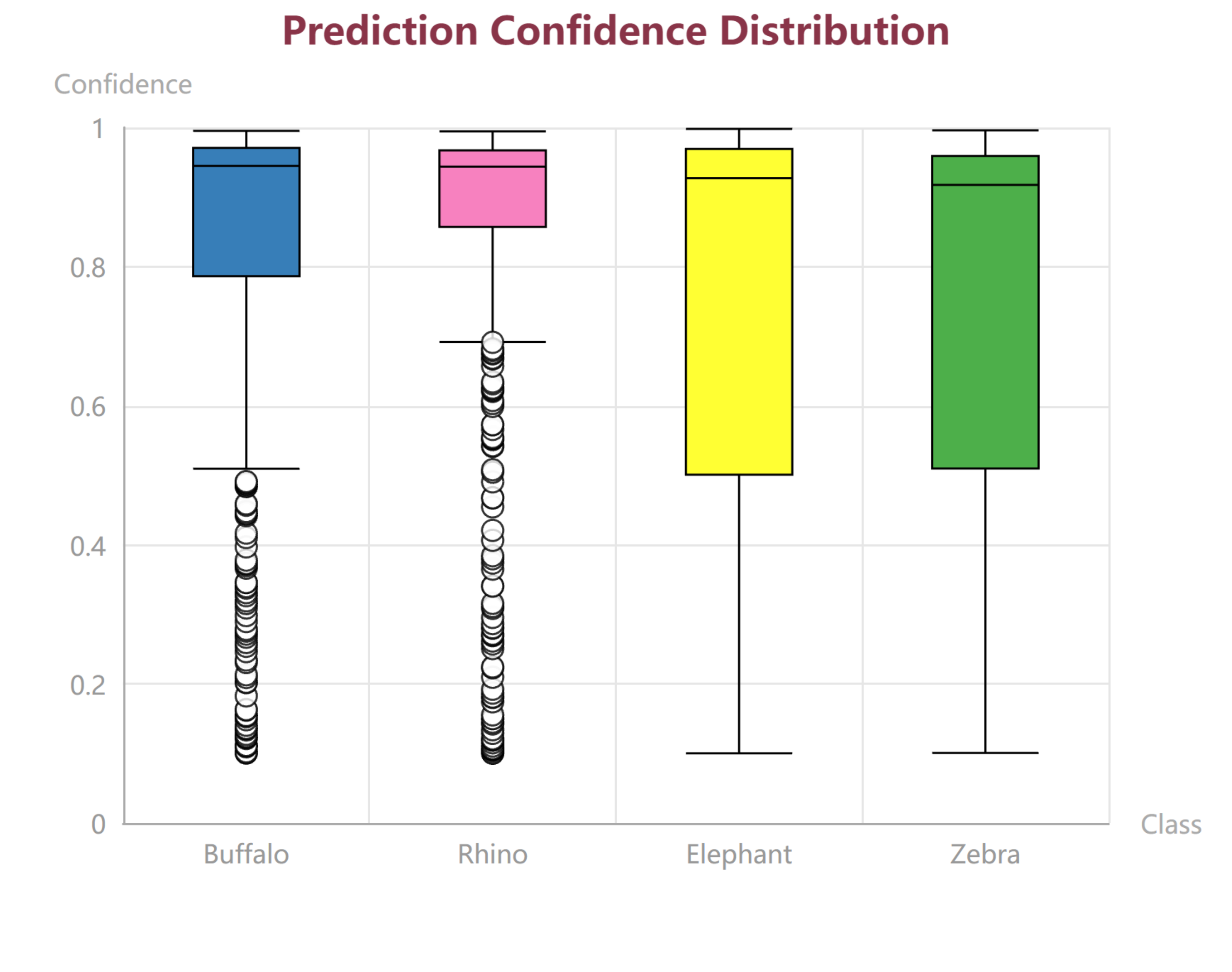}
  \end{minipage}
  \begin{minipage}[c]{0.4\textwidth}
    \centering
    \includegraphics[width=\linewidth]{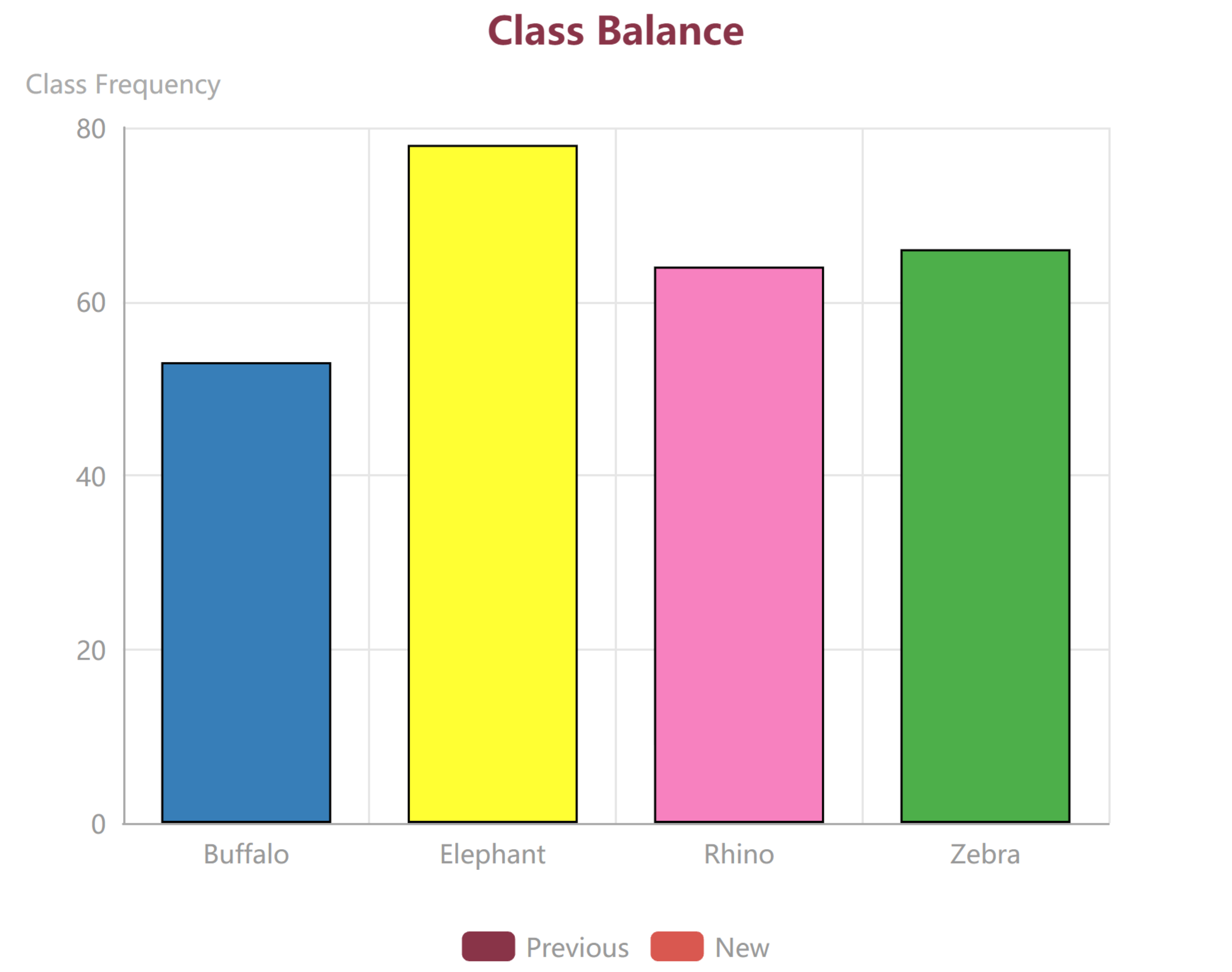}
  \end{minipage}
  \caption{Model View start of iteration 4.}
  \label{c3it4_modelview}
\end{figure}

The heatmap landscape in Figure \ref{c3it4_dataview} however, does look better than in the previous iteration, with only one region of the most intense red color. The confidence Distribution in \ref{c3it4_modelview} for classes Buffalo and rhino has also continued to improve, while elephants and zebras continue to lag a bit behind. However, it is important to consider all the provided information. The unlabeled dataset could simply be more biased towards easier samples of buffalo and rhinos. As noticed in previous iterations, there seem to exist many samples containing multiple instances of elephants and zebras, which could contribute to this discrepancy.

\begin{figure}[ht!]
  \centering
  \begin{minipage}[c]{0.40\textwidth}
    \centering
    \includegraphics[width=\linewidth]{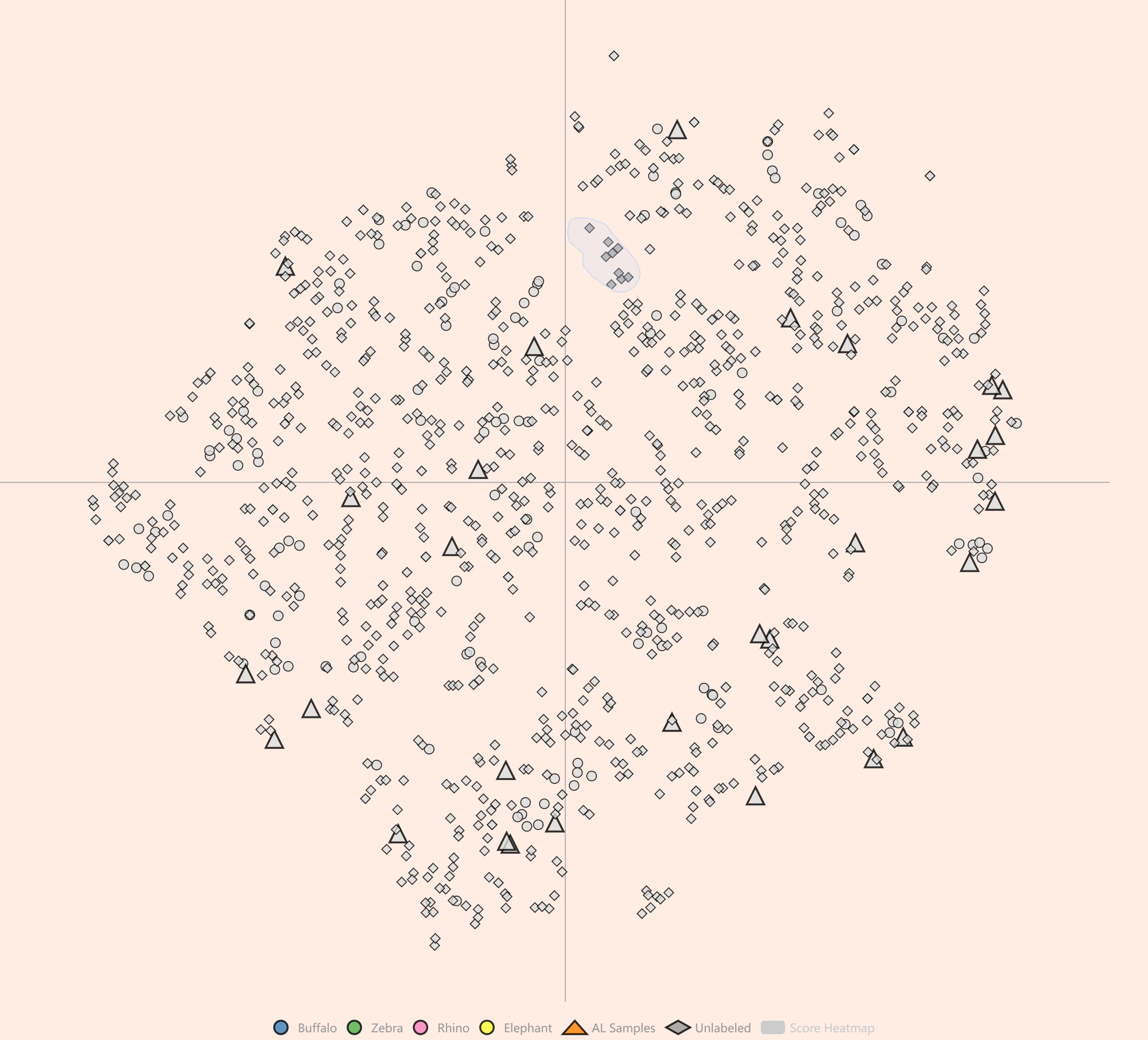}
  \end{minipage}
  \begin{minipage}[c]{0.48\textwidth}
    \centering
    \includegraphics[width=\linewidth]{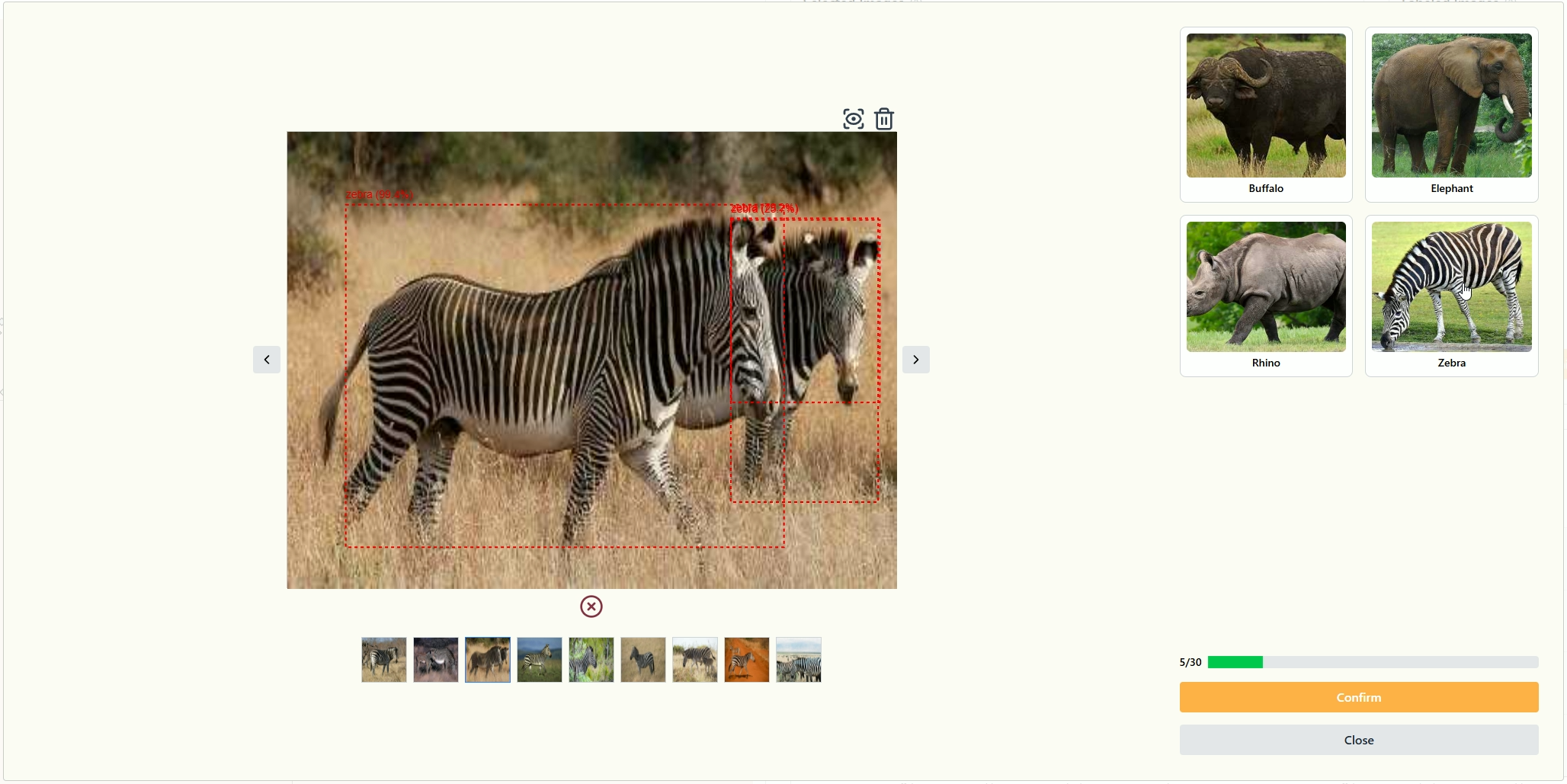}
  \end{minipage}
  \caption{A selection of points in the fourth iteration and the inspection of a sample of zebras stacked together.}
  \label{c3it4s1}
\end{figure}

\newpage
By momentarily turning off the uncertainty heatmap, sections that have not been labeled in the previous iterations can be more easily seen. A selection of a smaller cluster of samples is made in the first quadrant, loosely separated from other points (see Figure \ref{c3it4s1}).  This cluster depicts images of zebras. Upon inspection, the model is, in large already doing quite well in detecting these zebras. However, it struggles with localization when multiple zebras are stacked together, as also seen in Figure \ref{c3it4s1}. It seems reasonable that these kinds of samples are harder for the model. The colors of the zebras naturally blend very well together, making them harder to separate.

\begin{figure}[ht!]
  \centering
  \begin{minipage}[c]{0.40\textwidth}
    \centering
    \includegraphics[width=\linewidth]{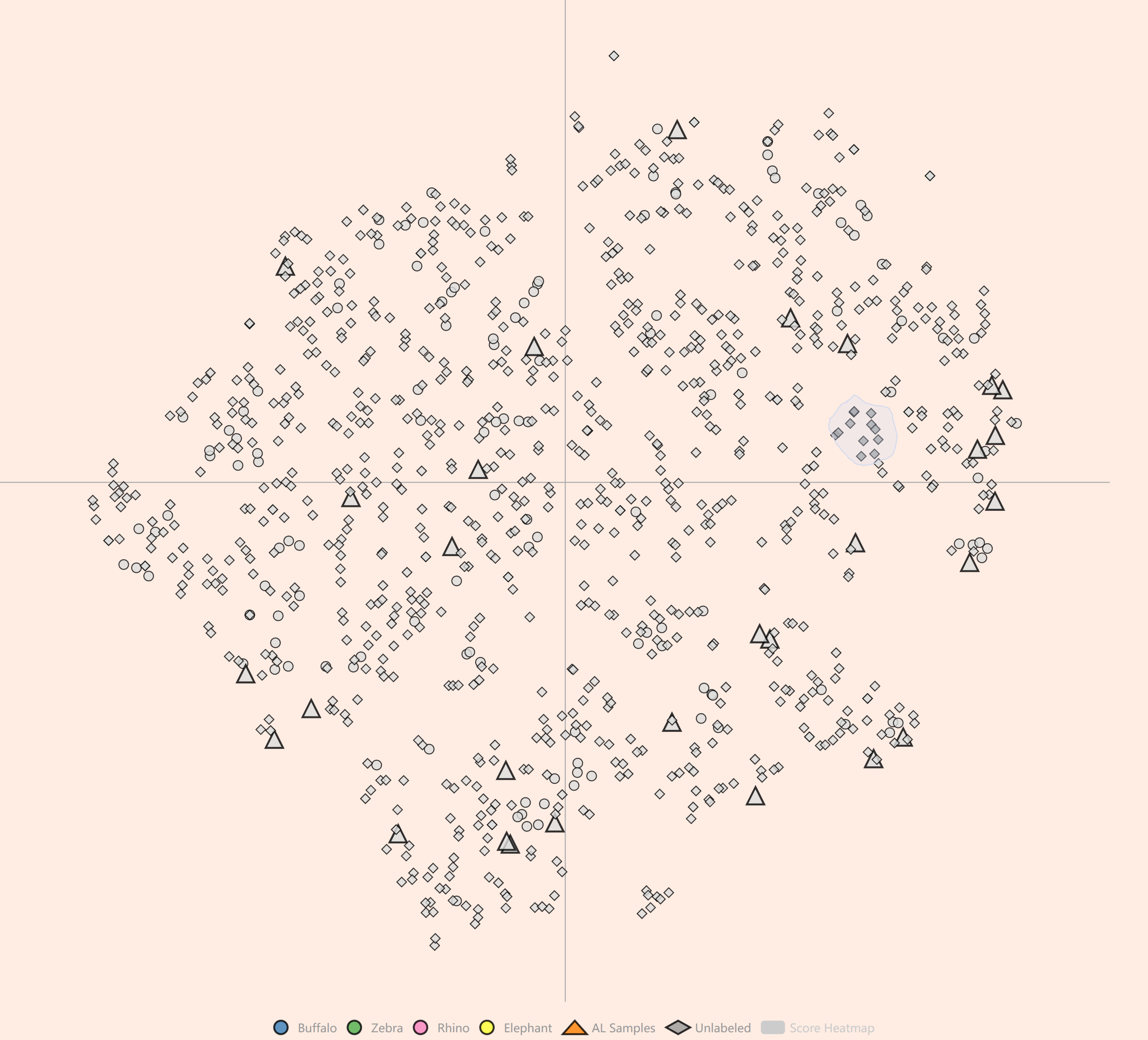}
  \end{minipage}
  \begin{minipage}[c]{0.48\textwidth}
    \centering
    \includegraphics[width=\linewidth]{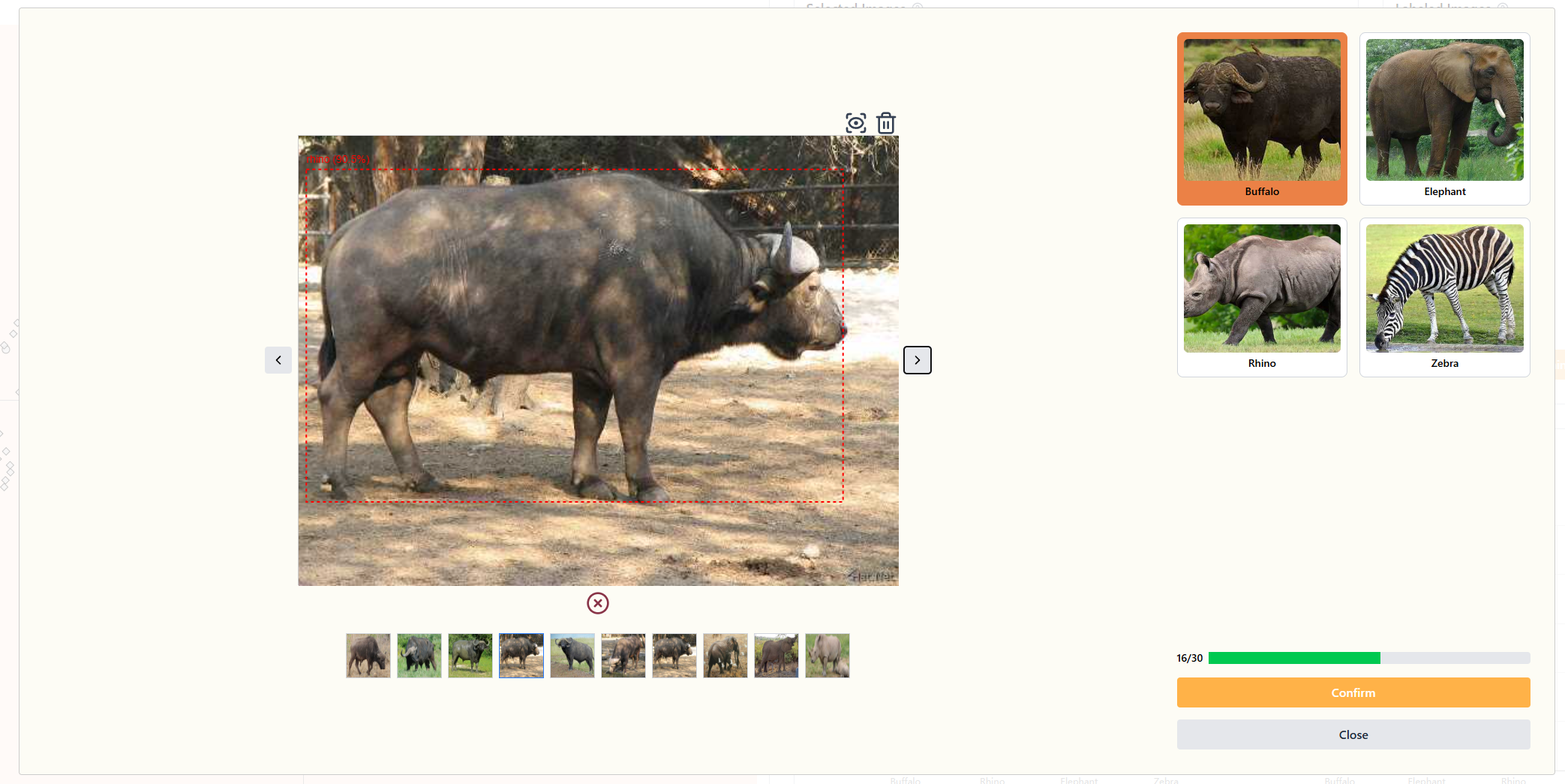}
  \end{minipage}
  \caption{A selection of points in the fourth iteration and the inspection and correction of a confidently misclassified sample.}
  \label{c3t4s2}
\end{figure}

Figure \ref{c3t4s2} depicts another selection made in iteration 4, along with the inspection of one of these samples. It is noticed that even though the detections made in these samples have a high confidence score, there exist examples where the model confidently makes the wrong classification. The buffalo in this sample is being predicted as a rhino with 90\% confidence. Correcting such samples can be important for further improvement of the model. 

\subsubsection{Iteration 5}
\begin{figure}[ht!]
    \centering
    \includegraphics[width=0.7\textwidth]{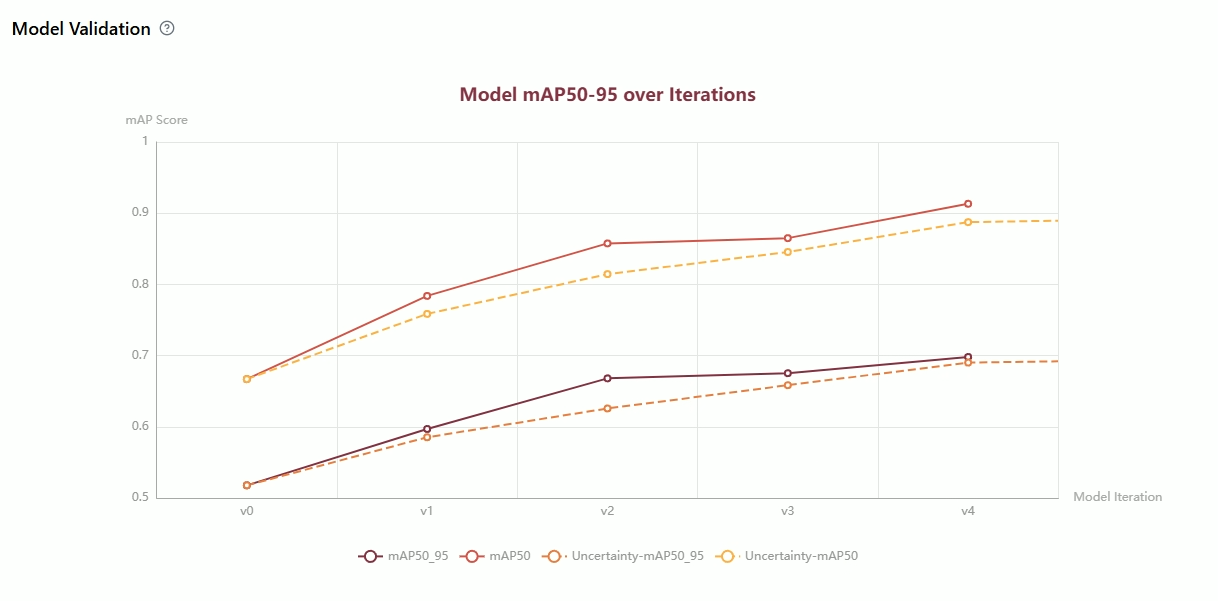}
    \caption{Model Validation View at the start of iteration 5.}
    \label{model_valc3i5}
\end{figure}

After the training of iteration 4, the final training iteration has been reached. First, the Model Validation view in Figure \ref{model_valc3i5} is consulted to see if the downward trend seen in iteration 3 could be turned around. It looks like it was, the mAP50-95 and the less strict mAP50 metrics are showing a better upwards trend again. The trajectory is not as steep as from iteration 0 to iteration 2, but that is per expectations.

\begin{figure}[ht!]
    \centering
    \includegraphics[width=0.4\textwidth]{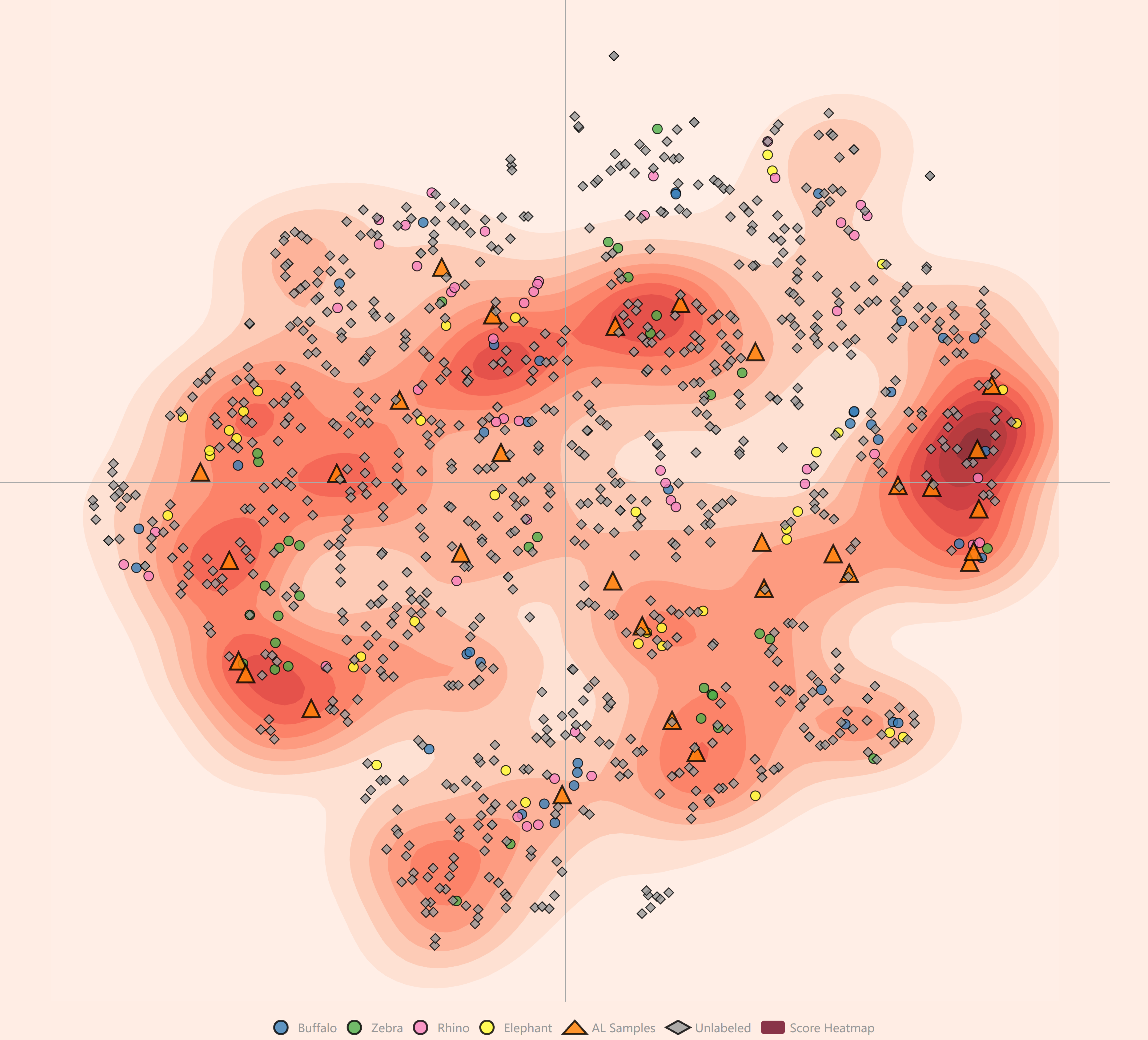}
    \caption{Data View start of iteration 5.}
    \label{c3it5_dataview}
\end{figure}

\begin{figure}[ht!]
  \centering
  \begin{minipage}[c]{0.4\textwidth}
    \centering
    \includegraphics[width=\linewidth]{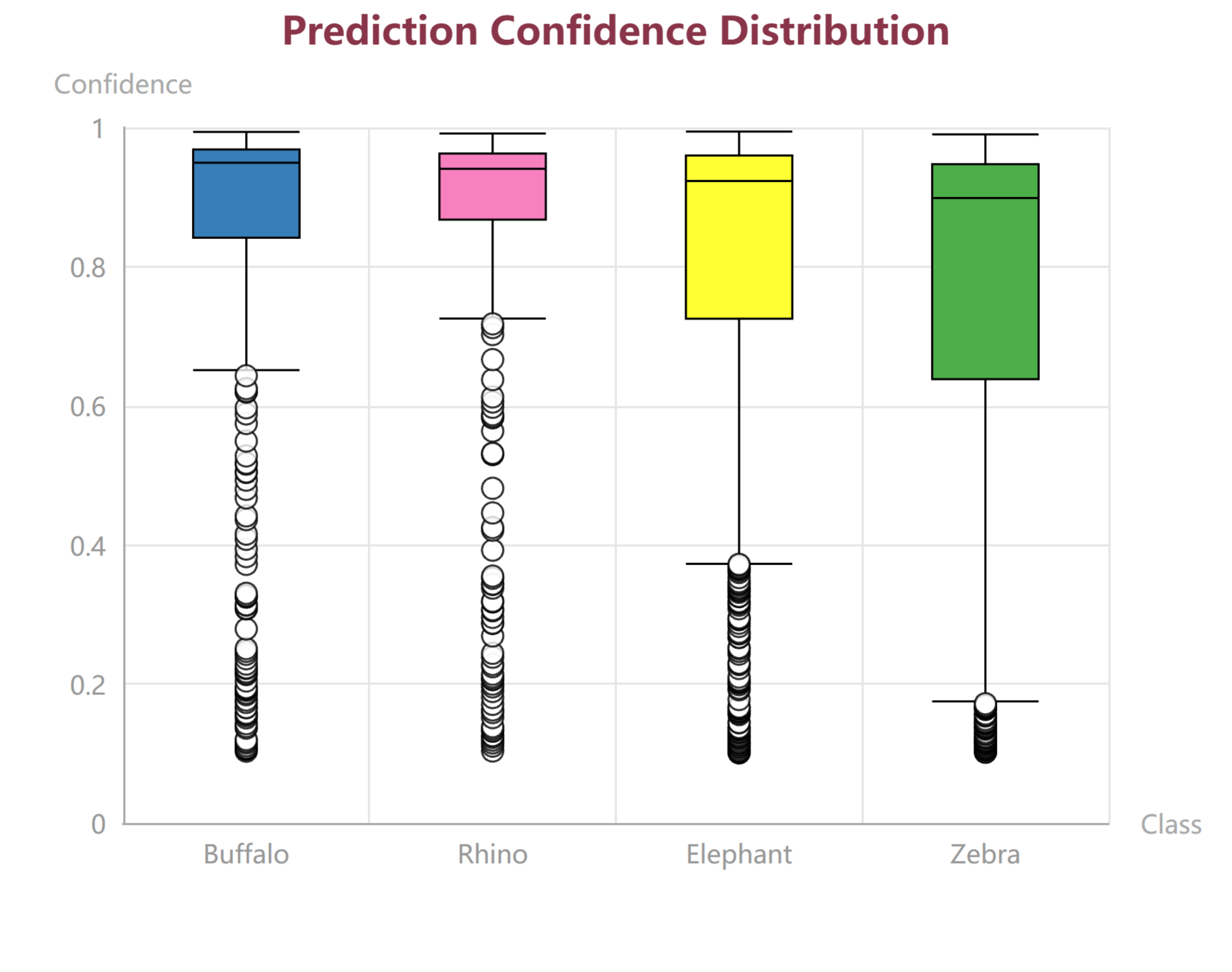}
  \end{minipage}
  \begin{minipage}[c]{0.4\textwidth}
    \centering
    \includegraphics[width=\linewidth]{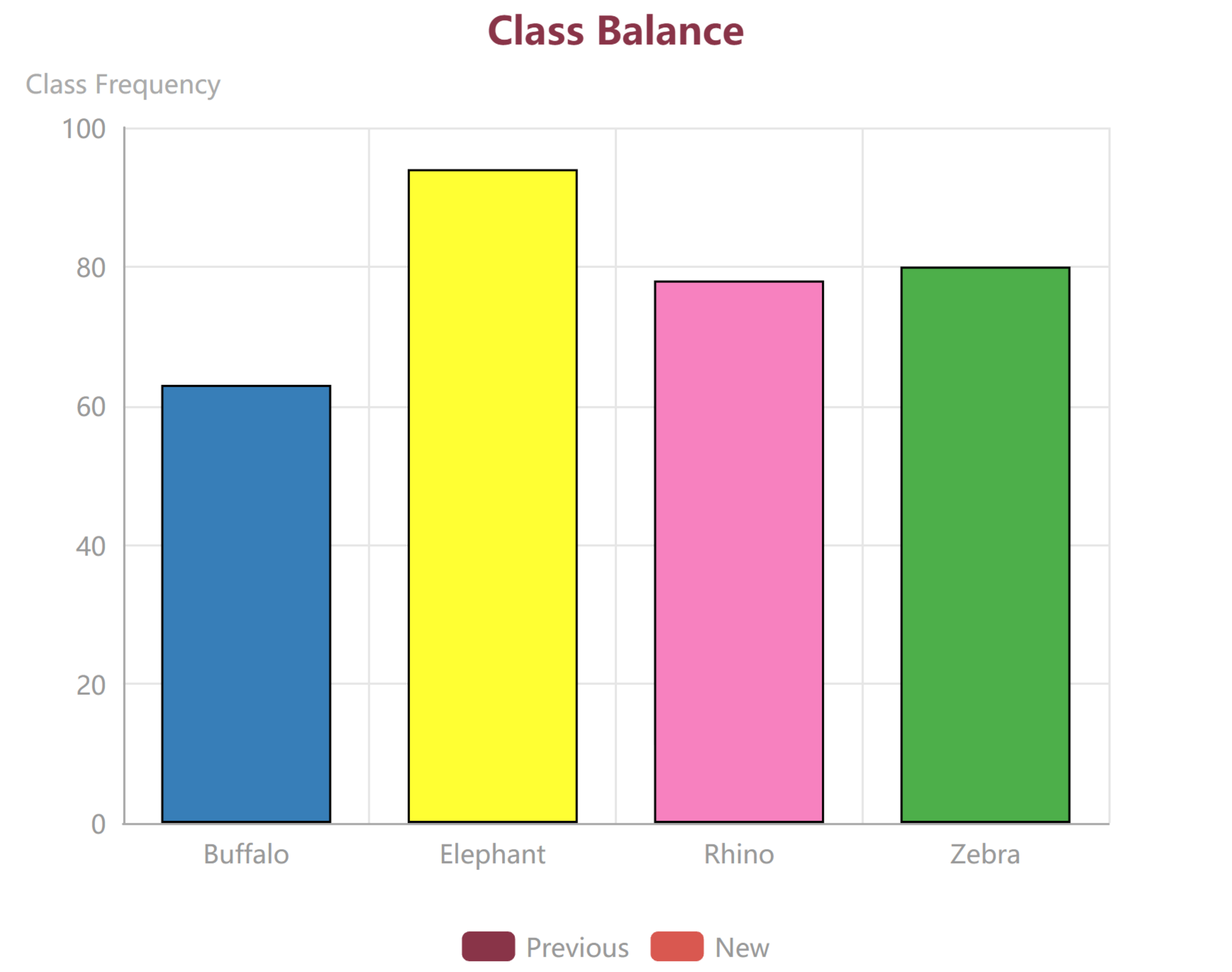}
  \end{minipage}
  \caption{Model View start of iteration 5.}
  \label{c3it5_modelview}
\end{figure}

The final Data View in \ref{c3it5_dataview} is largely displaying a similar uncertainty landscape. It seems that the dark red region towards the right of the visualization contains images that are much harder for the model to predict, even though labeling efforts have been made in and around this region. However, considering the confidence distribution of Figure \ref{c3it5_modelview} in conjunction with the heatmap reveals an important aspect to take into consideration. Many of the low confidence predictions are now considered outliers, meaning they are very rare in occurrence in comparison to the high confidence predictions. This is information that is mostly lost in the uncertainty heatmap. The low confidence predictions (high uncertainty) will still exist and contribute more. The model could still make correct detections of the actual objects present in the image, but add noisy detections as well. This highlights the complexity of OD in an active learning setting and the importance of considering multiple sources of information.

\begin{figure}[ht!]
  \centering
  \begin{minipage}[c]{0.4\textwidth}
    \centering
    \includegraphics[width=\linewidth]{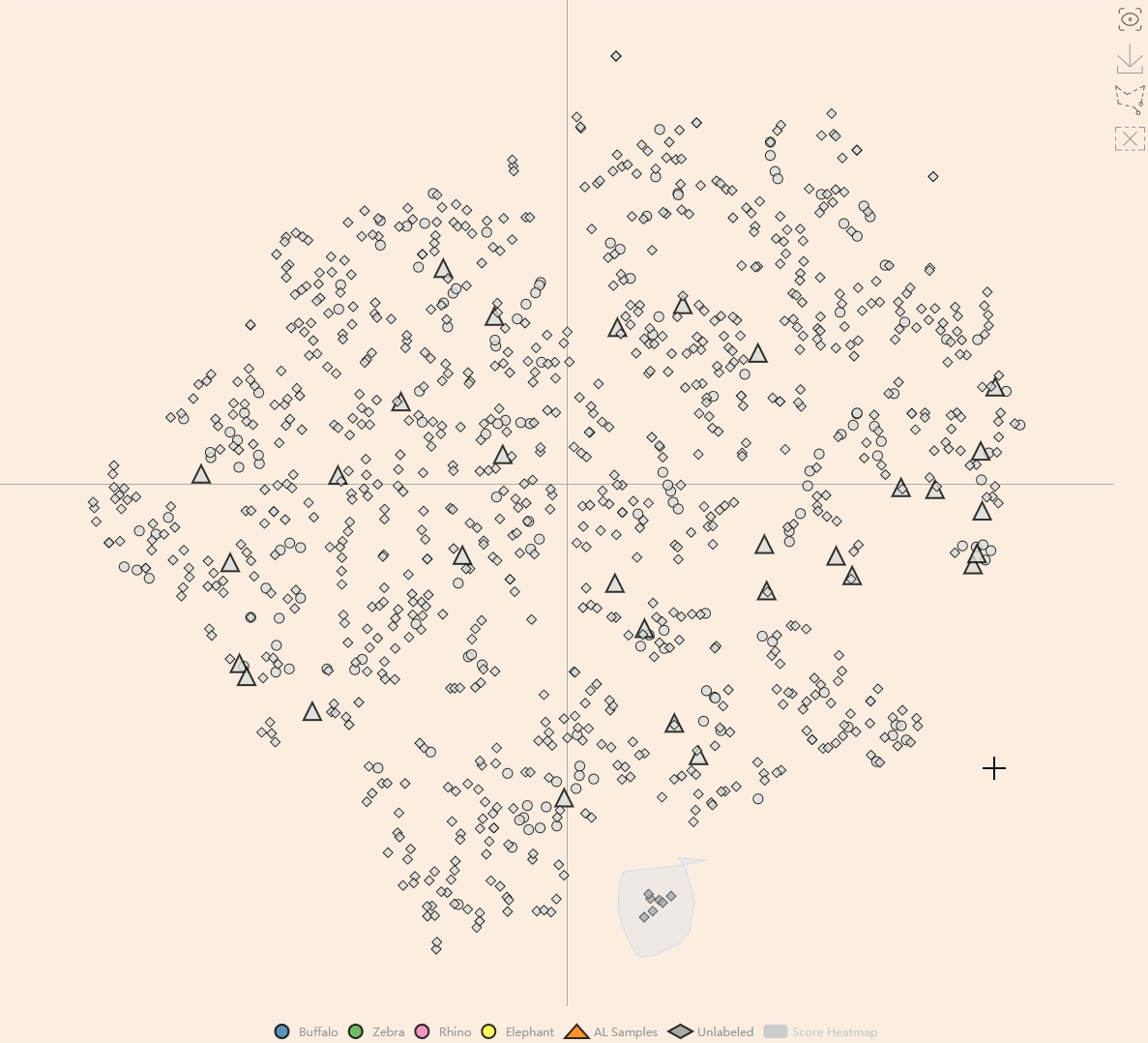}
  \end{minipage}
  \begin{minipage}[c]{0.4\textwidth}
    \centering
    \includegraphics[width=\linewidth]{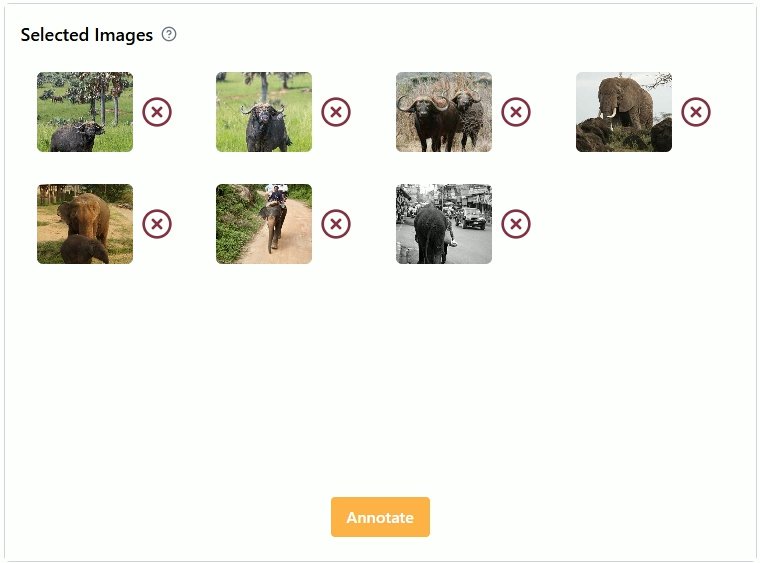}
  \end{minipage}
  \caption{Selection of a cluster separated from other points.}
  \label{c3i5s1}
\end{figure}

\newpage
The sampling strategy for this iteration will be similar to that of iteration 4. An attempt will be made to find underrepresented regions, but AL samples will also be inspected, and those deemed to indicate a structural confusion of the model, and not just detections with low confidence noise, will be annotated. The first selection is made at the bottom of the fourth quadrant of Figure \ref{c3i5s1}. This is a small cluster of samples that has gone unnoticed until now. They are well separated from any other points, and none of these have been labeled previously. The samples here are quite different from each other, and it is hard to pinpoint any common characteristics that might have separated these from the other points. One sample is a black and white image of an elephant in a city environment, which understandably sticks out from most other samples.

\begin{figure}[ht!]
  \centering
  \begin{minipage}[c]{0.4\textwidth}
    \centering
    \includegraphics[width=\linewidth]{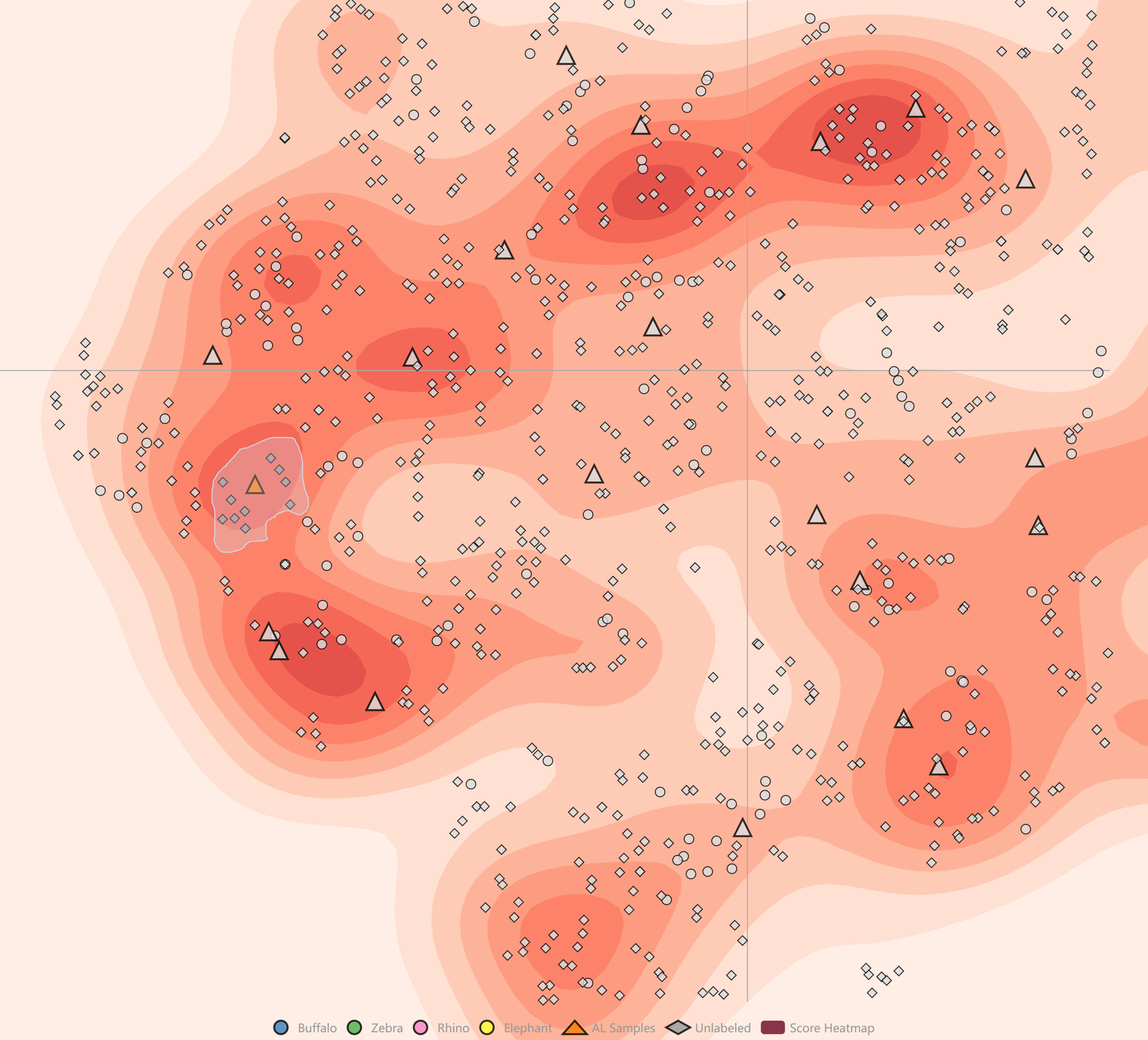}
    \caption{Second selection in iteration 5.}
    \label{c3s2}
  \end{minipage}
  \begin{minipage}[c]{0.4\textwidth}
    \centering
    \includegraphics[width=\linewidth]{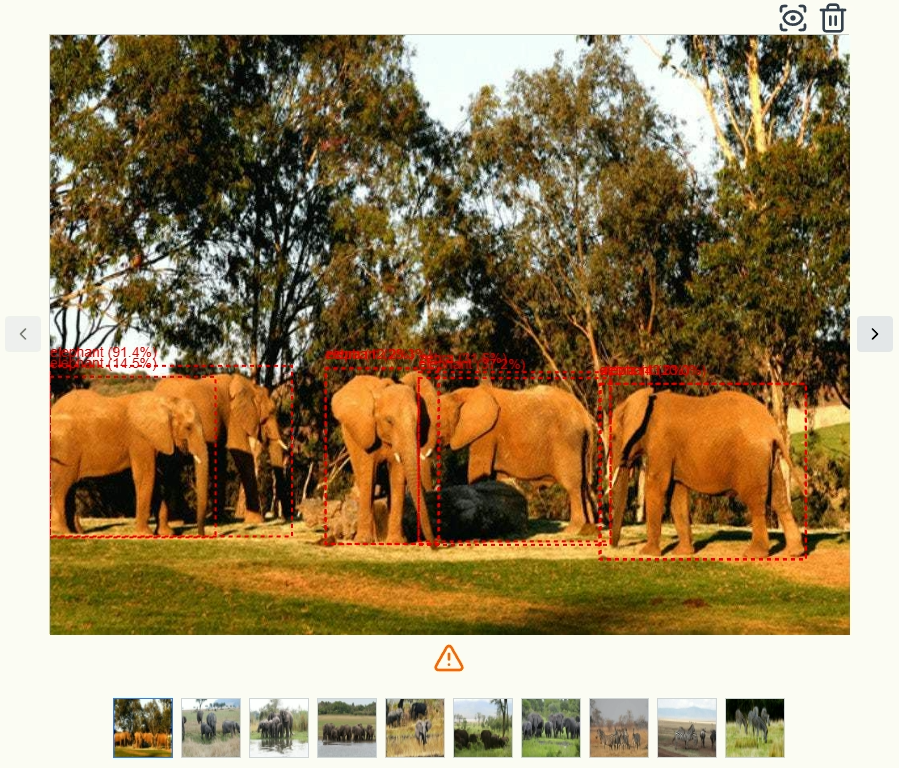}
    \caption{AL sample with noisy detections stacked upon each other.}
    \label{c3s2_AL}
  \end{minipage}
\end{figure}

Furthermore, another smaller region is found in the third quadrant (Figure \ref{c3s2}). These are also in a high uncertainty region, hence they are inspected to see if there are some noisy detections, as suspected might exist in these regions. The AL samples in this region certainly has some noisy predictions, as seen in \ref{c3s2_AL}. It looks like many high uncertainty prediction are stacked on top of each other. 

\begin{figure}[ht!]
  \centering
  \begin{minipage}[c]{0.3\textwidth}
    \centering
    \includegraphics[width=\linewidth]{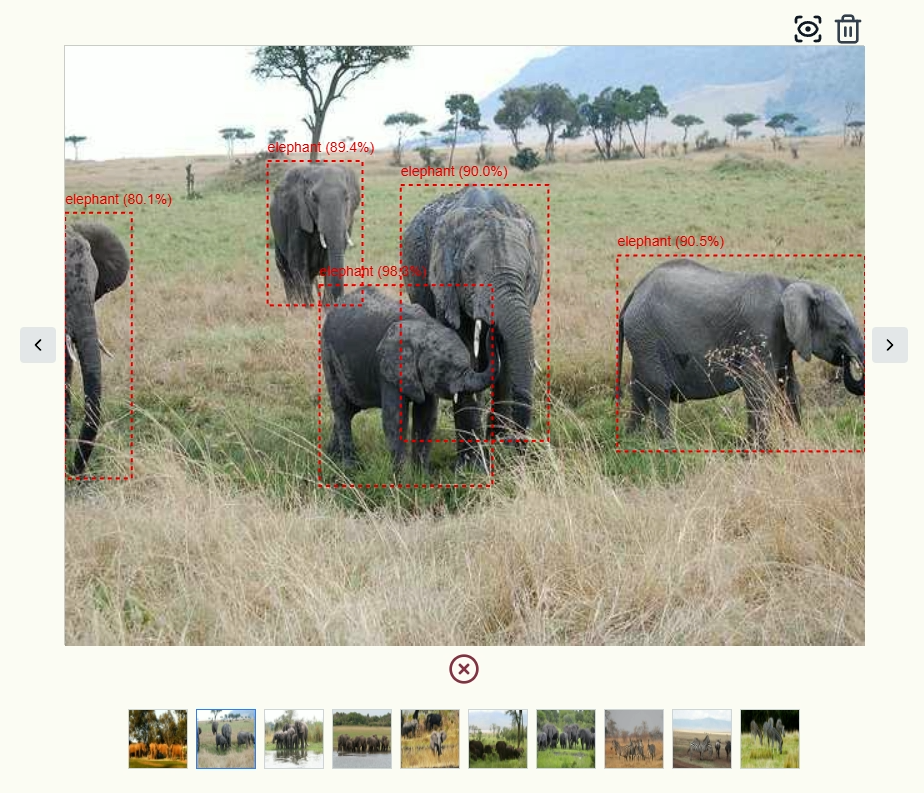}
  \end{minipage}
  \begin{minipage}[c]{0.3\textwidth}
    \centering
    \includegraphics[width=\linewidth]{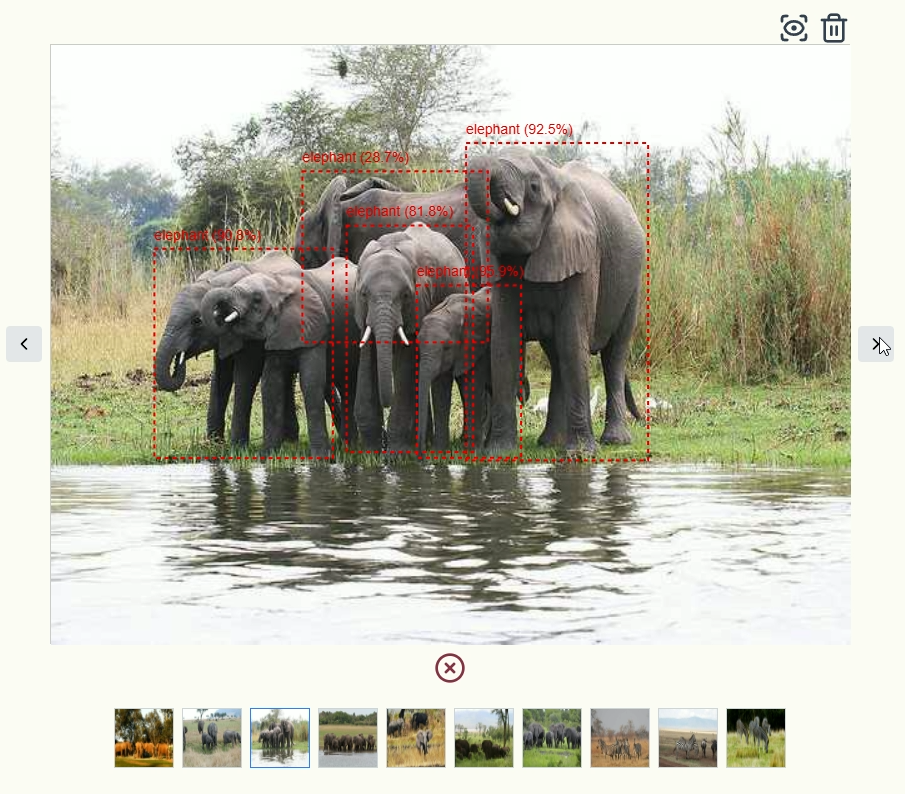}
  \end{minipage}
  \vspace{0.25em} \\
  \begin{minipage}[c]{0.3\textwidth}
    \centering
    \includegraphics[width=\linewidth]{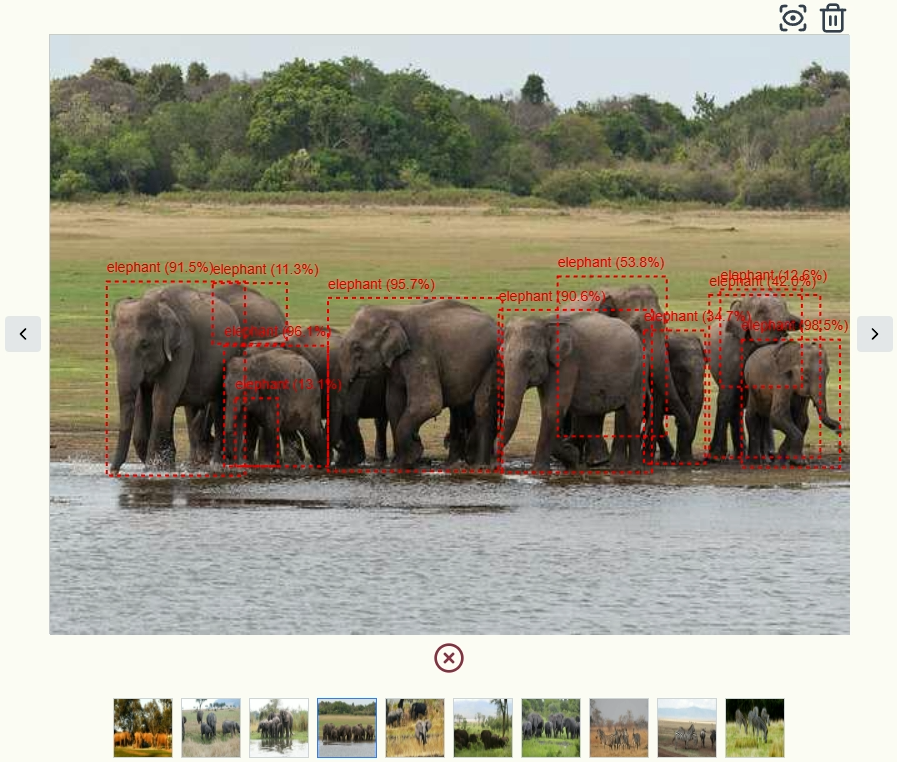}
  \end{minipage}
  \begin{minipage}[c]{0.3\textwidth}
    \centering
    \includegraphics[width=\linewidth]{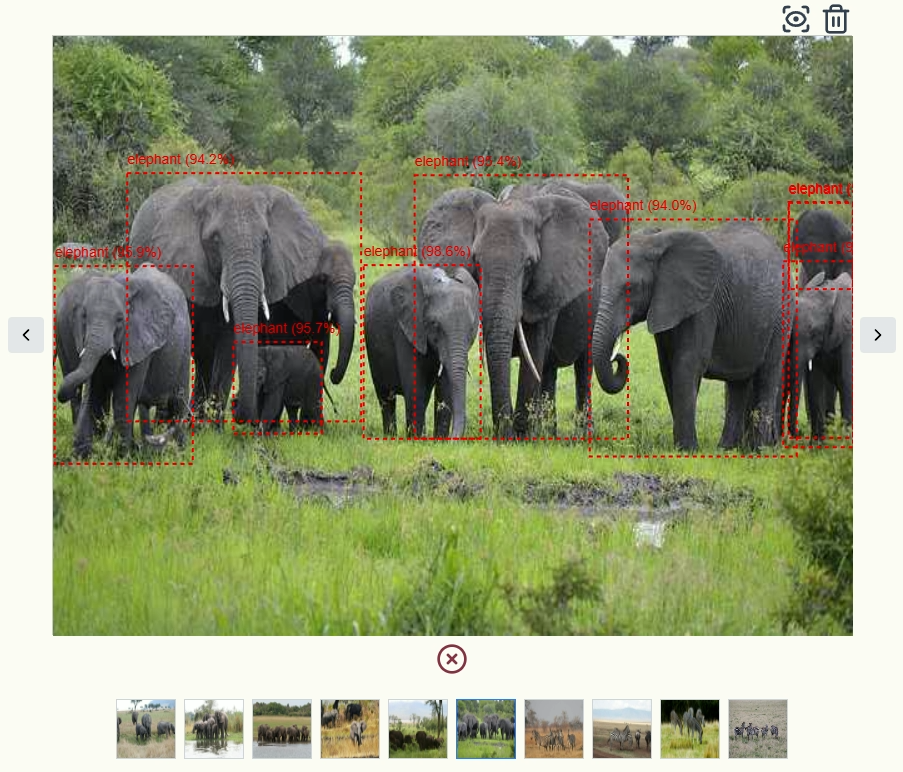}
  \end{minipage}
  \caption{Four examples of samples with multiple instances of the elephant class and their respective box detections made by the model.}
  \label{box_detections_i5}
\end{figure}

The rest of these samples also contain multiple instances of elephants. For reference, four of these samples and the bounding box detections made by the model are displayed in Figure \ref{box_detections_i5}. It can be seen that model predictions are fairly accurate, especially in the first image (top left), where the elephants are more separated, but localization becomes a lot harder when they are closer to each other, like in the other three examples. These four samples are annotated.

The last iteration is finalized by inspecting and annotating a few of the AL-suggested samples scattered across the visualization. 

This iterative process, employing the \textit{Balanced Guidance Integration} strategy within the VILOD system, has demonstrated how an expert user can leverage multiple coordinated views and interactive functionalities to navigate the complex object detection annotation task. A comprehensive quantitative evaluation of the models trained during this use case, comparing their performance trajectories against other strategies and the automated baseline, will be presented in Section \ref{analysis}. The full list of samples annotated during this \textit{Balanced Guidance Integration} use case is provided in Appendix \ref{appendixA}.


\subsection{Analysis}
\label{analysis}
This section interprets the collective findings from the three VILOD-guided labeling strategies, the \textit{Exploration \& Structure Focus} in Section \ref{use_case_stuctural}, the \textit{Uncertainty-Driven Focus} in Section \ref{use_case_uncertainty}, and finally the \textit{Balanced Guidance Integration} in Section \ref{use_case_balanced}, together with the automated AL baseline. The aim is to identify trends in model performance, compare these performance trends, and discuss the qualitative aspects of using VILOD for each strategy, thereby addressing the research questions regarding user enhancement (RQ1) and comparative performance (RQ2).

\subsubsection{Quantitative Analysis}
\label{quantitativeAnalysis}

\begin{figure}[ht!]
    \centering
    \includegraphics[width=0.6\textwidth]{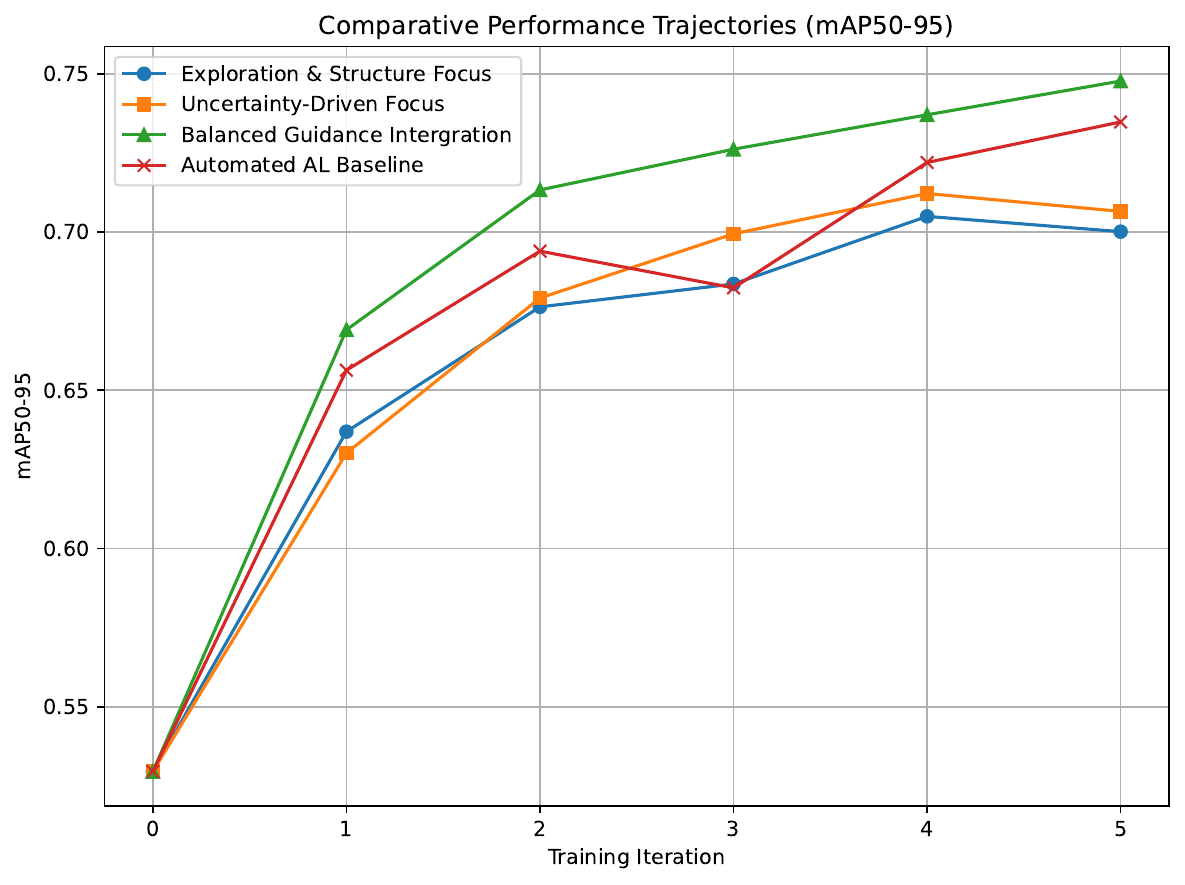}
    \caption{mAP50-95 performance trajectory comparison between the different labeling strategies against the number of training iterations.}
    \label{performance_trajectories}
\end{figure}

The quantitative performance of the object detection models, iteratively trained under each of the three VILOD-guided strategies and the automated AL baseline approach, reveals characteristics in their learning efficiencies and overall effectiveness. These results, primarily focusing on the mAP50-95 metric, are summarized in Table \ref{tab:perf_trajectories} and visualized as learning curves in Figure \ref{performance_trajectories}. All strategies began from an initial model ($M_0$) mAP50-95 of 0.52954.

\begin{table}[ht!]
\centering
\caption{Comparative Performance Trajectories (mAP50-95) by Iteration.}
\label{tab:perf_trajectories}
\begin{tabular}{crrrr}
\toprule
Strategy & Balanced Guidance & Exploration & Uncertainty & AL Baseline\\
\midrule
Iteration & & & & \\
\midrule
0         & 0.5295          & 0.5295          & 0.5295          & 0.5295 \\
1         & 0.6691          & 0.6369          & 0.6302          & 0.6563 \\
2         & 0.7133          & 0.6764          & 0.6792          & 0.6940 \\
3         & 0.7262          & 0.6836          & 0.6994          & 0.6824 \\
4         & 0.7370          & \textbf{0.7050} & \textbf{0.7122} & 0.7220 \\
5         & \textbf{0.7477} & 0.7001          & 0.7065          & \textbf{0.7348} \\
\bottomrule
\end{tabular}
\end{table}

Examining the final performance at iteration 5, the \textit{Balanced Guidance Integration} strategy achieved the highest overall mAP50-95 of 0.74771. This was followed by the \textit{Automated AL Baseline} at 0.73478, then the \textit{Uncertainty-Driven Focus} at 0.70652, and finally the \textit{Exploration \& Structure Focus} strategy at 0.70012. This suggests that the holistic approach of the Balanced Guidance strategy was most effective in achieving peak model performance within the given labeling budget.

The learning trajectories depicted in Figure \ref{performance_trajectories} further highlight the differences. The \textit{Balanced Guidance Integration} strategy not only achieved the highest final mAP50-95 but also learned the fastest, reaching 0.6691 by iteration 1. The highest among all strategies at that point and consistently improved to 0.7133 by iteration 2. Its sustained improvement across iterations indicates effective sample selection throughout the process. 

The Automated AL Baseline also showed competitive performance, reaching 0.6940 by iteration 2 and ultimately scoring the second-highest final mAP50-95. An interesting observation in the learning curve is a sudden steep drop in performance at iteration 3, which could highlight the downfall of using a fully automatic active learning algorithm. Noisy samples might have been added here, which deteriorates the model's performance. In comparison, the trajectory of the Uncertainty Driven focus strategy, which employs a similar approach as the AL baseline but with basic quality checks, has a more stable trajectory throughout the five iterations. However, it ultimately was outperformed by the AL baseline. This highlights the inherent strength of uncertainty sampling but also the signs of its weaknesses.

The Uncertainty-Driven Focus strategy reached an mAP50-95 of 0.6792 by iteration 2 and peaked at 0.7122 in iteration 4, before a slight dip to 0.7065 in iteration 5. While it showed steady gains initially, its performance plateaued and slightly decreased towards the end, suggesting that a pure focus on uncertainty might yield diminishing returns or introduce challenging samples that the model struggles to generalize from effectively in later stages, this is however in contradiction by the good performance showed by the Automated AL Baseline in these later iterations. 

The Exploration \& Structure Focus strategy exhibited a more gradual but consistent improvement in its early to mid-stages, reaching 0.6764 by iteration 2 and 0.7050 by iteration 4. However, it experienced a slight dip in performance between iteration 4 (0.7050) and iteration 5 (0.7001), indicating that while broad feature space coverage is beneficial, it might not always translate to monotonic increases in mAP, especially when focusing on subtle structural details or edge cases in later iterations as described qualitatively.

Considering other metrics from Table \ref{tab:final_performance}, the \textit{Balanced Guidance Integration} strategy also achieved the highest final Recall at 0.88662 and Precision 0.95506, suggesting a well-rounded model. The Exploration \& Structure Focus strategy also yielded a high recall of 0.88429, nearly matching the Balanced approach, which aligns with its goal of capturing diverse samples. The Uncertainty-Driven Focus strategy, while having a lower recall at 0.83594, maintained high precision of 0.93357, suggesting it made fewer incorrect positive predictions but might have missed more objects.

\begin{table}[ht!]
\centering
\caption{Final Model Performance Metrics (Iteration 5) for Different Labeling Strategies.}
\label{tab:final_performance}
\begin{tabular}{lrrrrr}
\toprule
 Strategy &  mAP50-95 &  mAP50 &  mAP75 &  Precision &  Recall \\
\midrule
AL Baseline & 0.7348 & 0.9283 & 0.8197 &     0.9001 &  0.8558 \\
Balanced Guidance & \textbf{0.7477} & \textbf{0.9540} & \textbf{0.8439} &     \textbf{0.9551} &  \textbf{0.8866} \\
Uncertainty &    0.7065 & 0.9210 & 0.7978 &     0.9336 &  0.8359 \\
Exploration  &    0.7001 & 0.9184 & 0.7841 &     0.8764 &  0.8843 \\
\bottomrule
\end{tabular}
\end{table}

\newpage
\subsubsection{Qualitative Analysis}
\label{qualitativeAnalysis}

Beyond the numerical metrics, the qualitative experiences detailed in Sections \ref{use_case_stuctural}, \ref{use_case_uncertainty}, and \ref{use_case_balanced} provide important context for understanding these performance trends and directly address how VILOD's visual framework facilitates the implementation of expert-driven labeling strategies (RQ1).

The better performance of the \textit{Balanced Guidance Integration} strategy (Section \ref{use_case_balanced}) can be attributed to its comprehensive use of VILOD's features, allowing for an adaptive approach. The user's ability to synthesize information from the Data View's uncertainty heatmap and t-SNE structure with the Model View's class balance and confidence distributions enabled strategic interventions, such as prioritizing underrepresented classes or addressing specific model weaknesses like confidently misclassified samples. This strategy's success underscores the benefit of a flexible, human-AI collaborative approach where the user can dynamically weigh different aspects of sample informativeness.

The Automated AL Baseline, serving as a benchmark, demonstrates the effectiveness of uncertainty sampling. However, its performance being slightly surpassed by the \textit{Balanced Guidance Integration} suggests that human oversight and strategic depth can offer an edge. The VILOD-guided strategies, particularly the Uncertainty-Driven Focus, highlighted another advantage of human involvement which is quality control.

In the Uncertainty-Driven Focus strategy (Section \ref{use_case_uncertainty}), while closely following AL suggestions, the user's intervention to discard very noisy or poor-quality AL samples likely prevented the introduction of detrimental data that a purely automated baseline could include. This human filtering, even within a largely algorithmic strategy, is a significant qualitative benefit. The observed plateau or slight dip in its performance in later iterations might be linked to the nature of samples selected by uncertainty alone, which, as noted, became increasingly challenging or potentially less broadly informative over time.

The Exploration \& Structure Focus strategy in Section \ref{use_case_stuctural} emphasized VILOD's capability to support data understanding independent of model uncertainty. The ability to disable the heatmap and focus on the t-SNE scatterplot's structure allowed for deliberate diverse sampling across identified "blobs" and peripheral regions. This method, while not achieving the highest final mAP50-95 in this study, is valuable for building a dataset with broad feature representation, potentially leading to more robust generalization, a quality not fully captured by mAP alone. The slight dip in its final iteration could be due to focusing on increasingly subtle structural details or specific edge cases that offered limited immediate gain in overall mAP. It's also worth noting that while the VILOD system does facilitate this approach in some ways, more help and guidance could be added in further iterations of the system to further guide the user.

Collectively, the three VILOD-guided use cases demonstrate that the system can effectively support the implementation of varied labeling strategies. VILOD's interactive visualizations (Data View, Model View) and tools (lasso selection, annotation interface with prediction overlays) provide the necessary transparency and control for the user to interpret model states such as understanding confidence distributions, identifying areas of high uncertainty, or recognizing specific model errors and to make strategically informed decisions that go beyond simple uncertainty sampling. This directly supports the premise of RQ1 by demonstrating how VILOD's visual framework provides the necessary components for an expert to successfully implement these distinct strategies. The differing performance trajectories, related to RQ2, further suggest that the choice of strategy, enabled by such an interactive tool, has a measurable impact on model development.

\subsection{Summary of Findings}
\label{summaryoffindings}
This chapter has presented the results and analysis from an empirical investigation into the VILOD system, comparing three different visually guided labeling strategies against each other and an automated AL baseline. The findings, encompassing both quantitative performance metrics and qualitative user interaction insights, provide a comprehensive understanding of VILOD's capabilities and the impact of different human-in-the-loop approaches to object detection annotation.

The quantitative analysis revealed notable differences in model performance trajectories and final outcomes across the tested strategies. The \textit{Balanced Guidance Integration} strategy emerged as the most effective, achieving the highest final mAP50-95 of 0.74777. This approach also demonstrated strong initial learning, outperforming other methods. The \textit{Automated AL Baseline} also showed strong performance, having the second highest mAP50-95 of 0.7348. The \textit{Uncertainty-Driven Focus} strategy achieved a final mAP50-95 of 0.7065, showing steady initial gains but experiencing a slight plateau and dip in later stages. Lastly, the \textit{Exploration \& Structure} Focus yielded a final mAP50-95 of 0.7001. Analysis of final precision and recall metrics further differentiated the strategies, with \textit{Balanced Guidance Integration} showing strong results in both. \textit{Exploration \& Structure Focus} also achieved high recall, and \textit{Uncertainty-Driven Focus} maintained a high precision.

Qualitatively, the investigation highlighted VILOD's effectiveness in supporting varied and nuanced labeling strategies. The better performance of the \textit{Balanced Guidance Integration} strategy can be attributed to the comprehensive use of VILOD's features, enabling the user to adaptively synthesize information from different views to make strategic decisions. The \textit{Uncertainty-Driven Focus} strategy demonstrated VILOD's capacity to facilitate a more algorithm-centric approach while still underscoring the significant value of human-in-the-loop quality control, particularly in filtering out noisy or uninformative AL-suggested samples. The \textit{Exploration \& Structure Focus} strategy showcased VILOD's support for data understanding independent of model uncertainty, allowing for deliberate diverse sampling aimed at building broad feature representation within the dataset.

Collectively, these findings indicate that while automated uncertainty sampling provides a strong baseline, human guidance, when effectively supported by an interactive visual analytics tool like VILOD, can offer an edge in performance and provides crucial benefits such as quality control and strategic depth. The VILOD system, through its interactive visualizations and tools, demonstrably facilitated the user's process of interpreting model states and dataset characteristics, which in turn enabled the implementation of diverse, informed labeling strategies, thereby directly answering RQ1. The observed differences in performance trajectories and final model metrics across the various strategies further affirm that the choice of labeling strategy, as enabled through an interactive tool, has a measurable impact on the efficiency and effectiveness of object detection model development, aligning with RQ2. These insights pave the way for further discussion on the implications of such human-AI collaborative systems.

\newpage
	
\section{Discussion}
This chapter discussed the findings presented in Chapter \ref{ResultsAnalysis}, interpreting their significance concerning the research questions and objectives outlined in Chapter \ref{introduction}. The results are contextualized within the existing literature reviewed in Chapter \ref{relatedWork}, and the implications and limitations of this study are critically discussed. The primary aim is to assess whether the research problem concerning the need for more transparent, interpretable, and strategically manageable HITL systems for OD annotation has been effectively addressed.

The empirical investigation produced several findings regarding the utility of the VILOD system and the comparative performance of different visually guided labelling strategies. The quantitative analysis revealed that the Balanced Guidance Integration strategy achieved the highest final mAP50-95, outperforming the other VILOD guided strategies, but also the AL Baseline. The Uncertainty-Driven and Exploration \& Structure Focus strategies have an overall lower performance than the AL baseline and the Balanced Guidance Integration. These results highlight that while a strong algorithmic baseline like uncertainty sampling is effective, a human-guided approach, considering various sources of information, can achieve higher performance. 

The findings directly address the research questions posed in Section \ref{problemFormulation}. Regarding RQ1, the qualitative analysis in \ref{qualitativeAnalysis} and the detailed use case walkthrough in Sections, \ref{use_case_stuctural}, \ref{use_case_uncertainty} and \ref{use_case_balanced} provide evidence that through the use of the VILOD system, including its integrated Data View and Model View demonstrably facilitated the user's process of interpreting the model's current state and the dataset's characteristics, which was crucial for implementing the various labeling strategies. For instance, the balanced Guidance Integration use case relied on interpreting several views to make informed decisions, such as correcting confidently misclassified samples, prioritizing underrepresented classes, or sampling from high uncertainty regions. The Exploration \& Structure focus use case showed how users could leverage the t-SNE projection to understand data clusters and ensure diverse coverage. Furthermore, the Uncertainty-Driven Focus use case shows how a more algorithmically centered approach can benefit from the VILOD interface for visualizing the AL suggestions and allow for human quality control in filtering out noisy or uninformative samples. The clear execution of these distinct strategies speaks to VILOD's success in facilitating an expert-driven process within a visually guided AL sampling pipeline in an OD setting.

Regarding RQ2, the quantitative results presented in \ref{quantitativeAnalysis} show clear differences in model performance based on three distinct use cases facilitated by a varying level of visual cues and guidance of the tool. The fact that the Balanced Guidance Integration approach, which incorporate all different information as provided by the system, surpassed the automatic AL baseline suggests that including the human-in-the-loop can indeed yield model performance gains. The other VILOD-guided strategies, namely Exploration \& Structure and Uncertainty-Driven, did not outperform the AL baseline in final mAP50-95 in this specific study, but their trajectories and qualitative benefits, such as dataset coverage for Exploration and quality control for Uncertainty, highlight different trade-offs. The dip in the AL Baseline's performance at iteration 3 somewhat reinforces the notion that automated strategies are not immune to selecting problematic samples, potentially due to accumulated noise or specific dataset characteristics that a human might identify and mitigate. The slight performance dips in the Exploration \& Structure Focus and Uncertainty-Driven Focus strategies in their final iterations might indicate that these more specialized strategies could reach points of diminishing returns or require further nuanced interaction as the model matures.

The findings of this thesis can be contextualized within the broader landscape of VIL, AL, and HITL systems, as reviewed in Chapter \ref{relatedWork}. The VILOD system and the Balanced Guidance Integration use case align with the principles of VIL systems such as VisGil in \cite{grimmeisen2023visgil}, which also aim to guide users towards informative samples. While VisGil focuses on classifications and uses specific visual cues like icon size for utility, VILOD adapts similar principles in a multi-class OD setting, using heatmaps for uncertainty and t-SNE projection for representing the data, along with explicit model state feedback. The study performed in \cite{bernard2017comparing} found that a user-centered VIL can match or outperform standard AL, which resonates with the findings in this work. Although a larger user study would have to be performed to establish any significance in the indicated performance gains within the VILOD system. This thesis extends upon the findings of such work by focusing specifically on the complexities of object detection and exploring multiple, distinct HITL strategies within a unified interactive system.

Compared to purely algorithmic AL approaches for OD, such as those in \cite{brust2018activelearningdeepobject, chen2023semi}, VILOD offers enhanced transparency and user control. While algorithmic methods focus on sophisticated query strategies, they often operate as "black boxes". VILOD, by visualizing AL suggestions and model state, allows users to understand why samples might be chosen and to strategically intervene. The qualitative benefits observed, such as filtering noisy AL samples in the Uncertainty-Driven Focus strategy, directly address a limitation of purely automated systems and demonstrate a practical advantage of the HITL approach facilitated by VILOD.

This work contributes to addressing some of the identified gaps in Section \ref{relatedWork}, particularly the lack of OD-specific interactive AL tools that support diverse expert strategies and integrate annotation with iterative exploration. VILOD provides a concrete example of such a system, and the comparative evaluation of different strategies offers empirical insights into their process and outcomes within an OD context.

The findings of this research have several implications for researchers and practitioners in the fields of VA, ML, and particularly for those involved in developing and deploying OD models. For researchers, the study reinforces the value of human-centered AI and the potential of interactive visual interfaces to mediate complex machine learning workflows. It suggests that moving beyond purely optimizing automatic AL query strategies and instead focusing on how to effectively empower human experts with interpretable information and control, can lead to both improved model performance and a more nuanced understanding of the data and model. The design and evaluation of the VILOD tool contribute to the knowledge on designing effective HITL systems, particularly for a more complex task like OD annotation. The varying performance and qualitative experiences of the different strategies also suggests areas for future research into adaptive interfaces that might guide users towards the most suitable strategy based on the task at hand. For practitioners involved in OD model training and deployment, this thesis demonstrates that tools like VILOD can offer an efficient and manageable alternative to traditional mass annotation or purely algorithmic AL. 

Although the empirical study conducted in this work shows promising results, it's important to note the limitations and the generalization of its findings. 

Firstly, the evaluation involved a single user acting as an expert. While this ensured consistent application of predefined strategies and allowed for a deep dive into decision making within the VILOD system, it could introduce potential subjectivity and limit the generalization to users with different levels of expertise and familiarity with the tool or similar systems. 

Secondly, the study was conducted using a single dataset and a specific object detection architecture. The effectiveness of the strategies and VILOD's features might vary with datasets possessing different characteristics or with other OD model architectures. 

Thirdly, the Active Learning component within VILOD was based solely on uncertainty sampling. While users could employ other strategic considerations, the system's inherent AL suggestions were limited. The interplay with more diverse AL query strategies was not explicitly explored.

Finally, the primary goal was to investigate how VILOD facilitates different strategic approaches and compare these within a HITL context, rather than solely optimizing for state-of-the-art mAP against other possible annotation methods. The findings indicate relative performance among the tested strategies but do not claim global optimality. 

These limitations suggest that further research with more users, diverse datasets, and potentially different underlying models and AL techniques would be beneficial to validate and extend the findings of this thesis. The reflections on the VILOD prototype itself, such as the note in Section \ref{analysis} about potentially adding more guidance for the "Exploration \& Structure Focus" strategy, also point towards areas for future tool refinement based on the insights gained.

\newpage
		
\section{Conclusion and Future Works}
This final chapter concludes the research conducted in this work. It summarizes the key findings concerning the development and evaluation of VILOD, a Visual Interactive Labeling tool for Object Detection. Furthermore, it outlines potential directions for future research and development, building upon the insights gained and the limitations identified in this work.

\subsection{Conclusion}
This thesis set out to address the challenges associated with data annotation for object detection (OD) models, especially focusing on how VA could enhance the HITL Active Learning process. The primary contribution of this work is the implementation and empirical investigation of the VILOD system. The system was evaluated through a comparative study of three distinct author-led use cases.

The author-led use cases, detailed in Chapter \ref{ResultsAnalysis}, showcase how the interplay between several different subfields within computer science can be leveraged to combat the complex and often expensive task of data annotation for training object detection models. This was evidenced by the varied ways specific VILOD features supported each distinct strategy, such as the holistic use of several views and visual guidance in the \textit{Balanced Guidance Integration} use case the focused use of the t-SNE projection in \textit{Exploration \& Structure Focus} and the facilitated oversight of AL suggestions in \textit{Uncertainty-Driven Focus}. This directly answers RQ1 as formulated in \ref{problemFormulation}.

The quantitative analysis of the object detection models in each use case scenario, detailed in Section \ref{quantitativeAnalysis} study successfully answers RQ2 regarding how different labeling strategies, as enabled by the VILOD system, have a measurable impact on model performance trajectories and final metrics, with different trade-offs in terms of learning speed, final accuracy, precision, and recall. The \textit{Balanced Guidance Integration} approach achieved the highest overall model performance across all captured metrics by the last iteration, slightly surpassing the automated AL baseline.

The importance of these results lies in their contribution to understanding and advancing human-AI collaboration in complex machine learning tasks. While the principles demonstrated are broadly applicable, the generalizability of specific performance outcomes is subject to the study's limitations, including the single-user evaluation, the use of a single dataset ("African Wildlife"), and a specific OD model (YOLOv11n). These factors mean that while the benefits of such an interactive approach are evident, outcomes may vary in different contexts. In summary, this thesis successfully demonstrated that an interactive visual labeling tool like VILOD can empower users to engage more strategically and effectively in the OD annotation process, leading to substantial benefits in model development.

\subsection{Future Work}
The insights gained and limitations identified in this research open up several promising areas for future work, which can be broadly categorized into VILOD system enhancements, extended empirical evaluations, and the exploration of new research questions.

Further development of the VILOD system itself could focus on several areas. One direction is the integration of more diverse and sophisticated AL query strategies beyond the current uncertainty-based suggestions, potentially including diversity sampling or other representativeness measures. This could also include giving users the ability to apply clustering algorithms to the unlabeled data points, and allowing the user to interactively define the initial data view by providing a set of dimensionality reduction methods and the ability to adjust the parameters of those methods. Another direction is to provide a more model-centric view of the training set. The Data View in this iteration focuses on providing a view of the original feature space and provides model state updates through the likes of an uncertainty heatmap. Future iterations could examine the impact of updating the view based on reduced embeddings from the newly trained models each iteration. Long term, future iteration of the VILOD tool could include adaptability of the task, such as simpler image classification or more complex OD task such as segmentation.

In terms of extended empirical evaluation, conducting broader user studies with multiple participants of varying expertise levels is crucial for assessing VILOD's usability and effectiveness more comprehensively and for understanding how different user characteristics influence strategy interaction and outcomes. Furthermore, a larger sample size in a user study would allow for a statistical analysis of the performance of the trained object detection models, investigation if there is any significant performance increase in models trained through VILOD in comparison to other state-of-the-art approaches.

By pursuing these directions, the field can continue to advance the development of more effective, efficient, and human-centered AI systems, transforming data-intensive tasks such as object detection annotation into more insightful, manageable, and ultimately more successful endeavors.

\newpage

%
\newpage

\hypersetup{urlcolor=black}
\bibliographystyle{IEEEtran}
\bibliography{references}
\newpage

\pagenumbering{Alph}
\setcounter{page}{1} 
\appendix

\section{Appendix 1} 
\label{appendixA}
\subsection*{Use Case 1: Exploration \& Structure Focus Annotated Samples}

\subsubsection*{Iteration 1}
\begin{table}[h!]
\centering
\caption{Exploration \& Structure Focus - Iteration 1 Annotated Samples}
\label{tab:exp_iter1_samples}
\begin{tabular}{cl}
\toprule
 Sample No. & Image Name \\
\midrule
         1 &  4 (9).jpg \\
         2 & 4 (321).jpg \\
         3 &  4 (27).jpg \\
         4 & 2 (330).jpg \\
         5 & 2 (242).jpg \\
         6 &  2 (21).jpg \\
         7 &  2 (12).jpg \\
         8 & 3 (195).jpg \\
         9 & 2 (364).jpg \\
        10 &   2 (3).jpg \\
        11 & 4 (273).jpg \\
        12 &  4 (26).jpg \\
        13 & 4 (195).jpg \\
        14 &  3 (57).jpg \\
        15 & 2 (147).jpg \\
        16 & 1 (213).jpg \\
        17 & 1 (166).jpg \\
        18 & 4 (352).jpg \\
        19 & 3 (270).jpg \\
        20 & 3 (237).jpg \\
        21 & 3 (224).jpg \\
        22 & 2 (223).jpg \\
        23 & 2 (163).jpg \\
        24 & 2 (162).jpg \\
        25 & 2 (145).jpg \\
        26 & 1 (158).jpg \\
        27 & 1 (157).jpg \\
        28 & 1 (140).jpg \\
        29 & 1 (117).jpg \\
        30 & 1 (100).jpg \\
\bottomrule
\end{tabular}
\end{table}
\clearpage 

\subsubsection*{Iteration 2}
\begin{table}[h!]
\centering
\caption{Exploration \& Structure Focus - Iteration 2 Annotated Samples}
\label{tab:exp_iter2_samples}
\begin{tabular}{cl}
\toprule
 Sample No. & Image Name \\
\midrule
         1 & 4 (151).jpg \\
         2 & 4 (133).jpg \\
         3 &  2 (15).jpg \\
         4 & 2 (124).jpg \\
         5 & 2 (257).jpg \\
         6 & 2 (112).jpg \\
         7 & 3 (292).jpg \\
         8 &  3 (21).jpg \\
         9 & 3 (159).jpg \\
        10 &  4 (38).jpg \\
        11 & 4 (349).jpg \\
        12 & 3 (202).jpg \\
        13 & 2 (349).jpg \\
        14 & 2 (302).jpg \\
        15 & 4 (236).jpg \\
        16 & 4 (109).jpg \\
        17 & 3 (310).jpg \\
        18 & 3 (309).jpg \\
        19 & 1 (293).jpg \\
        20 & 3 (339).jpg \\
        21 & 2 (211).jpg \\
        22 & 2 (180).jpg \\
        23 & 2 (155).jpg \\
        24 & 2 (277).jpg \\
        25 &  2 (54).jpg \\
        26 &  1 (64).jpg \\
        27 & 1 (335).jpg \\
        28 & 1 (318).jpg \\
        29 & 1 (311).jpg \\
        30 & 1 (294).jpg \\
\bottomrule
\end{tabular}
\end{table}
\clearpage 

\subsubsection*{Iteration 3}
\begin{table}[h!]
\centering
\caption{Exploration \& Structure Focus - Iteration 3 Annotated Samples}
\label{tab:exp_iter3_samples}
\begin{tabular}{cl}
\toprule
 Sample No. & Image Name \\
\midrule
         1 & 4 (121).jpg \\
         2 & 3 (356).jpg \\
         3 & 3 (183).jpg \\
         4 & 3 (180).jpg \\
         5 & 3 (151).jpg \\
         6 & 3 (290).jpg \\
         7 &  3 (23).jpg \\
         8 &   1 (7).jpg \\
         9 & 3 (359).jpg \\
        10 &  1 (46).jpg \\
        11 & 1 (371).jpg \\
        12 & 1 (271).jpg \\
        13 & 2 (281).jpg \\
        14 & 2 (168).jpg \\
        15 & 2 (125).jpg \\
        16 &  3 (70).jpg \\
        17 & 3 (194).jpg \\
        18 & 3 (126).jpg \\
        19 & 1 (307).jpg \\
        20 & 1 (280).jpg \\
        21 & 4 (203).jpg \\
        22 & 4 (202).jpg \\
        23 & 4 (194).jpg \\
        24 & 3 (258).jpg \\
        25 & 3 (252).jpg \\
        26 & 1 (164).jpg \\
        27 & 1 (133).jpg \\
        28 & 4 (341).jpg \\
        29 & 2 (360).jpg \\
        30 &  2 (31).jpg \\
\bottomrule
\end{tabular}
\end{table}
\clearpage 

\subsubsection*{Iteration 4}
\begin{table}[h!]
\centering
\caption{Exploration \& Structure Focus - Iteration 4 Annotated Samples}
\label{tab:exp_iter4_samples}
\begin{tabular}{cl}
\toprule
 Sample No. & Image Name \\
\midrule
         1 &  2 (65).jpg \\
         2 &  1 (28).jpg \\
         3 & 1 (224).jpg \\
         4 & 1 (265).jpg \\
         5 & 4 (248).jpg \\
         6 & 4 (178).jpg \\
         7 &   4 (1).jpg \\
         8 & 3 (247).jpg \\
         9 & 2 (352).jpg \\
        10 & 1 (302).jpg \\
        11 & 2 (276).jpg \\
        12 & 2 (229).jpg \\
        13 &  2 (33).jpg \\
        14 & 2 (179).jpg \\
        15 & 2 (177).jpg \\
        16 & 1 (217).jpg \\
        17 &  4 (94).jpg \\
        18 &  4 (33).jpg \\
        19 & 4 (182).jpg \\
        20 &  2 (42).jpg \\
        21 & 2 (275).jpg \\
        22 & 2 (160).jpg \\
        23 & 4 (190).jpg \\
        24 &   4 (7).jpg \\
        25 &  4 (63).jpg \\
        26 &  4 (39).jpg \\
        27 & 4 (154).jpg \\
        28 & 2 (151).jpg \\
        29 & 3 (369).jpg \\
        30 & 3 (225).jpg \\
\bottomrule
\end{tabular}
\end{table}
\clearpage 

\subsubsection*{Iteration 5}
\begin{table}[h!]
\centering
\caption{Exploration \& Structure Focus - Iteration 5 Annotated Samples}
\label{tab:exp_iter5_samples}
\begin{tabular}{cl}
\toprule
 Sample No. & Image Name \\
\midrule
         1 & 3 (216).jpg \\
         2 & 3 (177).jpg \\
         3 & 1 (350).jpg \\
         4 & 1 (322).jpg \\
         5 & 4 (319).jpg \\
         6 &  2 (44).jpg \\
         7 & 1 (189).jpg \\
         8 & 4 (358).jpg \\
         9 & 4 (277).jpg \\
        10 & 4 (183).jpg \\
        11 & 4 (174).jpg \\
        12 & 3 (212).jpg \\
        13 & 1 (312).jpg \\
        14 & 1 (260).jpg \\
        15 & 1 (232).jpg \\
        16 &  4 (48).jpg \\
        17 & 4 (308).jpg \\
        18 & 4 (171).jpg \\
        19 & 4 (141).jpg \\
        20 &  4 (67).jpg \\
        21 & 2 (164).jpg \\
        22 & 2 (110).jpg \\
        23 & 2 (116).jpg \\
        24 & 4 (139).jpg \\
        25 & 4 (117).jpg \\
        26 & 4 (145).jpg \\
        27 & 4 (134).jpg \\
        28 & 2 (137).jpg \\
        29 & 2 (102).jpg \\
        30 &  2 (30).jpg \\
\bottomrule
\end{tabular}
\end{table}

\subsection*{Use Case 2: Uncertainty-Driven Focus Annotated Samples}

\subsubsection*{Iteration 1}
\begin{table}[h!]
\centering
\caption{Uncertainty-Driven Focus - Iteration 1 Annotated Samples}
\label{tab:unc_iter1_samples}
\begin{tabular}{cl}
\toprule
 Sample No. & Image Name \\
\midrule
         1 &  2 (82).jpg \\
         2 & 1 (306).jpg \\
         3 & 1 (125).jpg \\
         4 & 1 (124).jpg \\
         5 & 1 (126).jpg \\
         6 & 4 (247).jpg \\
         7 & 4 (216).jpg \\
         8 &  4 (20).jpg \\
         9 & 4 (177).jpg \\
        10 &   4 (1).jpg \\
        11 &  3 (30).jpg \\
        12 & 2 (155).jpg \\
        13 & 1 (255).jpg \\
        14 & 4 (133).jpg \\
        15 &  3 (70).jpg \\
        16 & 3 (222).jpg \\
        17 & 2 (218).jpg \\
        18 & 2 (148).jpg \\
        19 & 3 (284).jpg \\
        20 & 2 (317).jpg \\
        21 & 1 (364).jpg \\
        22 & 1 (153).jpg \\
        23 &  4 (37).jpg \\
        24 & 4 (167).jpg \\
        25 & 1 (281).jpg \\
        26 & 1 (115).jpg \\
        27 &  1 (97).jpg \\
        28 & 1 (299).jpg \\
        29 & 1 (269).jpg \\
        30 & 1 (186).jpg \\
\bottomrule
\end{tabular}
\end{table}
\clearpage 

\subsubsection*{Iteration 2}
\begin{table}[h!]
\centering
\caption{Uncertainty-Driven Focus - Iteration 2 Annotated Samples}
\label{tab:unc_iter2_samples}
\begin{tabular}{cl}
\toprule
 Sample No. & Image Name \\
\midrule
         1 &  4 (94).jpg \\
         2 & 4 (182).jpg \\
         3 &  4 (33).jpg \\
         4 & 2 (258).jpg \\
         5 & 1 (155).jpg \\
         6 & 3 (184).jpg \\
         7 & 3 (181).jpg \\
         8 & 2 (295).jpg \\
         9 & 2 (220).jpg \\
        10 & 2 (121).jpg \\
        11 & 3 (183).jpg \\
        12 & 2 (157).jpg \\
        13 & 1 (213).jpg \\
        14 & 1 (166).jpg \\
        15 & 4 (205).jpg \\
        16 & 3 (162).jpg \\
        17 & 1 (132).jpg \\
        18 & 1 (106).jpg \\
        19 & 3 (167).jpg \\
        20 &  2 (94).jpg \\
        21 & 2 (183).jpg \\
        22 & 4 (117).jpg \\
        23 &  2 (58).jpg \\
        24 &  2 (40).jpg \\
        25 & 2 (163).jpg \\
        26 & 4 (311).jpg \\
        27 &  2 (21).jpg \\
        28 & 1 (335).jpg \\
        29 & 3 (280).jpg \\
        30 & 2 (367).jpg \\
\bottomrule
\end{tabular}
\end{table}
\clearpage 

\subsubsection*{Iteration 3}
\begin{table}[h!]
\centering
\caption{Uncertainty-Driven Focus - Iteration 3 Annotated Samples}
\label{tab:unc_iter3_samples}
\begin{tabular}{cl}
\toprule
 Sample No. & Image Name \\
\midrule
         1 & 3 (371).jpg \\
         2 &  4 (24).jpg \\
         3 & 4 (341).jpg \\
         4 & 4 (149).jpg \\
         5 & 1 (301).jpg \\
         6 &  4 (46).jpg \\
         7 & 2 (179).jpg \\
         8 & 1 (309).jpg \\
         9 &  1 (91).jpg \\
        10 & 4 (198).jpg \\
        11 & 1 (319).jpg \\
        12 & 3 (239).jpg \\
        13 &  1 (43).jpg \\
        14 &  3 (64).jpg \\
        15 &  2 (56).jpg \\
        16 &  2 (30).jpg \\
        17 &   4 (7).jpg \\
        18 & 4 (145).jpg \\
        19 &  3 (28).jpg \\
        20 & 3 (151).jpg \\
        21 &   1 (7).jpg \\
        22 & 1 (265).jpg \\
        23 & 1 (254).jpg \\
        24 & 4 (330).jpg \\
        25 & 1 (167).jpg \\
        26 & 4 (250).jpg \\
        27 & 4 (190).jpg \\
        28 & 4 (184).jpg \\
        29 & 4 (130).jpg \\
        30 & 4 (125).jpg \\
\bottomrule
\end{tabular}
\end{table}
\clearpage 

\subsubsection*{Iteration 4}
\begin{table}[h!]
\centering
\caption{Uncertainty-Driven Focus - Iteration 4 Annotated Samples}
\label{tab:unc_iter4_samples}
\begin{tabular}{cl}
\toprule
 Sample No. & Image Name \\
\midrule
         1 & 1 (146).jpg \\
         2 & 4 (207).jpg \\
         3 & 3 (217).jpg \\
         4 & 1 (374).jpg \\
         5 & 4 (302).jpg \\
         6 &  3 (41).jpg \\
         7 &  4 (39).jpg \\
         8 & 4 (154).jpg \\
         9 & 3 (172).jpg \\
        10 &  2 (12).jpg \\
        11 & 2 (200).jpg \\
        12 &   2 (9).jpg \\
        13 &  2 (89).jpg \\
        14 &  2 (54).jpg \\
        15 & 3 (231).jpg \\
        16 & 3 (277).jpg \\
        17 & 3 (230).jpg \\
        18 & 3 (242).jpg \\
        19 &  4 (67).jpg \\
        20 & 1 (164).jpg \\
        21 & 3 (264).jpg \\
        22 & 3 (322).jpg \\
        23 & 3 (135).jpg \\
        24 & 3 (130).jpg \\
        25 & 1 (287).jpg \\
        26 &  3 (81).jpg \\
        27 & 3 (196).jpg \\
        28 &  3 (15).jpg \\
        29 & 3 (199).jpg \\
        30 & 1 (190).jpg \\
\bottomrule
\end{tabular}
\end{table}
\clearpage 

\subsubsection*{Iteration 5}
\begin{table}[h!]
\centering
\caption{Uncertainty-Driven Focus - Iteration 5 Annotated Samples}
\label{tab:unc_iter5_samples}
\begin{tabular}{cl}
\toprule
 Sample No. & Image Name \\
\midrule
         1 & 4 (115).jpg \\
         2 & 2 (106).jpg \\
         3 &  4 (36).jpg \\
         4 & 4 (147).jpg \\
         5 & 1 (109).jpg \\
         6 & 4 (214).jpg \\
         7 & 4 (126).jpg \\
         8 &  1 (63).jpg \\
         9 &  3 (88).jpg \\
        10 & 3 (236).jpg \\
        11 & 2 (299).jpg \\
        12 & 2 (372).jpg \\
        13 & 2 (357).jpg \\
        14 & 2 (219).jpg \\
        15 & 2 (189).jpg \\
        16 & 1 (338).jpg \\
        17 & 1 (134).jpg \\
        18 & 1 (206).jpg \\
        19 & 3 (229).jpg \\
        20 & 2 (125).jpg \\
        21 & 2 (344).jpg \\
        22 & 1 (368).jpg \\
        23 &  4 (63).jpg \\
        24 &  4 (49).jpg \\
        25 & 4 (166).jpg \\
        26 & 2 (146).jpg \\
        27 & 1 (315).jpg \\
        28 & 3 (182).jpg \\
        29 & 3 (152).jpg \\
        30 & 1 (276).jpg \\
\bottomrule
\end{tabular}
\end{table}
\clearpage

\subsection*{Use Case 3: Balanced Guidance Integration Annotated Samples}

\subsubsection*{Iteration 1}
\begin{table}[h!]
\centering
\caption{Balanced Guidance Integration - Iteration 1 Annotated Samples}
\label{tab:bal_iter1_samples}
\begin{tabular}{cl}
\toprule
 Sample No. & Image Name \\
\midrule
         1 &  4 (286).jpg \\
         2 &  4 (236).jpg \\
         3 &  4 (151).jpg \\
         4 &  4 (109).jpg \\
         5 &  3 (285).jpg \\
         6 &  3 (171).jpg \\
         7 &  1 (364).jpg \\
         8 &  1 (153).jpg \\
         9 &  2 (168).jpg \\
        10 &  2 (148).jpg \\
        11 &  4 (349).jpg \\
        12 &  3 (202).jpg \\
        13 &  2 (302).jpg \\
        14 &  4 (133).jpg \\
        15 &  2 (206).jpg \\
        16 &  1 (281).jpg \\
        17 &  1 (115).jpg \\
        18 &  4 (205).jpg \\
        19 &  3 (153).jpg \\
        20 &  1 (213).jpg \\
        21 &  1 (166).jpg \\
        22 &  3 (291).jpg \\
        23 &  3 (195).jpg \\
        24 &  2 (364).jpg \\
        25 &    2 (3).jpg \\
        26 &  2 (141).jpg \\
        27 &  2 (129).jpg \\
        28 &  2 (111).jpg \\
        29 &  4 (177).jpg \\
        30 &    4 (1).jpg \\
\bottomrule
\end{tabular}
\end{table}
\clearpage

\subsubsection*{Iteration 2}
\begin{table}[h!]
\centering
\caption{Balanced Guidance Integration - Iteration 2 Annotated Samples}
\label{tab:bal_iter2_samples}
\begin{tabular}{cl}
\toprule
 Sample No. & Image Name \\
\midrule
         1 &  1 (355).jpg \\
         2 &  1 (100).jpg \\
         3 &   1 (91).jpg \\
         4 &  3 (133).jpg \\
         5 &  3 (125).jpg \\
         6 &  3 (101).jpg \\
         7 &   3 (95).jpg \\
         8 &  4 (330).jpg \\
         9 &    4 (3).jpg \\
        10 &  4 (166).jpg \\
        11 &  4 (110).jpg \\
        12 &  4 (100).jpg \\
        13 &  3 (181).jpg \\
        14 &  3 (185).jpg \\
        15 &  3 (184).jpg \\
        16 &  3 (179).jpg \\
        17 &  3 (357).jpg \\
        18 &  3 (280).jpg \\
        19 &  3 (260).jpg \\
        20 &   1 (87).jpg \\
        21 &  1 (294).jpg \\
        22 &   2 (54).jpg \\
        23 &  2 (357).jpg \\
        24 &  2 (317).jpg \\
        25 &  2 (295).jpg \\
        26 &  2 (187).jpg \\
        27 &  2 (112).jpg \\
        28 &  1 (297).jpg \\
        29 &  2 (220).jpg \\
        30 &  2 (189).jpg \\
\bottomrule
\end{tabular}
\end{table}
\clearpage

\subsubsection*{Iteration 3}
\begin{table}[h!]
\centering
\caption{Balanced Guidance Integration - Iteration 3 Annotated Samples}
\label{tab:bal_iter3_samples}
\begin{tabular}{cl}
\toprule
 Sample No. & Image Name \\
\midrule
         1 &  4 (207).jpg \\
         2 &    3 (3).jpg \\
         3 &  2 (258).jpg \\
         4 &   3 (34).jpg \\
         5 &  3 (108).jpg \\
         6 &   1 (97).jpg \\
         7 &  1 (349).jpg \\
         8 &  1 (269).jpg \\
         9 &  3 (194).jpg \\
        10 &  2 (221).jpg \\
        11 &  3 (368).jpg \\
        12 &  3 (301).jpg \\
        13 &  3 (267).jpg \\
        14 &  3 (236).jpg \\
        15 &  4 (198).jpg \\
        16 &   4 (48).jpg \\
        17 &   4 (40).jpg \\
        18 &   4 (67).jpg \\
        19 &   4 (96).jpg \\
        20 &  4 (220).jpg \\
        21 &  3 (331).jpg \\
        22 &  1 (290).jpg \\
        23 &   2 (98).jpg \\
        24 &  3 (201).jpg \\
        25 &  3 (162).jpg \\
        26 &  2 (186).jpg \\
        27 &  2 (176).jpg \\
        28 &   1 (42).jpg \\
        29 &  1 (286).jpg \\
        30 &  1 (242).jpg \\
\bottomrule
\end{tabular}
\end{table}
\clearpage

\subsubsection*{Iteration 4}
\begin{table}[h!]
\centering
\caption{Balanced Guidance Integration - Iteration 4 Annotated Samples}
\label{tab:bal_iter4_samples}
\begin{tabular}{cl}
\toprule
 Sample No. & Image Name \\
\midrule
         1 &  4 (130).jpg \\
         2 &   2 (89).jpg \\
         3 &  1 (311).jpg \\
         4 &  4 (145).jpg \\
         5 &  4 (117).jpg \\
         6 &   2 (30).jpg \\
         7 &  4 (182).jpg \\
         8 &   2 (56).jpg \\
         9 &  2 (157).jpg \\
        10 &   3 (74).jpg \\
        11 &   2 (82).jpg \\
        12 &    1 (8).jpg \\
        13 &    1 (7).jpg \\
        14 &  1 (306).jpg \\
        15 &  1 (246).jpg \\
        16 &  1 (118).jpg \\
        17 &  3 (300).jpg \\
        18 &  3 (193).jpg \\
        19 &   2 (80).jpg \\
        20 &  2 (215).jpg \\
        21 &   2 (15).jpg \\
        22 &  2 (126).jpg \\
        23 &   4 (35).jpg \\
        24 &  4 (265).jpg \\
        25 &  4 (238).jpg \\
        26 &  3 (319).jpg \\
        27 &   3 (21).jpg \\
        28 &  3 (128).jpg \\
        29 &  3 (113).jpg \\
        30 &  1 (181).jpg \\
\bottomrule
\end{tabular}
\end{table}
\clearpage 

\subsubsection*{Iteration 5}
\begin{table}[h!]
\centering
\caption{Balanced Guidance Integration - Iteration 5 Annotated Samples}
\label{tab:bal_iter5_samples}
\begin{tabular}{cl}
\toprule
 Sample No. & Image Name \\
\midrule
         1 &  2 (375).jpg \\
         2 &   4 (89).jpg \\
         3 &  3 (259).jpg \\
         4 &  1 (341).jpg \\
         5 &  1 (258).jpg \\
         6 &   4 (50).jpg \\
         7 &   3 (66).jpg \\
         8 &  1 (254).jpg \\
         9 &   3 (57).jpg \\
        10 &  3 (182).jpg \\
        11 &   4 (87).jpg \\
        12 &   4 (85).jpg \\
        13 &   4 (39).jpg \\
        14 &  2 (151).jpg \\
        15 &  2 (146).jpg \\
        16 &  1 (206).jpg \\
        17 &   2 (35).jpg \\
        18 &  2 (282).jpg \\
        19 &   4 (46).jpg \\
        20 &  4 (334).jpg \\
        21 &  4 (190).jpg \\
        22 &  4 (170).jpg \\
        23 &  4 (159).jpg \\
        24 &   2 (84).jpg \\
        25 &   2 (42).jpg \\
        26 &  2 (275).jpg \\
        27 &  2 (167).jpg \\
        28 &  2 (165).jpg \\
        29 &  2 (145).jpg \\
        30 &  1 (227).jpg \\
\bottomrule
\end{tabular}
\end{table}

\end{document}